\documentclass[10pt,twocolumn,letterpaper]{article}

\usepackage{booktabs,multirow}
\usepackage[table,dvipsnames]{xcolor}
\usepackage{graphicx}
\usepackage{titletoc}
\usepackage[pagenumbers]{cvpr} 

\definecolor{cvprblue}{rgb}{0.21,0.49,0.74}
\usepackage[pagebackref,breaklinks,colorlinks,allcolors=cvprblue]{hyperref}

\def\confName{CVPR}
\def\confYear{2027}

\title{iLens: Decoding and Editing Large Language and Vision-Language Model Representations through Isolated Computation}

\author{
Mohammadali Banayeeanzade\textsuperscript{1,*}\quad
Hosein Hasani\textsuperscript{1,*}\\
Reihaneh Zohrabi\textsuperscript{2}\quad
Mahdieh Soleymani Baghshah\textsuperscript{1}\\[3pt]
\textsuperscript{1}Sharif University of Technology\quad
\textsuperscript{2}Technical University of Darmstadt\\[3pt]
{\ttfamily\small \{mohammad.bana,hosein.hasani,soleymani\}@sharif.edu\quad reihaneh.zohrabi@tu-darmstadt.de}\\[2pt]
{\small \textsuperscript{*}Equal contribution.}
\vspace{-2.0mm}
}

\begin{document}
\maketitle

\begin{abstract}
Readout tools play an important role in mechanistic interpretability by mapping intermediate transformer representations to interpretable concepts. Existing linear readouts may miss concepts recoverable through nonlinear computation, while unrestricted nonlinear decoding can retrieve additional information from other prompt positions. We introduce iLens, a training-free mechanistic interpretability tool for reading and editing concepts in intermediate transformer representations. By applying attention knockout to block access to other token representations in the remaining layers, iLens makes the readout depend only on the selected residual embedding while retaining the model’s pretrained nonlinear transformations. This isolation strategy also supports targeted interventions that inject or suppress specified concepts through a regularized gradient update based on the first-order Taylor approximation of concept log-odds.
Extensive experiments on five textual and five visual concept-decoding tasks across 18 LLMs, 12 LVLMs, and two unified vision models show that iLens improves Pass@10 and Confidence over baseline readouts, with target concepts often recoverable at earlier layers. Intervention experiments across LLMs, LVLMs, and unified vision models further demonstrate that iLens enables more effective concept editing than baseline methods.
\end{abstract}

\vspace{-3.0mm}
\section{Introduction}

 \begin{figure}[!t]
    \centering
    \includegraphics[width=0.9\linewidth]{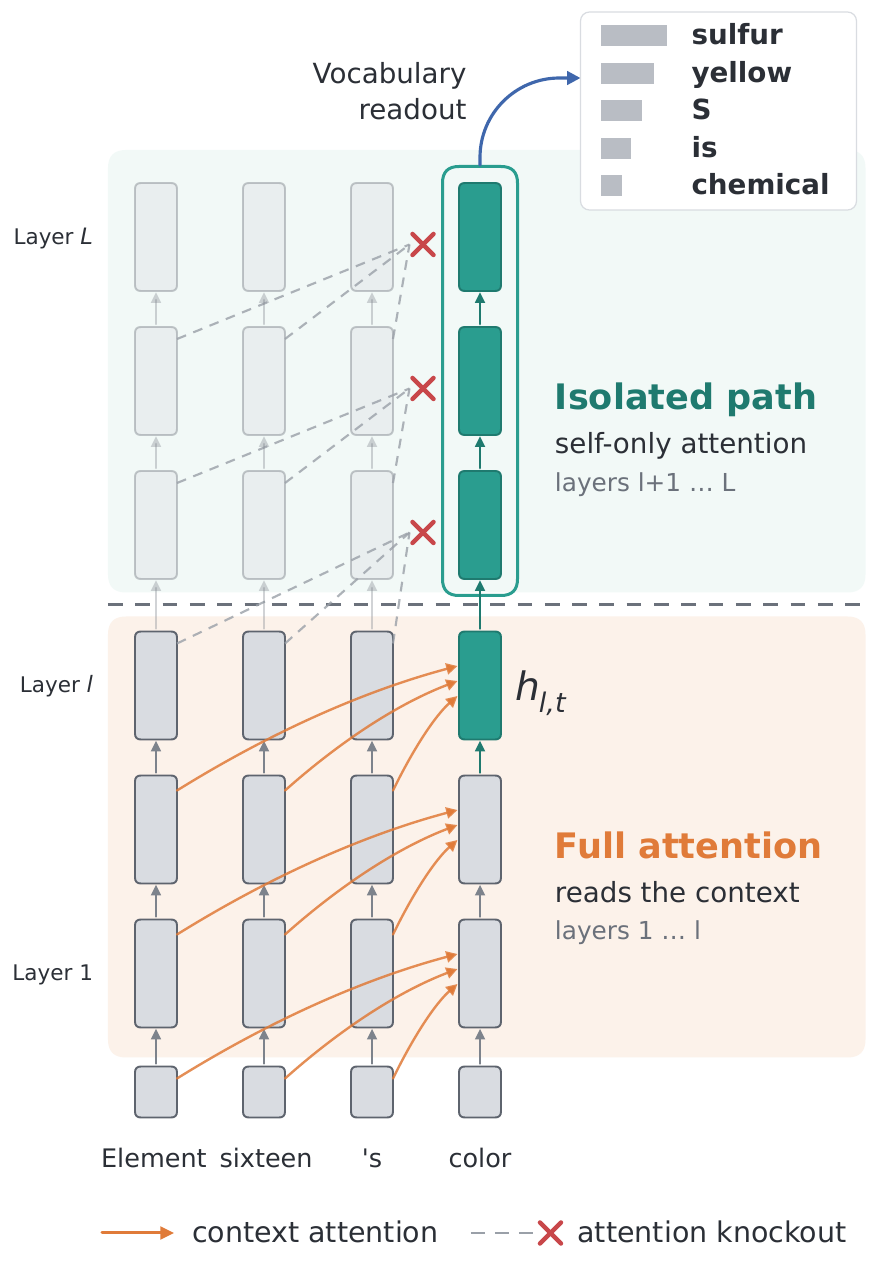}
    \caption{
    \textbf{Overview of iLens.} The selected residual embedding \(h_{\ell,t}\) is propagated through the remaining layers with self-only attention, while MLPs and residual connections remain active. Attention knockout blocks additional information from other token positions, and the final vocabulary readout reveals concepts recoverable from the selected embedding.
    This schematic example illustrates iLens’s goal of recovering intermediate concepts rather than merely predicting the next token.
    }
    \label{fig:ilens}
    \vspace{-4.0mm}
\end{figure}

\begin{figure*}[!t]
    \centering
    \includegraphics[width=0.99\linewidth]{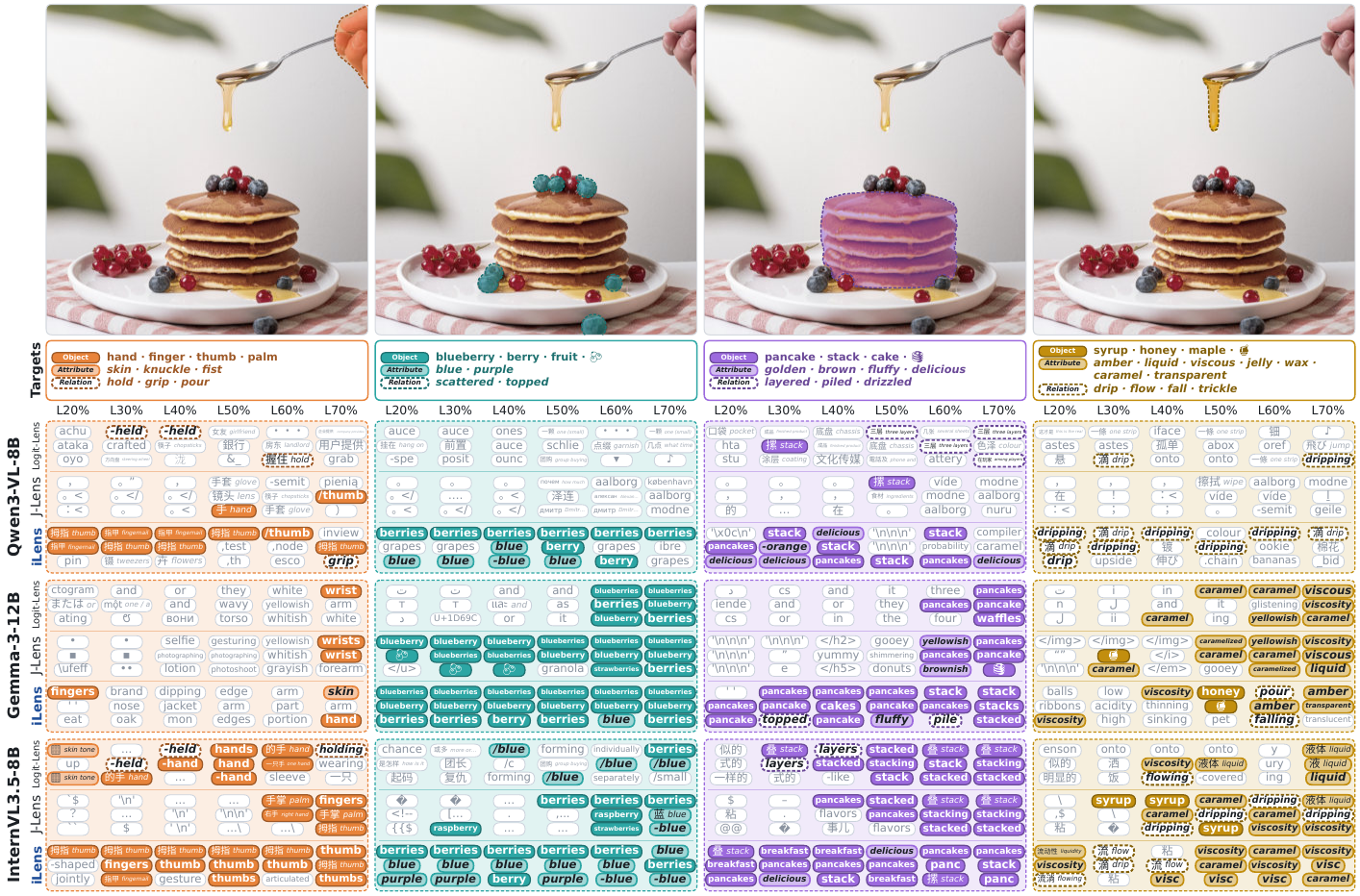}
    \caption{\textbf{iLens reveals meaningful visual concepts in early layers, where linear readouts are often less reliable.} It recovers object identities, colors, attributes, and non-spatial relations from intermediate image-token representations. We show top-3 vocabulary readouts from Logit Lens, J-Lens, and iLens across 20\%--70\% of model depth for Qwen3-VL-8B~\cite{qwen3vl}, Gemma-3-12B~\cite{gemma3}, and InternVL3.5-8B~\cite{wang2025internvl3_5}. Highlighted predictions identify semantically relevant concepts.}
    \label{fig:main-qualitative-example}    
    \vspace{-4.0mm}
\end{figure*}

Transformer-based language models progressively transform their internal representations, which may encode retrieved facts, entities, and partial computations that are not directly reflected in the model's final predictions \citep{gurnee2026verbalizable}. Understanding these representations and their contribution to model predictions is an important goal of mechanistic interpretability \citep{elhage2021mathematical}.
However, intermediate activations are high-dimensional vectors that are difficult to interpret directly. This has motivated the development of methods that map internal representations into more interpretable forms.

One common approach is to map intermediate representations into the model's vocabulary space using readout tools. The Logit Lens applies the model's final normalization and output projection directly to intermediate residual embeddings \citep{nostalgebraist2020logitlens}, while the Embedding Lens ranks vocabulary tokens by the cosine similarity between intermediate representations and the model's input token embeddings \citep{embedding_lens}.
These simple readouts, while being applicable, do not explicitly account for shifts in representation distributions across layers. Tuned Lens tries to reduce this mismatch by learning a separate affine transformation for each layer before applying the Logit Lens \citep{belrose2023tunedlens}.
J-Lens, a more recent linear readout, uses Jacobian-based approximations of downstream computation to identify verbalizable concepts within intermediate representations \citep{gurnee2026verbalizable}.
These methods have provided evidence of intermediate variables involved in factual recall and reasoning, including concepts that do not appear in the model's generated output \citep{gurnee2026verbalizable}.

Linear readouts, while easy to use, have limited decoding power because of the approximations they make. Moreover, methods such as Tuned Lens and J-Lens require constructing a readout for each layer of each model \citep{belrose2023tunedlens,gurnee2026verbalizable}. Their performance also depends on the dataset and the procedure used to construct these readouts, making it difficult to use them as a consistent reference for benchmarking across studies.
Methods such as Patchscopes and SelfIE use the nonlinear computation of large language models (LLMs) to decode concepts from internal token representations \citep{ghandeharioun2024patchscopes,chen2024selfie}. They use activation patching to transfer selected token representations into a target context and continue the forward pass in that context to produce outputs.

The complexity of the activation patching procedure makes these methods less widely used than simple readouts such as Logit Lens.
Additionally, the decoded outputs depend on the target context into which these representations are transferred. During the subsequent unrestricted forward pass in that context, attention allows a token to access information from other positions \citep{vaswani2017attention}, including information that may not be contained in the selected intermediate representation.
Resolving this ambiguity requires a method that retains downstream nonlinear computation while controlling access to other token representations.

To address this problem, we introduce iLens, a training-free method for probing intermediate transformer representations through isolated downstream computation (Figure~\ref{fig:ilens}).
Given the residual embedding at a selected token position and layer, iLens continues the forward pass through the remaining layers, applying attention knockout to block access to other token representations while retaining the selected position's self-attention path.
By blocking alternative routes from the context, this intervention isolates the causal contribution of the selected embedding to the readout while preserving the model's learned nonlinear transformations.
The resulting decoder does not rely on another context or tool and requires no training, fitted parameters, or external data.

Beyond probing, iLens supports targeted interventions by extending the isolated path to the final readout position and differentiating a candidate's log-odds there with respect to the selected residual embedding.
Concept editing is performed via a regularized gradient update derived from a first-order Taylor expansion of the readout score.
The update can suppress a source concept, inject a target concept, or combine both to replace one concept with another.
The edited residual embedding is then returned to the ordinary full-context model to measure its effect during the transformer's natural forward pass.

iLens effectively decodes internal concepts, particularly in vision-language models and at lower layers where linear readouts are often less reliable (Figure~\ref{fig:main-qualitative-example}). Our extensive quantitative evaluations on five textual and five visual concept-decoding tasks across 18 LLMs and 14 vision-capable models, including LVLMs and unified vision models, further demonstrate this capability. In the main comparisons, iLens achieves the highest task-averaged Pass@10 and decoding Confidence across most evaluated models in each modality, and generally recovers concepts at earlier layers. Intervention experiments on four textual tasks across LLMs and a CLEVR-based visual task across vision-capable models show notably stronger average shifts toward injected concepts in full-context output predictions.
These findings support iLens as a complementary training-free mechanistic interpretability tool for concept decoding and targeted editing of residual embeddings across textual and visual modalities.
Code is available at \url{https://github.com/sharif-ml-lab/iLens}.

\vspace{-2.0mm}
\section{iLens}
\label{sec:method}

In this section, we present iLens, a nonlinear readout that relies solely on the transformer's native operations, without requiring additional tools, data, or contexts. We first introduce the transformer notation and operations used in our formulation. We then specify the isolated computation for concept decoding and extend it to derive gradient-based activation edits. We use simplified notation to focus on the main methodological aspects; architecture-specific details are provided in Appendix~\ref{app:implementation_details}.

\subsection{Transformer Computation}
Consider an autoregressive transformer with $L$ blocks and vocabulary $\mathcal{V}$. For input $x_{1:T}$, let $H_\ell=(h_{\ell,1},\ldots,h_{\ell,T})$ contain the residual embeddings after block $\ell$, before final normalization. In vision-language models, this notation also covers visual-token representations in the language decoder. We seek to decode or edit $h_{\ell_0,t}$ at a selected layer $\ell_0$ and position $t$. A concept is specified by a candidate token $c\in\mathcal{V}$, which need not be the input or next token.

For a normalized input $X$, each attention head forms queries $Q=XW_\ell^Q$, keys $K=XW_\ell^K$, and values $V=XW_\ell^V$ using learned projections. The head computes
\begin{equation}
    \operatorname{Attention}(Q,K,V;M)
    =\operatorname{softmax}\!\left(\frac{QK^\top}{\sqrt{d_h}}+M\right)V.
    \label{eq:standard-attention}
\end{equation}
Here $d_h$ is the head dimension, and the additive mask $M$ is zero at allowed connections and $-\infty$ elsewhere. Ordinary causal attention allows each position to attend to itself and earlier positions. Combining the heads and applying the output projection gives the attention branch $\operatorname{Attn}_\ell$.

For a representative pre-normalized block, attention and the token-wise feedforward network (MLP) update the residual stream as
\begin{equation}
    \begin{aligned}
        R_\ell &= H_{\ell-1}+\operatorname{Attn}_\ell\!\left(
            \mathrm{Norm}_\ell^{\mathrm{attn}}(H_{\ell-1});M\right),\\
        H_\ell &= R_\ell+\operatorname{MLP}_\ell\!\left(
            \mathrm{Norm}_\ell^{\mathrm{mlp}}(R_\ell)\right).
    \end{aligned}
    \label{eq:standard-block}
\end{equation}
The $\mathrm{Norm}$ operators normalize each token's representation, while the additions of $H_{\ell-1}$ and $R_\ell$ are residual (skip) connections. The MLP applies a learned nonlinear transformation independently to each token. Final logits are obtained with the model's final normalization $\mathrm{Norm}_f$ and vocabulary projection $U$, including any output bias.

\subsection{Decoding}
\label{sec:decoding}

\paragraph{Isolating the selected representation.}
Starting from $h_{\ell_0,t}$, we apply attention knockout~\cite{geva2023dissecting} in every remaining block, retaining only each position's self-attention connection:
\begin{equation}
    M_{sj}^{\mathrm{self}}=
    \begin{cases}
        0, & j=s,\\
        -\infty, & j\ne s,
    \end{cases}
    \label{eq:knockout-mask}
\end{equation}
which yields $\operatorname{Attention}(Q,K,V;M^{\mathrm{self}})=V$. With only one allowed key, the softmax weight is exactly one, so query--key scores no longer affect the head output. This renormalizes the surviving connection rather than preserving its original attention weight. The value and output transformations remain active, together with normalization, MLPs, and skip connections. Information already accumulated in $h_{\ell_0,t}$ remains available, while other token positions cannot supply additional information after layer $\ell_0$. The readout can therefore still depend on the input context, but only through $h_{\ell_0,t}$.

Under the self-only mask, softmax attention preserves normalized value aggregation. Each head’s output remains in the convex hull of its current value vectors, although this property alone does not guarantee that downstream activations follow the distribution encountered during ordinary full-context inference.

\paragraph{Late-layer attenuation.}
Late layers can shift representations toward next-token prediction~\cite{gurnee2026verbalizable,lad2025stages}. To reduce their influence on intermediate-concept recovery, iLens optionally scales attention and MLP updates by a depth-dependent weight $w_\ell$. Starting from $v_{\ell_0}=h_{\ell_0,t}$, the isolated computation for $\ell=\ell_0+1,\ldots,L$ is
\begin{equation}
    \begin{aligned}
        u_\ell &= v_{\ell-1}+w_\ell\operatorname{Attn}_\ell^{\mathrm{self}}\!\left(
            \mathrm{Norm}_\ell^{\mathrm{attn}}(v_{\ell-1})\right),\\
        v_\ell &= u_\ell+w_\ell\operatorname{MLP}_\ell\!\left(
            \mathrm{Norm}_\ell^{\mathrm{mlp}}(u_\ell)\right),
    \end{aligned}
    \label{eq:isolated-block}
\end{equation}
where $\operatorname{Attn}_\ell^{\mathrm{self}}$ is the attention branch under Eq.~\eqref{eq:knockout-mask}. The weights scale completed branch outputs, after any branch-output normalization, while skip connections retain their native form. Setting $w_\ell=1$ at every remaining layer disables attenuation and retains the full isolated computation.

One option is linear attenuation beyond a chosen model depth. The weight remains one through normalized depth $\tau\in[0,1)$ and then decreases linearly to a small floor:
\begin{equation}
    w_\ell =
    \begin{cases}
        1, & \ell/L \leq \tau, \\
        \rho+(1-\rho)\dfrac{1-\ell/L}{1-\tau}, & \ell/L > \tau.
    \end{cases}
    \label{eq:attenuation}
\end{equation}
The effects of the attenuation threshold and schedule design vary across tasks, with no single setting performing best across all evaluated tasks. We use $\tau=0.8$ and $\rho=0.01$ as shared defaults across models and tasks in our experiments. Appendix~\ref{sec:ablation} examines threshold sensitivity and alternative schedules.

\vspace{-3.0mm}
\paragraph{Concept readout.}
After the isolated computation, the model's output head gives
\begin{equation}
    \begin{aligned}
        z_{\ell_0,t}^{\mathrm{iLens}}
        &= U\!\left(\mathrm{Norm}_f(v_L)\right),\\
        p_{\ell_0,t}^{\mathrm{iLens}}
        &= \operatorname{softmax}(z_{\ell_0,t}^{\mathrm{iLens}}).
    \end{aligned}
    \label{eq:ilens-readout}
\end{equation}
The entry $p_{\ell_0,t,c}^{\mathrm{iLens}}$ scores candidate $c$. At $\ell_0=L$, no blocks remain, so the readout equals the Logit Lens~\cite{nostalgebraist2020logitlens}. At earlier layers, it measures concept recoverability through isolated nonlinear computation.
With model parameters, selected layer and position, and attenuation schedule held fixed, the readout distribution $p_{\ell_0,t}^{\mathrm{iLens}}$ is a function of $h_{\ell_0,t}$ alone. Consequently, in the computation graph modified by attention knockout, $h_{\ell_0,t}$ $d$-separates this readout from the input $x_{1:T}$, giving the conditional independence
\begin{equation}
    p_{\ell_0,t}^{\mathrm{iLens}}
    \perp\!\!\!\perp x_{1:T} \mid h_{\ell_0,t}.
    \label{eq:context-isolation}
\end{equation}

\subsection{Intervention}
\paragraph{Connecting the edit to an output token.}
The goal of intervention is to edit a selected residual embedding $h_{\ell_0,t}$ by suppressing or injecting concepts and to measure the resulting changes in output-token predictions.
The proposed path isolation procedure naturally extends across multiple token positions to retain downstream token interactions, a property not explicitly captured by direct linear readouts.
To guide an edit by its effect on the selected output token, we extend the isolated path from $h=h_{\ell_0,t}$ to the final prompt position $T$. We first run the actual continuation tokens $x_{t+1:T}$ through layer $\ell_0$ on their own, using causal attention within that segment and their original position IDs.
At layer $\ell_0$, we prepend $h$ to these continuation states. Through the remaining blocks, position $t$ attends only to itself, while later positions can attend to $t$ and the continuation up to themselves.
Excluded prefix content can influence the readout only through $h$.
If $t=T$, this reduces to the single-token decoding path.
\vspace{-3.0mm}
\paragraph{Computing the concept edit.}
Let $\bar z_T(h)$ denote the resulting logits at the response prediction site, obtained using the model's final normalization and vocabulary projection. We score candidate $c$ as the next token after $T$ using full-vocabulary log-odds and compute the score's gradient with respect to $h$:
\begin{equation}
    \begin{aligned}
        S_c(h) &= \bar z_{T,c}(h)
        - \operatorname{LSE}_{j\ne c}\bar z_{T,j}(h), \\
        g_c(h) &= \nabla_h S_c(h),
    \end{aligned}
    \label{eq:response-objective}
\end{equation}
Here $\operatorname{LSE}$ is log-sum-exp over the other vocabulary tokens. We compute $g_c(h)$ through the remaining blocks, treating the continuation states at layer $\ell_0$ as fixed inputs. For a desired probability $q^*$, the target log-odds are $S^*=\log(q^*/(1-q^*))$.

We propose \emph{Isolated Taylor Gradient Intervention} (ITGI) for concept editing. It targets $S^*$ using the first-order approximation
\begin{equation}
    S^*=S_c(h+\Delta h)\approx S_c(h)+g_c(h)^\top\Delta h.
\end{equation}
We obtain the edited representation $\widetilde h=h+\Delta h$ using a regularized version of the minimum Euclidean norm solution to this linearized constraint:
\begin{equation}
    \widetilde h
    = h + \frac{S^*-S_c(h)}{\|g_c(h)\|_2^2+\epsilon}\,g_c(h),
    \qquad \epsilon=10^{-4}.
    \label{eq:concept-update}
\end{equation}
To replace a source concept with a target concept, we first suppress the source toward $q^*=0.01$, then inject the target toward $q^*=0.99$. These two objectives can also be applied simultaneously using matrix operations. More generally, the matrix extension of Eq.~\eqref{eq:concept-update} allows multiple concepts to be suppressed and multiple concepts to be injected in a single update.

\vspace{-3.0mm}
\paragraph{Evaluating the edit in the ordinary model.}
We replace $h_{\ell_0,t}$ with $\widetilde h$ in the ordinary full-context trajectory and resume the model's native computation without attention knockout or attenuation. The resulting output measures the edit's behavioral effect separately from its isolated objective. This property could be useful for causal mediation analysis. If multiple layers are edited, after editing the first selected layer, the ordinary model propagates to the next selected layer, where the isolated objective is rebuilt from the newly reached representation.

\begin{figure*}[t]
    \centering
    \includegraphics[width=0.96\linewidth]{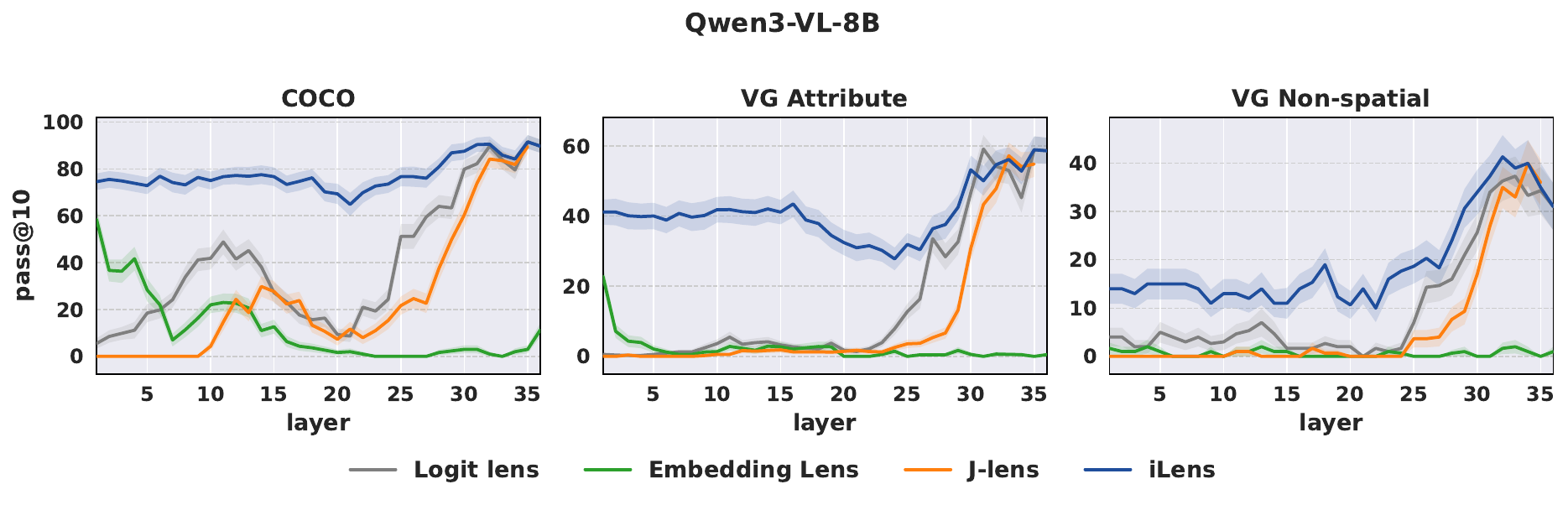}
    \caption{Layer-wise Pass@10 comparison of iLens against existing decoding methods on three visual concept-decoding tasks for Qwen3-VL-8B~\cite{qwen3}. The results are reported for COCO~\cite{coco}, Visual Genome Attribute, and Visual Genome Non-Spatial~\cite{visual_genome} tasks.}
    \label{fig:main-all-layers-curve}    
    \vspace{-3.0mm}
\end{figure*}

\vspace{-1.0mm}
\section{Experiments}

In this section, we describe the experimental setup and metrics, then evaluate iLens on concept decoding and intervention tasks in both vision and language. Appendix~\ref{sec:ablation} presents the ablation study.

\subsection{Experimental Setup}
\label{sec:experimental_setup}

\paragraph{Models.}
We evaluate iLens across diverse model families and parameter scales to assess its applicability across different architectures and modalities. For visual concept decoding, we consider two categories of vision-capable models: (i) large vision-language models (LVLMs) and (ii) unified vision models. For LVLMs, we consider Qwen3.5 (4B, 9B, and 27B)~\cite{qwen3.5}, Qwen3.6-27B~\cite{qwen3.6-27b}, Qwen3-VL-Instruct (4B and 8B)~\cite{qwen3vl}, Gemma3-Instruct (4B and 12B)~\cite{gemma3}, Gemma4-Instruct (12B, 26B-A4B, and 31B)~\cite{gemmateam2026gemma4}, and InternVL3.5-8B~\cite{wang2025internvl3_5}. For unified vision models, we include OmniGen2~\cite{omnigen2} and Show-o2-7B~\cite{xie2025showo2}, both of which support image understanding and generation, and evaluate them through their respective image-understanding pathways.

For text-token decoding, we consider a diverse set of large language models (LLMs), including Qwen3 (0.6B, 4B, 8B, and 14B)~\cite{qwen3}, Qwen3.5 (4B, 9B, and 27B)~\cite{qwen3.5}, Qwen3.6 (27B)~\cite{qwen3.6-27b}, Qwen2.5-7B-Instruct~\cite{qwen2.5}, Gemma3-Instruct models (4B and 12B)~\cite{gemma3}, Gemma-4-Instruct (12B, 26B-A4B, and 31B)~\cite{gemmateam2026gemma4}, OLMo3 (7B and 32B)~\cite{olmo2026olmo3}, Mistral-Small-24B-Instruct~\cite{mistral2025small}, and Gpt-oss-20B~\cite{openai2025gptoss120bgptoss20bmodel}.

\vspace{-3.0mm}
\paragraph{Baselines.}
We compare iLens with Logit Lens~\cite{nostalgebraist2020logitlens}, J-Lens~\cite{gurnee2026verbalizable}, Embedding Lens~\cite{embedding_lens}, and Patchscopes~\cite{ghandeharioun2024patchscopes}. 
We exclude \emph{Tuned Lens}~\cite{belrose2023tunedlens} from our comparisons for two reasons. First, Tuned Lens is trained to approximate the model's next-token distribution at each layer, an objective that prioritizes predicting the model's eventual output rather than recovering intermediate concepts. Consequently, it may overlook intermediate computations in favor of concepts associated with the final prediction. Second,~\citet{gurnee2026verbalizable} report that Tuned Lens underperforms both Logit Lens and J-Lens on intermediate-concept recovery benchmarks, particularly when the next-token prediction does not directly reveal the underlying intermediate concept. These limitations make Tuned Lens less suitable for the concept-decoding objectives considered in our experiments. 

\subsection{Evaluation Metrics}
\label{sec:evaluation_metrics}

We evaluate decoding methods using three complementary metrics: (i) Pass@$k$, which measures whether the target concept appears among the top-$k$ decoded tokens; (ii) Confidence, which measures the probability mass assigned to the target concept; and (iii) First Emergence Depth (FED), a metric we introduce to measure how early target concepts become recoverable in intermediate representations. We retain the Pass@$k$ terminology used in prior work~\cite{gurnee2026verbalizable}, distinguishing it from generation-based Pass@$k$ in reasoning benchmarks. For LLMs, a target is recovered if any surface form appears among the top-$k$ tokens at the probed position; for LVLMs, if it appears among the top-$k$ tokens of at least one corresponding image token. Confidence is computed at the probed position for LLMs and averaged over annotated object patches for LVLMs. We exclude Embedding Lens from Confidence evaluation because it produces cosine similarities rather than probabilities. We report Pass@10 in the main tables and Pass@20 in Appendix~\ref{app:additional_results}.

FED is the normalized index of the first layer where Pass@10 becomes $1$, with $100\%$ assigned if the target is never recovered; lower values indicate earlier recovery. We evaluate Pass@$k$ and Confidence at every layer and summarize their performance curves using the area under the curve (AUC), averaged across examples over normalized layer depth. FED complements AUC by distinguishing methods with similar aggregate performance but different recovery depths. Further implementation details are provided in Appendix~\ref{app:implementation_details}.

\vspace{-3.0mm}
\paragraph{Intervention Metrics.}
To evaluate the effectiveness of interventions, we measure how modifying an intermediate representation changes the recoverability of the injected concept relative to the original source concept. Specifically, for each example $i$, we intervene independently at each layer $\ell$.
For $M \in \{\text{Pass@10}, \text{Pass@20}, \text{Confidence}\}$,
let $\delta M_{w,i,\ell}$ be the post-intervention score minus the
pre-intervention score for word $w$. The intervention score is
\[
\Delta M_{i,\ell}
= \tfrac{1}{2}\bigl(
\delta M_{\mathrm{inj},i,\ell}
- \delta M_{\mathrm{src},i,\ell}
\bigr),
\]
where $\mathrm{inj}$ and $\mathrm{src}$ denote the injected and source
words. Higher scores indicate a stronger shift toward the injected
concept relative to the source. For Verbal Introspection, which has no source concept, we set the source-change term to zero. Finally, we compute the AUC of $\Delta M_{i,\ell}$
across layers.
Appendix~\ref{app:additional_results} gives task- and layer-wise results.

\subsection{Decoding Experiments}

We evaluate iLens on visual and textual concept decoding tasks to test its ability to recover concepts from intermediate transformer representations across different modalities.

\subsubsection{Visual Concept Decoding}

\paragraph{Tasks.}

We evaluate iLens on five visual probing tasks: CLEVR Shape~\cite{clevr}, CLEVR Color~\cite{clevr}, COCO~\cite{coco}, Visual Genome Attribute~\cite{visual_genome}, and Visual Genome Non-Spatial~\cite{visual_genome}, covering object shapes, colors, identities, attributes, and non-spatial relations. These tasks assess the recoverability of diverse object-centric concepts from intermediate image-token representations. Detailed descriptions of the datasets and individual tasks are provided in Appendix~\ref{app:task_definitions}.

\vspace{-2.0mm}
\paragraph{Results.}


\newcommand{\modelblockrule}{%
    \specialrule{0.75pt}{1.5pt}{1.5pt}%
}

\begin{table*}[t]
\centering
\caption{Comparison of iLens with existing decoding methods across five visual probing tasks: CLEVR Shape and CLEVR Color~\cite{clevr}, COCO~\cite{coco}, Visual Genome Attribute, and Visual Genome Non-spatial~\cite{visual_genome}. Evaluated models include Qwen3.5-9B~\cite{qwen3.5}, Qwen3.5-27B~\cite{qwen3.5}, Qwen3-VL-8B~\cite{qwen3vl}, InternVL3.5-8B~\cite{wang2025internvl3_5}, and Show-o2-7B~\cite{xie2025showo2}. Results are reported using Pass@10 (P@10 $\uparrow$), First Emergence Depth (FED $\downarrow$), and Confidence $\uparrow$, together with their averages across tasks. The best results within each model and task are shown in bold.}
\label{tab:main_vlm_results}

\begingroup
\fontsize{7}{8.4}\selectfont
\setlength{\tabcolsep}{1.8pt}
\renewcommand{\arraystretch}{1.08}
\setlength{\aboverulesep}{1.4pt}
\setlength{\belowrulesep}{1.4pt}

\resizebox{\textwidth}{!}{%
\begin{tabular}{@{}ll*{5}{rrr}|rrr@{}}

\toprule[0.9pt]

\multirow{2}{*}{Model}
& \multirow{2}{*}{Method}
& \multicolumn{3}{c}{CLEVR Shape~\cite{clevr}}
& \multicolumn{3}{c}{CLEVR Color~\cite{clevr}}
& \multicolumn{3}{c}{COCO~\cite{coco}}
& \multicolumn{3}{c}{VG Attribute~\cite{visual_genome}}
& \multicolumn{3}{c}{VG Non-spatial~\cite{visual_genome}}
& \multicolumn{3}{|c}{Average} \\

\cmidrule[0.55pt](lr){3-5}
\cmidrule[0.55pt](lr){6-8}
\cmidrule[0.55pt](lr){9-11}
\cmidrule[0.55pt](lr){12-14}
\cmidrule[0.55pt](lr){15-17}
\cmidrule[0.55pt](lr){18-20}

& & P@10 $\uparrow$ & FED $\downarrow$ & Conf. $\uparrow$
& P@10 $\uparrow$ & FED $\downarrow$ & Conf. $\uparrow$
& P@10 $\uparrow$ & FED $\downarrow$ & Conf. $\uparrow$
& P@10 $\uparrow$ & FED $\downarrow$ & Conf. $\uparrow$
& P@10 $\uparrow$ & FED $\downarrow$ & Conf. $\uparrow$
& P@10 $\uparrow$ & FED $\downarrow$ & Conf. $\uparrow$ \\

\midrule[0.65pt]


\multirow{5}{*}{Qwen3-VL-8B~\cite{qwen3vl}}
& Logit-Lens
& 28.82 & 62.57 & 3.35
& 14.61 & 80.97 & 1.42
& 38.76 & 41.00 & 2.09
& 13.85 & 78.19 & 0.56
& 9.59 & 77.58 & 0.41
& 21.13 & 68.06 & 1.57 \\

& Emb-Lens
& 3.67 & 70.41 & --
& 3.30 & 49.16 & --
& 11.09 & 40.51 & --
& 1.88 & 77.32 & --
& 0.61 & 95.43 & --
& 4.11 & 66.57 & -- \\

& J-Lens
& 16.72 & 78.71 & 1.68
& 10.08 & 87.42 & 0.61
& 24.35 & 65.85 & 1.49
& 9.27 & 87.93 & 0.32
& 5.98 & 91.98 & 0.31
& 13.28 & 82.38 & 0.88 \\

& Patchscopes
& 26.47 & 42.60 & 0.62
& 17.49 & 57.79 & 0.26
& 51.86 & 29.28 & 1.09
& 25.27 & 62.56 & 0.31
& 4.52 & 88.68 & 0.05
& 25.12 & 56.18 & 0.47 \\

\rowcolor{blue!7}
\cellcolor{white} & iLens
& \textbf{77.62} & \textbf{13.01} & \textbf{15.40}
& \textbf{59.33} & \textbf{18.47} & \textbf{9.08}
& \textbf{77.06} & \textbf{13.69} & \textbf{11.22}
& \textbf{40.38} & \textbf{39.78} & \textbf{2.33}
& \textbf{18.92} & \textbf{68.19} & \textbf{0.81}
& \textbf{54.66} & \textbf{30.63} & \textbf{7.77} \\

\modelblockrule


\multirow{5}{*}{Qwen3.5-9B~\cite{qwen3.5}}
& Logit-Lens
& 30.68 & 63.46 & 4.16
& 24.45 & 66.84 & 1.85
& 32.75 & 50.18 & 1.94
& 19.62 & 71.79 & 0.54
& 10.23 & 82.00 & 0.46
& 23.55 & 66.85 & 1.79 \\

& Emb-Lens
& 15.26 & 62.54 & --
& 43.13 & 42.52 & --
& 35.58 & 41.07 & --
& 20.68 & 68.41 & --
& 3.05 & 88.57 & --
& 23.54 & 60.62 & -- \\

& J-Lens
& 51.28 & 46.94 & 6.10
& 31.53 & 62.21 & 2.73
& 63.98 & 27.65 & 3.77
& 28.67 & 65.32 & 1.01
& 18.36 & 73.88 & 0.79
& 38.76 & 55.20 & 2.88 \\

& Patchscopes
& 70.30 & 25.47 & 6.60
& 57.97 & 39.06 & 4.11
& 80.01 & 14.67 & 4.12
& 54.56 & 42.12 & \textbf{1.90}
& 24.41 & 69.68 & 0.54
& 57.45 & 38.20 & 3.45 \\

\rowcolor{blue!7}
\cellcolor{white} & iLens
& \textbf{77.54} & \textbf{17.74} & \textbf{10.36}
& \textbf{61.58} & \textbf{15.69} & \textbf{5.05}
& \textbf{80.79} & \textbf{8.91} & \textbf{5.43}
& \textbf{57.14} & \textbf{27.90} & 1.58
& \textbf{26.47} & \textbf{52.84} & \textbf{1.01}
& \textbf{60.70} & \textbf{24.62} & \textbf{4.69} \\

\modelblockrule


\multirow{5}{*}{Qwen3.5-27B~\cite{qwen3.5}}
& Logit-Lens
& 26.97 & 58.28 & 2.86
& 21.03 & 65.34 & 1.42
& 35.67 & 45.48 & 1.45
& 23.18 & 65.43 & 0.45
& 7.79 & 84.38 & 0.23
& 22.93 & 63.78 & 1.28 \\

& Emb-Lens
& 0.75 & 97.99 & --
& 4.56 & 85.11 & --
& 4.81 & 83.12 & --
& 4.30 & 89.28 & --
& 1.29 & 96.21 & --
& 3.14 & 90.34 & -- \\

& J-Lens
& 44.04 & 49.03 & 5.77
& 36.66 & 52.33 & \textbf{3.34}
& 66.52 & 25.75 & \textbf{7.03}
& 30.84 & 60.03 & \textbf{1.40}
& 14.32 & 78.19 & \textbf{0.61}
& 38.48 & 53.07 & 3.63 \\

& Patchscopes
& 35.97 & 56.35 & 3.02
& 29.06 & 67.68 & 2.83
& 53.60 & 28.31 & 1.71
& 28.51 & 67.20 & 1.09
& 10.87 & 85.32 & 0.21
& 31.60 & 60.97 & 1.77 \\

\rowcolor{blue!7}
\cellcolor{white} & iLens
& \textbf{72.70} & \textbf{23.62} & \textbf{9.09}
& \textbf{66.86} & \textbf{13.29} & 3.09
& \textbf{73.44} & \textbf{14.86} & 4.56
& \textbf{51.50} & \textbf{34.79} & 1.35
& \textbf{19.15} & \textbf{68.78} & 0.56
& \textbf{56.73} & \textbf{31.07} & \textbf{3.73} \\

\modelblockrule

\multirow{5}{*}{InternVL3.5-8B~\cite{wang2025internvl3_5}}
& Logit-Lens
& 21.67 & 71.51 & 1.64
& 22.39 & 71.27 & 2.48
& 57.08 & 26.61 & 6.41
& 37.02 & 52.42 & 3.63
& 27.76 & 53.74 & 1.35
& 33.18 & 55.11 & 3.10 \\

& Emb-Lens
& 9.05 & 76.58 & --
& 19.37 & 37.77 & --
& 14.24 & 49.92 & --
& 7.12 & 74.38 & --
& 1.75 & 91.94 & --
& 10.31 & 66.12 & -- \\

& J-Lens
& 33.51 & 59.10 & 2.06
& 38.02 & 53.41 & 3.78
& 63.35 & 34.12 & 12.85
& 36.95 & 58.64 & 4.83
& 21.51 & 72.30 & 1.46
& 38.67 & 55.51 & 5.00 \\

& Patchscopes
& 28.89 & 58.94 & 0.51
& 41.30 & 38.80 & 0.98
& 75.38 & 20.50 & 4.33
& 52.18 & 43.77 & 2.28
& 19.94 & 75.47 & 0.54
& 43.54 & 47.50 & 1.73 \\

\rowcolor{blue!7}
\cellcolor{white} & iLens
& \textbf{49.28} & \textbf{36.76} & \textbf{3.02}
& \textbf{83.03} & \textbf{9.51} & \textbf{8.46}
& \textbf{87.27} & \textbf{10.39} & \textbf{22.59}
& \textbf{63.36} & \textbf{29.57} & \textbf{7.35}
& \textbf{33.38} & \textbf{53.29} & \textbf{1.82}
& \textbf{63.26} & \textbf{27.90} & \textbf{8.65} \\

\modelblockrule


\multirow{5}{*}{Show-o2-7B~\cite{xie2025showo2}}
& Logit-Lens
& 12.52 & 70.28 & 0.52
& 11.01 & 71.53 & 0.37
& 24.45 & 61.30 & 4.23
& 17.82 & 72.51 & 1.58
& 12.93 & 81.13 & 1.86
& 15.75 & 71.35 & 1.71 \\

& Emb-Lens
& 10.11 & 65.97 & --
& 47.69 & 24.30 & --
& 39.09 & 16.85 & --
& 32.63 & 46.25 & --
& 3.13 & 83.08 & --
& 26.53 & 47.29 & -- \\

& J-Lens
& 38.17 & 57.97 & 2.69
& 45.04 & 51.72 & 3.21
& 65.64 & 35.01 & 19.17
& 56.17 & 43.81 & 9.11
& 32.79 & 64.42 & \textbf{5.05}
& 47.56 & 50.59 & 7.85 \\

& Patchscopes
& 34.40 & 61.35 & 1.42
& 53.65 & 40.33 & 1.61
& 88.81 & 8.02 & 14.81
& 60.92 & 38.10 & 3.71
& 34.30 & 62.93 & 2.79
& 54.42 & 42.15 & 4.87 \\

\rowcolor{blue!7}
\cellcolor{white} & iLens
& \textbf{78.40} & \textbf{12.85} & \textbf{4.62}
& \textbf{88.82} & \textbf{8.87} & \textbf{8.69}
& \textbf{95.96} & \textbf{5.27} & \textbf{26.66}
& \textbf{81.26} & \textbf{14.04} & \textbf{12.56}
& \textbf{50.36} & \textbf{39.75} & 5.00
& \textbf{78.96} & \textbf{16.16} & \textbf{11.51} \\

\modelblockrule
\end{tabular}%
}

\endgroup
\vspace{-2.0mm}
\end{table*}

\begin{figure*}[t]
    \centering
    \includegraphics[width=0.93\linewidth]{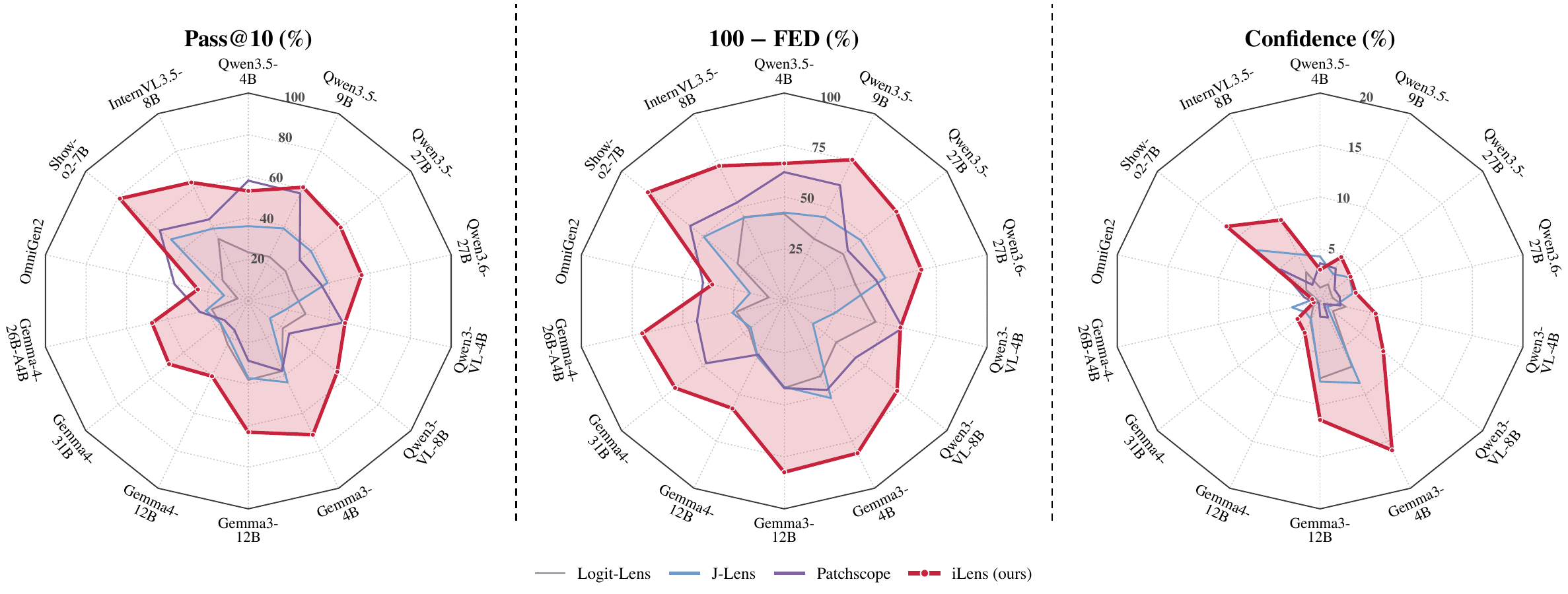}
    \caption{Visual concept-decoding performance across 14 vision-capable models, averaged over the five visual tasks. Each spoke corresponds to one model. The three panels report Pass@10, $100-\mathrm{FED}$, and Confidence, respectively, for Logit Lens, J-Lens, Patchscopes, and iLens. Pass@10 and Confidence are normalized layer AUCs. Higher values indicate better performance in all panels.}
    \label{fig:radar-vlm}
    \vspace{-3.0mm}
\end{figure*}

Table~\ref{tab:main_vlm_results} reports the results across five visual probing tasks. As illustrated, iLens consistently outperforms existing decoding methods in terms of Pass@10 and FED, achieving the best average performance across all models. For example, on Qwen3-VL-8B, iLens improves average Pass@10 from 25.12\% with Patchscopes to 54.66\%, a 29.54 percentage-point improvement, while reducing FED from 56.18 to 30.63. Furthermore, iLens achieves the highest average Confidence across all models.
Figure~\ref{fig:radar-vlm} summarizes task-averaged decoding performance across all 14 vision-capable models.

To provide a more detailed view of decoding performance across model depth, Figure~\ref{fig:main-all-layers-curve} presents the layer-wise Pass@10 results of iLens and three baseline methods on Qwen3-VL-8B across three visual decoding tasks.
iLens achieves higher Pass@10 across a broad range of layers, with a notable gap in \textit{earlier layers}.
This pattern suggests that early visual representations contain concept-relevant information that can be recovered through downstream nonlinear computation even when it is not readily accessible to the evaluated linear readouts.
The weak early-layer performance of Embedding Lens further indicates that direct similarity to input token embeddings does not resolve this gap.
These observations motivate the hypothesis that initial concept representations may be non-lexical yet recoverable through nonlinear decoding, although further targeted evaluation is needed to assess this hypothesis.

In addition, Figure~\ref{fig:main-qualitative-example} presents qualitative comparisons across three representative LVLMs: Qwen3-VL-8B, Gemma-3-12B, and InternVL3.5-8B. The examples cover diverse visual concepts, including object identities, colors, attributes, and non-spatial relations. We examine the decoded vocabulary predictions across intermediate layers, spanning 20\% to 70\% of model depth. As depicted, iLens recovers semantically relevant concepts at earlier layers, producing predictions that align more closely with the annotated object identities, visual attributes, and relations. 

Overall, these results demonstrate that iLens enables more accurate and earlier recovery of visual concepts across diverse vision-capable models and probing tasks. Additional quantitative and qualitative results and detailed comparisons are provided in Appendix~\ref{app:subsec:lvlm_results}.
\subsubsection{Textual Concept Decoding}

\paragraph{Tasks.}

We evaluate iLens on five text-probing tasks~\cite{gurnee2026verbalizable}: \emph{Typo}, \emph{Multilingual}, \emph{Multihop}, \emph{Order of Operations}, and \emph{Dual Task}, covering linguistic, multilingual, factual-retrieval, computational, and concurrent task-processing concepts. These tasks assess whether intermediate representations encode information that may not be explicitly expressed in the input or reflected in the model's final output. Detailed descriptions of the individual tasks are provided in Appendix~\ref{app:task_definitions}.
\vspace{-2.0mm}

\begin{table*}[t]
\centering
\caption{Comparison of iLens with existing decoding methods across five language-model probing tasks: Typo, Multilingual, Multihop, Order of Operations, and Dual Task. Evaluated models include Qwen3.5-9B~\cite{qwen3.5}, Qwen3.5-27B~\cite{qwen3.5}, Qwen3-4B-Instruct~\cite{qwen3}, Gemma3-12B-Instruct~\cite{gemma3}, and OLMo3-7B~\cite{olmo2026olmo3}. Results are reported using Pass@10 (P@10 $\uparrow$), First Emergence Depth (FED $\downarrow$), and Confidence (Conf. $\uparrow$), together with their averages across tasks. The best results within each model and task are shown in bold.}
\label{tab:main_llm_results}

\begingroup
\fontsize{6}{7.2}\selectfont
\setlength{\tabcolsep}{1.2pt}
\renewcommand{\arraystretch}{1.05}
\setlength{\aboverulesep}{1.2pt}
\setlength{\belowrulesep}{1.2pt}

\resizebox{\textwidth}{!}{%
\begin{tabular}{@{}ll*{5}{rrr}|rrr@{}}
\toprule[0.9pt]
\multirow{2}{*}{Model} & \multirow{2}{*}{Method}
& \multicolumn{3}{c}{Typo}
& \multicolumn{3}{c}{Multilingual}
& \multicolumn{3}{c}{Multihop}
& \multicolumn{3}{c}{Order of Operations}
& \multicolumn{3}{c}{Dual Task}
& \multicolumn{3}{|c}{Average} \\
\cmidrule[0.55pt](lr){3-20}
& & P@10 $\uparrow$ & FED $\downarrow$ & Conf. $\uparrow$
& P@10 $\uparrow$ & FED $\downarrow$ & Conf. $\uparrow$
& P@10 $\uparrow$ & FED $\downarrow$ & Conf. $\uparrow$
& P@10 $\uparrow$ & FED $\downarrow$ & Conf. $\uparrow$
& P@10 $\uparrow$ & FED $\downarrow$ & Conf. $\uparrow$
& P@10 $\uparrow$ & FED $\downarrow$ & Conf. $\uparrow$ \\
\midrule[0.65pt]


\multirow{5}{*}{Qwen3.5-9B~\cite{qwen3.5}}
& Logit-Lens
& 13.1 & 44.4 & 0.4
& 20.4 & 74.2 & 4.6
& 11.3 & 80.7 & 1.1
& 30.3 & \textbf{32.8} & 1.2
& 6.1 & 86.4 & 0.4
& 16.2 & 63.7 & 1.5 \\

& Emb-Lens
& \textbf{82.1} & \textbf{9.6} & --
& 2.4 & 89.9 & --
& 3.1 & 90.2 & --
& 0.0 & 100.0 & --
& 7.2 & 86.0 & --
& 19.0 & 75.1 & -- \\

& J-Lens
& 18.3 & 64.2 & \textbf{2.3}
& \textbf{33.7} & 62.1 & \textbf{9.8}
& 24.5 & 67.8 & 4.3
& 24.0 & 56.1 & 1.4
& 15.4 & 78.4 & \textbf{1.9}
& 23.2 & 65.7 & 3.9 \\

& Patchscopes
& 1.1 & 92.7 & 0.1
& 19.5 & 73.4 & 2.0
& 8.7 & 82.8 & 0.4
& \textbf{62.0} & 39.9 & \textbf{2.7}
& \textbf{28.4} & \textbf{50.6} & 0.8
& 23.9 & 67.9 & 1.2 \\

\rowcolor{blue!7}
\cellcolor{white} & iLens
& 15.6 & 21.5 & 1.9
& 33.4 & \textbf{60.2} & 8.5
& \textbf{30.3} & \textbf{59.5} & \textbf{6.2}
& 32.3 & 48.3 & 2.0
& 18.2 & 60.4 & 1.4
& \textbf{25.9} & \textbf{50.0} & \textbf{4.0} \\

\modelblockrule


\multirow{5}{*}{Qwen3.5-27B~\cite{qwen3.5}}
& Logit-Lens
& 22.6 & 43.0 & 0.4
& 17.9 & 71.4 & 3.7
& 14.0 & 71.2 & 1.1
& 34.1 & \textbf{32.2} & 0.9
& 19.3 & 64.6 & 2.1
& 21.6 & 56.5 & 1.6 \\

& Emb-Lens
& \textbf{43.7} & \textbf{19.5} & --
& 0.0 & 100.0 & --
& 2.1 & 88.9 & --
& 13.7 & 69.7 & --
& 11.6 & 64.6 & --
& 14.2 & 68.6 & -- \\

& J-Lens
& 12.1 & 72.5 & \textbf{2.0}
& 22.5 & 64.9 & 7.8
& \textbf{30.9} & \textbf{56.9} & \textbf{8.4}
& 12.1 & 75.9 & 1.6
& 26.9 & 62.7 & \textbf{7.3}
& 20.9 & 66.6 & \textbf{5.4} \\

& Patchscopes
& 1.5 & 93.8 & 0.1
& 11.2 & 83.6 & 1.1
& 9.6 & 83.4 & 0.2
& \textbf{62.0} & 39.0 & \textbf{2.6}
& 24.3 & \textbf{58.2} & 1.0
& 21.8 & 71.6 & 1.0 \\

\rowcolor{blue!7}
\cellcolor{white} & iLens
& 23.9 & 25.9 & 1.8
& \textbf{30.4} & \textbf{64.7} & \textbf{8.4}
& 26.8 & 61.1 & 5.5
& 37.5 & 37.6 & 1.3
& \textbf{28.6} & 63.1 & 4.4
& \textbf{29.4} & \textbf{50.5} & 4.3 \\

\modelblockrule


\multirow{5}{*}{Qwen3-4B-Inst.~\cite{qwen3}}
& Logit-Lens
& 16.7 & 45.8 & 2.2
& \textbf{23.0} & \textbf{70.0} & \textbf{7.1}
& 7.8 & 81.1 & 0.5
& 13.6 & 77.8 & 2.7
& 8.1 & 79.6 & 1.5
& 13.8 & 70.9 & 2.8 \\

& Emb-Lens
& 8.4 & 51.2 & --
& 12.8 & 84.3 & --
& 3.6 & 86.2 & --
& 3.6 & 91.8 & --
& 5.7 & 91.6 & --
& 6.8 & 81.0 & -- \\

& J-Lens
& 4.7 & 86.7 & 0.3
& 15.2 & 82.4 & 5.7
& 11.5 & 79.6 & 1.8
& 8.8 & 63.3 & 0.6
& 6.8 & 79.4 & 1.5
& 9.4 & 78.3 & 2.0 \\

& Patchscopes
& 0.1 & 98.1 & 0.0
& 17.5 & 75.7 & 2.4
& 7.9 & 86.4 & 0.3
& \textbf{61.9} & \textbf{39.7} & \textbf{2.8}
& \textbf{27.0} & \textbf{47.0} & 1.2
& \textbf{22.9} & 69.4 & 1.3 \\

\rowcolor{blue!7}
\cellcolor{white} & iLens
& \textbf{28.7} & \textbf{21.9} & \textbf{5.4}
& 18.7 & 78.2 & 5.0
& \textbf{25.4} & \textbf{49.1} & \textbf{5.2}
& 14.2 & 65.5 & 1.7
& 11.7 & 66.6 & \textbf{2.5}
& 19.7 & \textbf{56.3} & \textbf{3.9} \\

\modelblockrule









\multirow{5}{*}{Gemma3-12B-Inst.~\cite{gemma3}}
& Logit-Lens
& 11.2 & 70.1 & 0.9
& \textbf{40.2} & 55.7 & \textbf{18.0}
& 14.2 & 77.6 & 3.0
& 26.5 & 59.8 & \textbf{6.8}
& 22.9 & 68.3 & 5.8
& 23.0 & 66.3 & 6.9 \\

& Emb-Lens
& 0.1 & 98.0 & --
& 1.3 & 95.5 & --
& 0.1 & 99.6 & --
& 0.0 & 100.0 & --
& 0.0 & 100.0 & --
& 0.3 & 98.6 & -- \\

& J-Lens
& 7.2 & 78.2 & 1.3
& 30.2 & 64.6 & 13.5
& 20.1 & 68.2 & 5.7
& 17.5 & 68.3 & 6.3
& 17.7 & 77.6 & 6.6
& 18.5 & 71.4 & 6.7 \\

& Patchscopes
& 0.0 & 100.0 & 0.0
& 19.2 & 74.9 & 1.5
& 9.8 & 85.0 & 0.2
& \textbf{55.3} & \textbf{39.6} & 3.5
& 13.1 & 72.6 & 0.7
& 19.5 & 74.4 & 1.2 \\

\rowcolor{blue!7}
\cellcolor{white} & iLens
& \textbf{20.8} & \textbf{20.8} & \textbf{3.4}
& 38.5 & \textbf{49.8} & 15.1
& \textbf{27.9} & \textbf{41.7} & \textbf{6.3}
& 21.8 & 42.1 & 5.5
& \textbf{30.4} & \textbf{48.9} & \textbf{8.4}
& \textbf{27.9} & \textbf{40.7} & \textbf{7.8} \\

\modelblockrule


\multirow{5}{*}{Olmo3-7B~\cite{olmo2026olmo3}}
& Logit-Lens
& 11.1 & 71.8 & 0.5
& \textbf{29.2} & \textbf{66.3} & \textbf{11.4}
& 5.5 & 88.9 & 0.4
& 25.2 & 63.5 & 3.9
& 10.6 & 77.8 & 0.5
& 16.3 & 73.7 & 3.3 \\

& Emb-Lens
& \textbf{89.2} & \textbf{12.0} & --
& 0.3 & 98.8 & --
& \textbf{18.6} & 72.5 & --
& 0.1 & 96.1 & --
& 2.9 & 81.3 & --
& 22.2 & 72.2 & -- \\

& J-Lens
& 14.3 & 69.6 & 0.9
& 25.7 & 71.2 & 7.5
& 18.4 & \textbf{70.9} & \textbf{3.2}
& 27.2 & 54.0 & 4.5
& 10.6 & 87.6 & \textbf{1.0}
& 19.2 & 70.6 & 3.4 \\

& Patchscopes
& 0.1 & 98.4 & 0.0
& 9.1 & 81.6 & 0.5
& 7.0 & 88.8 & 0.5
& 49.6 & 43.9 & 4.5
& \textbf{31.3} & \textbf{51.6} & \textbf{1.0}
& 19.4 & 72.9 & 1.3 \\

\rowcolor{blue!7}
\cellcolor{white} & iLens
& 34.9 & 29.6 & \textbf{4.0}
& 26.3 & 70.6 & 9.6
& 12.3 & 73.6 & 1.4
& \textbf{51.6} & \textbf{26.1} & \textbf{6.3}
& 16.1 & 66.1 & 0.6
& \textbf{28.2} & \textbf{53.2} & \textbf{4.4} \\

\modelblockrule

\end{tabular}%
}
\endgroup
\vspace{-2.0mm}
\end{table*}

\begin{figure*}[t]
    \centering
    \includegraphics[width=0.93\linewidth]{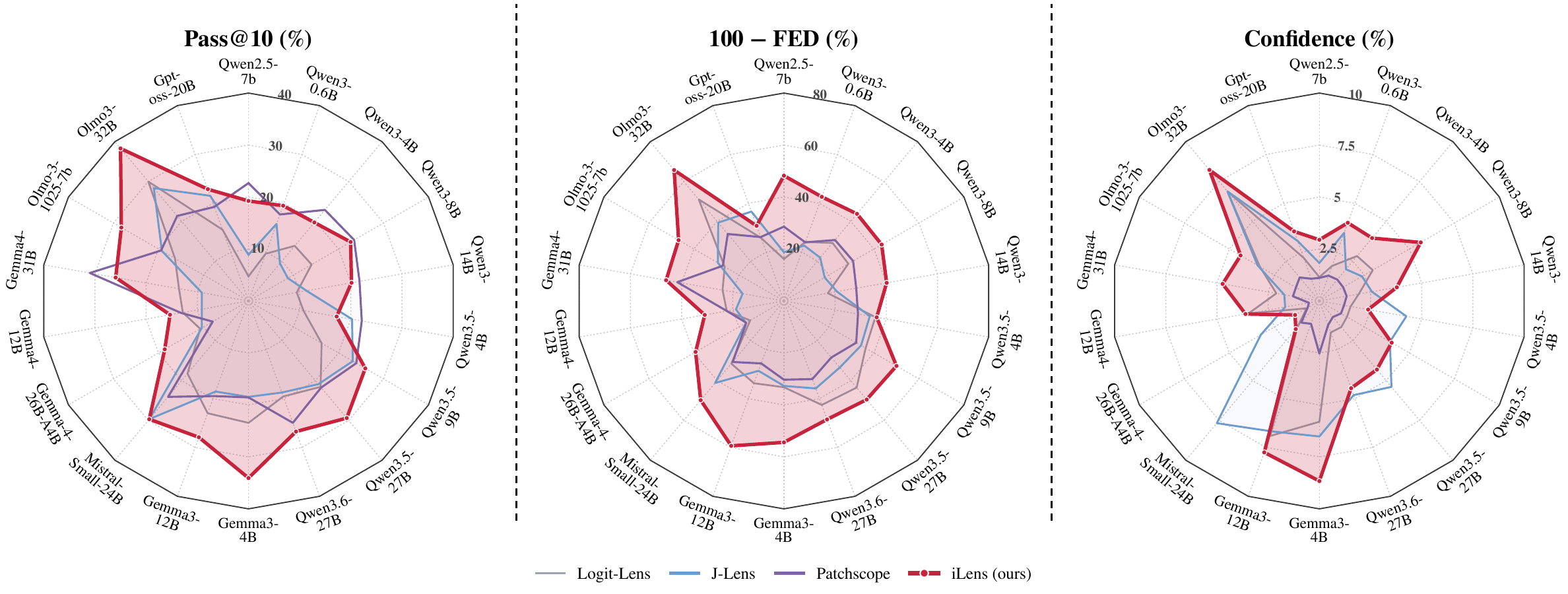}
    \caption{Textual concept-decoding performance across 18 LLMs, averaged over the five textual tasks separately for each model. Each spoke corresponds to one model. The three panels report Pass@10, $100-\mathrm{FED}$, and Confidence, respectively, for Logit Lens, J-Lens, Patchscopes, and iLens. Pass@10 and Confidence are normalized layer AUCs. Higher values indicate better performance in all panels.}
    \label{fig:radar-llm}
    \vspace{-3.0mm}
\end{figure*}

\paragraph{Results.}
Table~\ref{tab:main_llm_results} presents the textual concept-decoding results across five tasks. iLens achieves the best average Pass@10 on four out of five models and the lowest average FED across all five models, indicating more effective and earlier concept recovery. It also achieves the highest average Confidence on four models. For example, on Qwen3.5-9B~\cite{qwen3.5}, iLens achieves an average Pass@10 of 25.9\%, outperforming the strongest baseline, Patchscopes with 23.9\%. Furthermore, iLens reduces the average FED to 50.0\%, compared with 63.7\% for Logit-Lens, and achieves an average Confidence of 4.0\%, surpassing the strongest baseline, J-Lens (3.9\%). Notable improvements are also observed on Typo and Multihop, particularly for Qwen3-4B~\cite{qwen3} and Gemma3-12B~\cite{gemma3}. However, J-Lens performs better on Multihop for Qwen3.5-27B~\cite{qwen3.5} and OLMo3-7B~\cite{olmo2026olmo3}, while Patchscopes shows strong performance on Order of Operations across several models. Despite these task-specific exceptions, iLens achieves the strongest performance across diverse LLM architectures.
Figure~\ref{fig:radar-llm} summarizes task-averaged decoding performance across all 18 LLMs. Detailed task- and layer-wise results are provided in Appendix~\ref{app:subsec:llm_results}.

\begin{table*}[t]
\centering
\caption{Comparison of iLens with existing intervention methods across four
text-based intervention tasks: Probe Swap, Verbal Report, Flexible
Generalization, and Verbal Introspection. Evaluated models include
Qwen3.5-9B~\cite{qwen3.5},
Qwen3-4B~\cite{qwen3}, Gemma3-12B-Inst.~\cite{gemma3},
OLMo-3-7B~\cite{olmo2026olmo3}, and
GPT-oss-20B~\cite{openai2025gptoss120bgptoss20bmodel}.
Results are reported using $\Delta$Pass@10, $\Delta$Pass@20, and
$\Delta$Confidence, computed as one-half of the area under the layer-wise curve
of the difference between the mean injected-word change and the mean
source-word change. Average columns
report the mean across the four tasks. The best results for each
model and metric are shown in bold.}
\label{tab:llm_intervention_results}

\begingroup
\fontsize{6}{7.2}\selectfont
\setlength{\tabcolsep}{1.2pt}
\renewcommand{\arraystretch}{1.05}
\setlength{\aboverulesep}{1.2pt}
\setlength{\belowrulesep}{1.2pt}

\resizebox{\textwidth}{!}{%
\begin{tabular}{@{}ll*{4}{rrr}|rrr@{}}

\toprule[0.9pt]

\multirow{2}{*}{Model}
& \multirow{2}{*}{Method}
& \multicolumn{3}{c}{Probe Swap}
& \multicolumn{3}{c}{Verbal Report}
& \multicolumn{3}{c}{Flexible Generalization}
& \multicolumn{3}{c}{Verbal Introspection}
& \multicolumn{3}{|c}{Average} \\

\cmidrule[0.55pt](lr){3-17}

& &
$\Delta$P@10 $\uparrow$ & $\Delta$P@20 $\uparrow$ & $\Delta$Conf. $\uparrow$
& $\Delta$P@10 $\uparrow$ & $\Delta$P@20 $\uparrow$ & $\Delta$Conf. $\uparrow$
& $\Delta$P@10 $\uparrow$ & $\Delta$P@20 $\uparrow$ & $\Delta$Conf. $\uparrow$
& $\Delta$P@10 $\uparrow$ & $\Delta$P@20 $\uparrow$ & $\Delta$Conf. $\uparrow$
& $\Delta$P@10 $\uparrow$ & $\Delta$P@20 $\uparrow$ & $\Delta$Conf. $\uparrow$ \\

\midrule[0.65pt]


\multirow{4}{*}{Qwen3.5-9B~\cite{qwen3.5}}
& Logit Lens + ITGI
& 8.20 & 13.70 & 5.18
& 0.80 & 2.60 & 0.70
& 3.37 & 0.27 & 0.15
& 0.00 & 0.00 & 0.00
& 3.09 & 4.14 & 1.51 \\

& J-Lens
& 14.17 & 16.73 & 8.51
& 0.67 & 1.47 & 1.59
& 4.09 & 2.05 & 0.78
& 0.03 & 0.03 & 0.00
& 4.74 & 5.07 & 2.72 \\

& J-Lens + ITGI
& 26.07 & 27.07 & 19.60
& 4.67 & 6.50 & 6.48
& 10.15 & 4.20 & 2.42
& 0.00 & 0.40 & 0.01
& 10.22 & 9.54 & 7.13 \\

\rowcolor{blue!7}
\cellcolor{white} & iLens
& \textbf{47.13} & \textbf{45.60} & \textbf{27.10}
& \textbf{20.23} & \textbf{19.70} & \textbf{30.90}
& \textbf{26.29} & \textbf{16.52} & \textbf{15.20}
& \textbf{17.80} & \textbf{19.90} & \textbf{3.44}
& \textbf{27.86} & \textbf{25.43} & \textbf{19.16} \\

\specialrule{0.55pt}{1.2pt}{1.2pt}


\multirow{4}{*}{Qwen3-4B~\cite{qwen3}}
& Logit Lens + ITGI
& 20.89 & 22.25 & 11.40
& 3.29 & 0.82 & 1.02
& 0.00 & 0.00 & 0.19
& 0.00 & 0.00 & 0.00
& 6.05 & 5.77 & 3.15 \\

& J-Lens
& 16.21 & 17.38 & 14.40
& 0.50 & -0.21 & 1.07
& 1.27 & 1.34 & 2.81
& 0.00 & 0.00 & 0.00
& 4.50 & 4.63 & 4.57 \\

& J-Lens + ITGI
& 21.87 & 22.96 & 21.70
& 4.53 & 2.56 & 6.52
& 0.67 & 0.90 & 2.02
& 0.00 & 0.00 & 0.00
& 6.77 & 6.61 & 7.56 \\

\rowcolor{blue!7}
\cellcolor{white} & iLens
& \textbf{56.18} & \textbf{55.84} & \textbf{42.70}
& \textbf{19.79} & \textbf{14.15} & \textbf{37.50}
& \textbf{9.36} & \textbf{7.65} & \textbf{10.20}
& \textbf{10.00} & \textbf{11.29} & \textbf{3.83}
& \textbf{23.83} & \textbf{22.23} & \textbf{23.56} \\

\specialrule{0.55pt}{1.2pt}{1.2pt}


\multirow{4}{*}{Gemma3-12B-Inst.~\cite{gemma3}}
& Logit Lens + ITGI
& 5.30 & 6.98 & 7.11
& 3.78 & 5.09 & 5.71
& 1.73 & 1.46 & 0.36
& 0.15 & 0.15 & 0.02
& 2.74 & 3.42 & 3.30 \\

& J-Lens
& 4.98 & 7.33 & 4.66
& 0.13 & 3.00 & 0.38
& 1.85 & 2.17 & 0.26
& 0.00 & 0.00 & 0.00
& 1.74 & 3.13 & 1.32 \\

& J-Lens + ITGI
& 0.63 & 2.54 & 3.83
& 1.78 & 4.02 & 2.76
& 0.51 & 1.20 & 0.05
& 0.00 & 0.04 & 0.00
& 0.73 & 1.95 & 1.66 \\

\rowcolor{blue!7}
\cellcolor{white} & iLens
& \textbf{9.48} & \textbf{12.07} & \textbf{13.90}
& \textbf{6.72} & \textbf{6.57} & \textbf{8.79}
& \textbf{3.42} & \textbf{3.10} & \textbf{1.37}
& \textbf{5.09} & \textbf{7.96} & \textbf{0.32}
& \textbf{6.18} & \textbf{7.43} & \textbf{6.09} \\

\specialrule{0.55pt}{1.2pt}{1.2pt}


\multirow{4}{*}{Olmo3-7B~\cite{olmo2026olmo3}}
& Logit Lens + ITGI
& 10.47 & 7.69 & 5.30
& 3.77 & 2.47 & 0.81
& 1.29 & 0.00 & 0.48
& 0.00 & 0.00 & 0.00
& 3.88 & 2.54 & 1.65 \\

& J-Lens
& 14.10 & 12.65 & 7.03
& 2.57 & 1.93 & 1.06
& 2.23 & 2.23 & 1.17
& 0.00 & 0.00 & 0.00
& 4.73 & 4.20 & 2.32 \\

& J-Lens + ITGI
& 22.44 & 19.79 & 13.60
& 11.97 & 7.17 & 6.41
& 3.30 & 1.70 & 2.61
& 0.00 & 0.33 & 0.00
& 9.43 & 7.25 & 5.66 \\

\rowcolor{blue!7}
\cellcolor{white} & iLens
& \textbf{54.79} & \textbf{47.26} & \textbf{22.40}
& \textbf{43.00} & \textbf{24.53} & \textbf{46.40}
& \textbf{31.36} & \textbf{27.05} & \textbf{17.70}
& \textbf{16.30} & \textbf{17.67} & \textbf{1.21}
& \textbf{36.36} & \textbf{29.13} & \textbf{21.93} \\

\specialrule{0.55pt}{1.2pt}{1.2pt}


\multirow{4}{*}{GPT-oss-20B~\cite{openai2025gptoss120bgptoss20bmodel}}
& Logit Lens + ITGI
& 8.82 & 5.30 & 3.20
& 4.23 & 0.64 & 3.02
& 2.07 & 0.46 & 1.43
& 0.00 & 0.00 & 0.00
& 3.78 & 1.60 & 1.91 \\

& J-Lens
& 15.77 & 8.61 & 7.81
& -0.23 & -0.32 & 0.04
& 3.98 & 2.22 & 1.35
& 0.00 & 0.00 & 0.00
& 4.88 & 2.63 & 2.30 \\

& J-Lens + ITGI
& 22.11 & 14.33 & 11.70
& 6.91 & 3.05 & 9.38
& 9.40 & 7.95 & 4.17
& 0.09 & 0.18 & 0.01
& 9.63 & 6.38 & 6.31 \\

\rowcolor{blue!7}
\cellcolor{white} & iLens
& \textbf{40.84} & \textbf{30.72} & \textbf{21.20}
& \textbf{33.05} & \textbf{22.91} & \textbf{27.60}
& \textbf{30.27} & \textbf{23.92} & \textbf{30.40}
& \textbf{2.09} & \textbf{3.95} & \textbf{0.37}
& \textbf{26.56} & \textbf{20.38} & \textbf{19.89} \\

\bottomrule[0.9pt]

\end{tabular}%
}

\endgroup
\vspace{-2.0mm}
\end{table*}

\subsection{Intervention Experiments}

\paragraph{Tasks.}
We evaluate iLens on five intervention tasks spanning visual and textual modalities. For visual interventions, we use a CLEVR-based task~\cite{clevr} to examine whether modifying intermediate representations of object identities and colors influences generated descriptions. For textual interventions, we consider four tasks~\cite{gurnee2026verbalizable}: Verbal Report, Probe Swap, Flexible Generalization, and Verbal Introspection, which assess the effects of representation editing on verbal reports, internal reasoning, downstream generalization, and verbal identification of injected concepts, respectively. Detailed descriptions of the experimental settings and individual tasks are provided in Appendix~\ref{app:task_definitions}.

\vspace{-3.0mm}
\paragraph{Results.}

Table~\ref{tab:llm_intervention_results} presents the results of textual intervention experiments across four tasks and LLMs. We compare iLens against three baselines: \emph{Logit Lens + ITGI}, \emph{J-Lens}, and \emph{J-Lens + ITGI}. J-Lens refers to the intervention method introduced in the original J-Lens work~\cite{gurnee2026verbalizable}. The two ITGI-based baselines use Logit Lens and J-Lens for probing, respectively, while employing ITGI for intervention.

As demonstrated, iLens consistently achieves the best performance across different model architectures and intervention tasks, obtaining the highest average scores for all three evaluation metrics across all models. Specifically, on Qwen3.5-9B~\cite{qwen3.5}, iLens achieves average scores of 27.86, 25.43, and 19.16 for $\Delta$Pass@10, $\Delta$Pass@20, and $\Delta$Confidence, respectively, substantially outperforming J-Lens + ITGI, which achieves 10.22, 9.54, and 7.13.

\begin{figure}[t]
    \centering
    \includegraphics[width=\linewidth]{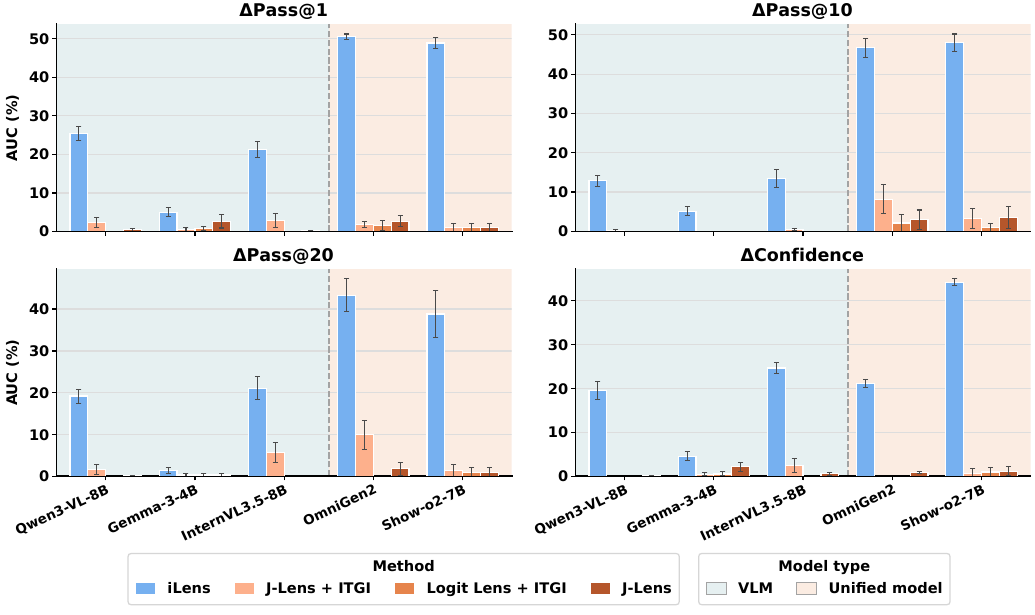}
    \caption{Intervention-effect AUC (\%) across five LVLMs and unified vision models for $\Delta$Pass@1, $\Delta$Pass@10, $\Delta$Pass@20, and $\Delta$Confidence. iLens consistently achieves the highest AUC, demonstrating its strong and great effectiveness in steering intermediate visual representations.}
    \label{fig:main-vlm-intervention}
    \vspace{-3.0mm}
\end{figure}

Furthermore, Figure~\ref{fig:main-vlm-intervention} presents the results of visual intervention experiments across three LVLMs (Qwen3-VL-8B, Gemma-3-4B, and InternVL3.5-8B) and two unified vision models (OmniGen2 and Show-o2-7B). We compare iLens against the same three baselines using four evaluation metrics: $\Delta$Pass@1, $\Delta$Pass@10, $\Delta$Pass@20, and $\Delta$Confidence. In particular, since each CLEVR experiment's scene contains a single object, we additionally report $\Delta$Pass@1 to assess whether the injected concept becomes the top-ranked prediction while the original concept is suppressed.

iLens substantially outperforms the baselines across both model families, achieving the highest AUC in nearly all model-metric comparisons. For example, on OmniGen2, iLens achieves a $\Delta$Pass@1 AUC of approximately 50\%, compared with less than 10\% for the strongest baseline. This advantage is also evident in $\Delta$Pass@10, $\Delta$Pass@20, and $\Delta$Confidence, for which iLens likewise achieves substantially higher AUCs. Moreover, among LVLMs, iLens demonstrates particularly strong improvements on Qwen3-VL-8B and InternVL3.5-8B, with gains also observed on Gemma-3-4B. Overall, these results demonstrate the effectiveness of iLens for visual concept intervention across diverse model architectures.

Taken together, these experiments demonstrate that iLens steers model predictions more effectively toward injected concepts and away from the original source concepts across textual interventions in LLMs and visual interventions in both LVLMs and unified vision models. These findings highlight iLens as a more effective approach to targeted concept editing across modalities. Appendix~\ref{app:subsec:intervention_results} provides additional results for both textual and visual interventions.

\section{Related Work}
\label{sec:related_work}

\paragraph{Mechanistic interpretability.}
Mechanistic interpretability seeks to explain model behavior by reverse-engineering the internal representations and computations that produce it~\citep{elhage2021mathematical}.
Intervention-based analyses help investigate how particular components and information pathways contribute to predictions.
Causal mediation analysis examines the effects transmitted through internal neurons and attention heads~\citep{vig2020causalmediation}, while activation interventions such as causal tracing identify states that influence factual predictions~\citep{meng2022locating}.
Attention knockout provides a complementary approach by blocking selected attention connections to investigate information flow during factual recall~\citep{geva2023dissecting}.
Such interventions also support the discovery and validation of behavior-specific circuits, through both manual analysis~\citep{wang2023interpretability} and automated search~\citep{conmy2023automated}.
Developing efficient tools for decoding and editing concepts in selected residual embeddings can further facilitate mechanistic interpretability studies.

\vspace{-2mm}
\paragraph{Concept decoding.}
To decode intermediate concepts, vocabulary readouts map hidden states to token scores.
Logit Lens applies the model's final normalization and output projection to the selected state~\citep{nostalgebraist2020logitlens}, while our Embedding Lens baseline ranks tokens by cosine similarity with their input embeddings~\citep{embedding_lens}.
These fixed interfaces do not adapt to layer-dependent distribution shifts.
Tuned Lens trains layer-specific affine maps to approximate the final output distribution~\citep{belrose2023tunedlens}, while Future Lens anticipates subsequent tokens using learned projections or patched continuations with fixed or trained prompts~\citep{pal2023futurelens}.
Their objectives focus on next-token or subsequent-token prediction and do not explicitly require recovering intermediate concepts.
Jacobian-based methods incorporate downstream nonlinear computation through first-order approximations: relation decoders construct relation-specific maps~\citep{hernandez2024linearity}, while J-Lens averages derivatives over reference prompts and token positions to obtain one linear map per layer~\citep{gurnee2026verbalizable}.
J-Lens provides efficient readouts and intervention directions once constructed, but requires derivative computation and depends on the reference corpus and averaging procedure.

Complementary approaches interpret intermediate representations through learned feature decompositions or natural-language descriptions.
Sparse autoencoders learn sparse feature dictionaries from model activations without prespecified concept labels, but require training and subsequent feature interpretation~\citep{cunningham2024sparse,templeton2024scaling}.
Patchscopes and SelfIE can use pretrained language generation to describe hidden states inserted into a target prompt via activation patching~\citep{ghandeharioun2024patchscopes,chen2024selfie}.
These descriptions depend on the prompt, insertion layer and position, and previously generated tokens, and may include additional inferences made during decoding.
Unrestricted attention also allows information from other target-context states to enter the readout, complicating attribution of the explanation solely to the selected representation.

\vspace{-2mm}
\paragraph{Concept editing.}
To examine how concepts influence model behavior, interventions modify intermediate activations without changing model weights.
CAA derives steering vectors from positive--negative activation differences, requiring suitable example pairs and a chosen intervention strength~\citep{rimsky2024steering}.
J-Lens supports concept injection, suppression, and replacement through vocabulary-associated directions~\citep{gurnee2026verbalizable}, while SAE interventions modify learned feature activations~\citep{templeton2024scaling}.
In vision-language models, ProjectAway uses linear orthogonalization of image features against selected object representations to suppress object-related content and reduce object hallucinations~\citep{jiang2025projectaway}.
Gradient-guided methods such as PPLM update internal activations to improve attribute scores, but require a specified attribute model or word set and additional inference-time optimization~\citep{dathathri2020pplm}.
Evaluating these interventions through changes in downstream model predictions complements measurements of the targeted internal readout~\citep{bhalla2024unifying}.

\vspace{-2mm}
\paragraph{Positioning iLens.}
iLens decodes intermediate concepts using the model's own operations, while applying attention knockout to isolate the effect of the selected residual embedding on the final readout.
Compared with linear readouts, iLens retains downstream nonlinear computation; it also avoids the data-dependent training or readout construction used by Tuned Lens, J-Lens, and sparse autoencoders.
Compared with prompt-based nonlinear generation, it requires no additional decoding context and reduces the influence of context-dependent confounders by blocking parallel contributions from other token representations to the final readout.
For concept editing, iLens uses gradients of concept log-odds through an isolated nonlinear path to effectively suppress or inject concepts within targeted residual embeddings.
Finally, while much of the prior work focuses on LLMs, our evaluation also spans diverse vision-language and unified vision models to study visual concept decoding and editing.

\section{Discussion and Limitations}

In this work, we proposed iLens, a nonlinear readout that relies solely on the model and its native operations, without additional context, data, or training.
It uses attention knockout to isolate the effect of the selected representation during readout.
Evaluations across 18 LLMs and 14 vision-capable models on five visual and five textual decoding tasks demonstrate its effectiveness, particularly for decoding visual concepts in early layers.
The same isolation principle also supports concept editing through an update derived from a first-order Taylor expansion of the isolated function.
Our experiments show a notable performance advantage in both concept suppression and injection.

Together, these results support iLens as a practical, training-free mechanistic interpretability tool for studying and editing internal concepts across language and vision-capable models. Compared with simple linear readouts, iLens has a relatively higher computational cost because it uses the models' own native nonlinear operations across the remaining transformer blocks. While this overhead is generally tolerable in interpretability studies, developing faster nonlinear probes remains a direction for future research.

\clearpage

{
    \small
    \bibliographystyle{ieeenat_fullname}
    \bibliography{main}
}

\newpage
\clearpage

\setlength{\textwidth}{6.875in}
\setlength{\oddsidemargin}{-0.1875in}
\setlength{\evensidemargin}{-0.1875in}

\appendix

\clearpage
\onecolumn

\startcontents[supplement]

\begingroup
\large

\printcontents[supplement]{l}{1}{
    \section*{Table of Contents}
    \setcounter{tocdepth}{3}
}

\endgroup

\clearpage
\twocolumn

\section{Experimental Setup and Implementation Details}
\label{app:experimental_details}

All experiments with iLens are conducted using the original pretrained model weights, which remain frozen throughout both decoding and intervention. Unlike approaches that learn additional transformations, such as Tuned Lens~\cite{belrose2023tunedlens}, iLens requires no model fine-tuning, separately trained probes, or learned readout or transport mappings. For ITGI, gradients are computed with respect to the selected residual representations to construct activation-level updates, without modifying any model parameters. Thus, iLens introduces no additional training stage or requirement for additional training data.

\subsection{Task Definitions}
\label{app:task_definitions}

\subsubsection{Visual Concept Decoding Tasks}
\label{app:visual_decoding_tasks}

We evaluate iLens on five visual probing tasks to assess its ability to recover object-centric visual concepts from intermediate image-token representations. These tasks cover a range of visual concepts, from basic object properties (shape and color) to object identities, attributes, and relationships. For each example, we provide the model with a general instruction to describe and analyze the scene, followed by the input image, without explicitly asking about a particular object or visual concept. We then probe the image tokens corresponding to the annotated object or, for relational annotations, the tokens covering both participating objects. Specifically, we consider the following tasks:

\begin{itemize}
    \item \textbf{CLEVR Shape.}
    We use synthetic scenes of simple 3D objects from CLEVR~\cite{clevr} to evaluate whether iLens can recover object shapes from their corresponding image tokens. For example, given a red cube, the target concept is \emph{cube}.

    \item \textbf{CLEVR Color.}
    Using the same CLEVR scenes, we probe for the color of each annotated object. For example, given the same red cube, the target concept is \emph{red}.

    \item \textbf{COCO.}
    We use natural images from MS-COCO~\cite{coco} to evaluate object-level semantic recognition. Specifically, we probe for the category of each annotated object. For example, for an image region containing a cat, the target is \emph{cat}, including its accepted synonyms.

    \item \textbf{Visual Genome Attribute.}
    We use the object-attribute annotations from Visual Genome~\cite{visual_genome} to evaluate whether iLens can recover visual attributes associated with individual objects. These attributes describe properties such as appearance and physical characteristics. For example, an annotated tree may have the attributes \emph{green} and \emph{tall}, which serve as target concepts.

    \item \textbf{Visual Genome Non-Spatial.}
    We use non-spatial relationship annotations from Visual Genome to evaluate whether iLens can recover interactions between objects. In contrast to the previous tasks, which focus on individual objects, this task probes the image tokens covering both objects involved in an annotated relation. We use the relation's verb as the target concept. For example, for the relation \emph{man holding pole}, the target is \emph{hold}.
\end{itemize}

\subsubsection{Textual Concept Decoding Tasks}
\label{app:textual_decoding_tasks}

We evaluate iLens on five text-probing tasks~\cite{gurnee2026verbalizable}, covering a diverse range of linguistic, factual, and computational capabilities. These tasks examine whether intermediate representations encode information that may not be explicitly expressed in the input or reflected in the model's final output. Specifically, we consider the following tasks:

\begin{itemize}
    \item \textbf{Typo.}
    This task investigates whether intermediate representations recover the intended correct word from a misspelled input. We provide a sentence ending with a common misspelling and probe at the final fragment of the misspelled token. For example, given \emph{langauge}, the target concept is \emph{language}.

    \item \textbf{Multilingual.}
    This task investigates whether non-English inputs activate English-aligned concepts and linguistic representations. For example, given the Spanish prompt ``\emph{Lo opuesto de grande es},'' we probe for representations of the input words, their semantic relationship, and their English equivalents, as well as the corresponding output concepts.

    \item \textbf{Multihop.}
    This task examines whether intermediate representations capture implicit bridge entities required to answer multihop factual questions. For example, answering ``the color of the planet fourth from the Sun is'' requires retrieving the intermediate entity \emph{Mars} before arriving at the final answer \emph{red}.

    \item \textbf{Order of Operations.}
    This task evaluates whether intermediate representations encode partial arithmetic results and pending operations during mathematical reasoning. For example, when evaluating $(2+3)\times4$, the target concepts include the intermediate result \emph{5} and the pending operation \emph{multiplication}.

    \item \textbf{Dual Task.}
    This task investigates whether intermediate representations can encode and manipulate concepts under explicit instructions, independently of the content of the generated output. For example, when instructed to compute $3^2-2$ while writing an unrelated sentence, we probe for intermediate concepts associated with the calculation, including \emph{calculation}, \emph{nine}, \emph{subtraction}, \emph{seven}, and \emph{equality}.
\end{itemize}

\subsubsection{Visual and Textual Intervention Tasks}
\label{app:intervention_tasks}

We evaluate iLens on five intervention tasks spanning two modalities: visual and textual interventions.

\paragraph{Visual Interventions.}

For LVLMs, we use the CLEVR dataset~\cite{clevr}, selecting scenes containing a single 3D object with a specific shape and color. This experiment investigates whether interventions on intermediate visual representations influence the model's generated descriptions. For each image, we ask the model to describe the object in the scene, including its color. We then intervene on the intermediate representations associated with both the object's identity and its color. By examining the resulting descriptions, we evaluate whether modifying these representations leads to corresponding changes in the model's output.

\paragraph{Textual Interventions.}

For LLMs, we evaluate iLens on four text-based intervention experiments from prior work~\cite{gurnee2026verbalizable}. These experiments investigate whether modifying intermediate representations can influence the model's verbal reports, reasoning processes, and subsequent predictions. Specifically, we consider the following experiments:

\begin{itemize}
    \item \textbf{Verbal Report.}
    This experiment examines whether intervening on an internally selected concept changes the model's subsequent verbal report. The model is first instructed to think of an item from a given category. Before it reports its choice, we intervene on the corresponding intermediate representation, replacing the original item with another category member. For example, replacing \emph{soccer} with \emph{rugby} allows us to test whether the model reports the intervened sport.

    \item \textbf{Probe Swap (Internal Reasoning).}
    This experiment evaluates whether intervening on an intermediate reasoning step changes the model's subsequent answer. Given a two-hop factual question, we replace the representation of the implicit bridge entity with an alternative entity and examine whether the final answer changes accordingly. This tests whether the intervened representation is functionally involved in the model's reasoning process.

    \item \textbf{Flexible Generalization.}
    This experiment investigates whether an intervened concept representation can be reused across different downstream reasoning tasks. We replace a concept with another and evaluate whether the model consistently uses the modified representation when answering different questions. For example, replacing \emph{France} with \emph{China} allows us to test whether subsequent answers about the capital, language, and continent reflect the substituted entity.

    \item \textbf{Verbal Introspection.}
    This experiment examines whether the model can identify concepts injected directly into its intermediate activations. We inject a target concept into the model's representations and subsequently ask the model to identify any injected thought. We evaluate whether its verbal response correctly identifies the injected concept.
\end{itemize}

\subsection{Implementation Details}
\label{app:implementation_details}
\subsubsection{Decoding Details}

\paragraph{Target tokens.} All methods score a single next-token distribution, so each target concept must be represented by tokens in the model's vocabulary. For every target word, we consider six surface forms (lowercase, title case, and uppercase, each with and without a leading space) and keep only those that the model's tokenizer encodes as exactly one token. Multi-token forms are discarded rather than approximated by their first sub-token, since a shared prefix such as \texttt{Re} would match many unrelated words. Synonyms and inflections of a concept (e.g., \emph{refrigerator}/\emph{fridge}, \emph{5}/\emph{five}) are expanded in the same way, and all surviving token IDs are pooled into the target set. 

\paragraph{Metrics.} Pass@$k$ uses the rank of the best-ranked target token, and Confidence sums the softmax probabilities of all target tokens. For models with a final logit soft-cap (Gemma Families~\cite{gemma3, gemmateam2026gemma4}), ranks are computed from the logits before the cap to avoid saturation ties. When a textual task defines several candidate probe positions for one example, we take the maximum of each metric over those positions. For visual tasks, an object's region is mapped to visual tokens, and every token that overlaps the mask is included. Per-token scores are first aggregated within an annotated unit, then averaged over units within an image, and finally averaged over images. All AUCs use the trapezoidal rule over layers $1,\ldots,L-1$, normalized by the span. The final layer is excluded because J-Lens has no translator there, and using the same span for every method keeps the areas comparable. FED is computed over the same span.

\paragraph{Visual prompts.} Images are $336\times336$ pixels and are passed through each model's native processor and chat template, together with a task-level instruction that does not name any particular object: ``analysis and describe the objects of the scene precisely'' (COCO), ``\ldots the attributes and properties of the objects \ldots'' (VG Attribute), ``\ldots the actions and interactions of the objects \ldots'' (VG Non-Spatial), and ``Describe the shape and color of the objects in the scene.'' (CLEVR).

\paragraph{Baselines.} Logit Lens applies $\mathrm{Norm}_f$ and $U$ directly to $h_{\ell,t}$. Embedding Lens ranks vocabulary items by the cosine similarity between $h_{\ell,t}$ and each row of the input embedding matrix, without final normalization; For tied-embedding models, applying $\mathrm{Norm}_f$ first would align the hidden-state preprocessing with Logit Lens, but cosine similarity would still normalize each vocabulary embedding by its norm. J-Lens applies a per-layer translator $J_\ell\in\mathbb{R}^{d\times d}$ and decodes $U(\mathrm{Norm}_f(J_\ell h_{\ell,t}))$.
For Patchscopes, we use its default prompt.

\paragraph{iLens implementation.} Because each isolated attention branch has a single key, its softmax weight is exactly one. The branch therefore reduces to the value and output projections applied to the selected state, and queries, keys, and rotary embeddings do not affect the result. Linear-attention layers (Gated DeltaNet in Qwen series~\cite{qwen3.5, qwen3.6-27b}) are evaluated on a length-one sequence with zero convolution and recurrent state, which realizes the same isolation. In architectures that normalize each branch output before the residual addition (Gemma models~\cite{gemma3, gemmateam2026gemma4}), $w_k$ multiplies the branch after this post-normalization, since RMSNorm is scale-invariant, and model-specific residual scalars are retained. We verify each port by checking that, with $w_k=1$, the isolated step reproduces the model's own decoder block applied to a length-one sequence at every layer. The attenuation schedule uses $\tau=0.8$ and reaches $w_L=0.01$ at the final block.

\subsubsection{Intervention Details}
\paragraph{Trials.} Trials are built from the released datasets of \citet{gurnee2026verbalizable}. The source and injected concepts must each be a single token and must share casing and leading-space form, so that a swap does not also change the surface form. For Probe Swap, we keep only items for which the unedited model's greedy next token is already the correct answer. Items are selected round-robin across relation categories, with up to 50 trials per task. Since the Verbal Introspection has no source concept, only the injection step is applied, and the $\Delta M_{src}$, which is the source term of $\Delta M$, is zero.

\paragraph{Update rule.} We use $q^*=0.01$ for the source and $q^*=0.99$ for the injected concept, $\epsilon=10^{-4}$. Logit Lens + ITGI applies the same update, with the gradient taken through $U(\mathrm{Norm}_f(h_{\ell,t}))$ at the edited position. J-Lens + ITGI takes it through $U(\mathrm{Norm}_f(J_\ell h_{\ell,t}))$. J-Lens uses the published coordinate swap along $J_\ell^\top u$ at strength one.

\paragraph{Visual intervention.} We use a fixed set of 50 single-object CLEVR scenes, restricted for each model to the scenes it answers correctly. The model is asked ``What is the color and the object in the scene?'' and must answer in a fixed \texttt{[Color] [Shape]} format. We edit the visual tokens covering the object and their 4-connected neighbours toward a randomly drawn (seeded) different color and shape. Color is suppressed and injected first, then shape. The color is read from the first answer step. The shape is read from the second step, with the color teacher-forced; for the injected pair, the forced color is the injected one. Pass@$k$ requires both words to appear in the top $k$, and Confidence is the product of their probabilities ($P(Shape, Color) = P(Color) \times P(Shape | Color)$).

\subsection{Hardware Configurations.}

All experiments are conducted using  NVIDIA Quadro RTX and L40S GPUs, with CPU resources are used for data preprocessing, metric computation, and result aggregation.

\section{Ablation Study and Analysis}
\label{sec:ablation}

We conduct ablation studies along five dimensions to investigate the decoding and intervention capabilities of iLens: (i) the sensitivity of the attenuation threshold $\tau$ across different tasks and model architectures, (ii) the role of late-layer attenuation in target-concept recovery and next-token prediction, (iii) the individual and joint contributions of attention and MLP residual updates, including the choice of branches to attenuate, (iv) the effect of different attenuation schedules, comparing linear, cosine, and exponential decay, and (v) the effect of the position at which the ITGI objective is evaluated, comparing the current (edited) token position with the final prompt position.

Overall, our results show that late-layer attenuation generally improves intermediate concept recovery, although its effectiveness depends on the task and model architecture. While no single attenuation threshold or schedule consistently outperforms the alternatives across all tasks, the default configuration ($\tau=0.8$ with linear attenuation) provides a strong and balanced choice across modalities. Our branch ablations further demonstrate the importance of retaining both attention and MLP transformations while attenuating their contributions, rather than suppressing them entirely. Finally, the intervention ablation shows that computing the ITGI objective at the final prompt position consistently yields stronger concept-editing effects than computing it at the edited token position. Together, these findings provide empirical support for the principal design choices underlying iLens and its intervention framework.

\begin{figure}[t]
    \centering
    \includegraphics[width=\linewidth]{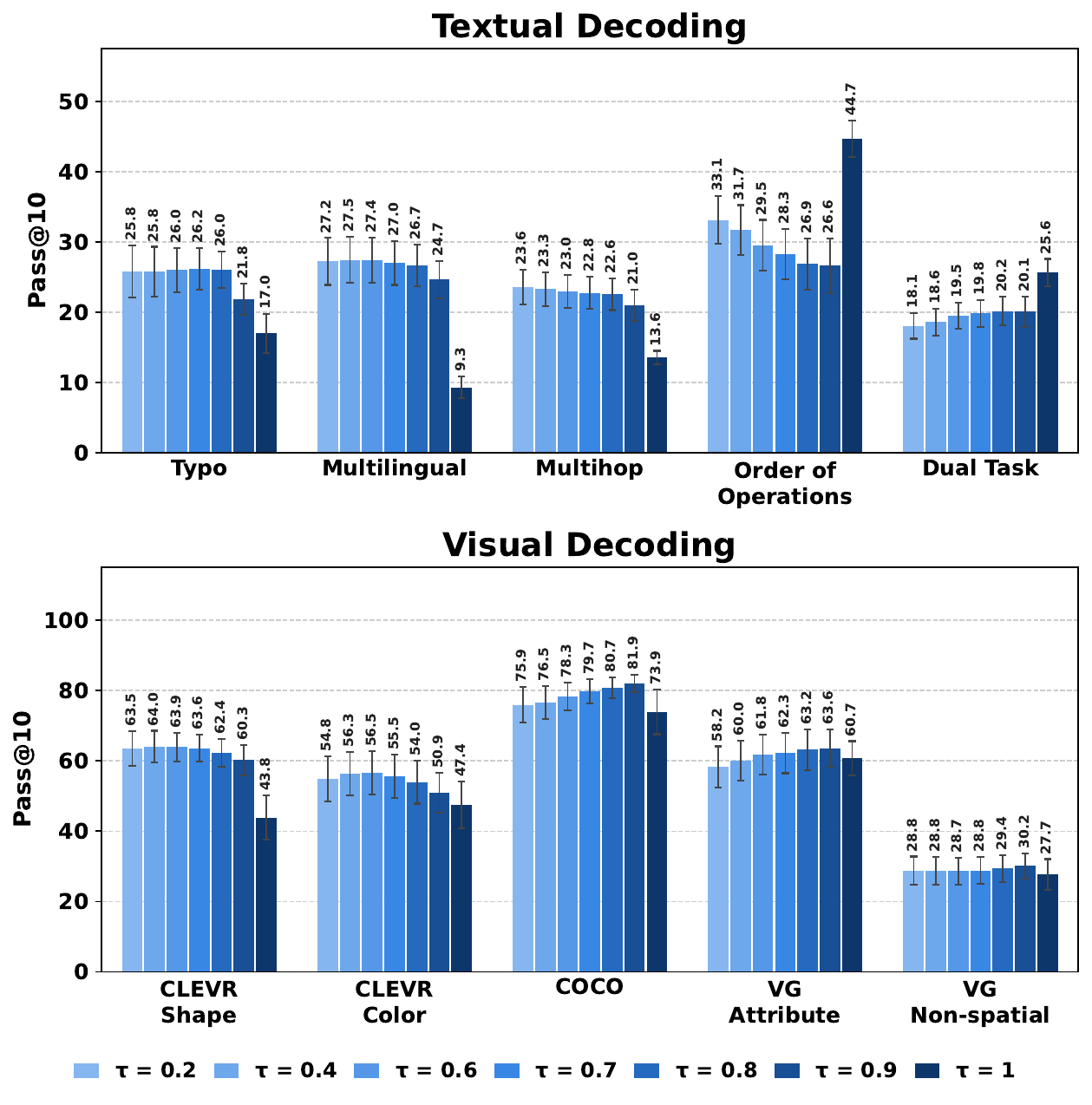}
        \caption{Ablation of the late-layer attenuation threshold $\tau$ on Pass@10 across 18 LLMs and 14 LVLMs. For each task, bars report Pass@10 averaged across models. The results reveal task-dependent performance trends, with no single value of $\tau$ consistently outperforming the others. We therefore fix $\tau = 0.8$ across all experiments.}
    \label{fig:tau-ablation}
    \vspace{-2.0mm}
\end{figure}

\paragraph{Sensitivity to the attenuation threshold $\tau$.}
We first examine the sensitivity of iLens to the attenuation threshold $\tau$. Figure~\ref{fig:tau-ablation} compares Pass@10 across different values of $\tau$ for all LLM and LVLM decoding tasks, averaged over models. Notably, $\tau=1$, which disables late-layer attenuation, often results in weaker performance, particularly across several textual and visual decoding tasks. This observation is consistent with our motivation for attenuating late-layer updates: as later transformer blocks increasingly shape representations toward next-token prediction~\cite{gurnee2026verbalizable}, retaining their full contribution may hinder the recovery of intermediate concepts. However, the relative performance of different $\tau$ values varies across tasks, with no single value consistently outperforming the others.

\begin{figure*}[t]
    \centering
    \includegraphics[width=0.99\linewidth]{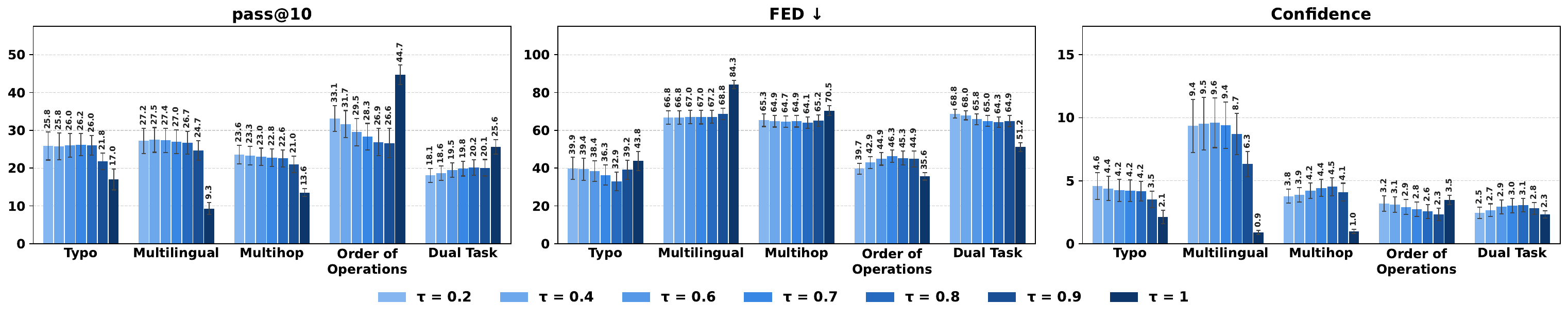}
    \caption{Ablation of the attenuation threshold $\tau$ across five textual concept decoding tasks, averaged over 18 LLMs. We report Pass@10, FED, and Confidence for different values of $\tau$.}
    \label{fig:appendix-llm-ablation}
    \vspace{-2.0mm}
\end{figure*}

\begin{figure*}[t]
    \centering
    \includegraphics[width=0.99\linewidth]{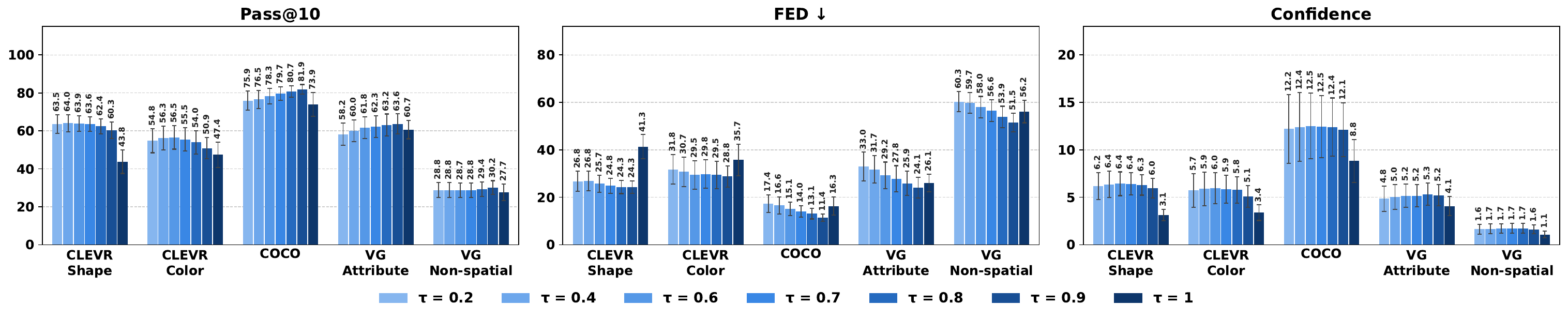}
    \caption{Ablation of the attenuation threshold $\tau$ across five visual concept decoding tasks, averaged over 14 LVLMs. We report Pass@10, FED, and Confidence for different values of $\tau$.}
    \label{fig:appendix-vlm-ablation}
    \vspace{-2.0mm}
\end{figure*}

\begin{figure*}[t]
    \centering
    \includegraphics[width=0.99\linewidth]{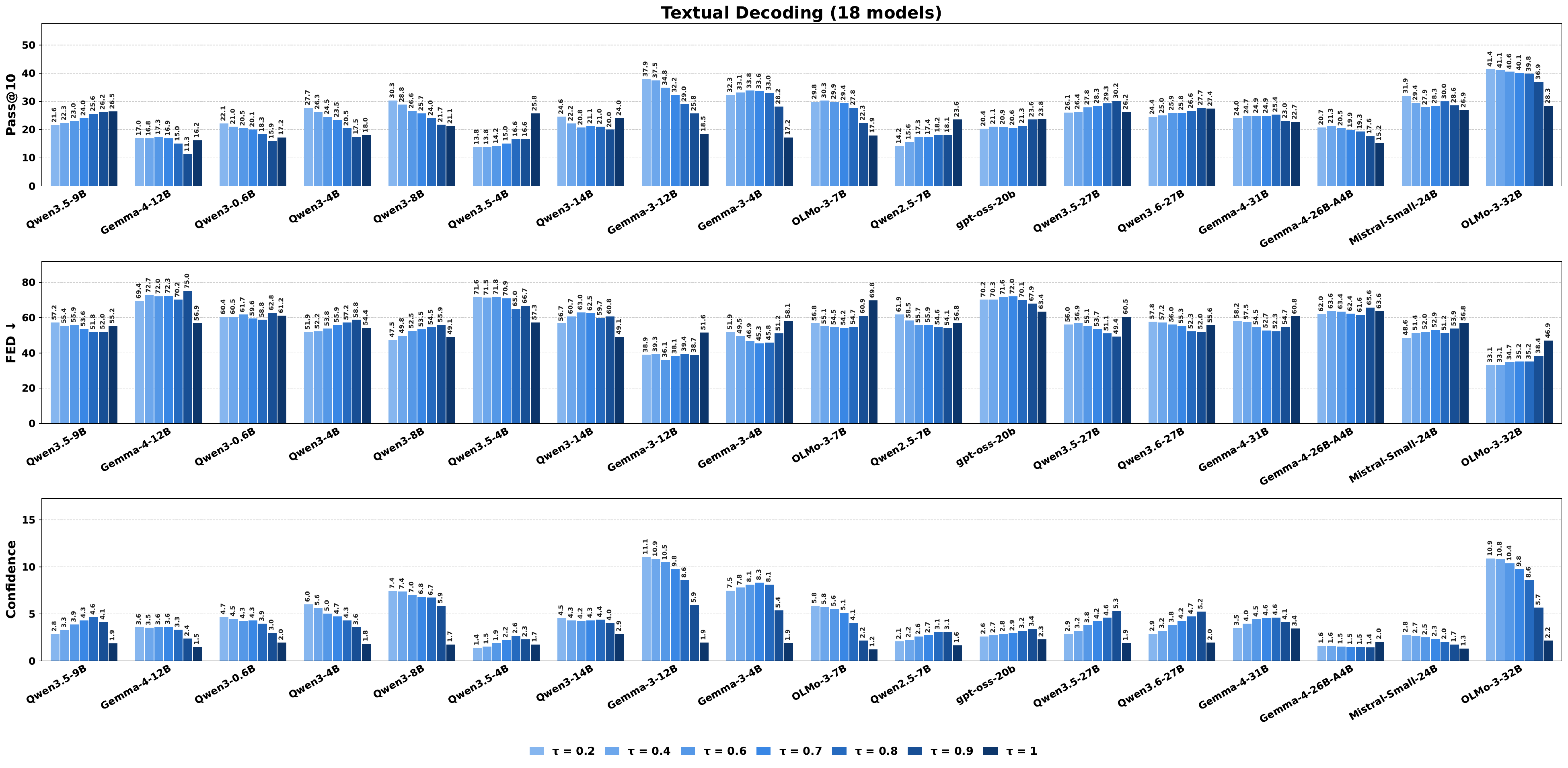}
    \caption{Per-model ablation of the attenuation threshold $\tau$ across various LLMs. For each model, Pass@10, FED, and Confidence are averaged across the corresponding evaluation tasks (five textual decoding tasks). Results are shown separately for 18 LLMs.}
    \label{fig:appendix-per-model-llm-ablation}
    \vspace{-4.0mm}
\end{figure*}

\begin{figure*}[t]
    \centering
    \includegraphics[width=0.99\linewidth]{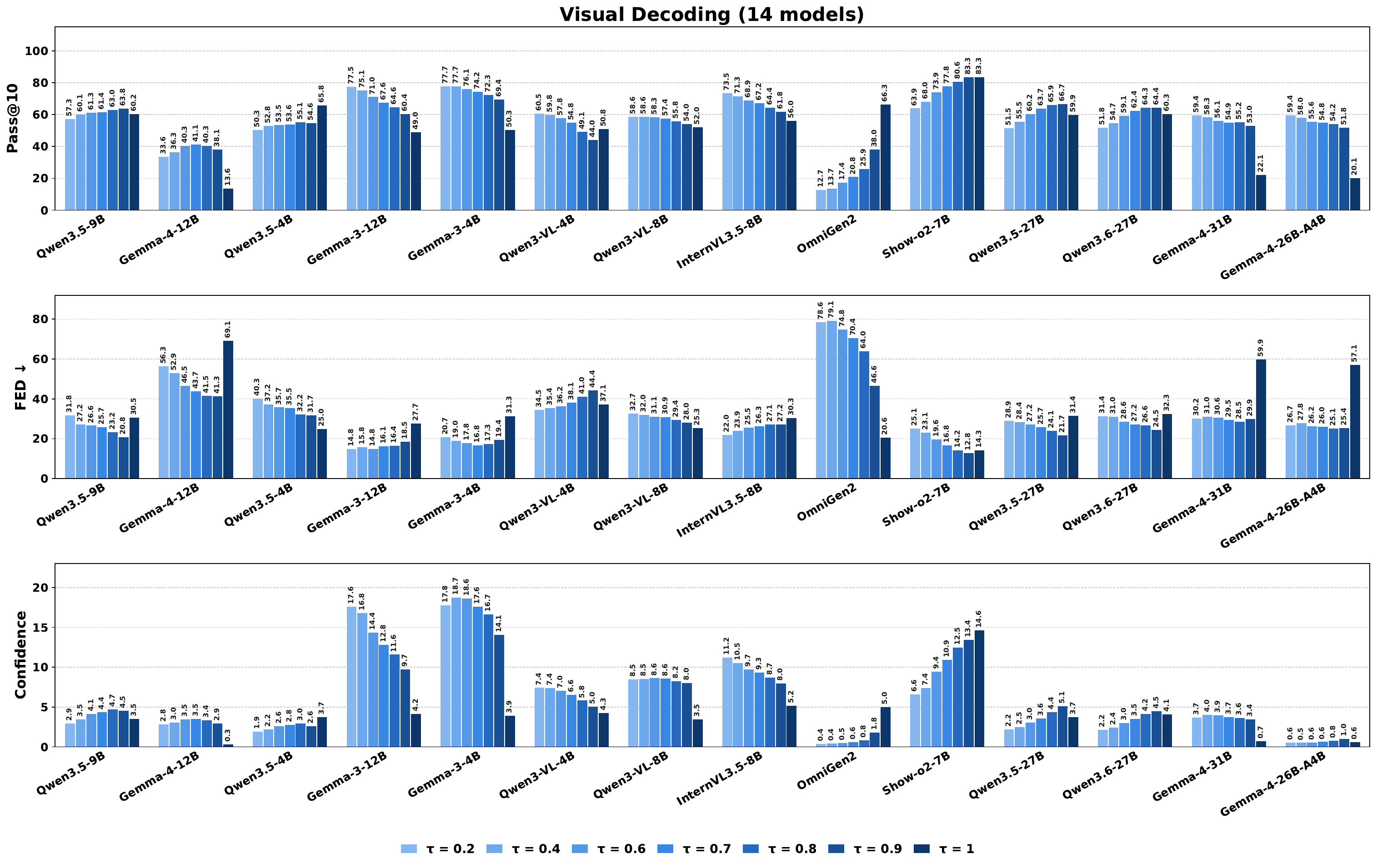}
    \caption{Per-model ablation of the attenuation threshold $\tau$ across various LVLMs. For each model, Pass@10, FED, and Confidence are averaged across the corresponding evaluation tasks (five visual decoding tasks). Results are shown separately for 14 LVLMS.}
    \label{fig:appendix-per-model-vlm-ablation}
    \vspace{-4.0mm}
\end{figure*}

We further illustrate this variation across modalities and individual models. Figures~\ref{fig:appendix-llm-ablation} and~\ref{fig:appendix-vlm-ablation} provide a comprehensive comparison of Pass@10, FED, and Confidence across the five textual and five visual decoding tasks, respectively, averaged over all models within each modality. Figures~\ref{fig:appendix-per-model-llm-ablation}
and~\ref{fig:appendix-per-model-vlm-ablation}
present the corresponding results for individual models,
with metrics averaged across tasks separately for each model. These results further demonstrate that the effectiveness of $\tau$ depends on both the decoding task and the underlying model architecture. Therefore, to avoid task- or model-specific hyperparameter tuning, we adopt $\tau=0.8$ as a shared default across all experiments and report our results using this setting.

\begin{figure}[t]
    \centering
    \includegraphics[width=\linewidth]{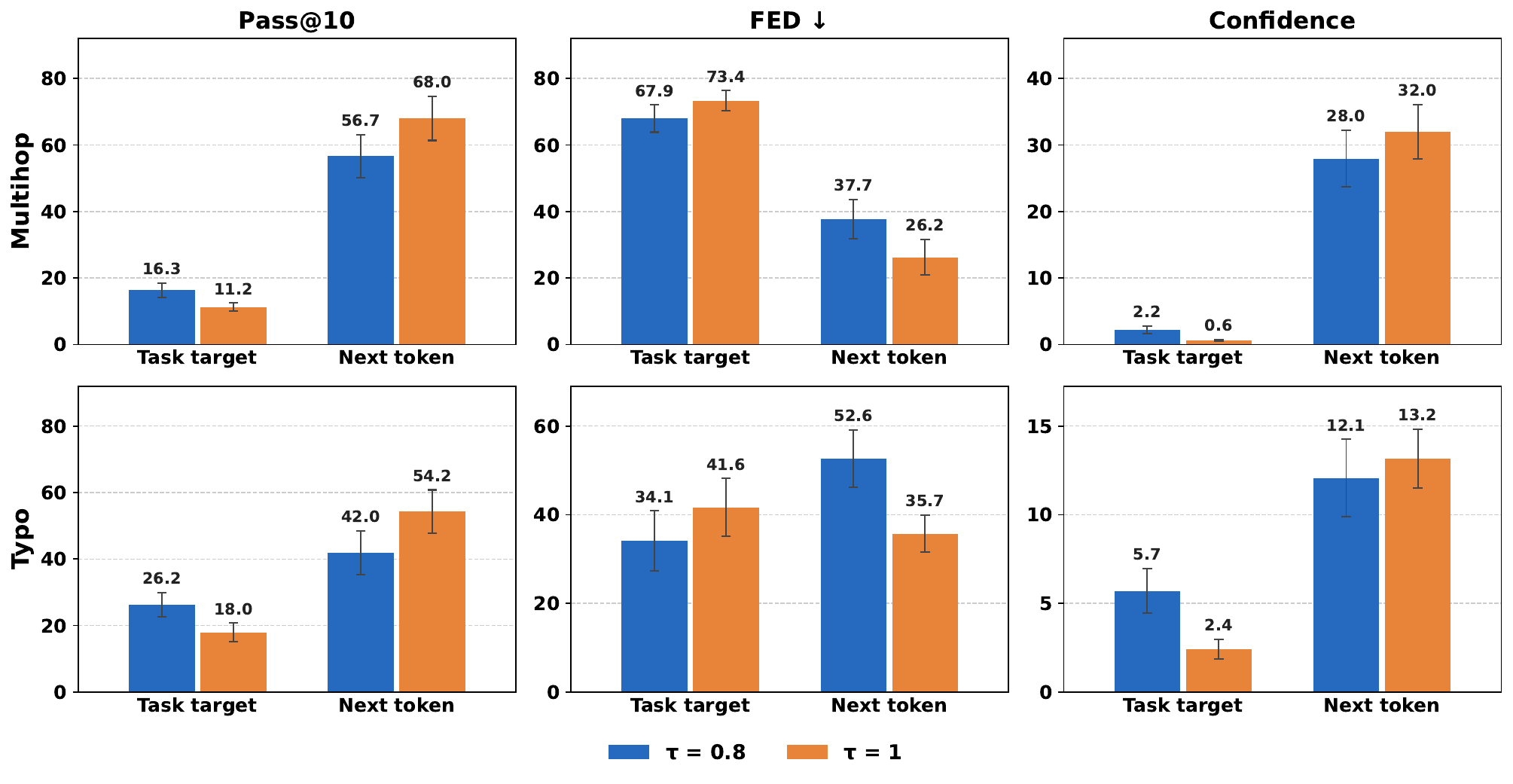}
        \caption{Comparison of $\tau=0.8$ and $\tau=1$ for target-concept recovery and next-token prediction on the Multihop and Typo tasks, averaged across 18 LLMs. We report Pass@10, First Emergence Depth (FED; lower is better), and Confidence. Late-layer attenuation improves target-concept recovery, whereas retaining full-strength downstream updates benefits next-token prediction.}
    \label{fig:next-token-ablation}
    \vspace{-2.0mm}
\end{figure}

\paragraph{Role of late-layer attenuation on different decoding objectives.}
We further compare $\tau=0.8$ and $\tau=1$ on the Typo and Multihop tasks under two decoding objectives: target-concept recovery and next-token prediction. Figure~\ref{fig:next-token-ablation} reports the results averaged across 18 LLMs. We observe a clear distinction between the two objectives: attenuating late-layer updates ($\tau=0.8$) consistently improves target-concept recovery, whereas retaining the full downstream computation ($\tau=1$) yields better next-token prediction~\cite{gurnee2026verbalizable}. This finding supports our motivation that late transformer layers increasingly specialize in next-token prediction. Consequently, reducing their contribution helps preserve the recoverability of intermediate concepts that might otherwise be shadowed by the model's next-token prediction objective.

\begin{figure}[t]
    \centering
    \includegraphics[width=\linewidth]{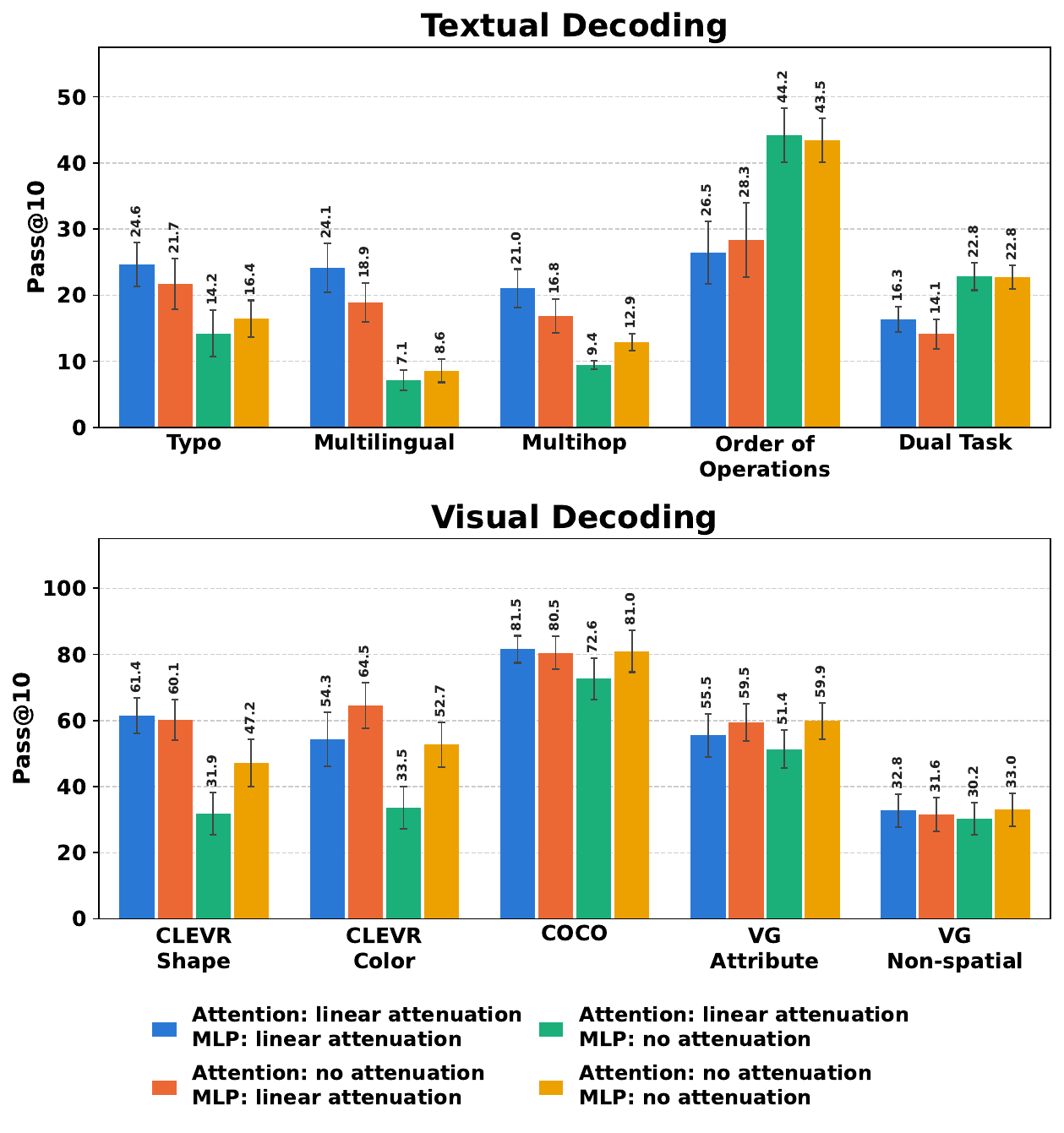}
    \caption{Comparison of four late-layer attenuation configurations: attenuating both attention and MLP branches, only the attention branch, only the MLP branch, or neither (attenuating all of them at $\tau=0.8$). A scaling factor of $1$ indicates that the corresponding branch is not attenuated. Pass@10 is averaged over 18 LLMs for textual decoding and 14 LVLMs for visual decoding. Attenuating both branches provides the best overall balance across textual and visual concept decoding tasks.}
    \label{fig:ramp-target-ablation}
    \vspace{-2.0mm}
\end{figure}

\paragraph{Choice of branches to attenuate.}
We investigate whether late-layer attenuation should be applied to the attention branch, the MLP branch, or both. Specifically, we compare four configurations: attenuating both branches using the default schedule $w_k$, attenuating only the MLP branch while retaining full-strength attention updates, attenuating only the attention branch while retaining full-strength MLP updates, and disabling attenuation for both branches. Figure~\ref{fig:ramp-target-ablation} shows that attenuating both branches achieves the highest average Pass@10 on textual decoding (22.5), while attenuating only the MLP achieves slightly better performance on visual decoding (59.2 vs.\ 57.1). In contrast, attenuating attention alone substantially degrades performance, particularly on CLEVR. The benefits of MLP attenuation are also task-dependent: it improves Typo, Multilingual, and Multihop, whereas Order of Operations and Dual Task benefit from retaining full-strength MLP updates. Overall, applying the same attenuation schedule to both branches provides the best balance across tasks and modalities without introducing branch-specific hyperparameters.

\begin{figure}[t]
    \centering
    \includegraphics[width=\linewidth]{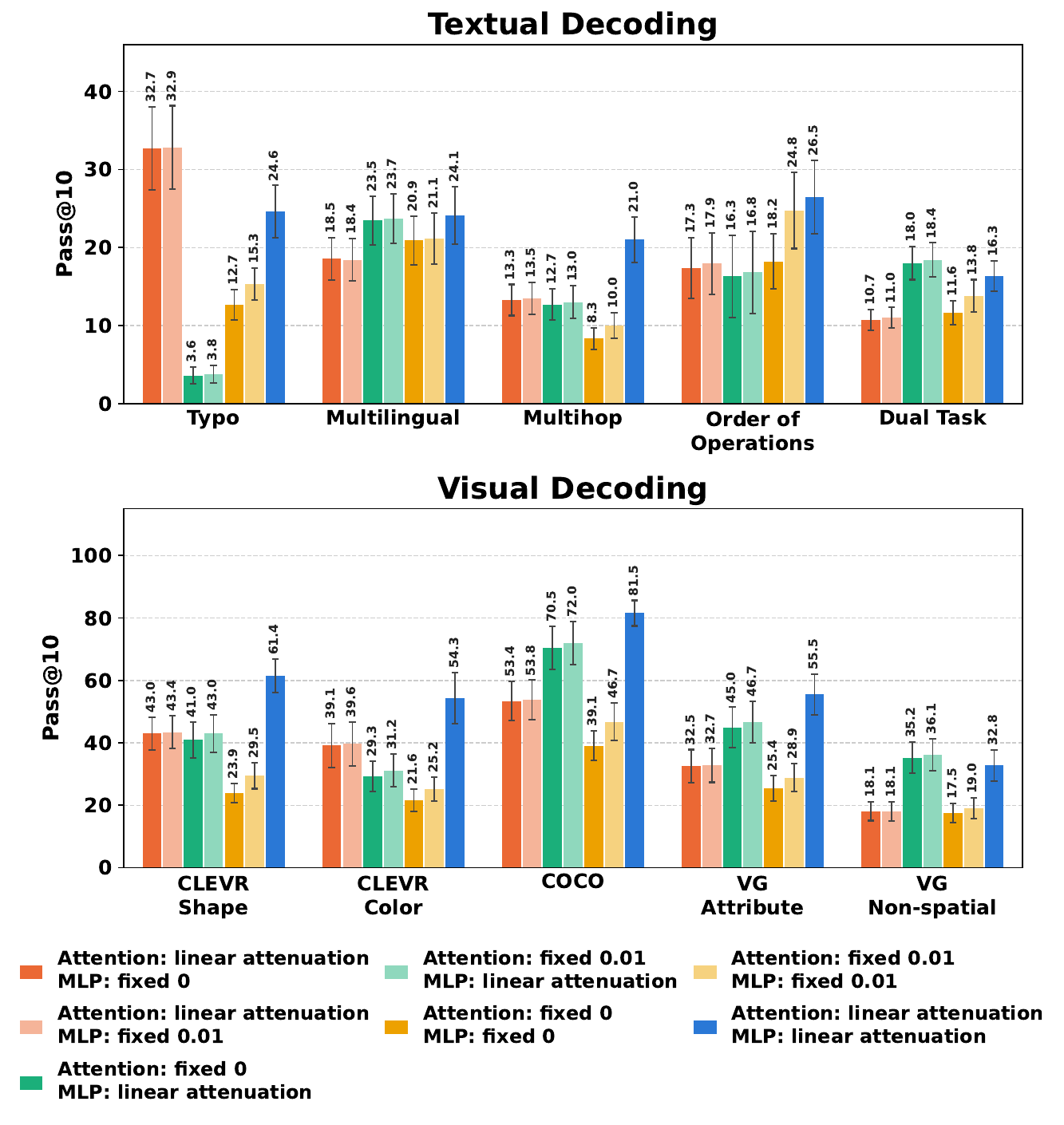}
    \caption{Ablation of the individual and joint contributions of attention and MLP residual updates in the isolated suffix. Each branch either follows the default attenuation schedule ($\tau=0.8$) or uses a fixed scaling factor of $0$ or $0.01$ across all isolated blocks. Suppressing both branches reduces the isolated suffix to an identity mapping, recovering the Logit Lens readout; a scaling factor of $0.01$ provides a close approximation. Pass@10 is averaged over 18 LLMs for textual decoding and 14 LVLMs for visual decoding.}
    \label{fig:branch-ablation}
    \vspace{-2.0mm}
\end{figure}

\paragraph{Contribution of attention and MLP updates.}
We further investigate the individual and joint contributions of attention and MLP residual updates by independently suppressing each branch within the isolated suffix. Specifically, we compare the default configuration, in which both branches follow the attenuation schedule, against configurations in which either or both branches are assigned a fixed scaling factor of $0$ or $0.01$. Figure~\ref{fig:branch-ablation} shows that retaining both branches with the default attenuation schedule generally provides the strongest performance. Notably, configurations with a scaling factor of $0.01$ for both branches generally outperform those with both branches set to zero. This difference highlights the importance of retaining even a small contribution from downstream nonlinear transformations. When both branches are assigned zero weight, their residual updates are completely removed, effectively turning each corresponding isolated block into an identity mapping and reducing the overall computation to a direct readout of the original residual representation, as in the Logit Lens. In contrast, a scaling factor of $0.01$ preserves a small but nonzero contribution from both attention and MLP transformations, allowing the isolated suffix to retain a subtle effect on the representation rather than skipping these transformations entirely. Although this contribution is heavily attenuated, it can still provide useful nonlinear transformations for intermediate concept recovery. The substantial performance degradation of both configurations relative to the default setting further highlights the importance of retaining downstream computation. Overall, these results demonstrate the complementary contributions of attention and MLP updates and support attenuating, rather than removing, both branches.

\paragraph{Different Attenuation schedules.}

We also investigate the effect of different attenuation schedules on iLens by comparing linear, cosine, and two exponential decay variants. Specifically, all schedules share the plateau and the endpoint of Eq.~\eqref{eq:attenuation}: $w_k=1$ for
$k/L\le\tau$, and $w_L=\varepsilon$ with a small floor $\varepsilon=0.01$.
With the normalized ramp position
\begin{equation}
    t_k = \min\!\left(1,\ \max\!\left(0,\ \frac{k/L-\tau}{1-\tau}\right)\right),
    \label{eq:ramp-position}
\end{equation}
they differ only in how $w_k$ falls from $1$ at $t_k=0$ to $\varepsilon$ at $t_k=1$:
\begin{equation}
    w_k =
    \begin{cases}
        1-(1-\varepsilon)\,t_k, & \text{Linear (default)},\\[2pt]
        \varepsilon+(1-\varepsilon)\,\tfrac{1}{2}\bigl(1+\cos(\pi t_k)\bigr), & \text{Cosine},\\[2pt]
        \varepsilon^{\,t_k}=\exp\!\bigl(t_k\ln\varepsilon\bigr), & \text{Early EXP},\\[2pt]
        1+\varepsilon-\varepsilon^{\,1-t_k}, & \text{Late EXP}.
    \end{cases}
    \label{eq:ramp-shapes}
\end{equation}
\emph{Linear} lowers the weight by the same amount at every block of the ramp.
\emph{Cosine} follows a half cosine period: it leaves the plateau and approaches the floor with
zero slope, and descends fastest mid-ramp.
\emph{Early EXP} is a geometric decay, lowering the weight by the same factor
$\varepsilon^{1/(L(1-\tau))}$ at every block; most of the attenuation therefore happens
immediately after the cutoff ($w=0.5$ already at $t_k\approx0.15$, and $w=0.1$ at $t_k=0.5$).
\emph{Late EXP} is its point reflection, $w^{\mathrm{late}}(t)=1+\varepsilon-w^{\mathrm{early}}(1-t)$:
it keeps the weight close to one for most of the ramp ($w\approx0.91$ at $t_k=0.5$) and
concentrates the attenuation in the final blocks ($w=0.5$ only at $t_k\approx0.85$).
Linear and Cosine are symmetric about the ramp's midpoint ($w=(1+\varepsilon)/2$ at $t_k=0.5$).
In every variant the same $w_k$ scales both the attention and the MLP update of
Eq.~\eqref{eq:isolated-block}.

\begin{figure}[t]
    \centering
    \includegraphics[width=\linewidth]{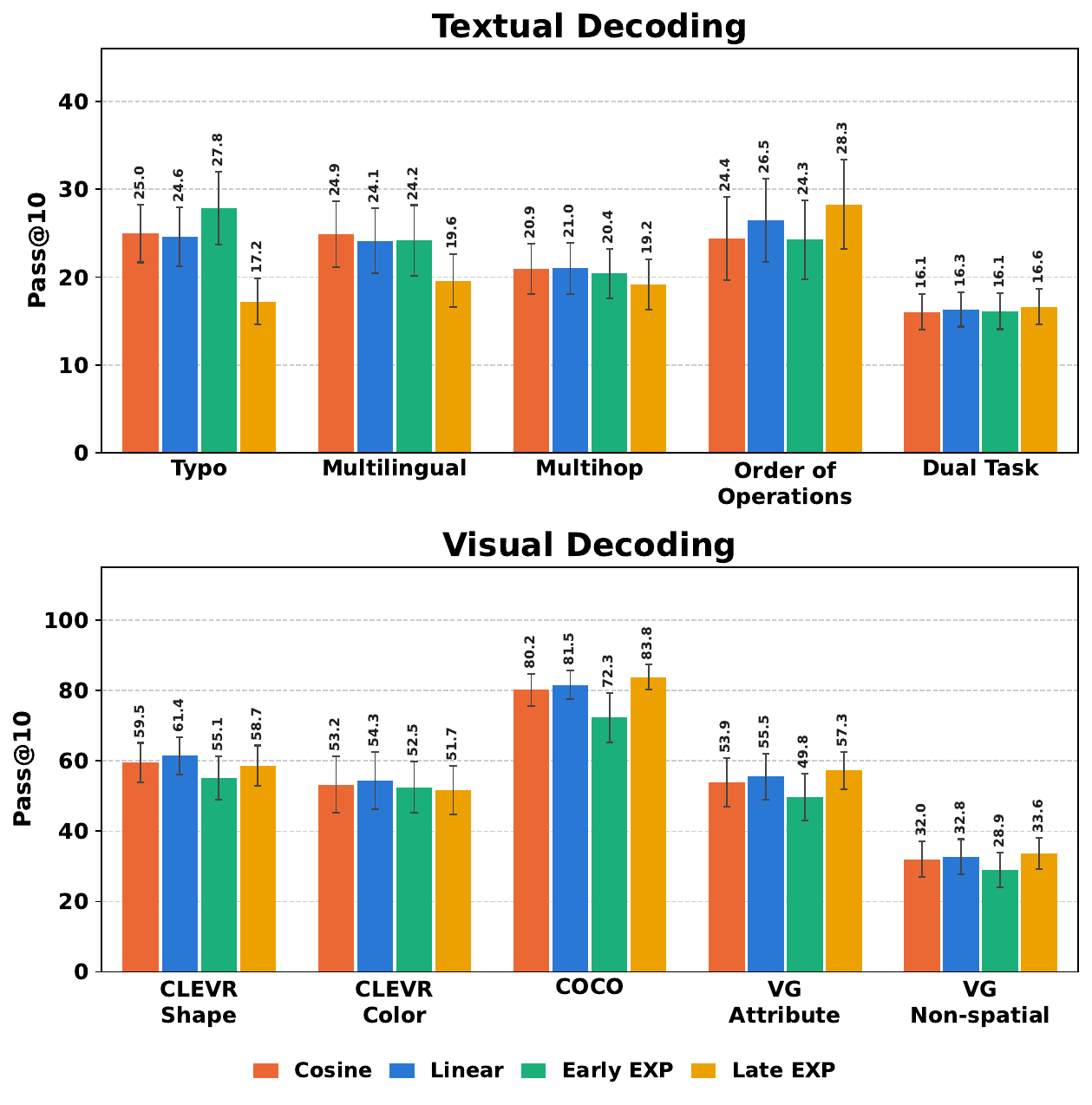}
    \caption{Comparison of four attenuation schedules (Linear, Cosine, Early EXP, and Late EXP) across five textual and five visual concept decoding tasks. Pass@10 is averaged over 18 LLMs for textual decoding and 14 LVLMs for visual decoding. Overall, the schedules achieve comparable performance, with no consistent or substantial advantage across all tasks.}
    \label{fig:ramp-schedule-ablation}
    \vspace{-2.0mm}
\end{figure}

Figure~\ref{fig:ramp-schedule-ablation} reports Pass@10 across the five textual and five visual decoding tasks. The results reveal that the effectiveness of an attenuation schedule varies across tasks and modalities, with no single schedule consistently outperforming the others. We therefore choose linear decay as the default schedule due to its simplicity and consistently competitive performance across both modalities, avoiding the need for task-specific schedule selection.

\begin{figure*}[t]
    \centering
    \includegraphics[width=\linewidth]
    {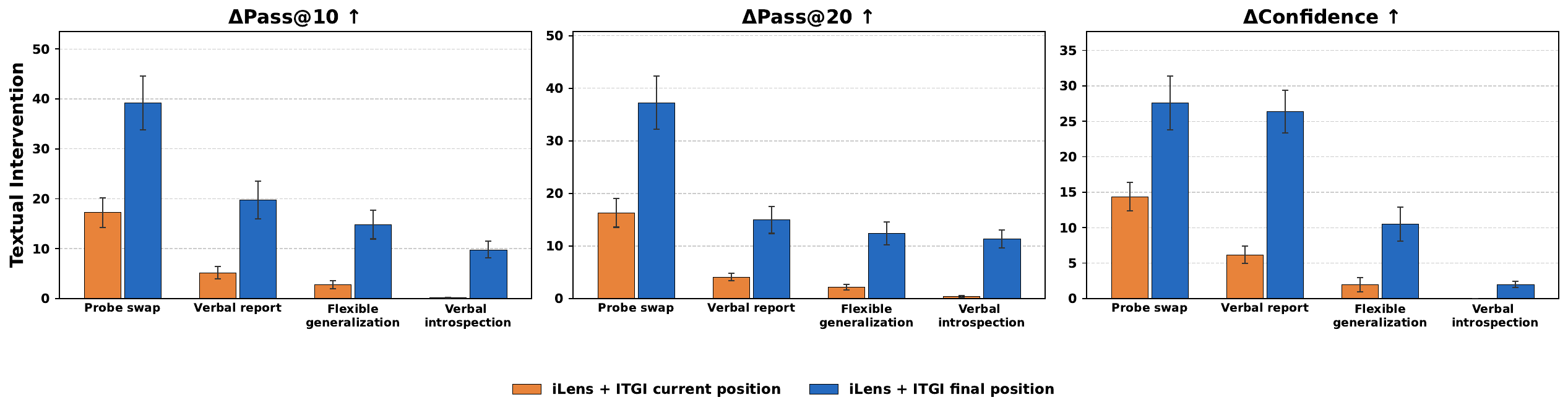}
    \caption{Ablation of the ITGI objective position for textual interventions, averaged across the evaluated language models. We compare concept interventions using gradients computed at the current (edited) token position and the final prompt position. Results are reported in terms of intervention-effect AUC for $\Delta$Pass@10, $\Delta$Pass@20, and $\Delta$Confidence across Probe Swap, Verbal Report, Flexible Generalization, and Verbal Introspection. Higher values indicate stronger shifts toward injected concepts relative to source concepts.}
    \label{fig:intervention-ablation}
    \vspace{-2.0mm}
\end{figure*}

\paragraph{Choice of the intervention objective position.}
We investigate how the position at which the ITGI objective is evaluated affects the effectiveness of concept interventions. Specifically, we compare two variants of iLens + ITGI: computing the concept log-odds at the current (edited) token position $t$, and computing them at the final prompt position $T$, corresponding to the response prediction site. In both settings, the intervention modifies the selected residual embedding $h_{\ell,t}$; the difference lies in the position used to evaluate the objective and compute its gradient. The final-position variant propagates the selected representation together with the actual continuation tokens and evaluates the resulting concept score at the response prediction site, thereby accounting for downstream interactions between the edited representation and subsequent tokens.

Figure~\ref{fig:intervention-ablation} reports the intervention-effect AUC across four textual intervention tasks, averaged over the evaluated language models. The results show that computing the objective at the final prompt position consistently outperforms the current-position variant across all three evaluation metrics: $\Delta$Pass@10, $\Delta$Pass@20, and $\Delta$Confidence. This advantage is particularly pronounced in Probe Swap and Verbal Report, while substantial improvements are also observed in Flexible Generalization and Verbal Introspection. This result supports our design choice of computing the ITGI objective at the final prompt position.

\section{Broader Experiments}
\label{app:additional_results}

In this section, we extend the experimental evaluation of iLens
beyond the representative models reported in the main paper.
We provide additional quantitative, qualitative, and layer-wise
results to further examine the generality of iLens across different
model architectures and visual concept-decoding tasks.

In particular, we extend the visual concept-decoding evaluation
to 14 vision-capable models, covering both large vision-language
models (LVLMs) and unified vision models. We report detailed
per-task comparisons, model-level aggregate results, and
layer-wise performance curves in Section~\ref{app:subsec:lvlm_results}.
We also provide additional textual decoding results in
Section~\ref{app:subsec:llm_results} and intervention results
in Section~\ref{app:subsec:intervention_results} to complement
the main experiments.

\subsection{Additional Visual Concept Decoding Experiments}
\label{app:subsec:lvlm_results}

We provide an extended evaluation of visual concept decoding
across 14 vision-capable models, including different model
families, parameter scales, and architectural designs.
The evaluation covers five visual benchmarks:
CLEVR Shape, CLEVR Color, COCO, Visual Genome Attribute,
and Visual Genome Non-Spatial.

These experiments allow us to examine whether the improvements
observed in the main comparisons generalize across different
architectures and visual concept categories. All methods are
evaluated using the same fixed set of 50 samples per task,
ensuring consistent comparisons across decoding methods.

\paragraph{Quantitative Results.}

Table~\ref{tab:appendix-vlm-full-results} presents the
comprehensive quantitative results across all five visual
decoding tasks and 14 vision-capable models. We report
Pass@10, Pass@20, Confidence, and First Emergence Depth (FED),
providing complementary measures of concept recovery,
probability assignment, and the depth at which target concepts
first become recoverable.

Taking the average across all 14 vision-capable models,
iLens achieves 56.6\% Pass@10 and 63.3\% Pass@20,
compared with 31.2\% and 36.5\% for J-Lens,
respectively. It also achieves a substantially lower
FED of 30.6\%, compared with 62.8\% for J-Lens,
indicating that visual concepts become recoverable
considerably earlier. For Confidence, iLens achieves
an average of 6.1\%, outperforming J-Lens (4.0\%)
and Logit Lens (2.4\%). These results demonstrate
consistent improvements in visual concept recovery
across diverse model architectures.



\providecommand{\se}[1]{$_{\textcolor{gray}{\scalebox{0.65}{$\scriptstyle\scalebox{0.6}{$\scriptstyle\pm$}#1$}}}$}

\begin{table*}[t]
\centering
\caption{Comparison of iLens with existing decoding methods across five visual probing tasks: CLEVR Shape and CLEVR Color~\cite{clevr}, COCO~\cite{coco}, Visual Genome Attribute, and Visual Genome Non-spatial~\cite{visual_genome}. Results are reported using Pass@10 (P@10 $\uparrow$), Pass@20 (P@20 $\uparrow$), First Emergence Depth (FED $\downarrow$), and Confidence (Conf. $\uparrow$), together with their averages across tasks. The best results within each model and task are shown in bold.}
\label{tab:appendix-vlm-full-results}

\begingroup
\fontsize{7}{8.4}\selectfont
\setlength{\tabcolsep}{1.8pt}
\renewcommand{\arraystretch}{1.08}
\setlength{\aboverulesep}{1.4pt}
\setlength{\belowrulesep}{1.4pt}

\resizebox{\textwidth}{!}{%
\begin{tabular}{@{}ll*{5}{rrrr}|rrrr@{}}

\toprule[0.9pt]

\multirow{2}{*}{Model}
& \multirow{2}{*}{Method}
& \multicolumn{4}{c}{CLEVR Shape~\cite{clevr}}
& \multicolumn{4}{c}{CLEVR Color~\cite{clevr}}
& \multicolumn{4}{c}{COCO~\cite{coco}}
& \multicolumn{4}{c}{VG Attribute~\cite{visual_genome}}
& \multicolumn{4}{c}{VG Non-spatial~\cite{visual_genome}}
& \multicolumn{4}{|c}{Average} \\

\cmidrule[0.55pt](lr){3-6}
\cmidrule[0.55pt](lr){7-10}
\cmidrule[0.55pt](lr){11-14}
\cmidrule[0.55pt](lr){15-18}
\cmidrule[0.55pt](lr){19-22}
\cmidrule[0.55pt](lr){23-26}

& & P@10 $\uparrow$ & P@20 $\uparrow$ & FED $\downarrow$ & Conf. $\uparrow$
& P@10 $\uparrow$ & P@20 $\uparrow$ & FED $\downarrow$ & Conf. $\uparrow$
& P@10 $\uparrow$ & P@20 $\uparrow$ & FED $\downarrow$ & Conf. $\uparrow$
& P@10 $\uparrow$ & P@20 $\uparrow$ & FED $\downarrow$ & Conf. $\uparrow$
& P@10 $\uparrow$ & P@20 $\uparrow$ & FED $\downarrow$ & Conf. $\uparrow$
& P@10 $\uparrow$ & P@20 $\uparrow$ & FED $\downarrow$ & Conf. $\uparrow$ \\

\midrule[0.65pt]

\multirow{5}{*}{Qwen3.5-9B~\cite{qwen3.5}}
& Logit-Lens
& 30.7\se{1.1} & 35.5\se{1.3} & 63.5\se{1.3} & 4.2\se{0.2}
& 24.5\se{0.8} & 29.8\se{0.8} & 66.8\se{1.0} & 1.8\se{0.1}
& 32.7\se{2.3} & 38.4\se{2.5} & 50.2\se{3.5} & 1.9\se{0.2}
& 19.6\se{1.2} & 25.5\se{1.3} & 71.8\se{1.7} & 0.5\se{0.1}
& 10.2\se{1.3} & 13.0\se{1.5} & 82.0\se{2.2} & 0.5\se{0.1}
& 23.5\se{0.6} & 28.4\se{0.7} & 66.9\se{0.9} & 1.8\se{0.1} \\

& Emb-Lens
& 15.3\se{1.1} & 21.3\se{1.3} & 62.5\se{2.3} & --
& 43.1\se{2.3} & 48.3\se{2.2} & 42.5\se{2.3} & --
& 35.6\se{3.5} & 43.9\se{3.6} & 41.1\se{3.9} & --
& 20.7\se{2.3} & 25.3\se{2.5} & 68.4\se{2.8} & --
& 3.0\se{0.8} & 5.5\se{1.1} & 88.6\se{2.4} & --
& 23.5\se{1.0} & 28.9\se{1.0} & 60.6\se{1.2} & -- \\
\end{tabular}%
}

\endgroup
\end{table*}

\begin{figure*}[t]
    \centering
    \includegraphics[width=0.99\linewidth]{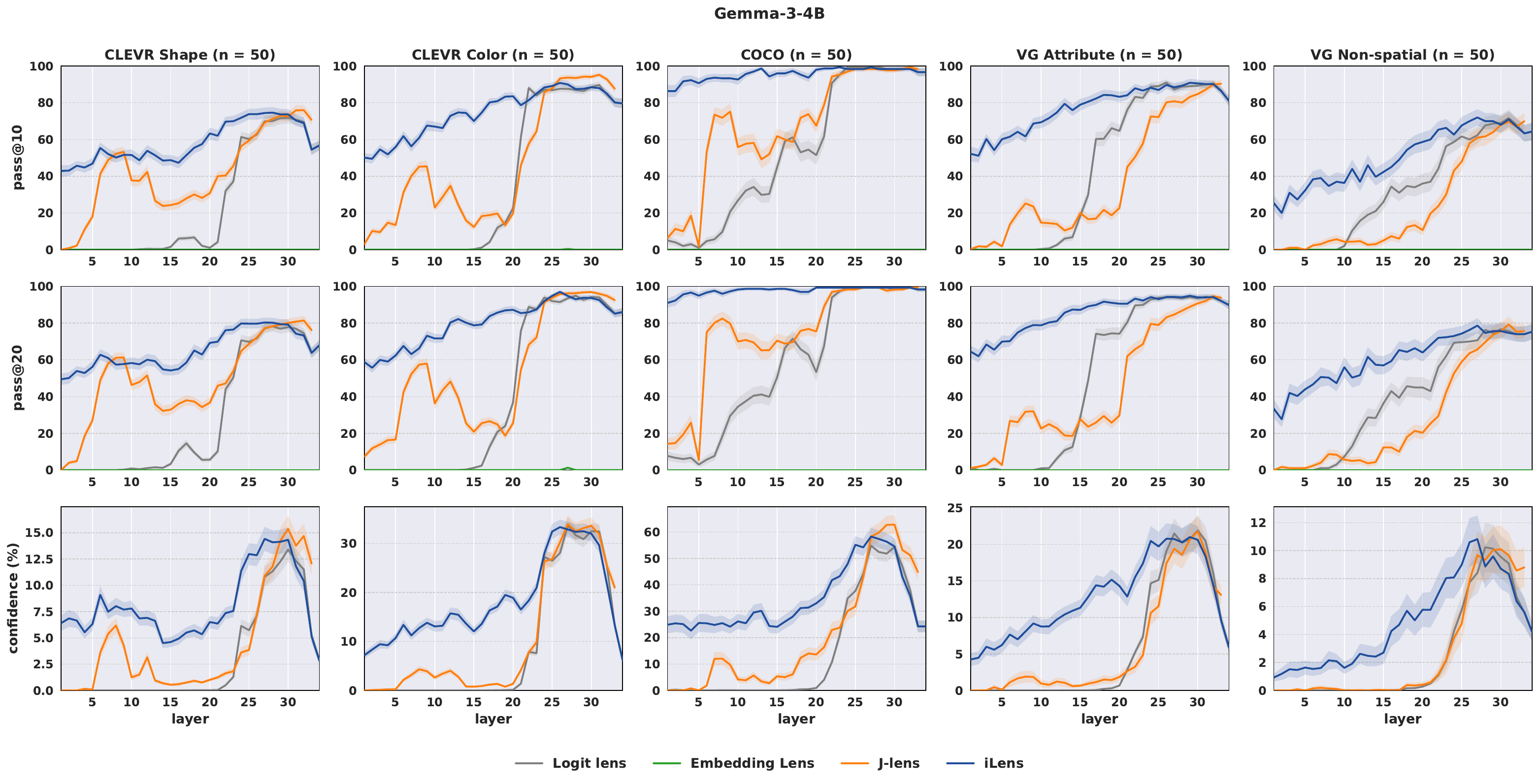}
    \caption{Layer-wise visual concept decoding performance of Gemma-3-4B across five evaluation datasets: CLEVR Shape ($n=50$), CLEVR Color ($n=50$), COCO ($n=50$), VG Attribute ($n=50$), and VG Non-spatial ($n=50$). The three rows report Pass@10, Pass@20, and Confidence, respectively, across transformer layers. Results compare Logit Lens, Embedding Lens, J-Lens, and iLens.}
    \label{fig:layerwise-vlm-gemma-3-4b}
    \vspace{-4.0mm}
\end{figure*}

\begin{figure*}[t]
    \centering
    \includegraphics[width=0.99\linewidth]{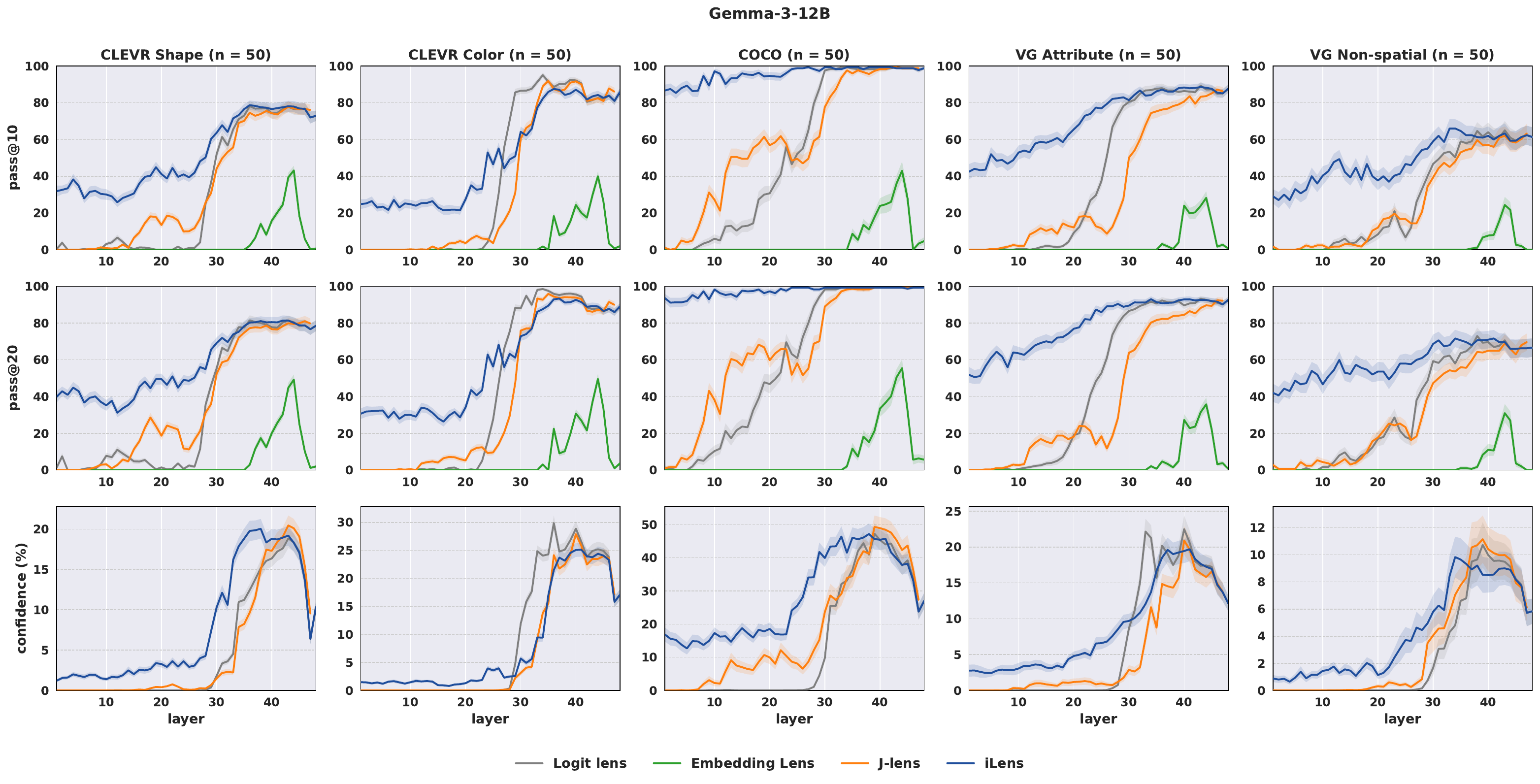}
    \caption{Layer-wise visual concept decoding performance of Gemma-3-12B across five evaluation datasets: CLEVR Shape ($n=50$), CLEVR Color ($n=50$), COCO ($n=50$), VG Attribute ($n=50$), and VG Non-spatial ($n=50$). The three rows report Pass@10, Pass@20, and Confidence, respectively, across transformer layers. Results compare Logit Lens, Embedding Lens, J-Lens, and iLens.}
    \label{fig:layerwise-vlm-gemma-3-12b}
    \vspace{-4.0mm}
\end{figure*}

\begin{figure*}[t]
    \centering
    \includegraphics[width=0.99\linewidth]{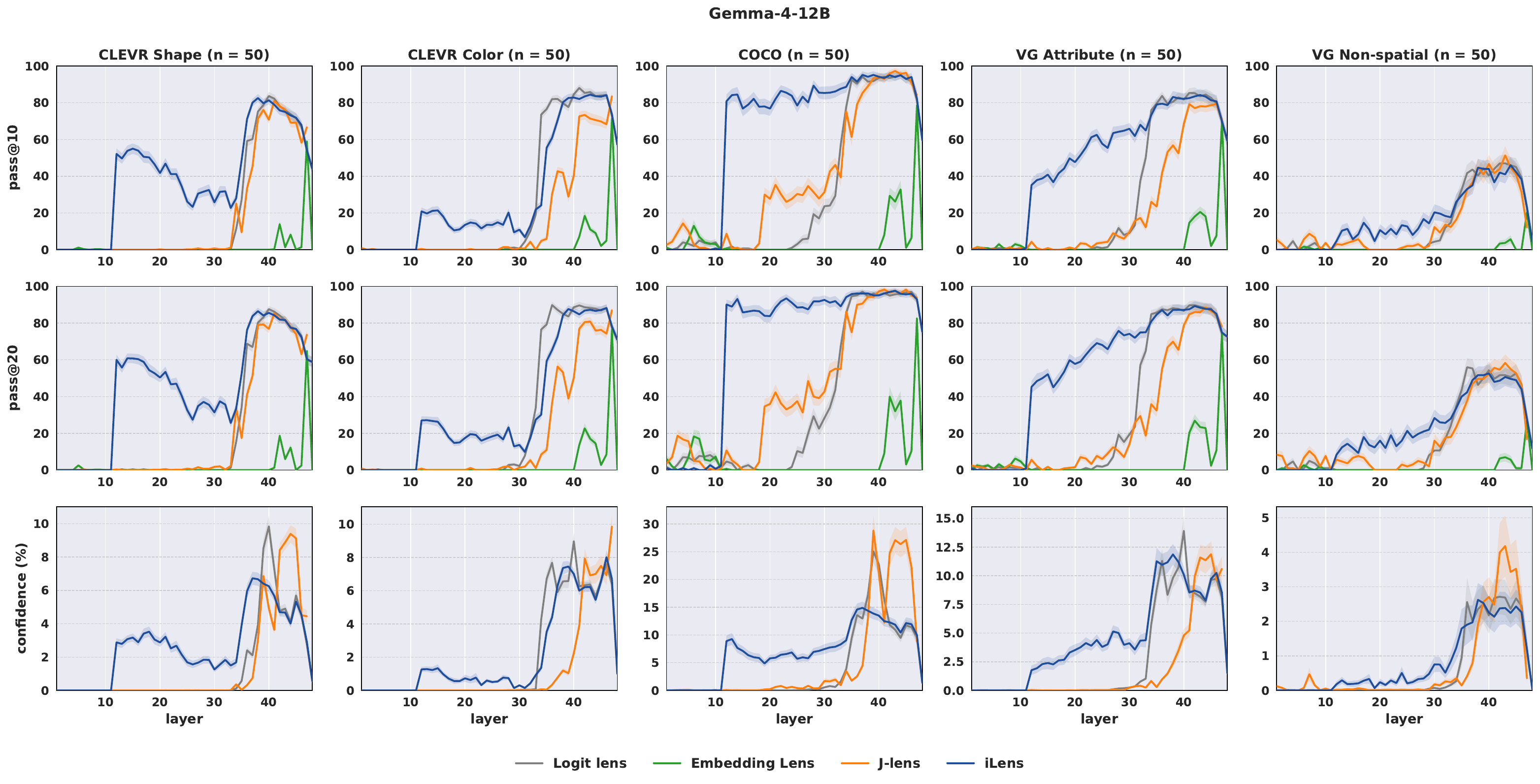}
    \caption{Layer-wise visual concept decoding performance of Gemma-4-12B across five evaluation datasets: CLEVR Shape ($n=50$), CLEVR Color ($n=50$), COCO ($n=50$), VG Attribute ($n=50$), and VG Non-spatial ($n=50$). The three rows report Pass@10, Pass@20, and Confidence, respectively, across transformer layers. Results compare Logit Lens, Embedding Lens, J-Lens, and iLens.}
    \label{fig:layerwise-vlm-gemma-4-12b}
    \vspace{-4.0mm}
\end{figure*}

\begin{figure*}[t]
    \centering
    \includegraphics[width=0.99\linewidth]{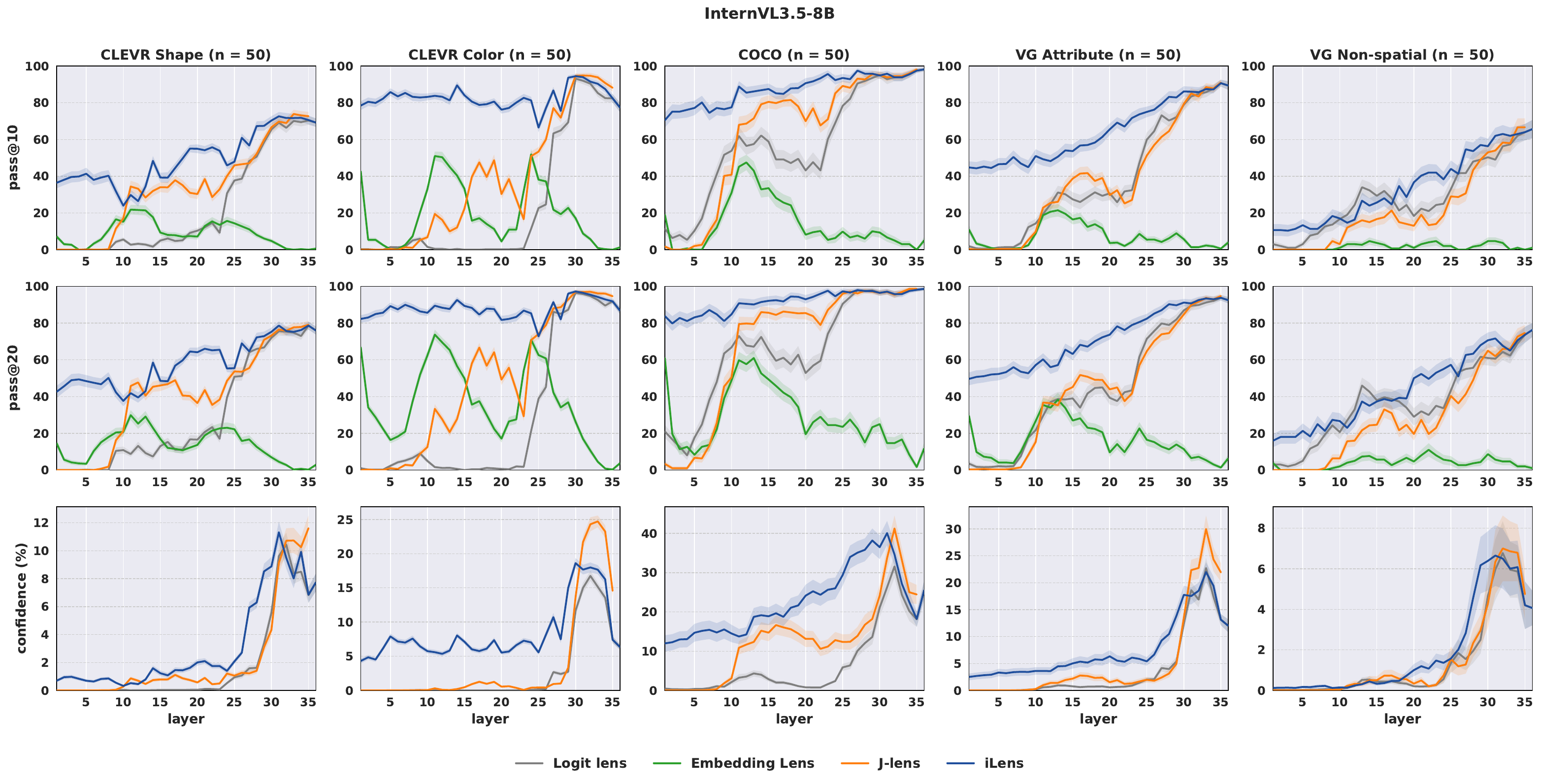}
    \caption{Layer-wise visual concept decoding performance of InternVL3.5-8B across five evaluation datasets: CLEVR Shape ($n=50$), CLEVR Color ($n=50$), COCO ($n=50$), VG Attribute ($n=50$), and VG Non-spatial ($n=50$). The three rows report Pass@10, Pass@20, and Confidence, respectively, across transformer layers. Results compare Logit Lens, Embedding Lens, J-Lens, and iLens.}
    \label{fig:layerwise-vlm-internvl3-5-8b}
    \vspace{-4.0mm}
\end{figure*}

\begin{figure*}[t]
    \centering
    \includegraphics[width=0.99\linewidth]{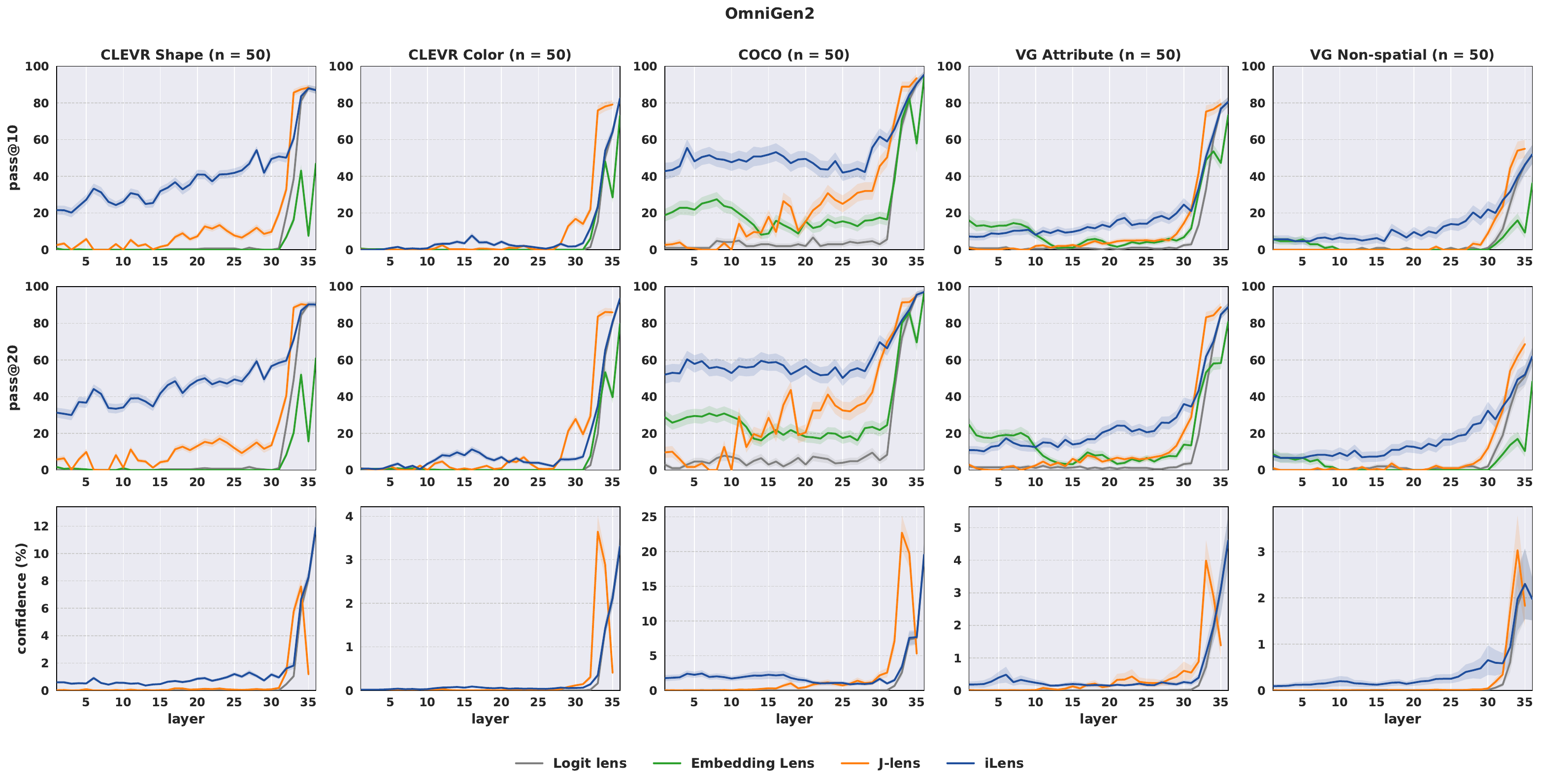}
    \caption{Layer-wise visual concept decoding performance of OmniGen2 across five evaluation datasets: CLEVR Shape ($n=50$), CLEVR Color ($n=50$), COCO ($n=50$), VG Attribute ($n=50$), and VG Non-spatial ($n=50$). The three rows report Pass@10, Pass@20, and Confidence, respectively, across transformer layers. Results compare Logit Lens, Embedding Lens, J-Lens, and iLens.}
    \label{fig:layerwise-vlm-omnigen2}
    \vspace{-4.0mm}
\end{figure*}

\begin{figure*}[t]
    \centering
    \includegraphics[width=0.99\linewidth]{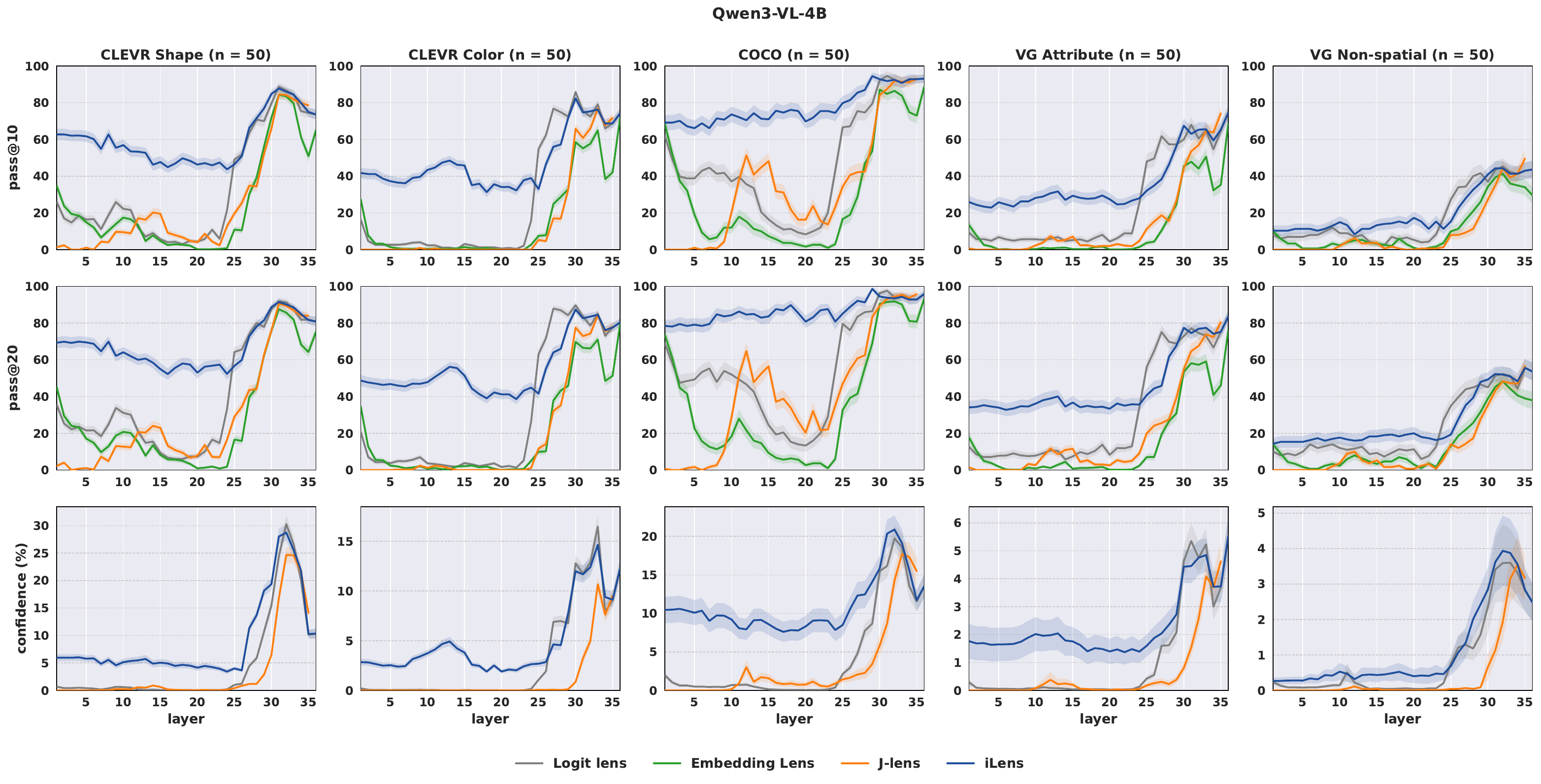}
    \caption{Layer-wise visual concept decoding performance of Qwen3-VL-4B across five evaluation datasets: CLEVR Shape ($n=50$), CLEVR Color ($n=50$), COCO ($n=50$), VG Attribute ($n=50$), and VG Non-spatial ($n=50$). The three rows report Pass@10, Pass@20, and Confidence, respectively, across transformer layers. Results compare Logit Lens, Embedding Lens, J-Lens, and iLens.}
    \label{fig:layerwise-vlm-qwen3-vl-4b}
    \vspace{-4.0mm}
\end{figure*}

\begin{figure*}[t]
    \centering
    \includegraphics[width=0.99\linewidth]{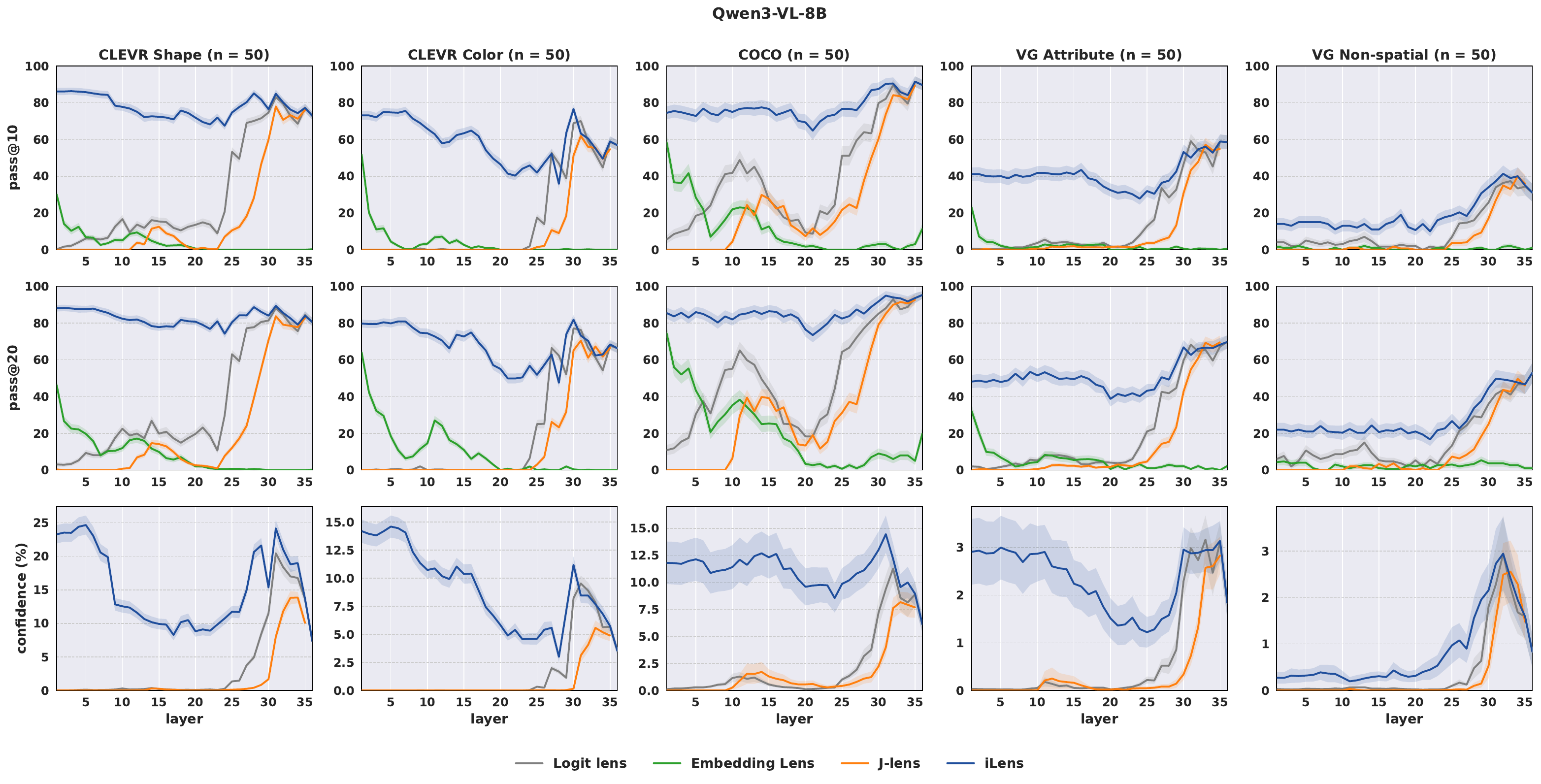}
    \caption{Layer-wise visual concept decoding performance of Qwen3-VL-8B across five evaluation datasets: CLEVR Shape ($n=50$), CLEVR Color ($n=50$), COCO ($n=50$), VG Attribute ($n=50$), and VG Non-spatial ($n=50$). The three rows report Pass@10, Pass@20, and Confidence, respectively, across transformer layers. Results compare Logit Lens, Embedding Lens, J-Lens, and iLens.}
    \label{fig:layerwise-vlm-qwen3-vl-8b}
    \vspace{-4.0mm}
\end{figure*}

\begin{figure*}[t]
    \centering
    \includegraphics[width=0.99\linewidth]{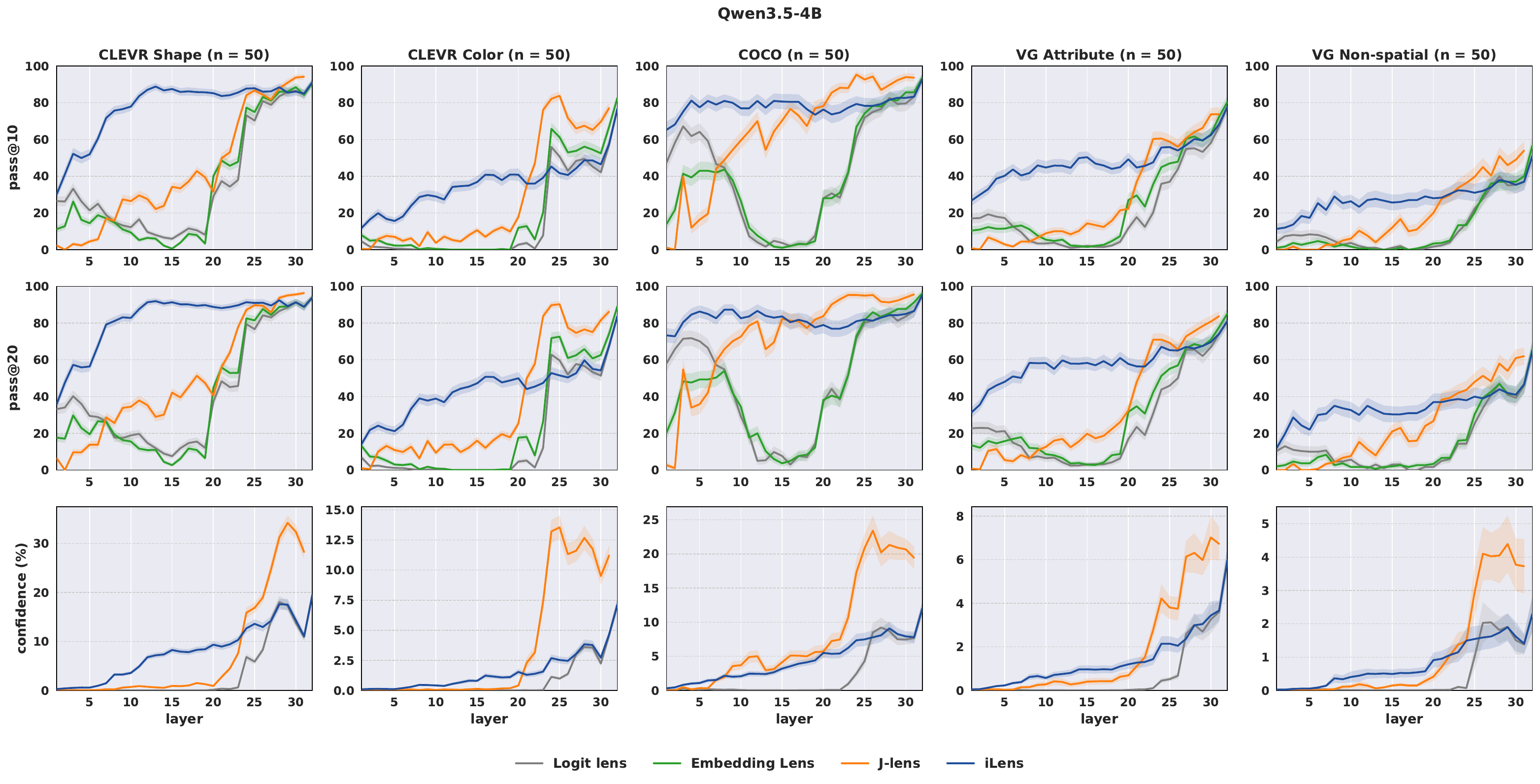}
    \caption{Layer-wise visual concept decoding performance of Qwen3.5-4B across five evaluation datasets: CLEVR Shape ($n=50$), CLEVR Color ($n=50$), COCO ($n=50$), VG Attribute ($n=50$), and VG Non-spatial ($n=50$). The three rows report Pass@10, Pass@20, and Confidence, respectively, across transformer layers. Results compare Logit Lens, Embedding Lens, J-Lens, and iLens.}
    \label{fig:layerwise-vlm-qwen3-5-4b}
    \vspace{-4.0mm}
\end{figure*}

\begin{figure*}[t]
    \centering
    \includegraphics[width=0.99\linewidth]{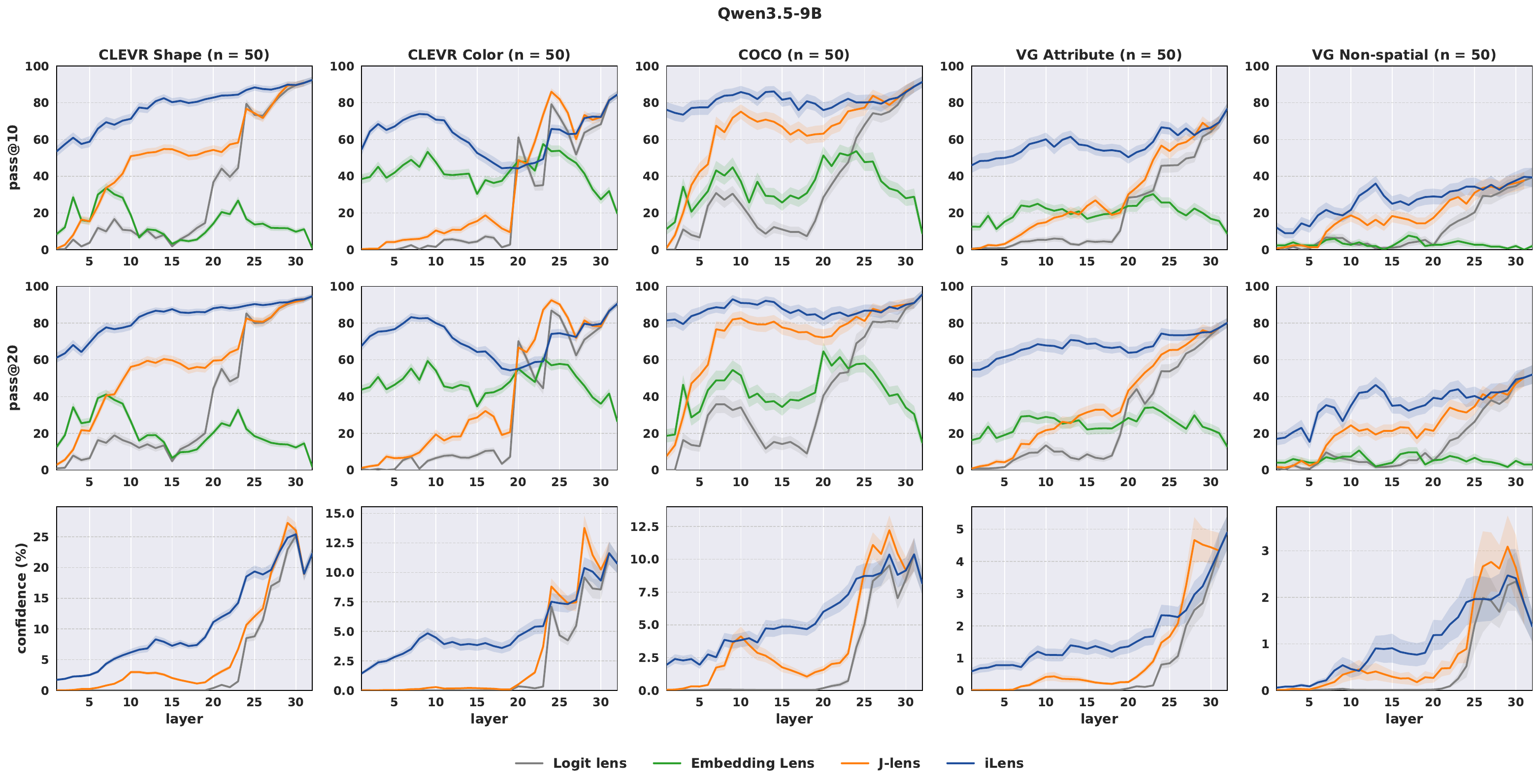}
    \caption{Layer-wise visual concept decoding performance of Qwen3.5-9B across five evaluation datasets: CLEVR Shape ($n=50$), CLEVR Color ($n=50$), COCO ($n=50$), VG Attribute ($n=50$), and VG Non-spatial ($n=50$). The three rows report Pass@10, Pass@20, and Confidence, respectively, across transformer layers. Results compare Logit Lens, Embedding Lens, J-Lens, and iLens.}
    \label{fig:layerwise-vlm-qwen3-5-9b}
    \vspace{-4.0mm}
\end{figure*}

\begin{figure*}[t]
    \centering
    \includegraphics[width=0.99\linewidth]{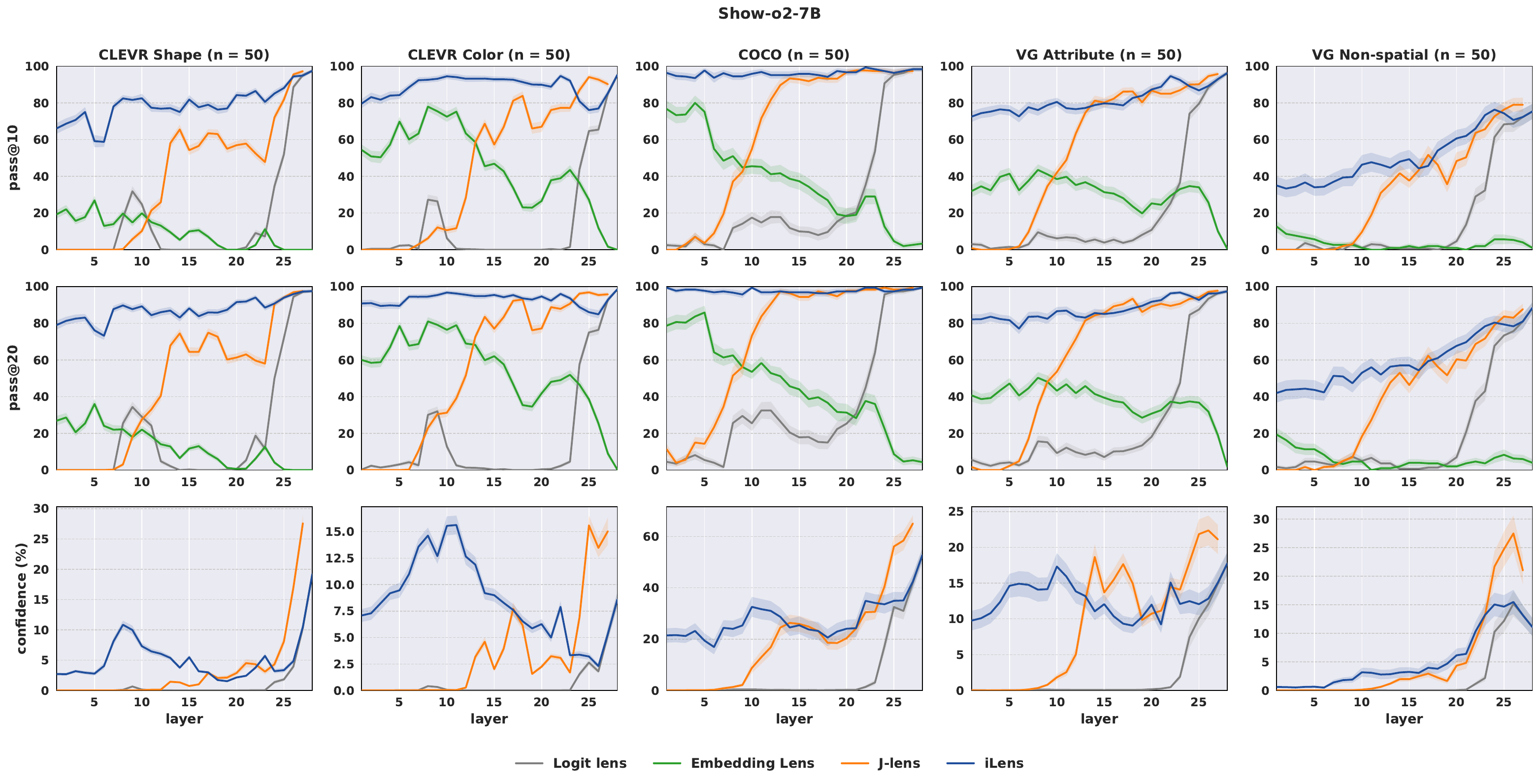}
    \caption{Layer-wise visual concept decoding performance of Show-o2-7B across five evaluation datasets: CLEVR Shape ($n=50$), CLEVR Color ($n=50$), COCO ($n=50$), VG Attribute ($n=50$), and VG Non-spatial ($n=50$). The three rows report Pass@10, Pass@20, and Confidence, respectively, across transformer layers. Results compare Logit Lens, Embedding Lens, J-Lens, and iLens.}
    \label{fig:layerwise-vlm-show-o2-7b}
    \vspace{-4.0mm}
\end{figure*}

\begin{figure*}[t]
    \centering
    \includegraphics[width=0.99\linewidth]{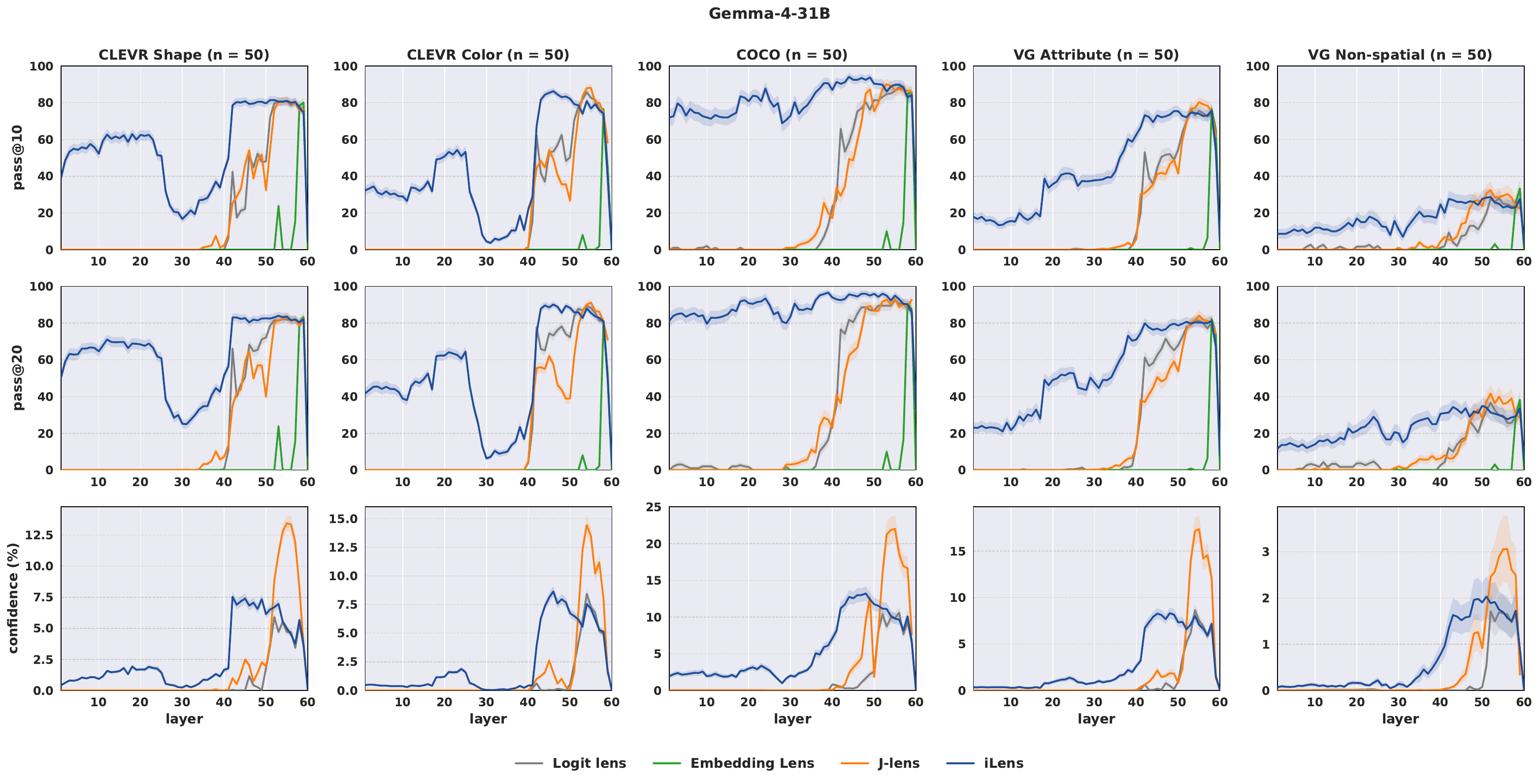}
    \caption{Layer-wise visual concept decoding performance of Gemma4-31B across five evaluation datasets: CLEVR Shape ($n=50$), CLEVR Color ($n=50$), COCO ($n=50$), VG Attribute ($n=50$), and VG Non-spatial ($n=50$). The three rows report Pass@10, Pass@20, and Confidence, respectively, across transformer layers. Results compare Logit Lens, Embedding Lens, J-Lens, and iLens.}
    \label{fig:layerwise-vlm-gemma-4-32b}
    \vspace{-4.0mm}
\end{figure*}

\begin{figure*}[t]
    \centering
    \includegraphics[width=0.99\linewidth]{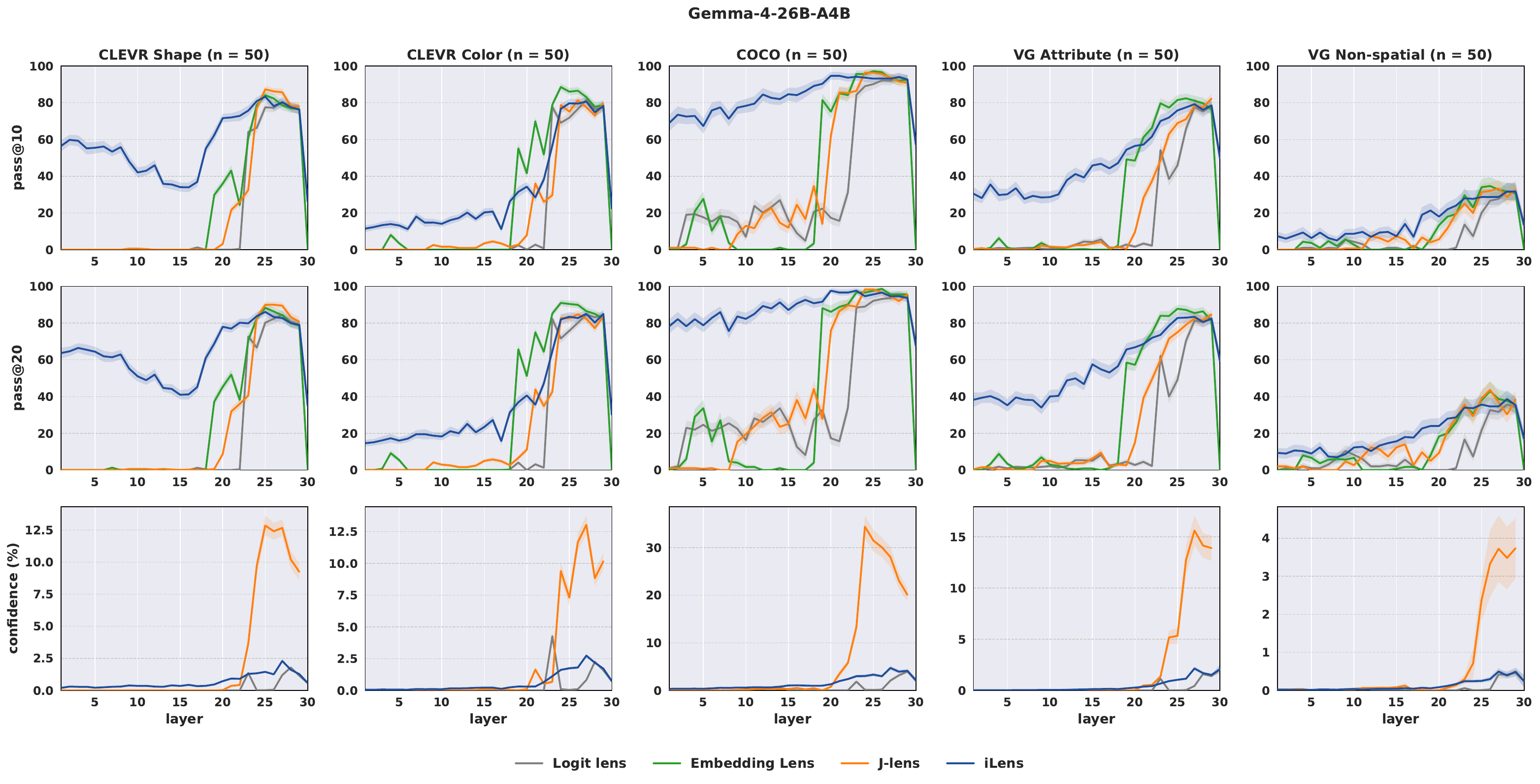}
    \caption{Layer-wise visual concept decoding performance of Gemma-4-26B-A4B across five evaluation datasets: CLEVR Shape ($n=50$), CLEVR Color ($n=50$), COCO ($n=50$), VG Attribute ($n=50$), and VG Non-spatial ($n=50$). The three rows report Pass@10, Pass@20, and Confidence, respectively, across transformer layers. Results compare Logit Lens, Embedding Lens, J-Lens, and iLens.}
    \label{fig:layerwise-vlm-gemma-4-26b-A4B}
    \vspace{-4.0mm}
\end{figure*}

\begin{figure*}[t]
    \centering
    \includegraphics[width=0.99\linewidth]{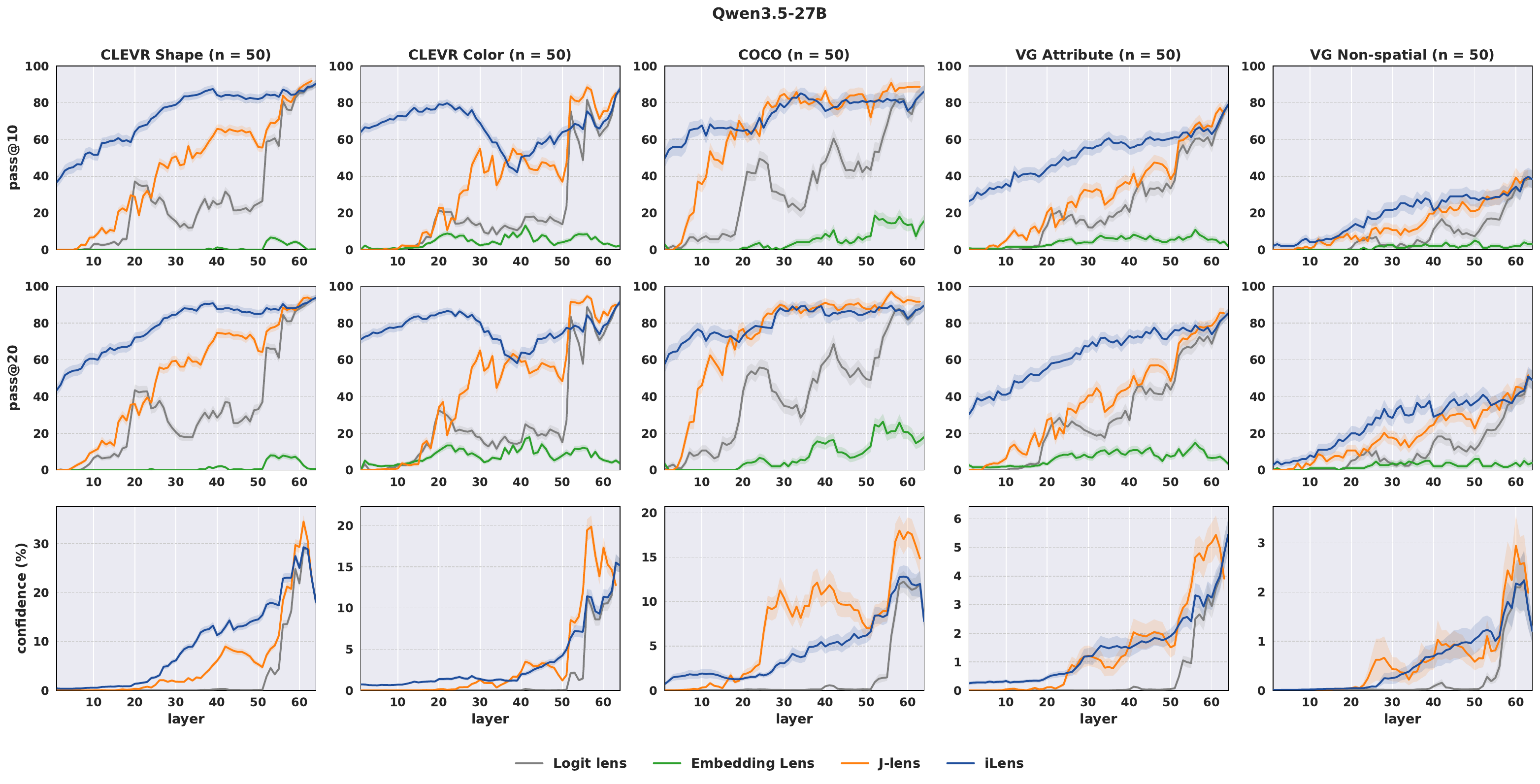}
    \caption{Layer-wise visual concept decoding performance of Qwen3.5-27B across five evaluation datasets: CLEVR Shape ($n=50$), CLEVR Color ($n=50$), COCO ($n=50$), VG Attribute ($n=50$), and VG Non-spatial ($n=50$). The three rows report Pass@10, Pass@20, and Confidence, respectively, across transformer layers. Results compare Logit Lens, Embedding Lens, J-Lens, and iLens.}
    \label{fig:layerwise-vlm-qwen3.5-27B}
    \vspace{-4.0mm}
\end{figure*}

\begin{figure*}[t]
    \centering
    \includegraphics[width=0.99\linewidth]{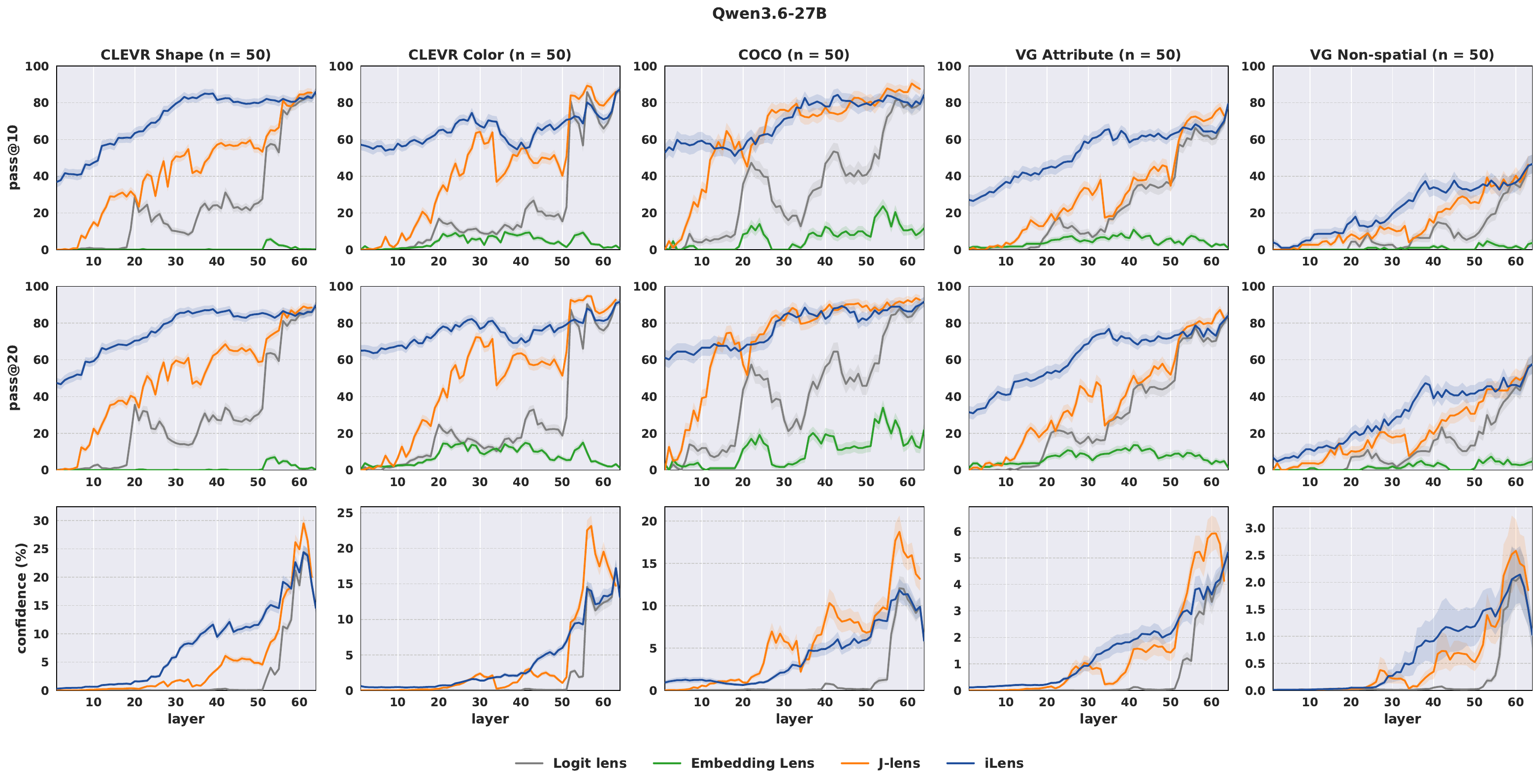}
    \caption{Layer-wise visual concept decoding performance of Qwen3.6-27B across five evaluation datasets: CLEVR Shape ($n=50$), CLEVR Color ($n=50$), COCO ($n=50$), VG Attribute ($n=50$), and VG Non-spatial ($n=50$). The three rows report Pass@10, Pass@20, and Confidence, respectively, across transformer layers. Results compare Logit Lens, Embedding Lens, J-Lens, and iLens.}
    \label{fig:layerwise-vlm-qwen3.6-27B}
    \vspace{-4.0mm}
\end{figure*}

To examine how visual concept recoverability evolves across
model depth, Figures~\ref{fig:layerwise-vlm-gemma-3-4b}--\ref{fig:layerwise-vlm-qwen3.6-27B}
present layer-wise performance curves for all 14 vision-capable
models across the five visual decoding benchmarks. For each
model and benchmark, we report Pass@10, Pass@20, and
Confidence at every transformer layer, comparing Logit Lens,
Embedding Lens, J-Lens, and iLens. These curves provide
a fine-grained view of how each method recovers visual
concepts throughout the model's depth and how its performance
evolves across different architectures.

Figure~\ref{fig:radar-vlm} in the main text summarizes the task-averaged results for all 14 vision-capable models.

\paragraph{Qualitative Results.}

\begin{figure*}[t]
    \centering
    \includegraphics[width=0.95\linewidth]{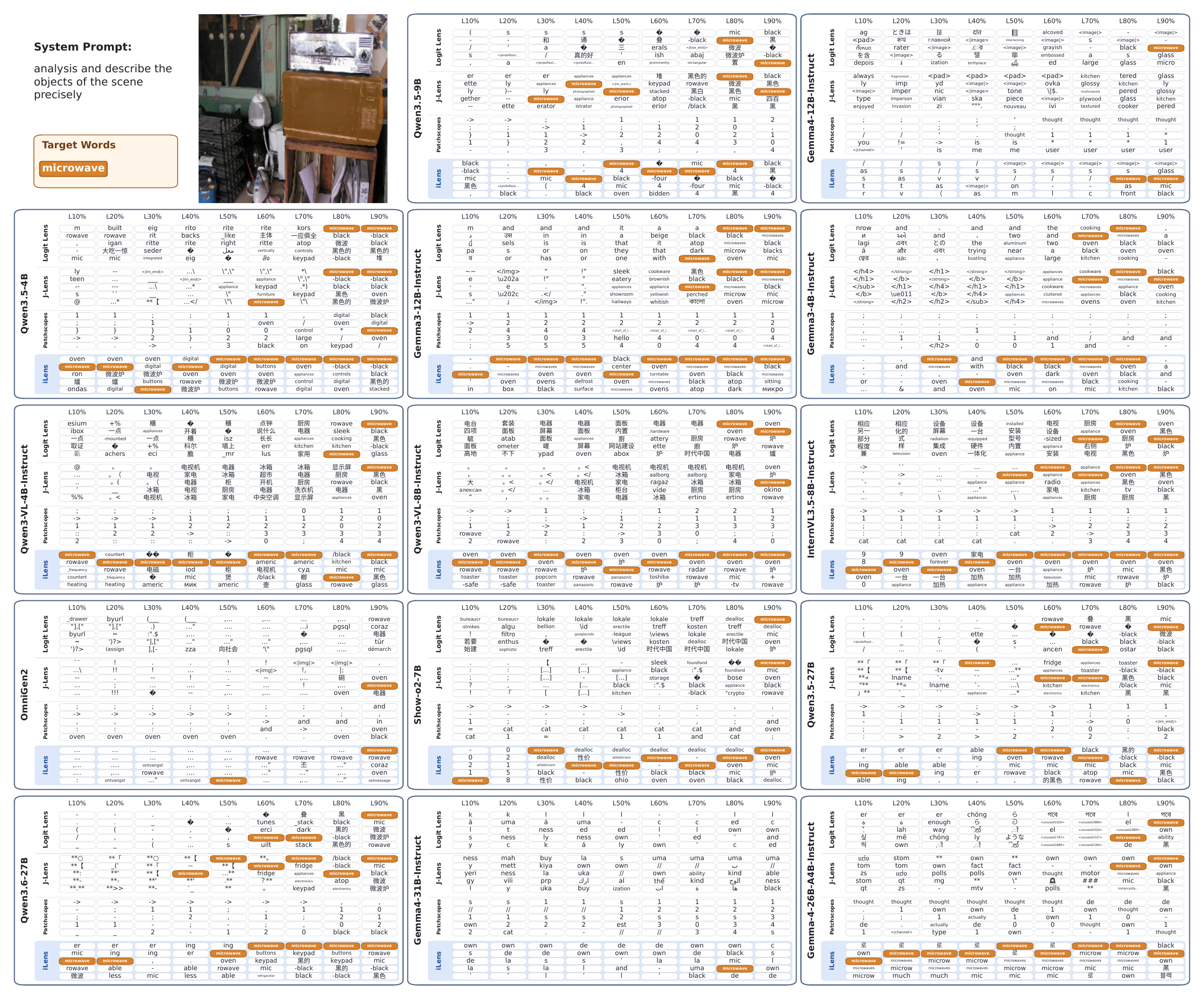}
    \caption{Qualitative example of the Visual-Decoding task across 14 vision-capable models, with predictions examined at nine normalized layer depths (10\%--90\% of model depth). Orange highlights indicate successful recovery of the target concepts.}
    \label{fig:qualitative-vlm-1}
    \vspace{-2.0mm}
\end{figure*}

\begin{figure*}[t]
    \centering
    \includegraphics[width=0.95\linewidth]{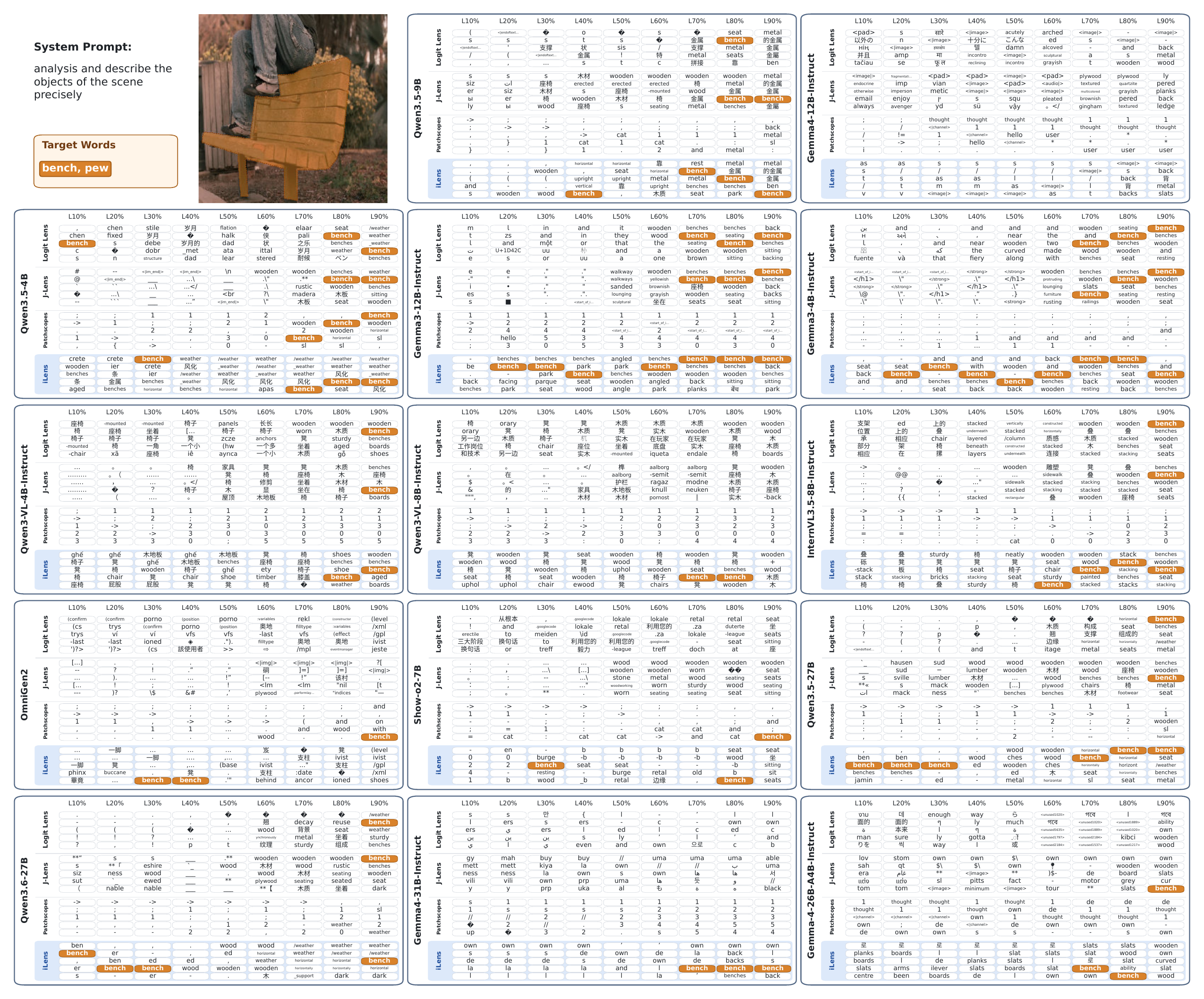}
    \caption{Qualitative example of the Visual-Decoding task across 14 vision-capable models, with predictions examined at nine normalized layer depths (10\%--90\% of model depth). Orange highlights indicate successful recovery of the target concepts.}
    \label{fig:qualitative-vlm-2}
    \vspace{-2.0mm}
\end{figure*}

\begin{figure*}[t]
    \centering
    \includegraphics[width=0.95\linewidth]{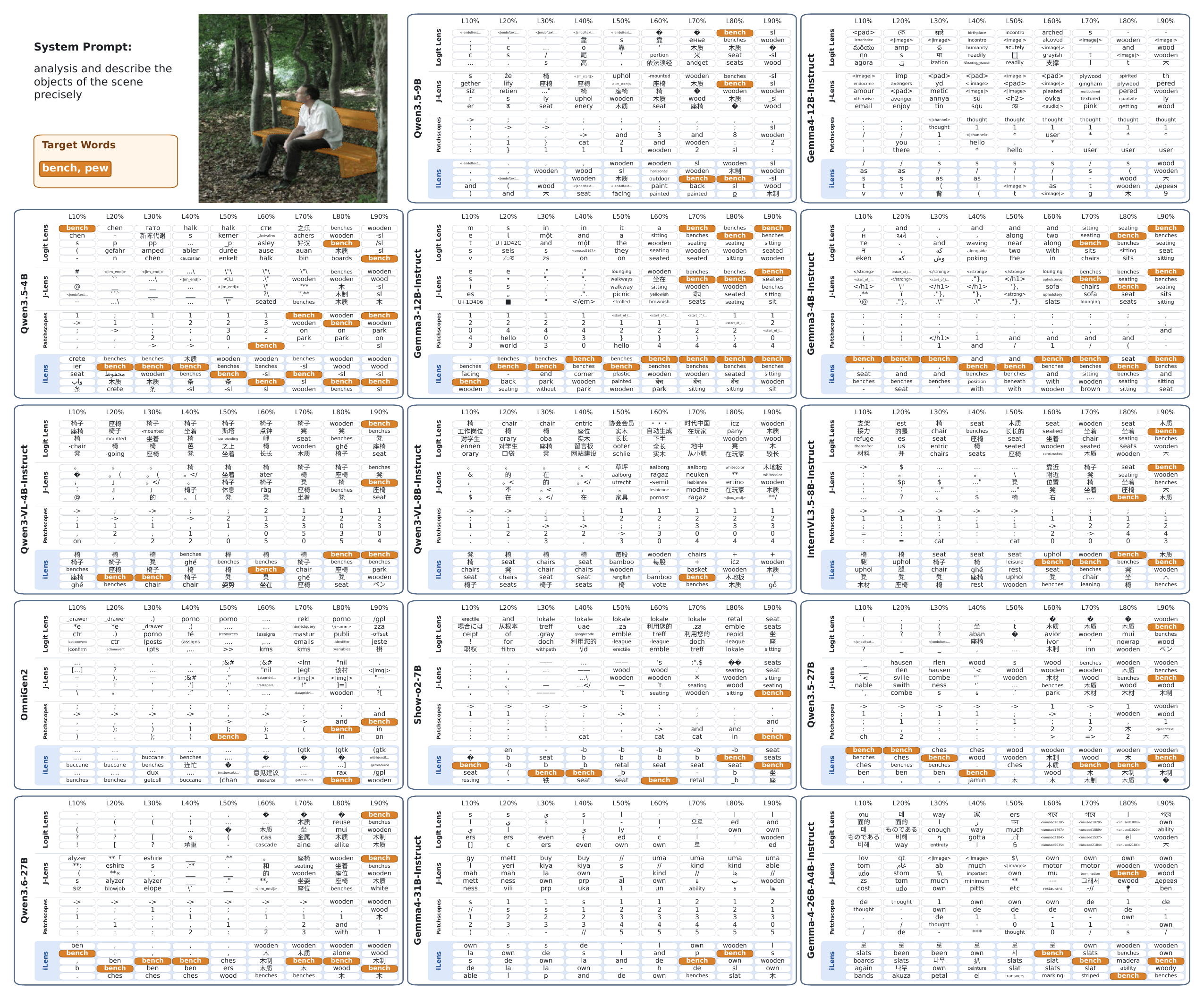}
    \caption{Qualitative example of the Visual-Decoding task across 14 vision-capable models, with predictions examined at nine normalized layer depths (10\%--90\% of model depth). Orange highlights indicate successful recovery of the target concepts.}
    \label{fig:qualitative-vlm-3}
    \vspace{-2.0mm}
\end{figure*}

\begin{figure*}[t]
    \centering
    \includegraphics[width=0.95\linewidth]{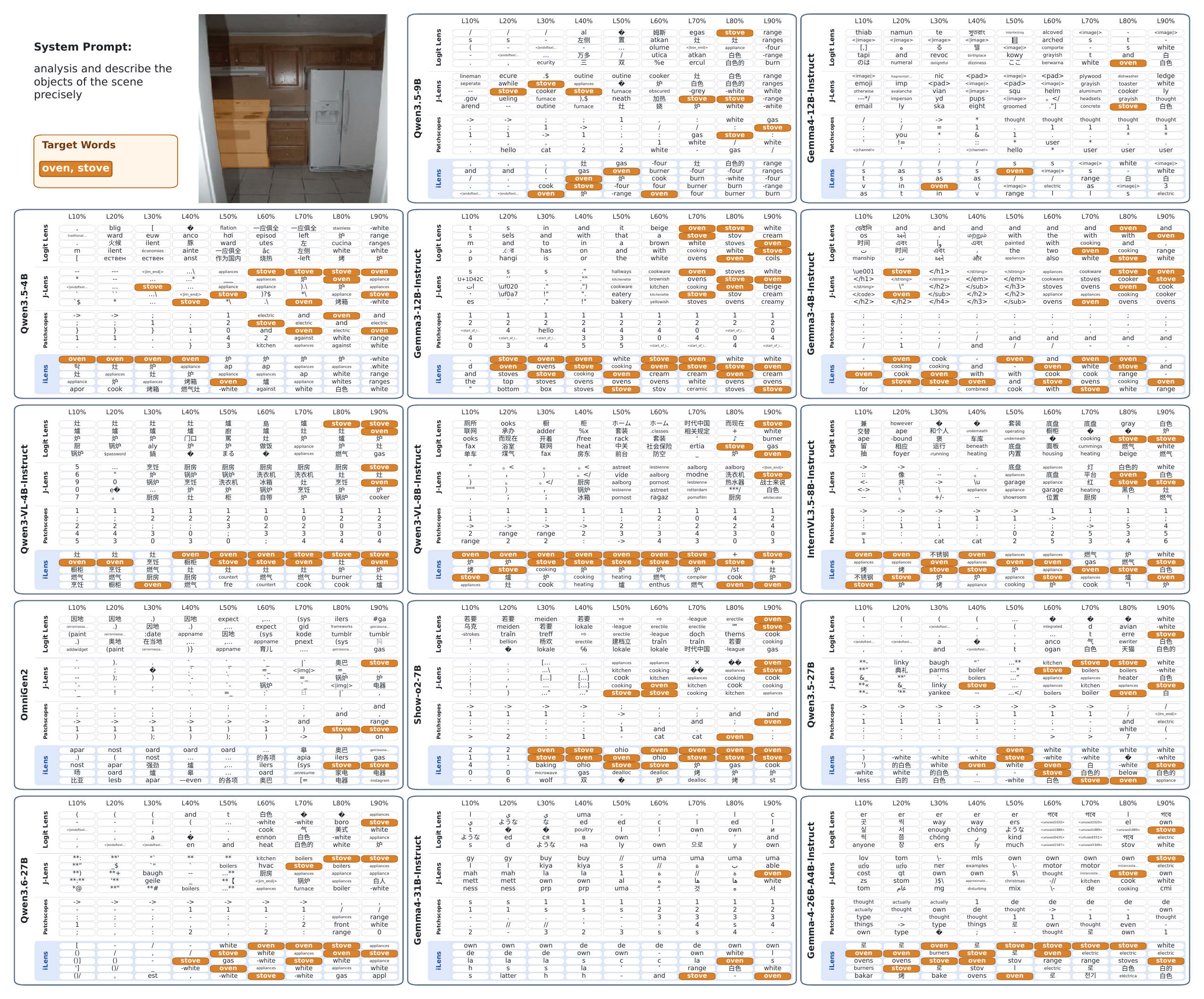}
    \caption{Qualitative example of the Visual-Decoding task across 14 vision-capable models, with predictions examined at nine normalized layer depths (10\%--90\% of model depth). Orange highlights indicate successful recovery of the target concepts.}
    \label{fig:qualitative-vlm-4}
    \vspace{-2.0mm}
\end{figure*}

\begin{figure*}[t]
    \centering
    \includegraphics[width=0.95\linewidth]{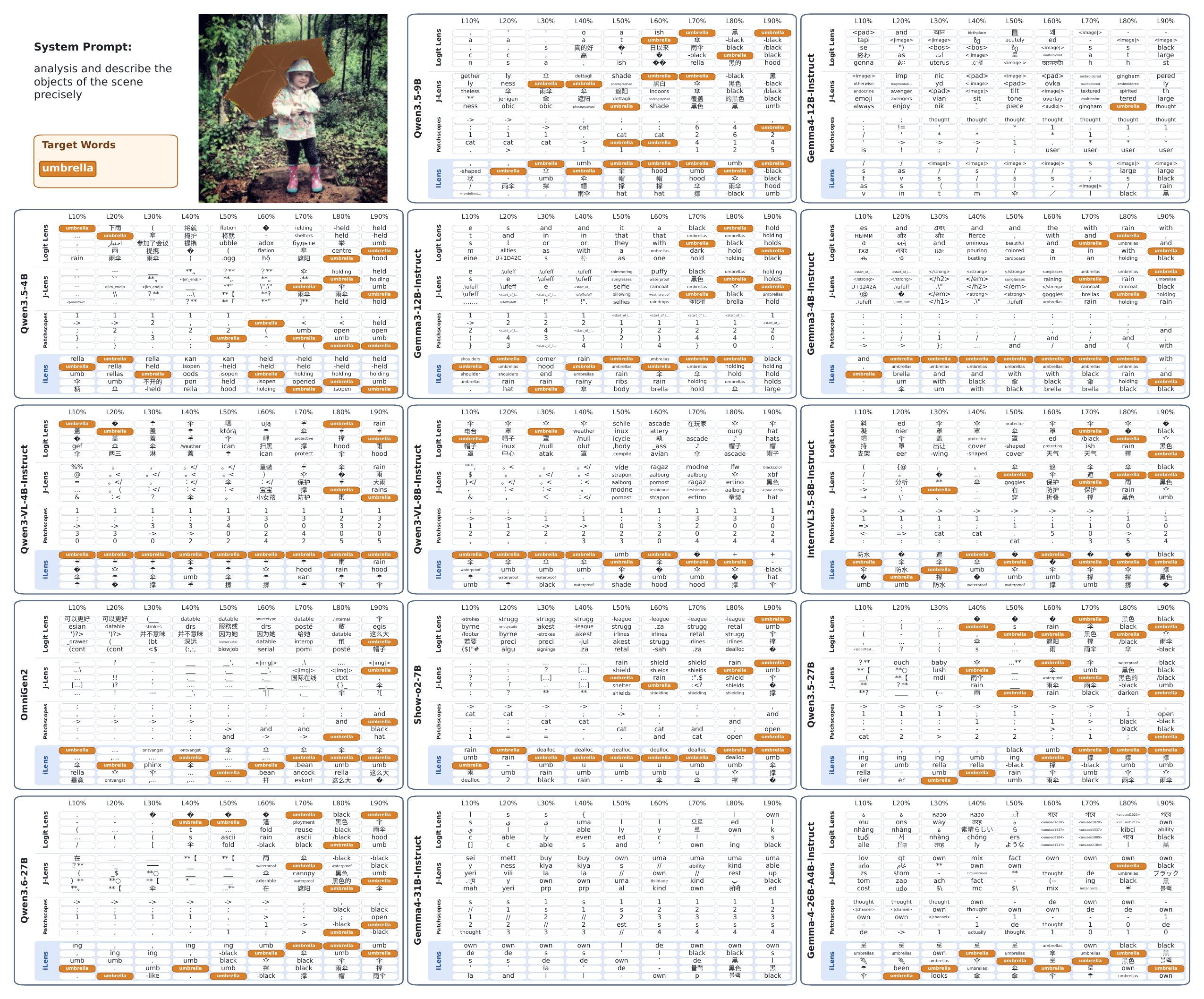}
    \caption{Qualitative example of the Visual-Decoding task across 14 vision-capable models, with predictions examined at nine normalized layer depths (10\%--90\% of model depth). Orange highlights indicate successful recovery of the target concepts.}
    \label{fig:qualitative-vlm-5}
    \vspace{-2.0mm}
\end{figure*}

\begin{figure*}[t]
    \centering
    \includegraphics[width=0.95\linewidth]{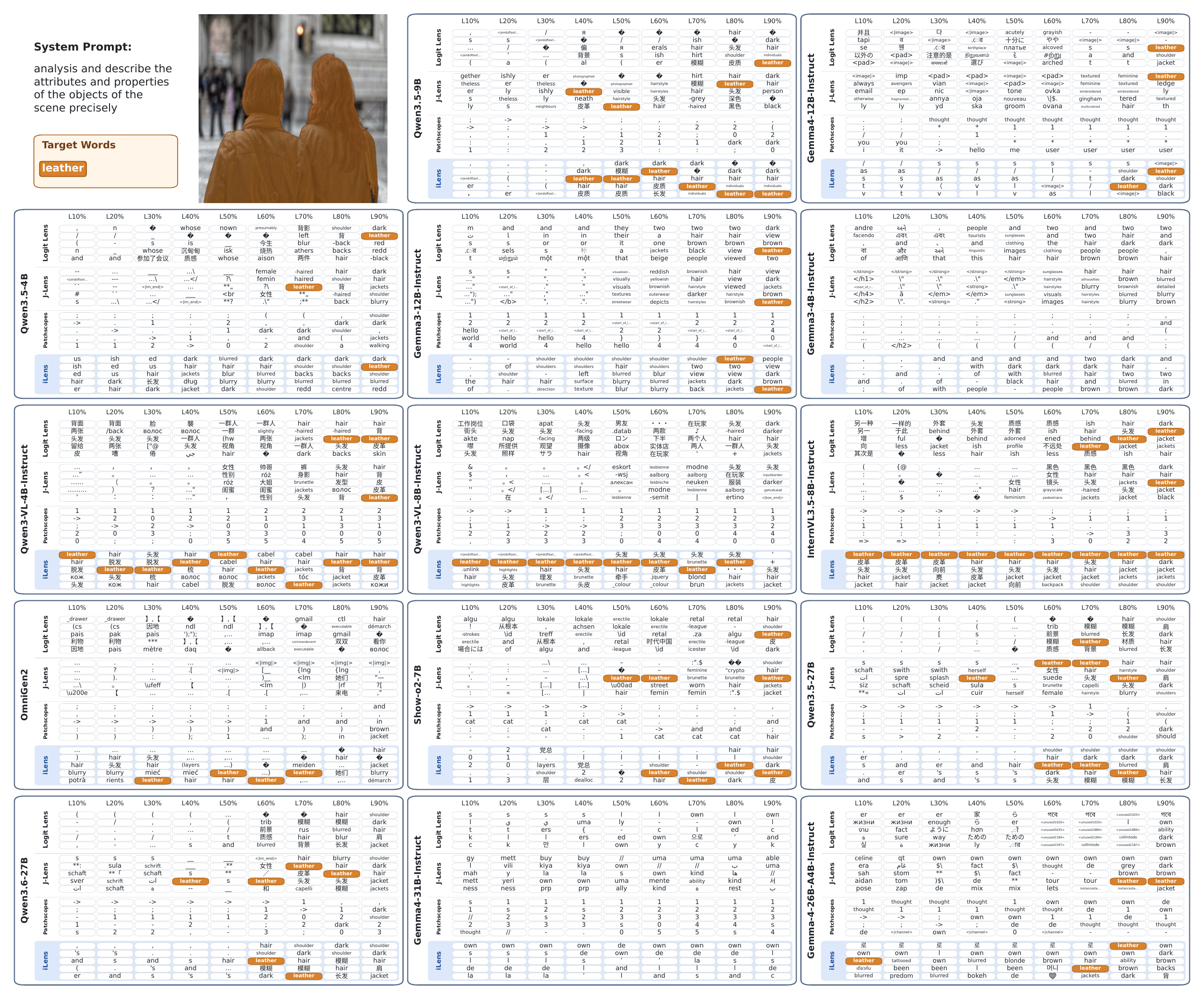}
    \caption{Qualitative example of the Visual-Decoding task across 14 vision-capable models, with predictions examined at nine normalized layer depths (10\%--90\% of model depth). Orange highlights indicate successful recovery of the target concepts.}
    \label{fig:qualitative-vlm-6}
    \vspace{-2.0mm}
\end{figure*}

\begin{figure*}[t]
    \centering
    \includegraphics[width=0.95\linewidth]{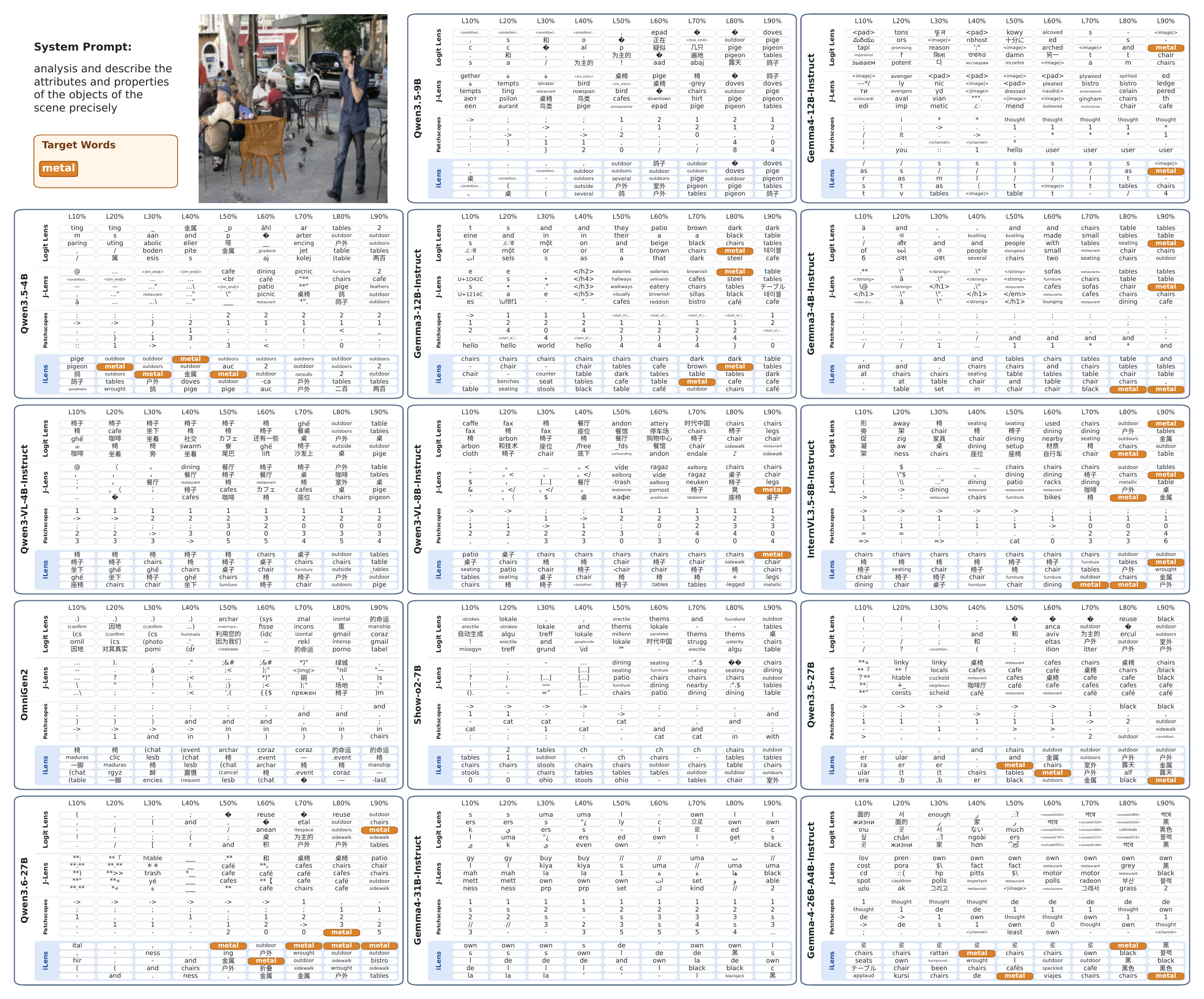}
    \caption{Qualitative example of the Visual-Decoding task across 14 vision-capable models, with predictions examined at nine normalized layer depths (10\%--90\% of model depth). Orange highlights indicate successful recovery of the target concepts.}
    \label{fig:qualitative-vlm-7}
    \vspace{-2.0mm}
\end{figure*}

\begin{figure*}[t]
    \centering
    \includegraphics[width=0.95\linewidth]{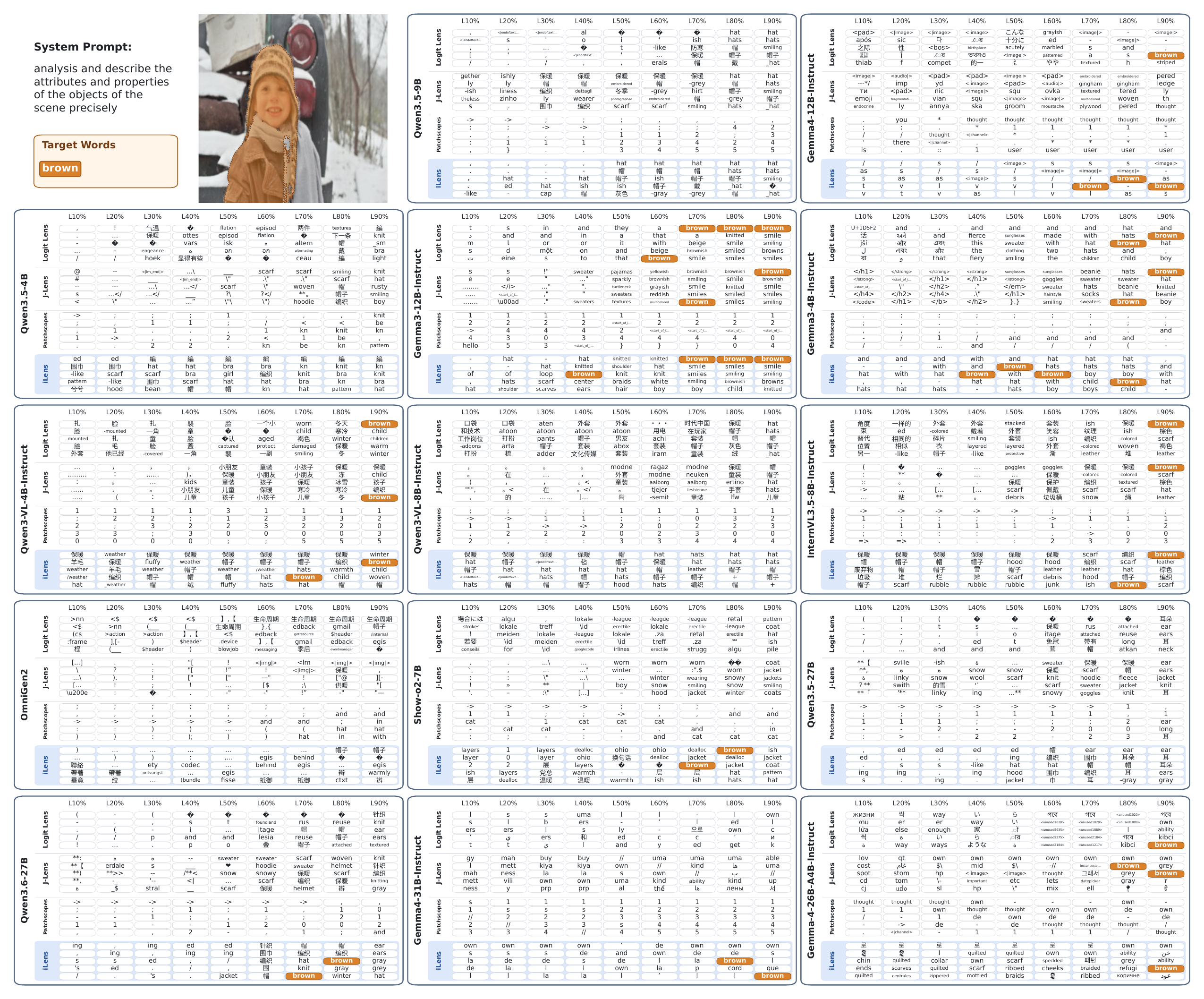}
    \caption{Qualitative example of the Visual-Decoding task across 14 vision-capable models, with predictions examined at nine normalized layer depths (10\%--90\% of model depth). Orange highlights indicate successful recovery of the target concepts.}
    \label{fig:qualitative-vlm-8}
    \vspace{-2.0mm}
\end{figure*}

\begin{figure*}[t]
    \centering
    \includegraphics[width=0.95\linewidth]{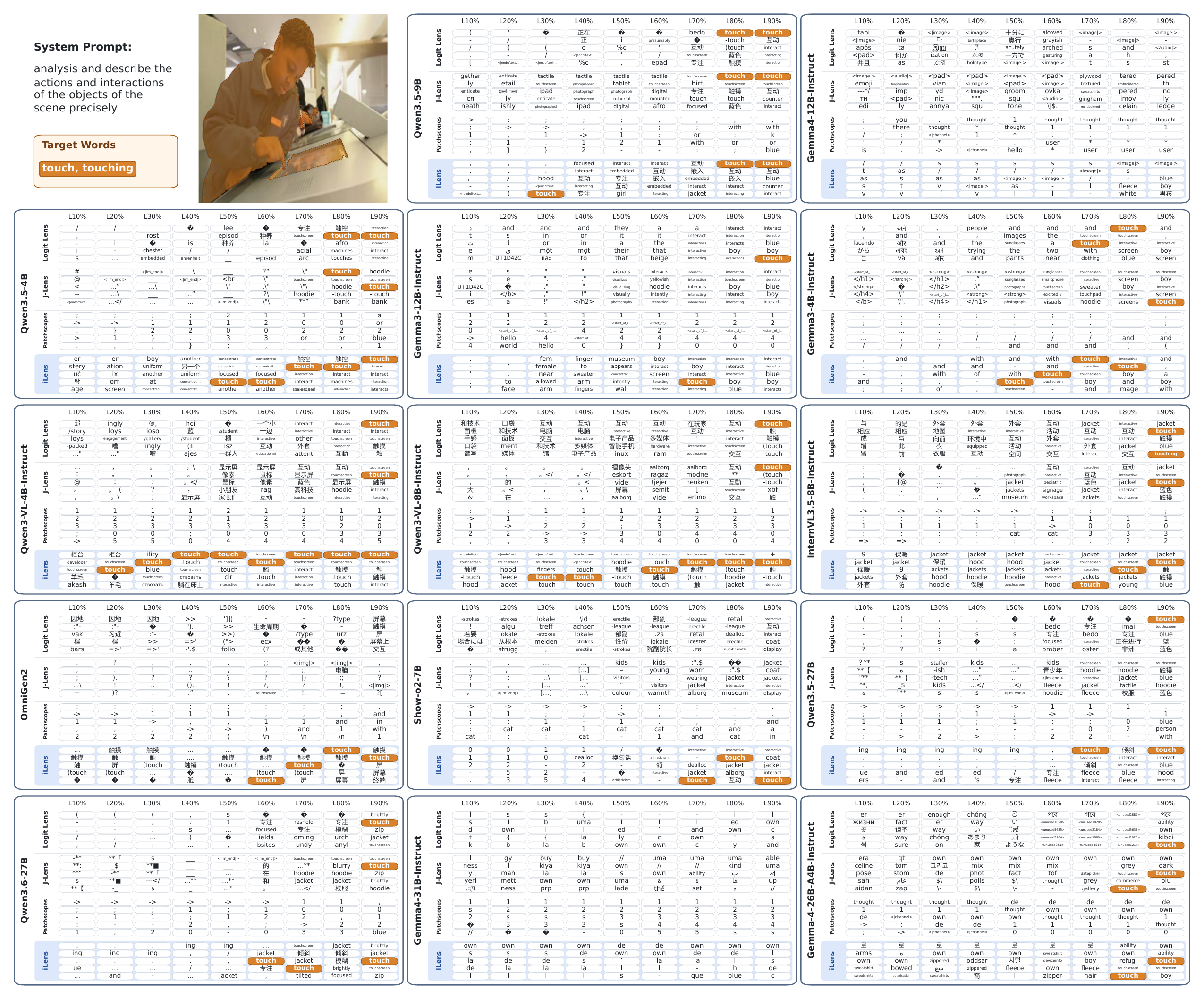}
    \caption{Qualitative example of the Visual-Decoding task across 14 vision-capable models, with predictions examined at nine normalized layer depths (10\%--90\% of model depth). Orange highlights indicate successful recovery of the target concepts.}
    \label{fig:qualitative-vlm-9}
    \vspace{-2.0mm}
\end{figure*}

\begin{figure*}[t]
    \centering
    \includegraphics[width=0.95\linewidth]{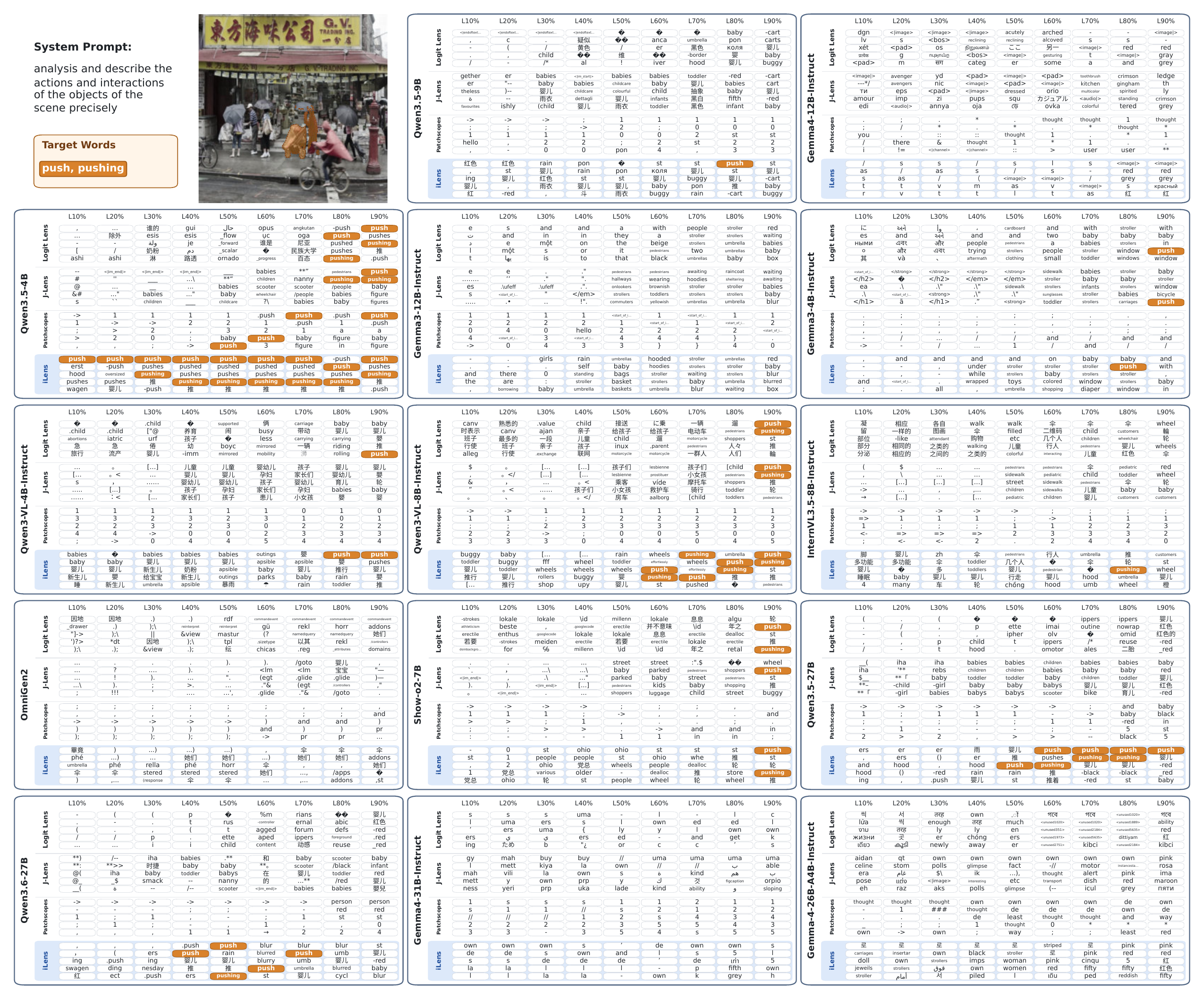}
    \caption{Qualitative example of the Visual-Decoding task across 14 vision-capable models, with predictions examined at nine normalized layer depths (10\%--90\% of model depth). Orange highlights indicate successful recovery of the target concepts.}
    \label{fig:qualitative-vlm-10}
    \vspace{-2.0mm}
\end{figure*}

\begin{figure*}[t]
    \centering
    \includegraphics[width=0.95\linewidth]{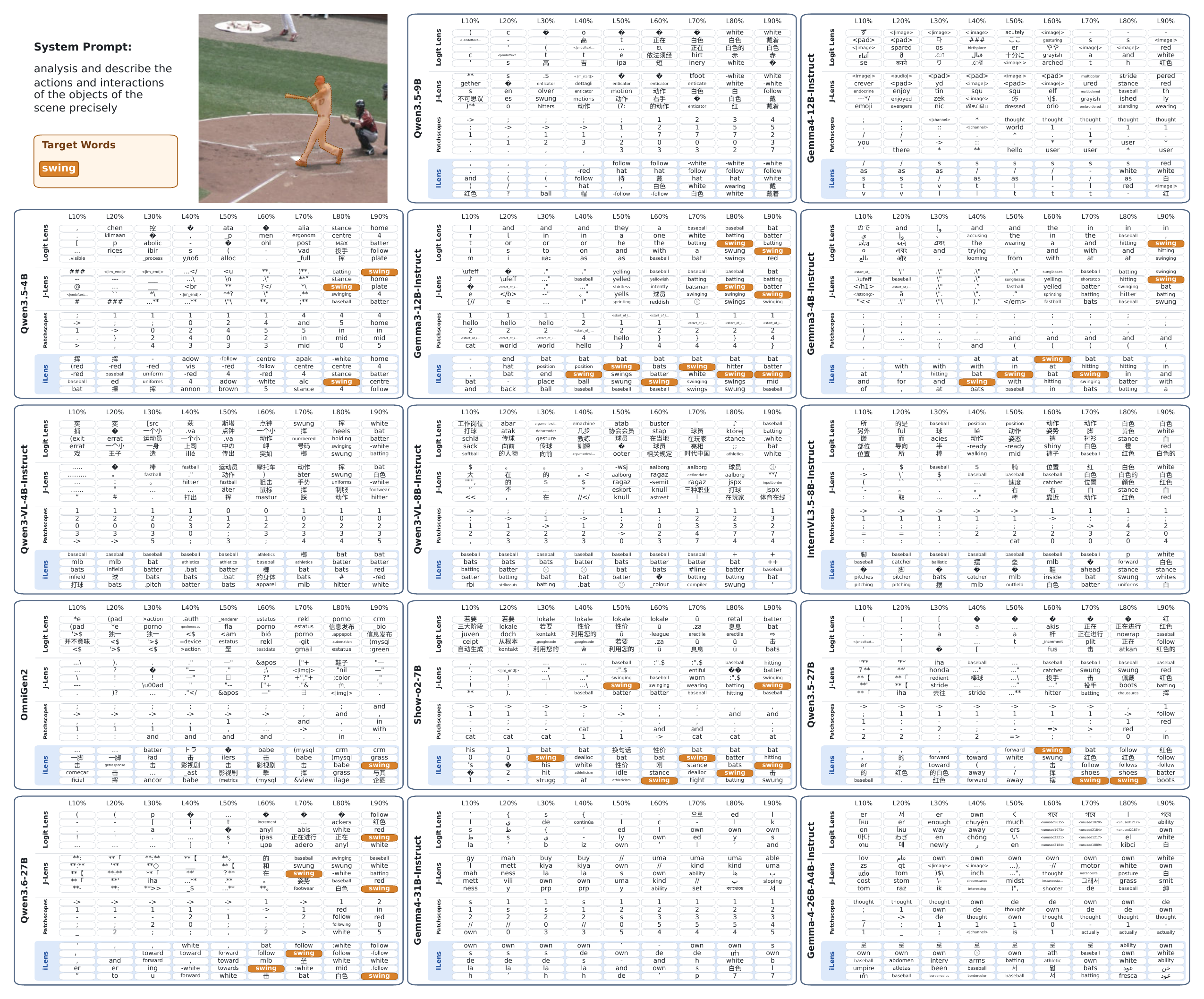}
    \caption{Qualitative example of the Visual-Decoding task across 14 vision-capable models, with predictions examined at nine normalized layer depths (10\%--90\% of model depth). Orange highlights indicate successful recovery of the target concepts.}
    \label{fig:qualitative-vlm-11}
    \vspace{-2.0mm}
\end{figure*}

\begin{figure*}[t]
    \centering
    \includegraphics[width=0.95\linewidth]{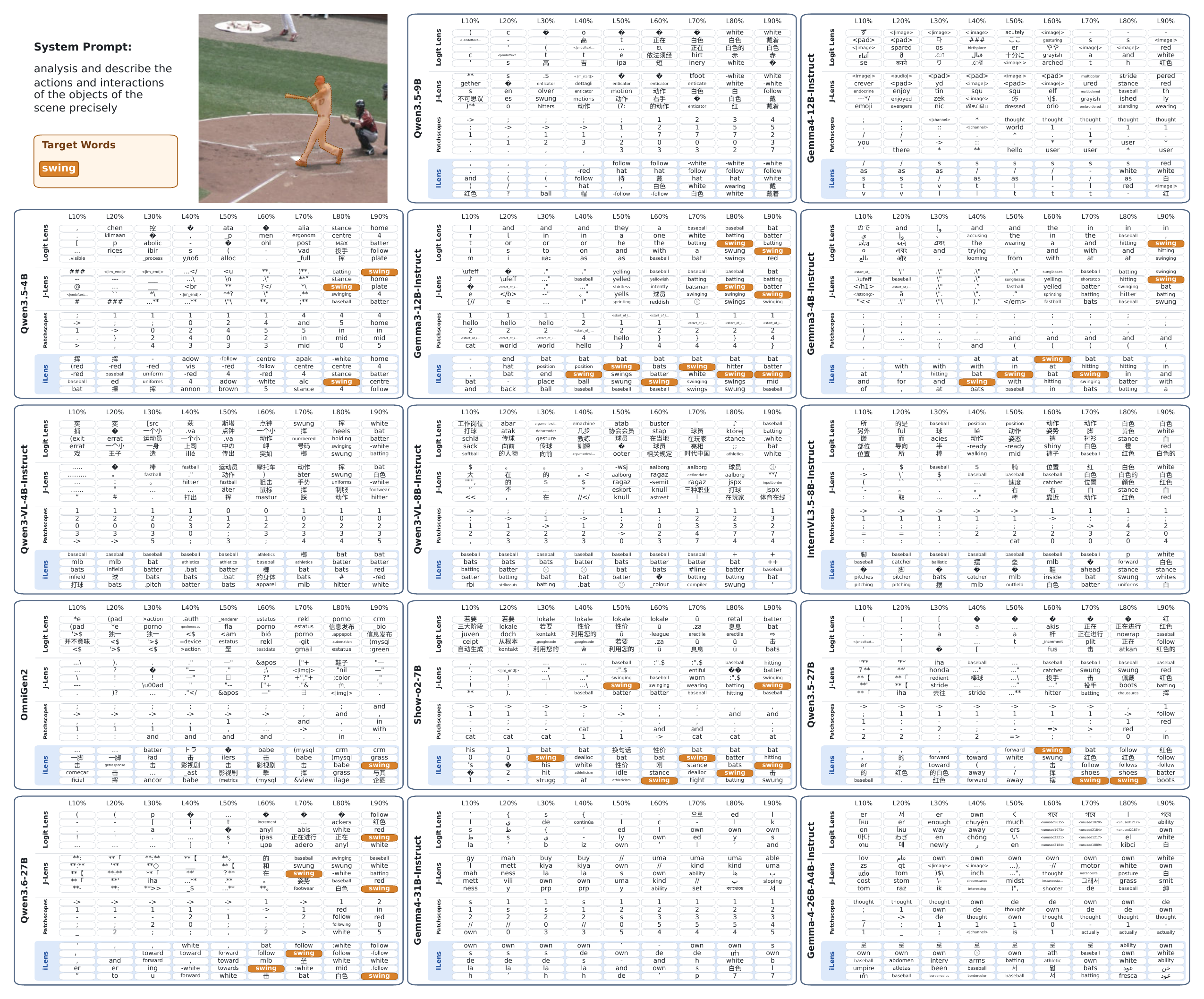}
    \caption{Qualitative example of the Visual-Decoding task across 14 vision-capable models, with predictions examined at nine normalized layer depths (10\%--90\% of model depth). Orange highlights indicate successful recovery of the target concepts.}
    \label{fig:qualitative-vlm-12}
    \vspace{-2.0mm}
\end{figure*}

\begin{figure*}[t]
    \centering
    \includegraphics[width=0.95\linewidth]{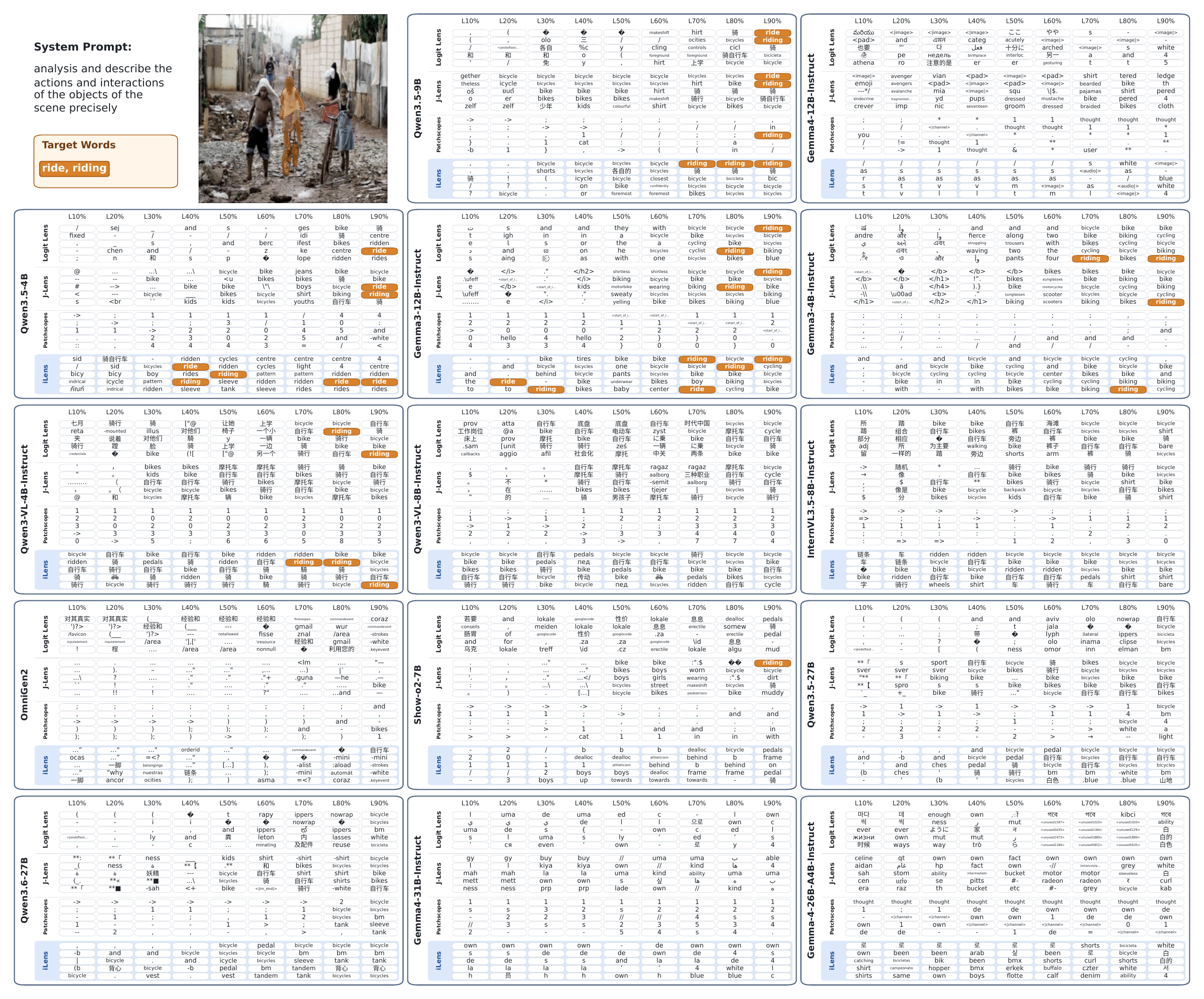}
    \caption{Qualitative example of the Visual-Decoding task across 14 vision-capable models, with predictions examined at nine normalized layer depths (10\%--90\% of model depth). Orange highlights indicate successful recovery of the target concepts.}
    \label{fig:qualitative-vlm-13}
    \vspace{-2.0mm}
\end{figure*}

\begin{figure*}[t]
    \centering
    \includegraphics[width=0.95\linewidth]{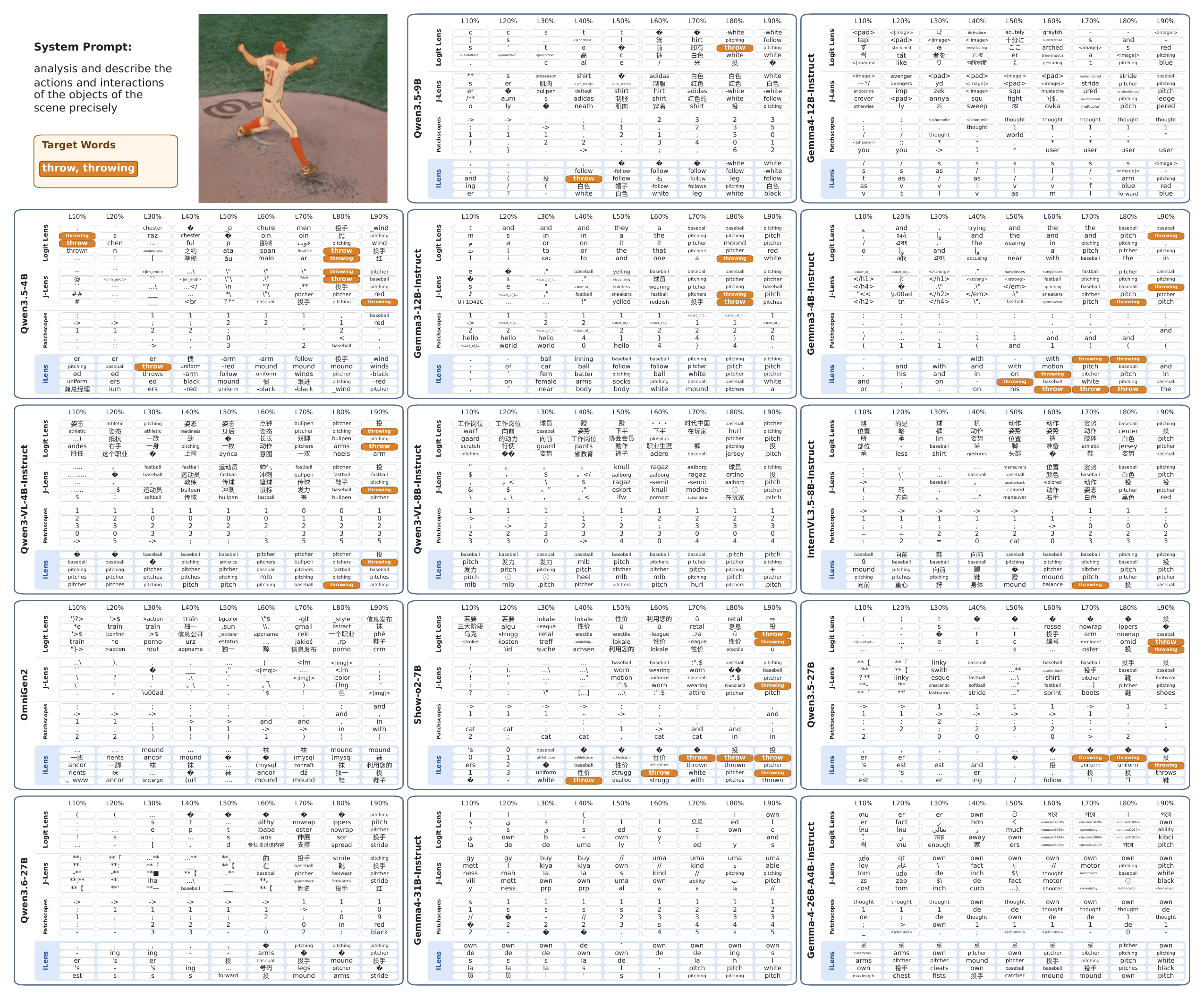}
    \caption{Qualitative example of the Visual-Decoding task across 14 vision-capable models, with predictions examined at nine normalized layer depths (10\%--90\% of model depth). Orange highlights indicate successful recovery of the target concepts.}
    \label{fig:qualitative-vlm-14}
    \vspace{-2.0mm}
\end{figure*}

\begin{figure*}[t]
    \centering
    \includegraphics[width=0.95\linewidth]{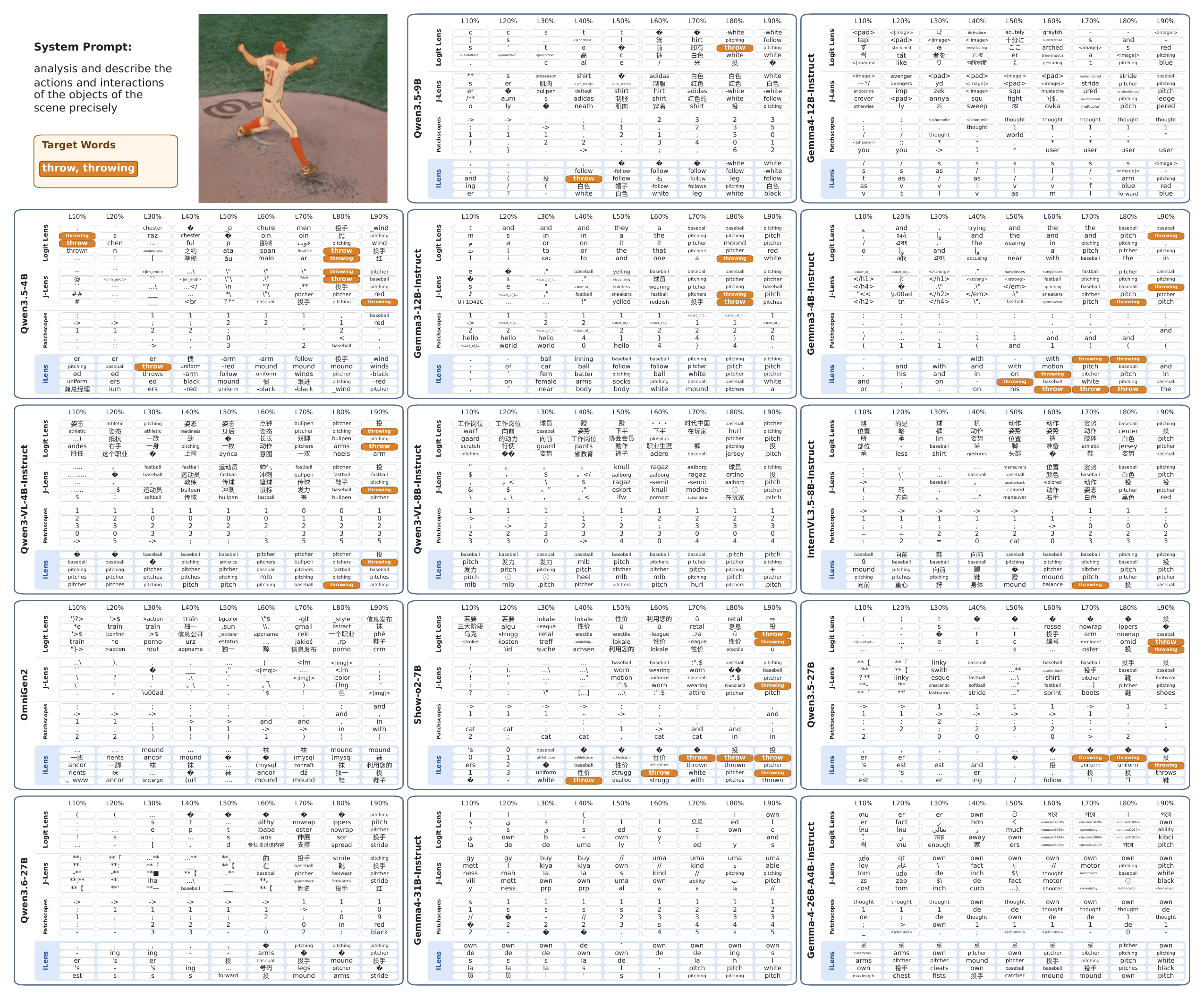}
    \caption{Qualitative example of the Visual-Decoding task across 14 vision-capable models, with predictions examined at nine normalized layer depths (10\%--90\% of model depth). Orange highlights indicate successful recovery of the target concepts.}
    \label{fig:qualitative-vlm-15}
    \vspace{-2.0mm}
\end{figure*}

\begin{figure*}[t]
    \centering
    \includegraphics[width=0.95\linewidth]{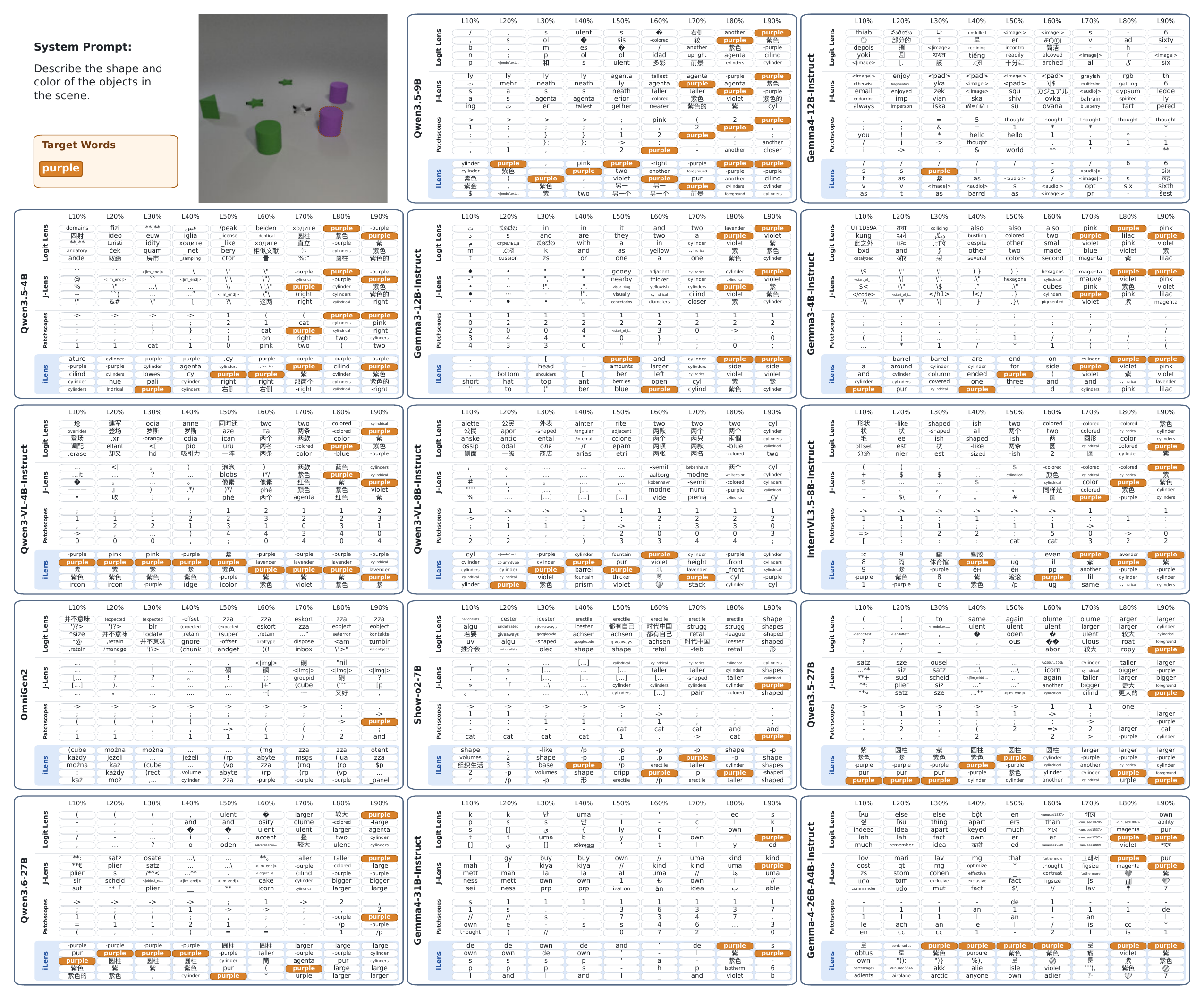}
    \caption{Qualitative example of the Visual-Decoding task across 14 vision-capable models, with predictions examined at nine normalized layer depths (10\%--90\% of model depth). Orange highlights indicate successful recovery of the target concepts.}
    \label{fig:qualitative-vlm-16}
    \vspace{-2.0mm}
\end{figure*}

\begin{figure*}[t]
    \centering
    \includegraphics[width=0.95\linewidth]{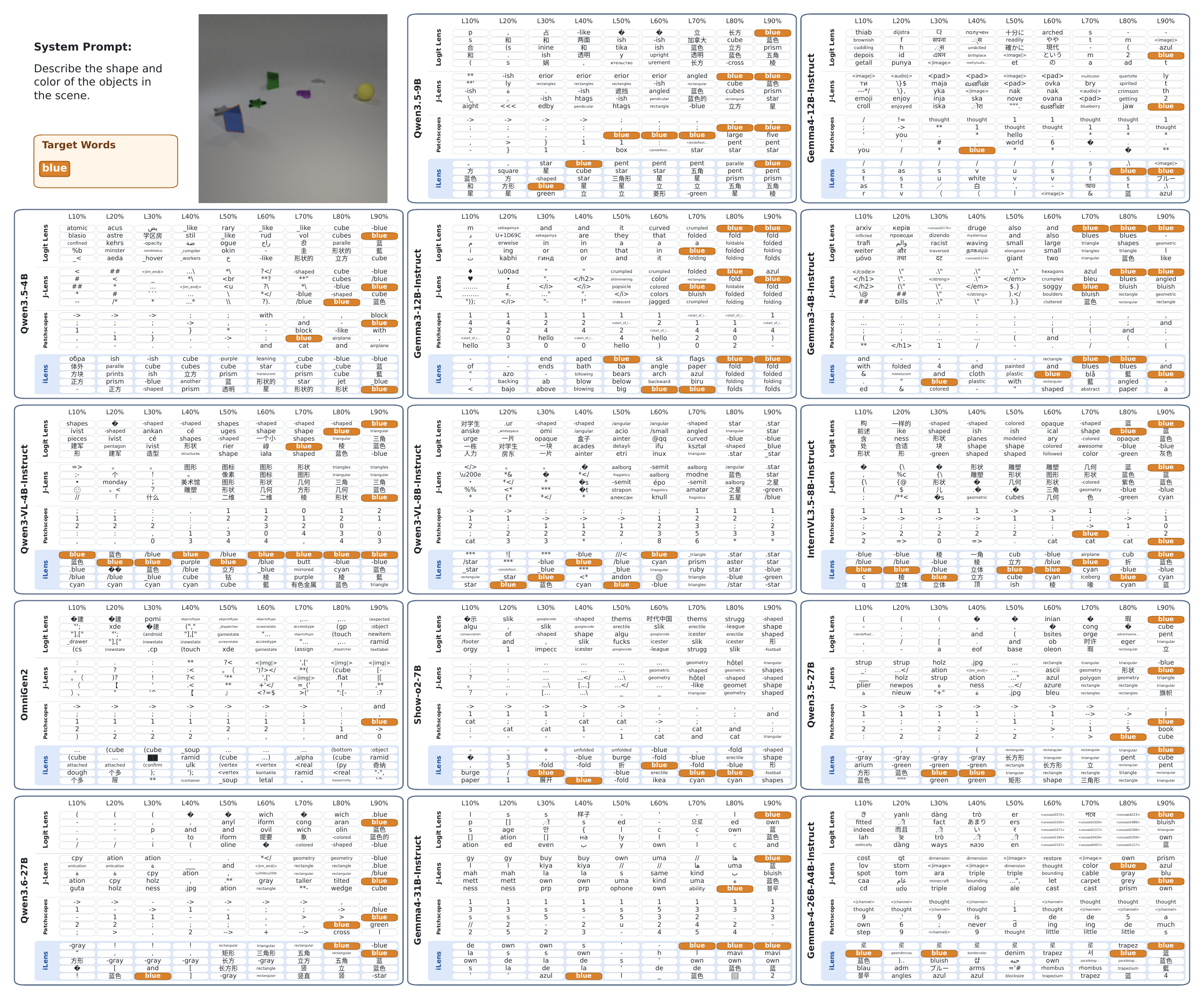}
    \caption{Qualitative example of the Visual-Decoding task across 14 vision-capable models, with predictions examined at nine normalized layer depths (10\%--90\% of model depth). Orange highlights indicate successful recovery of the target concepts.}
    \label{fig:qualitative-vlm-17}
    \vspace{-2.0mm}
\end{figure*}

\begin{figure*}[t]
    \centering
    \includegraphics[width=0.95\linewidth]{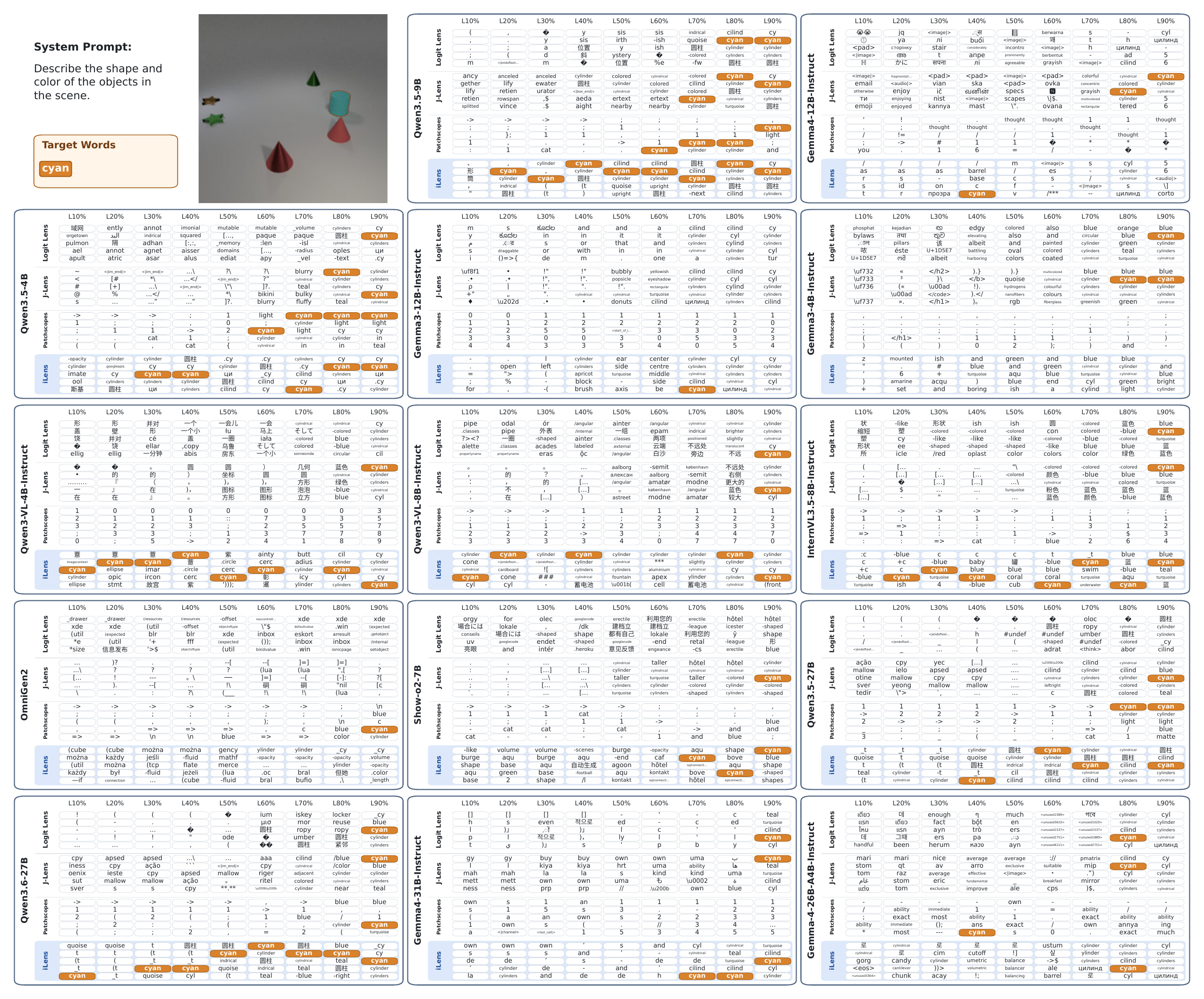}
    \caption{Qualitative example of the Visual-Decoding task across 14 vision-capable models, with predictions examined at nine normalized layer depths (10\%--90\% of model depth). Orange highlights indicate successful recovery of the target concepts.}
    \label{fig:qualitative-vlm-18}
    \vspace{-2.0mm}
\end{figure*}

\begin{figure*}[t]
    \centering
    \includegraphics[width=0.95\linewidth]{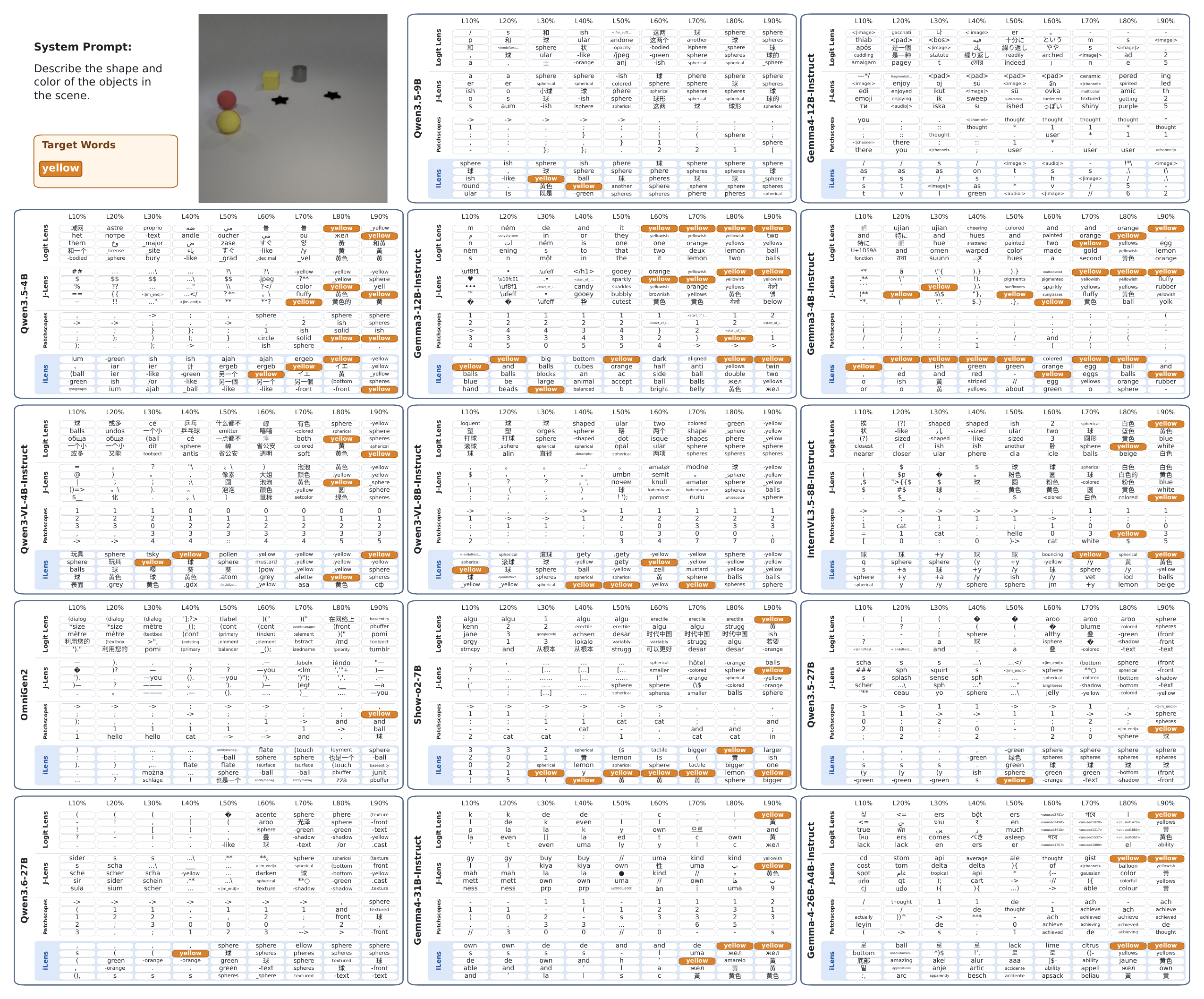}
    \caption{Qualitative example of the Visual-Decoding task across 14 vision-capable models, with predictions examined at nine normalized layer depths (10\%--90\% of model depth). Orange highlights indicate successful recovery of the target concepts.}
    \label{fig:qualitative-vlm-19}
    \vspace{-2.0mm}
\end{figure*}

\begin{figure*}[t]
    \centering
    \includegraphics[width=0.95\linewidth]{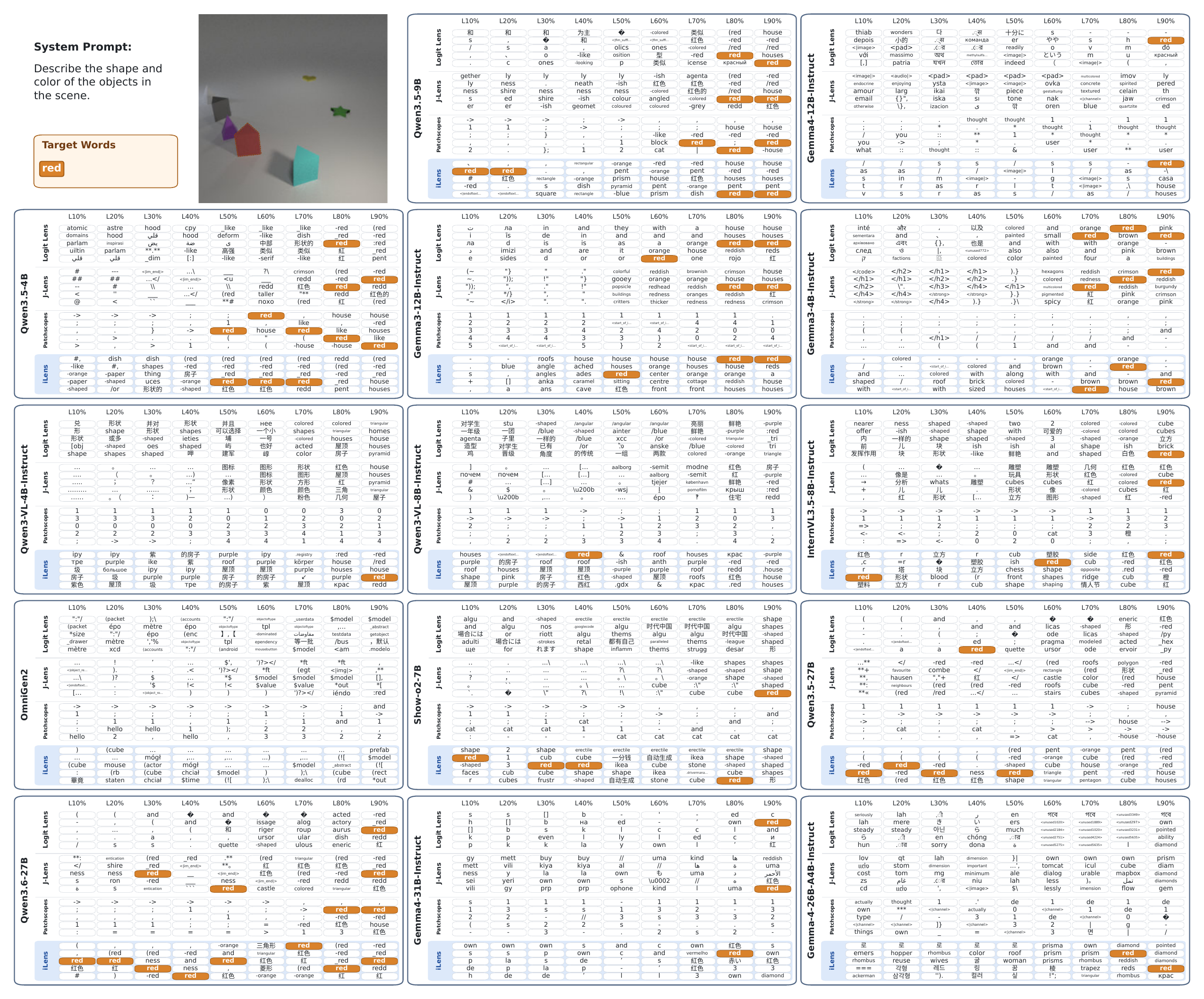}
    \caption{Qualitative example of the Visual-Decoding task across 14 vision-capable models, with predictions examined at nine normalized layer depths (10\%--90\% of model depth). Orange highlights indicate successful recovery of the target concepts.}
    \label{fig:qualitative-vlm-20}
    \vspace{-2.0mm}
\end{figure*}

\begin{figure*}[t]
    \centering
    \includegraphics[width=0.95\linewidth]{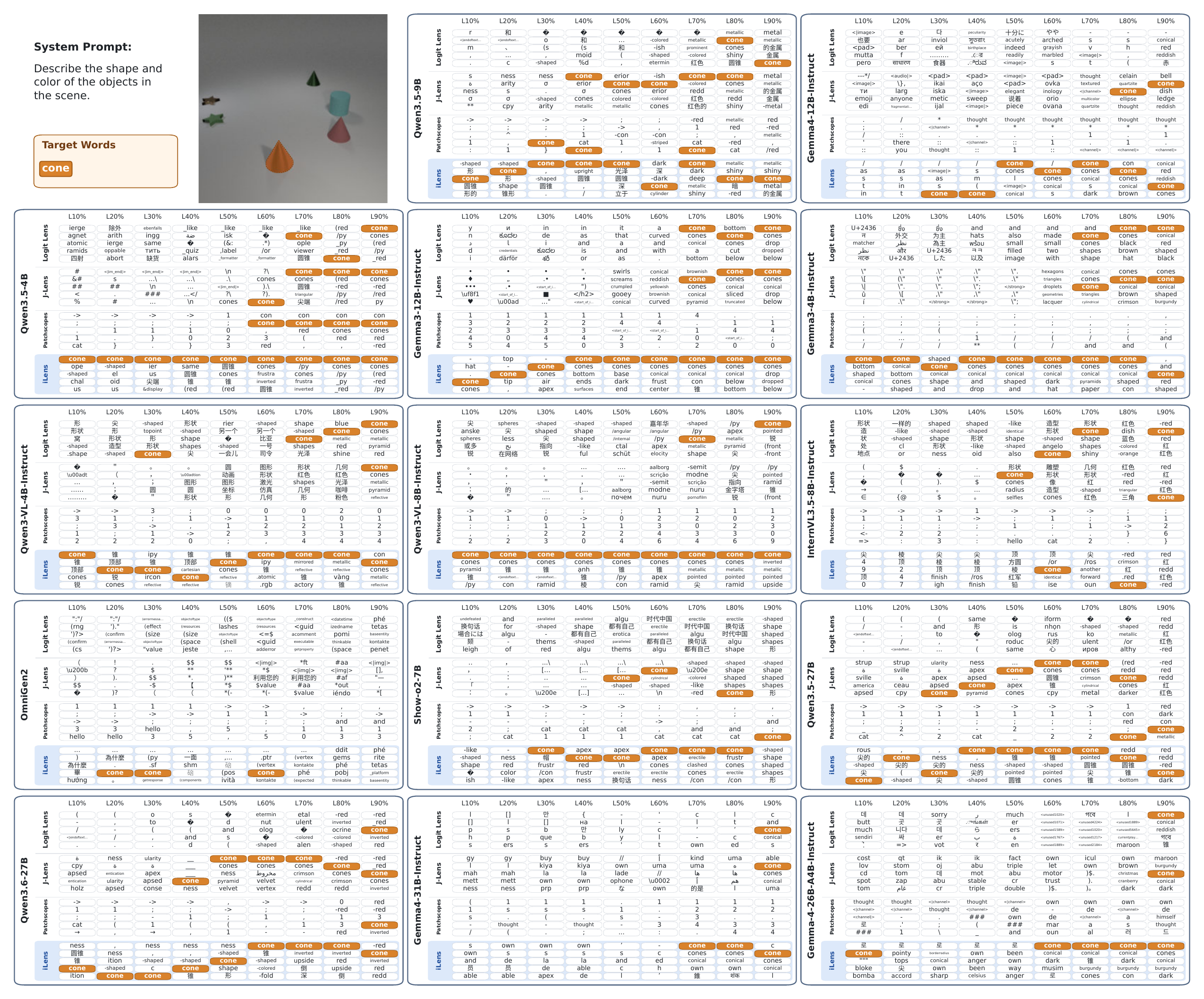}
    \caption{Qualitative example of the Visual-Decoding task across 14 vision-capable models, with predictions examined at nine normalized layer depths (10\%--90\% of model depth). Orange highlights indicate successful recovery of the target concepts.}
    \label{fig:qualitative-vlm-21}
    \vspace{-2.0mm}
\end{figure*}

\begin{figure*}[t]
    \centering
    \includegraphics[width=0.95\linewidth]{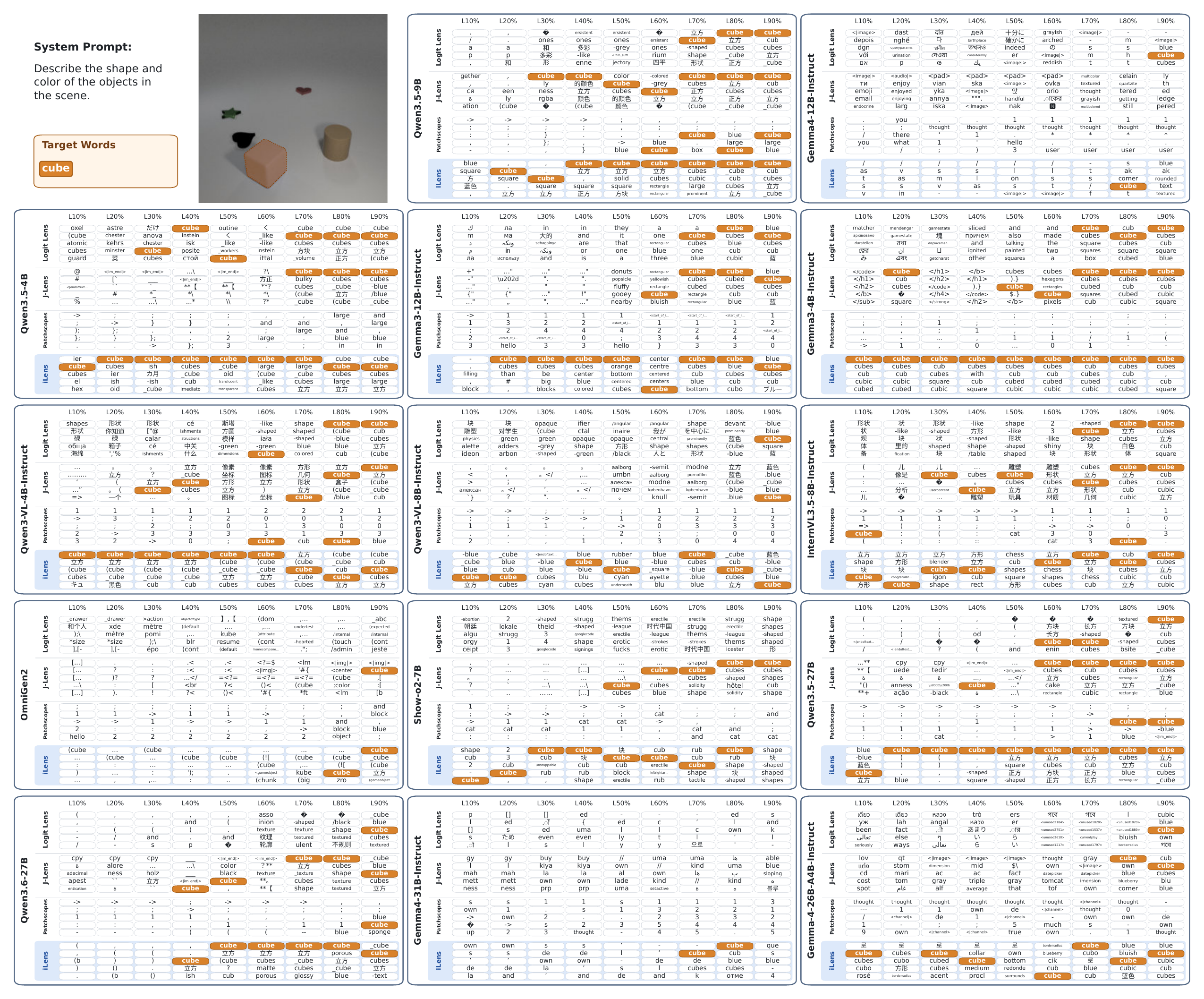}
    \caption{Qualitative example of the Visual-Decoding task across 14 vision-capable models, with predictions examined at nine normalized layer depths (10\%--90\% of model depth). Orange highlights indicate successful recovery of the target concepts.}
    \label{fig:qualitative-vlm-22}
    \vspace{-2.0mm}
\end{figure*}

\begin{figure*}[t]
    \centering
    \includegraphics[width=0.95\linewidth]{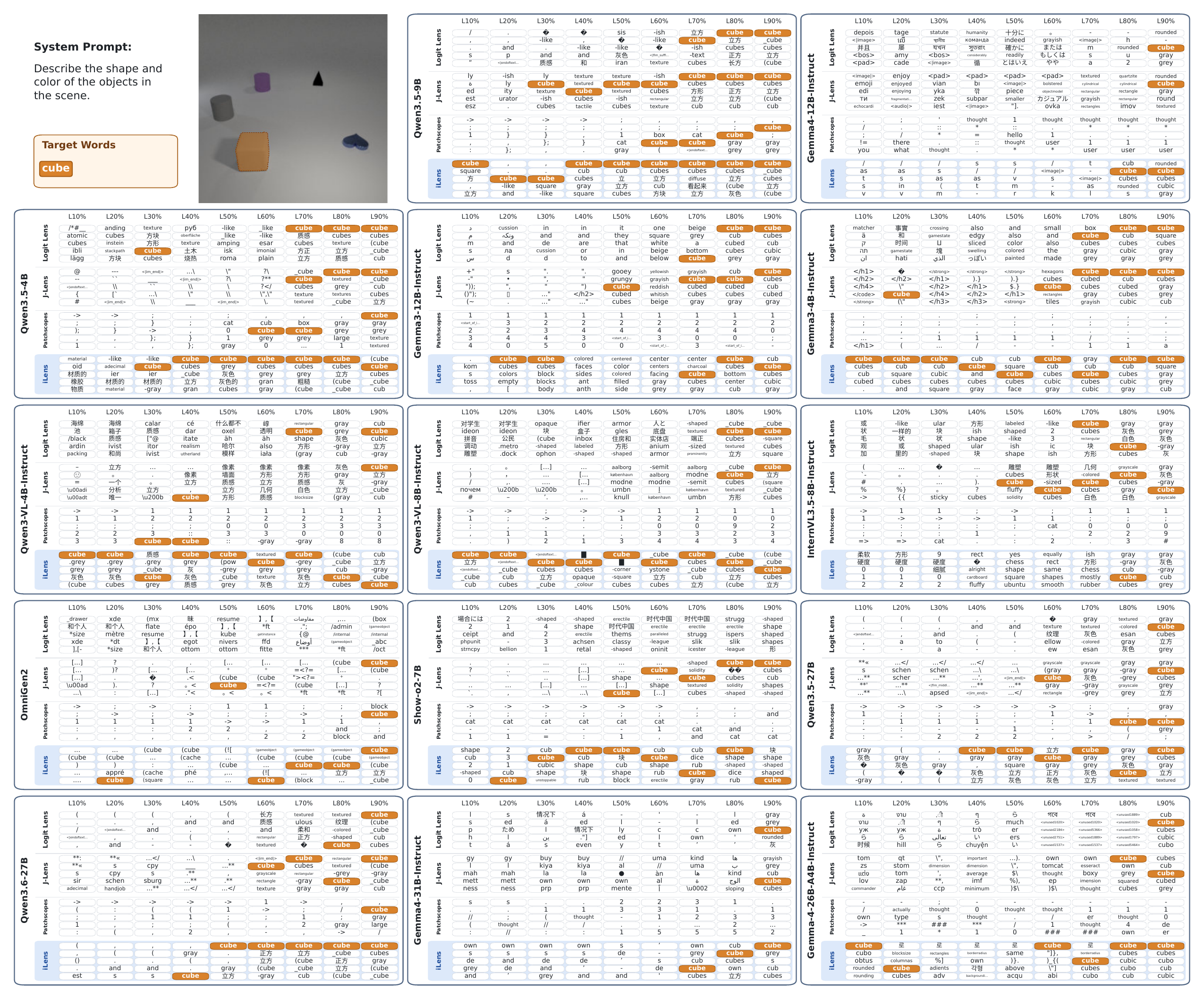}
    \caption{Qualitative example of the Visual-Decoding task across 14 vision-capable models, with predictions examined at nine normalized layer depths (10\%--90\% of model depth). Orange highlights indicate successful recovery of the target concepts.}
    \label{fig:qualitative-vlm-23}
    \vspace{-2.0mm}
\end{figure*}

\begin{figure*}[t]
    \centering
    \includegraphics[width=0.95\linewidth]{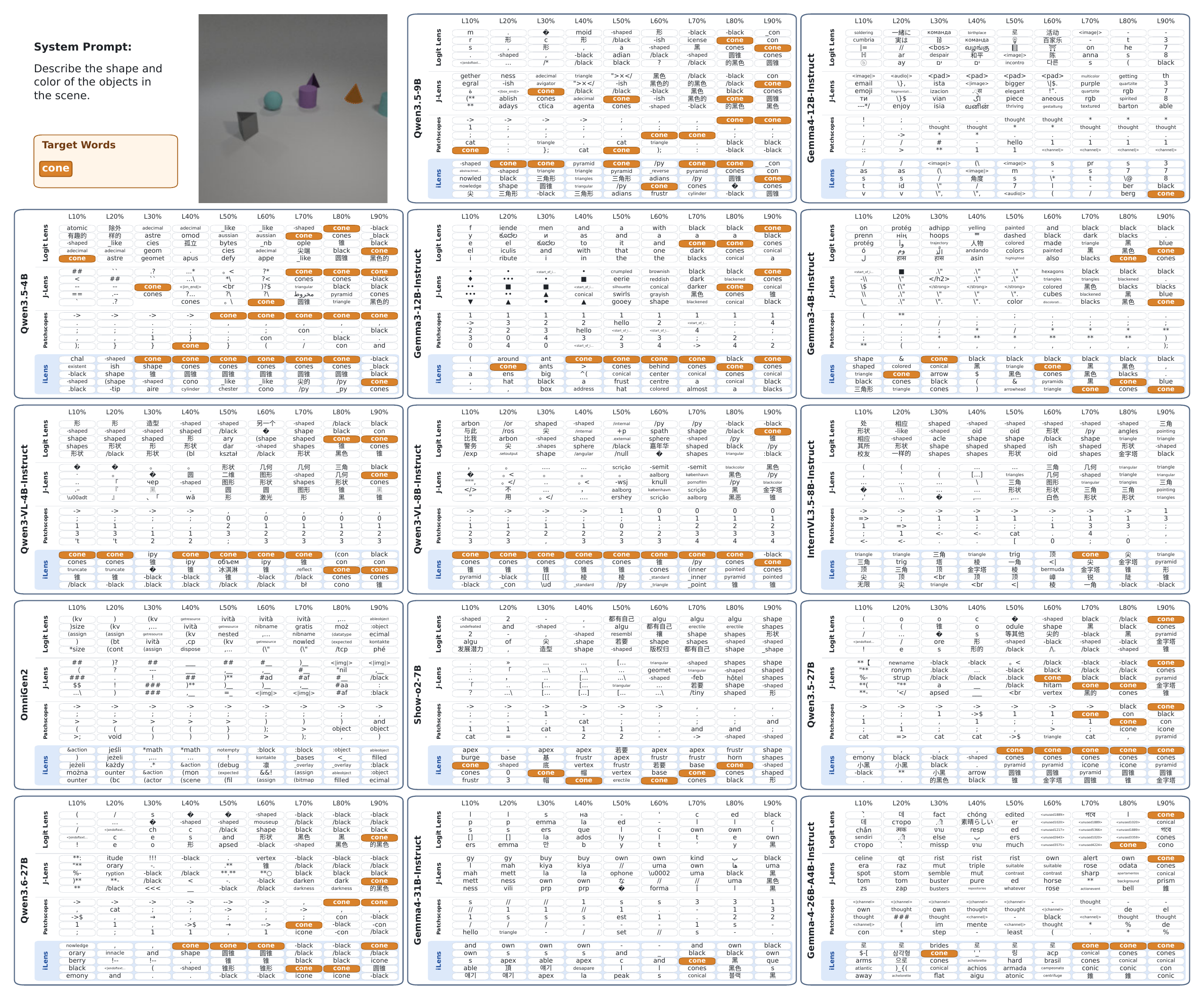}
    \caption{Qualitative example of the Visual-Decoding task across 14 vision-capable models, with predictions examined at nine normalized layer depths (10\%--90\% of model depth). Orange highlights indicate successful recovery of the target concepts.}
    \label{fig:qualitative-vlm-24}
    \vspace{-2.0mm}
\end{figure*}

\begin{figure*}[t]
    \centering
    \includegraphics[width=0.95\linewidth]{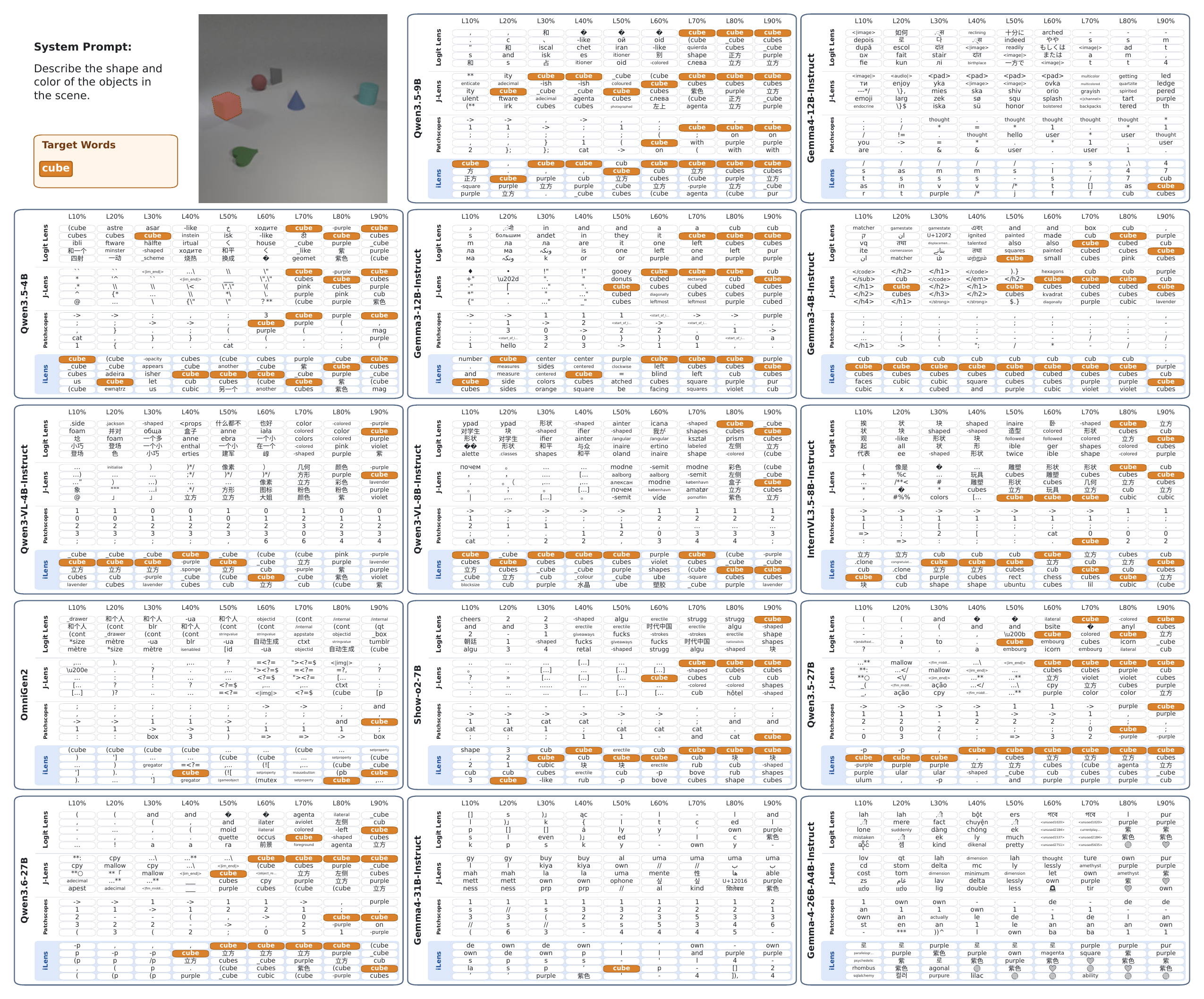}
    \caption{Qualitative example of the Visual-Decoding task across 14 vision-capable models, with predictions examined at nine normalized layer depths (10\%--90\% of model depth). Orange highlights indicate successful recovery of the target concepts.}
    \label{fig:qualitative-vlm-25}
    \vspace{-2.0mm}
\end{figure*}

We provide additional qualitative examples for all five
visual decoding tasks in Figures~\ref{fig:qualitative-vlm-1}--\ref{fig:qualitative-vlm-25},
including both successful predictions and failure cases.
As illustrated, iLens not only successfully recovers target
concepts but also consistently recovers them at earlier
layers than the baseline methods. This behavior is consistent with
the lower FED values observed in the quantitative results.

\providecommand{\se}[1]{$_{\textcolor{gray}{\scalebox{0.65}{$\scriptstyle\scalebox{0.6}{$\scriptstyle\pm$}#1$}}}$}

\begin{table*}[t]
\centering
\caption{Comparison of iLens with existing decoding methods across five language-model probing tasks: Typo, Multilingual, Multihop, Order of Operations, and Dual Task. Results are reported using Pass@10 (P@10 $\uparrow$), First Emergence Depth (FED $\downarrow$), and Confidence (Conf. $\uparrow$), together with their averages across tasks. The best results within each model and task are shown in bold.}
\label{tab:appendix-llm-full-results}

\begingroup
\fontsize{6}{7.2}\selectfont
\setlength{\tabcolsep}{1.2pt}
\renewcommand{\arraystretch}{1.05}
\setlength{\aboverulesep}{1.2pt}
\setlength{\belowrulesep}{1.2pt}

\resizebox{\textwidth}{!}{%
\begin{tabular}{@{}ll*{5}{rrrr}|rrrr@{}}
\toprule[0.9pt]
\multirow{2}{*}{Model}
& \multirow{2}{*}{Method}
& \multicolumn{4}{c}{Typo}
& \multicolumn{4}{c}{Multilingual}
& \multicolumn{4}{c}{Multihop}
& \multicolumn{4}{c}{Order of Operations}
& \multicolumn{4}{c}{Dual Task}
& \multicolumn{4}{|c}{Average} \\

\cmidrule[0.55pt](lr){3-26}

& & P@10 $\uparrow$ & P@20 $\uparrow$ & FED $\downarrow$ & Conf. $\uparrow$
& P@10 $\uparrow$ & P@20 $\uparrow$ & FED $\downarrow$ & Conf. $\uparrow$
& P@10 $\uparrow$ & P@20 $\uparrow$ & FED $\downarrow$ & Conf. $\uparrow$
& P@10 $\uparrow$ & P@20 $\uparrow$ & FED $\downarrow$ & Conf. $\uparrow$
& P@10 $\uparrow$ & P@20 $\uparrow$ & FED $\downarrow$ & Conf. $\uparrow$
& P@10 $\uparrow$ & P@20 $\uparrow$ & FED $\downarrow$ & Conf. $\uparrow$ \\

\midrule[0.65pt]

\multirow{5}{*}{Qwen3.5-9B~\cite{qwen3.5}}
& Logit-Lens
& 13.1\se{2.2} & 16.8\se{2.6} & 44.4\se{6.1} & 0.4\se{0.1}
& 20.4\se{1.5} & 24.2\se{1.7} & 74.2\se{2.1} & 4.6\se{0.7}
& 11.3\se{2.4} & 15.5\se{2.8} & 80.7\se{3.7} & 1.1\se{0.3}
& 30.3\se{2.6} & 41.8\se{2.8} & \textbf{32.8}\se{4.5} & 1.2\se{0.2}
& 6.1\se{1.1} & 10.8\se{1.7} & 86.4\se{2.8} & 0.4\se{0.1}
& 16.2\se{0.9} & 21.8\se{1.1} & 63.7\se{1.8} & 1.5\se{0.2} \\

& Emb-Lens
& \textbf{82.1}\se{3.2} & \textbf{84.7}\se{3.0} & \textbf{9.6}\se{1.9} & --
& 2.4\se{1.0} & 4.3\se{1.6} & 89.9\se{3.7} & --
& 3.1\se{1.5} & 6.2\se{2.1} & 90.2\se{3.2} & --
& 0.0\se{0.0} & 0.2\se{0.1} & 100.0\se{0.0} & --
& 7.2\se{1.9} & 14.7\se{2.9} & 86.0\se{2.7} & --
& 19.0\se{0.8} & 22.0\se{1.0} & 75.1\se{1.2} & -- \\

& J-Lens
& 18.3\se{3.2} & 25.2\se{3.4} & 64.2\se{4.3} & \textbf{2.3}\se{0.7}
& \textbf{33.7}\se{3.0} & 38.6\se{3.0} & 62.1\se{2.7} & \textbf{9.8}\se{0.9}
& 24.5\se{4.2} & 32.0\se{4.4} & 67.8\se{4.7} & 4.3\se{1.0}
& 24.0\se{2.0} & 33.2\se{2.3} & 56.1\se{3.1} & 1.4\se{0.3}
& 15.4\se{2.8} & 20.6\se{3.1} & 78.4\se{3.6} & \textbf{1.9}\se{0.6}
& 23.2\se{1.4} & 29.9\se{1.5} & 65.7\se{1.7} & 3.9\se{0.3} \\
\end{tabular}%
}
\endgroup
\end{table*}

\subsection{Additional Textual Concept Decoding Experiments}
\label{app:subsec:llm_results}

We extend the textual concept-decoding evaluation to 18 LLMs
covering diverse model families, parameter scales, and
architectural designs. The evaluation considers five textual
benchmarks: Typo, Multilingual, Multihop, Order of Operations,
and Dual Task.

These experiments allow us to examine whether the improvements
observed in the main comparisons generalize across different
architectures and types of textual concepts. Given the specialized nature of these benchmarks, which focus on specific and well-defined concepts and capabilities, we use a fixed set of 50 samples per task. All methods are evaluated using the same samples, ensuring consistent and fair comparisons across decoding methods.

\paragraph{Quantitative Results.}

Table~\ref{tab:appendix-llm-full-results} presents the
comprehensive quantitative results across all five textual
decoding tasks and 18 LLMs. We report Pass@10, Pass@20,
Confidence, and First Emergence Depth (FED), providing
complementary measures of concept recovery, probability
assignment, and the depth at which target concepts first
become recoverable.

Taking the average across all 18 LLMs, iLens achieves
24.5\% Pass@10 and 32.8\% Pass@20, compared with
16.8\% and 21.8\% for J-Lens, respectively.
Furthermore, iLens achieves an average FED of 54.2\%,
compared with 70.2\% for Logit Lens, indicating
earlier recovery of textual concepts across model
depth. For Confidence, iLens achieves an average
of 4.4\%, slightly outperforming J-Lens (4.0\%)
and Logit Lens (2.8\%). These results highlight
the effectiveness of iLens in recovering textual
concepts across different language models.

\begin{figure*}[t]
    \centering
    \includegraphics[width=0.99\linewidth]{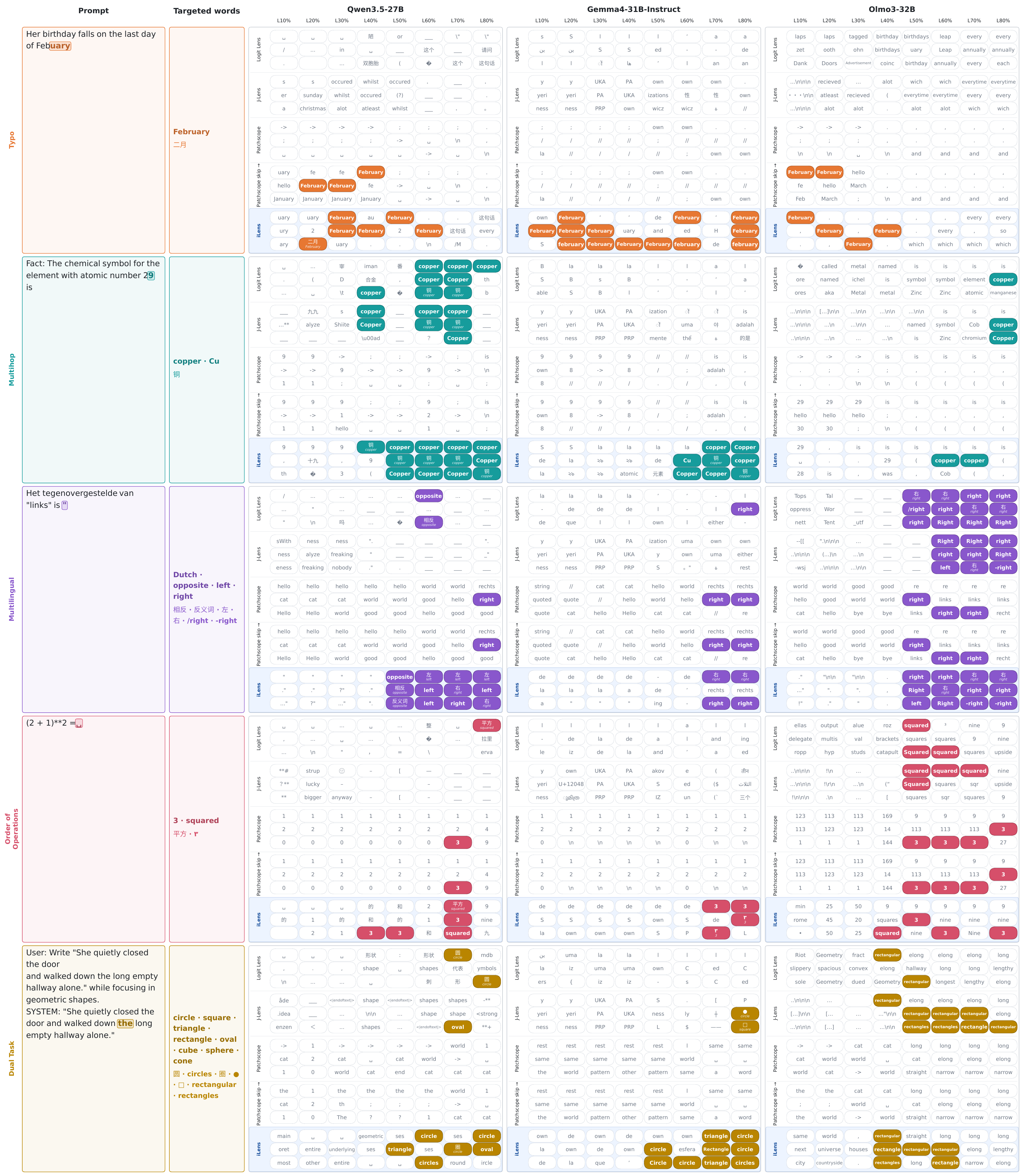}
    \caption{\textbf{iLens reveals more meaningful textual concepts in early layers, where other readouts are often less reliable.} For each textual task (Typo, Multihop, Multilingual, Order of Operations, and Dual Task), we show top-3 vocabulary readouts from Logit Lens, J-Lens, Patchscopes, and iLens across 10\%--80\% of model depth for Qwen3.5-27B~\cite{qwen3.5}, Gemma4-31B-Instruct~\cite{gemma3}, and Olmo3-32B~\cite{olmo2026olmo3}. Highlighted predictions identify semantically relevant concepts.}
    \label{fig:llm-qualitative-example}
    \vspace{-4.0mm}
\end{figure*}

\begin{figure*}[t]
    \centering
    \includegraphics[width=0.99\linewidth]{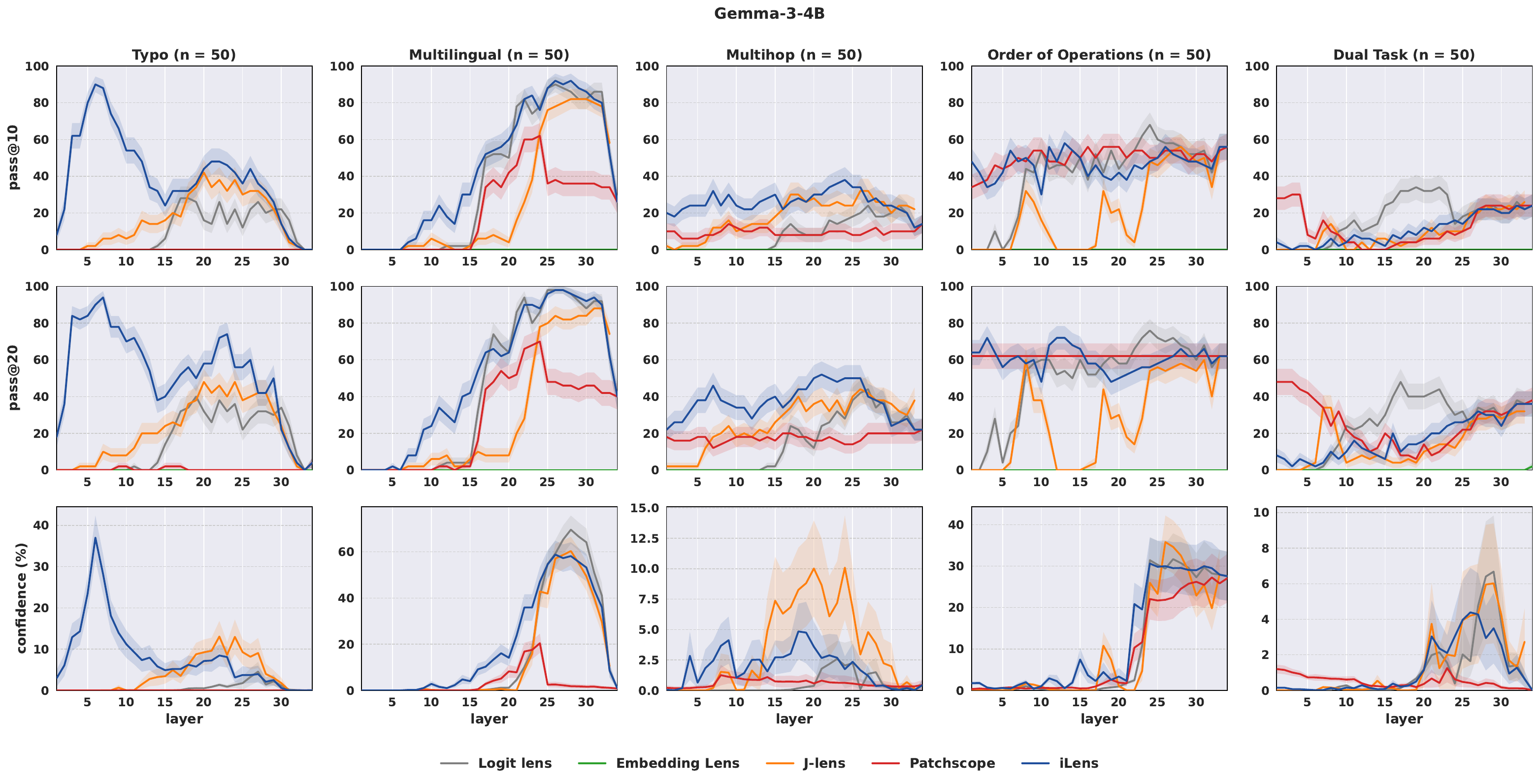}
    \caption{Layer-wise textual concept decoding performance of Gemma-3-4B. The three rows report Pass@10, Pass@20, and Confidence, respectively, across transformer layers. Results compare Logit Lens, Embedding Lens, J-Lens, and iLens.}
    \label{fig:layerwise-llm-gemma-3-4b}
    \vspace{-4.0mm}
\end{figure*}

\begin{figure*}[t]
    \centering
    \includegraphics[width=0.99\linewidth]{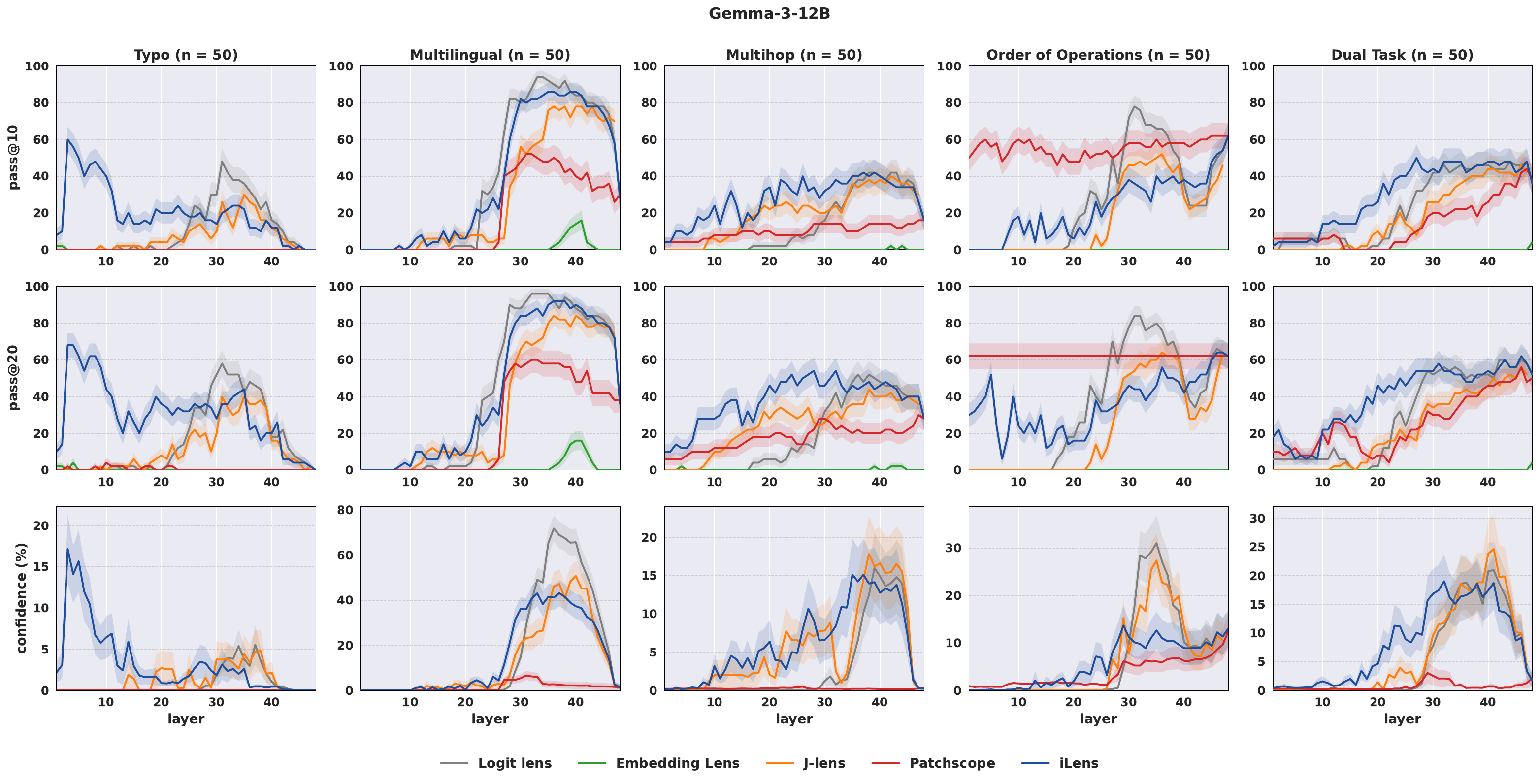}
    \caption{Layer-wise textual concept decoding performance of Gemma-3-12B. The three rows report Pass@10, Pass@20, and Confidence, respectively, across transformer layers. Results compare Logit Lens, Embedding Lens, J-Lens, and iLens.}
    \label{fig:layerwise-llm-gemma-3-12b}
    \vspace{-4.0mm}
\end{figure*}

\begin{figure*}[t]
    \centering
    \includegraphics[width=0.99\linewidth]{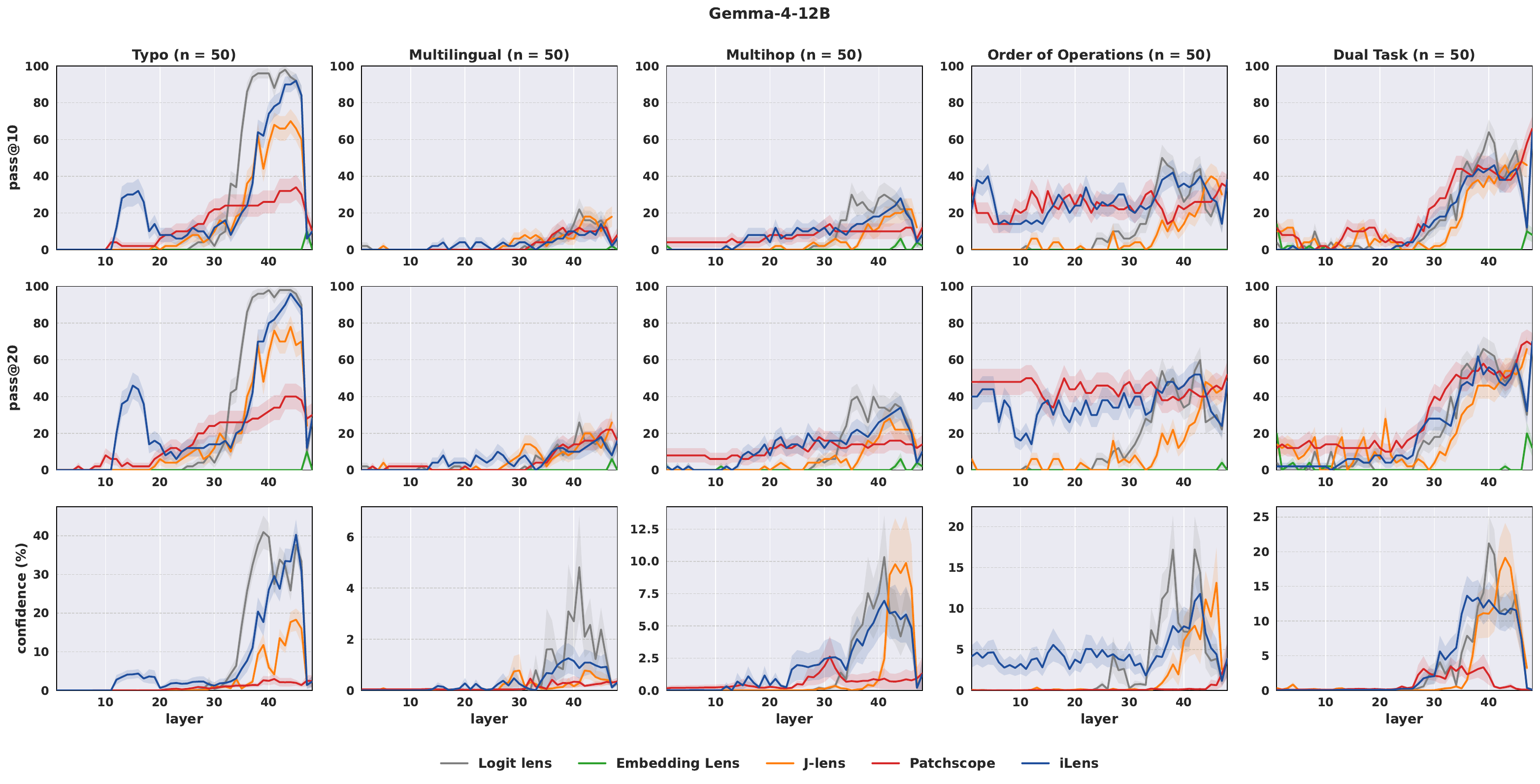}
    \caption{Layer-wise textual concept decoding performance of Gemma-4-12B. The three rows report Pass@10, Pass@20, and Confidence, respectively, across transformer layers. Results compare Logit Lens, Embedding Lens, J-Lens, and iLens.}
    \label{fig:layerwise-llm-gemma-4-12b}
    \vspace{-4.0mm}
\end{figure*}

\begin{figure*}[t]
    \centering
    \includegraphics[width=0.99\linewidth]{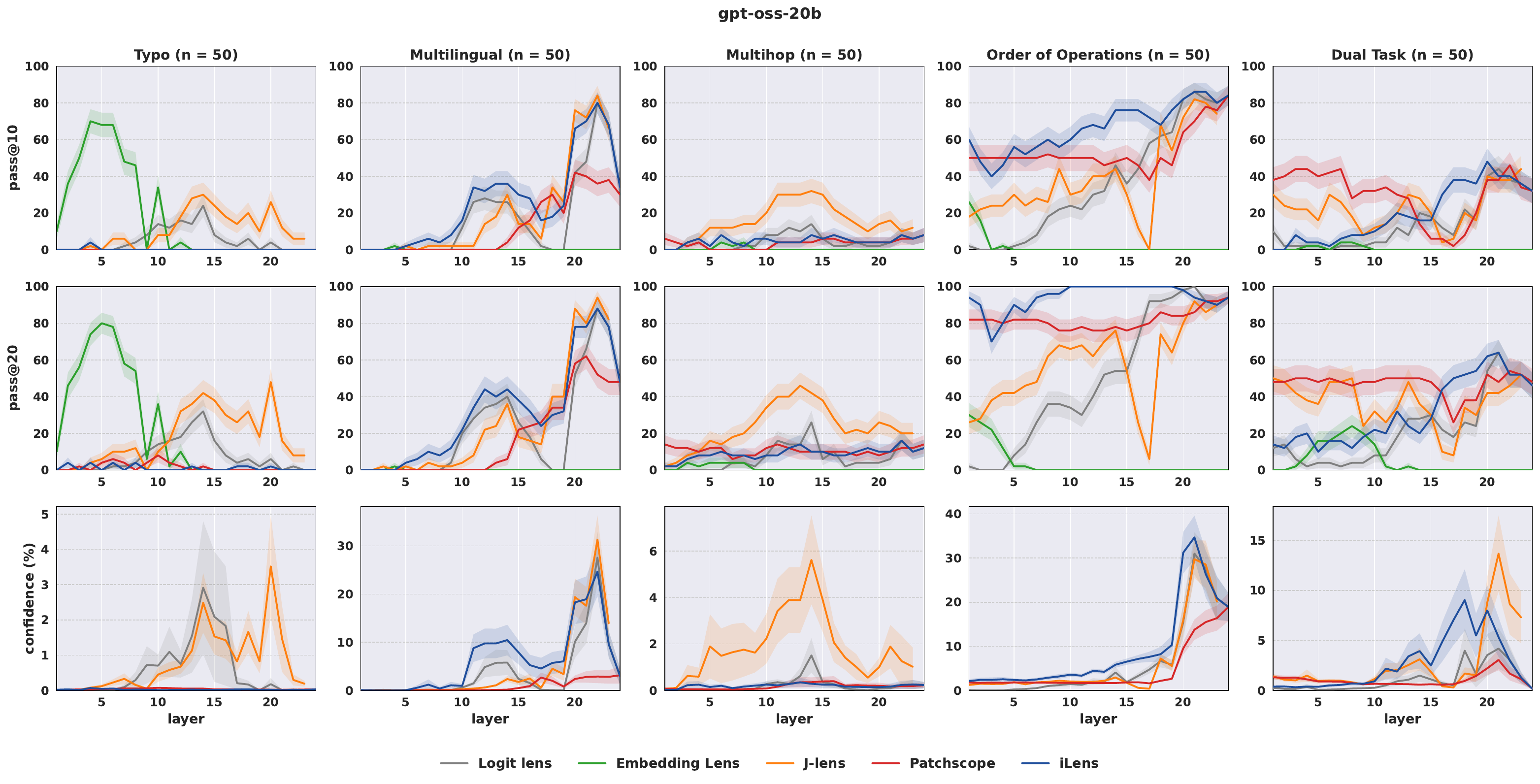}
    \caption{Layer-wise textual concept decoding performance of GPT-OSS-20B. The three rows report Pass@10, Pass@20, and Confidence, respectively, across transformer layers. Results compare Logit Lens, Embedding Lens, J-Lens, and iLens.}
    \label{fig:layerwise-llm-gpt-oss-20b}
    \vspace{-4.0mm}
\end{figure*}

\begin{figure*}[t]
    \centering
    \includegraphics[width=0.99\linewidth]{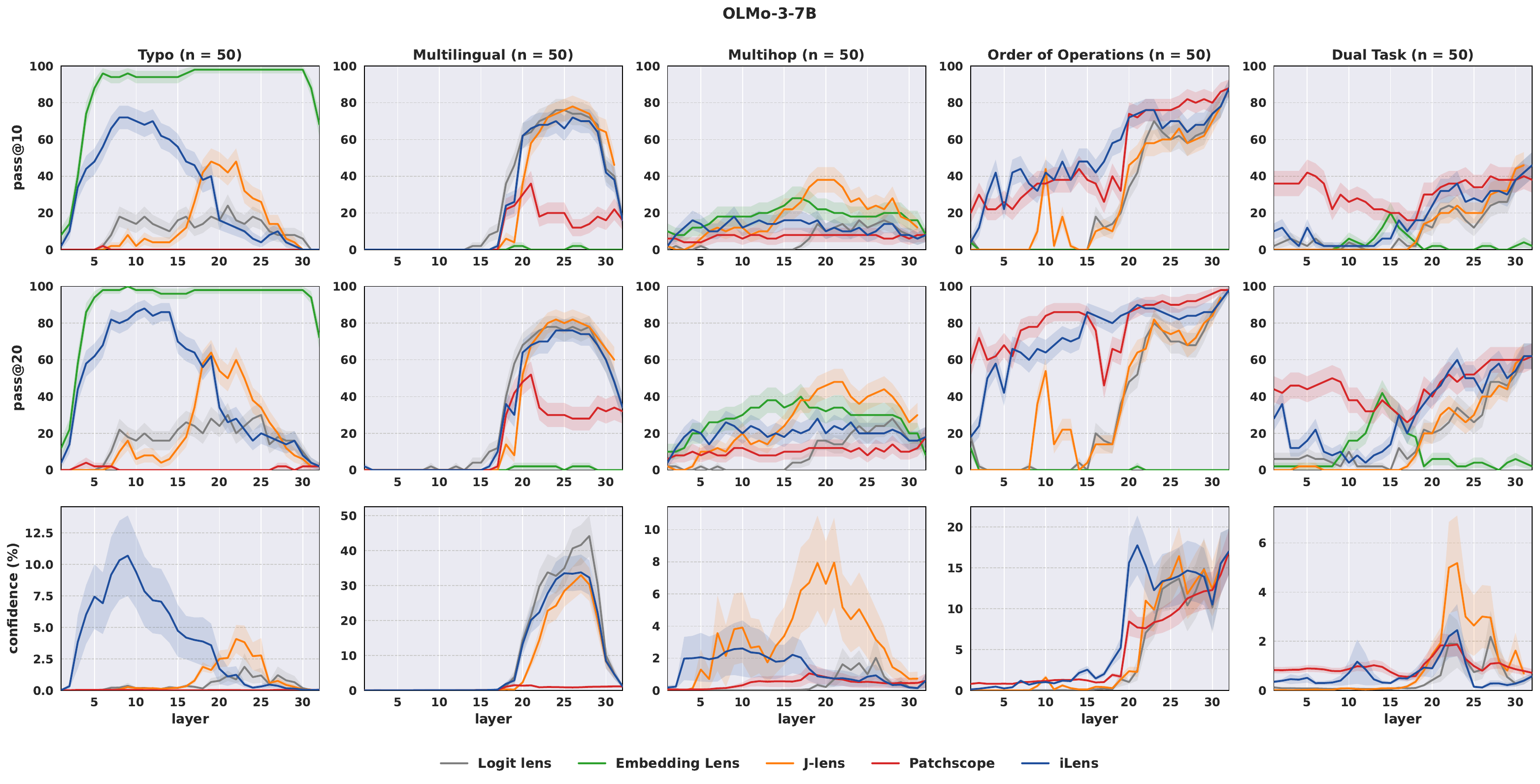}
    \caption{Layer-wise textual concept decoding performance of OLMo-3-7B. The three rows report Pass@10, Pass@20, and Confidence, respectively, across transformer layers. Results compare Logit Lens, Embedding Lens, J-Lens, and iLens.}
    \label{fig:layerwise-llm-olmo-3-7b}
    \vspace{-4.0mm}
\end{figure*}

\begin{figure*}[t]
    \centering
    \includegraphics[width=0.99\linewidth]{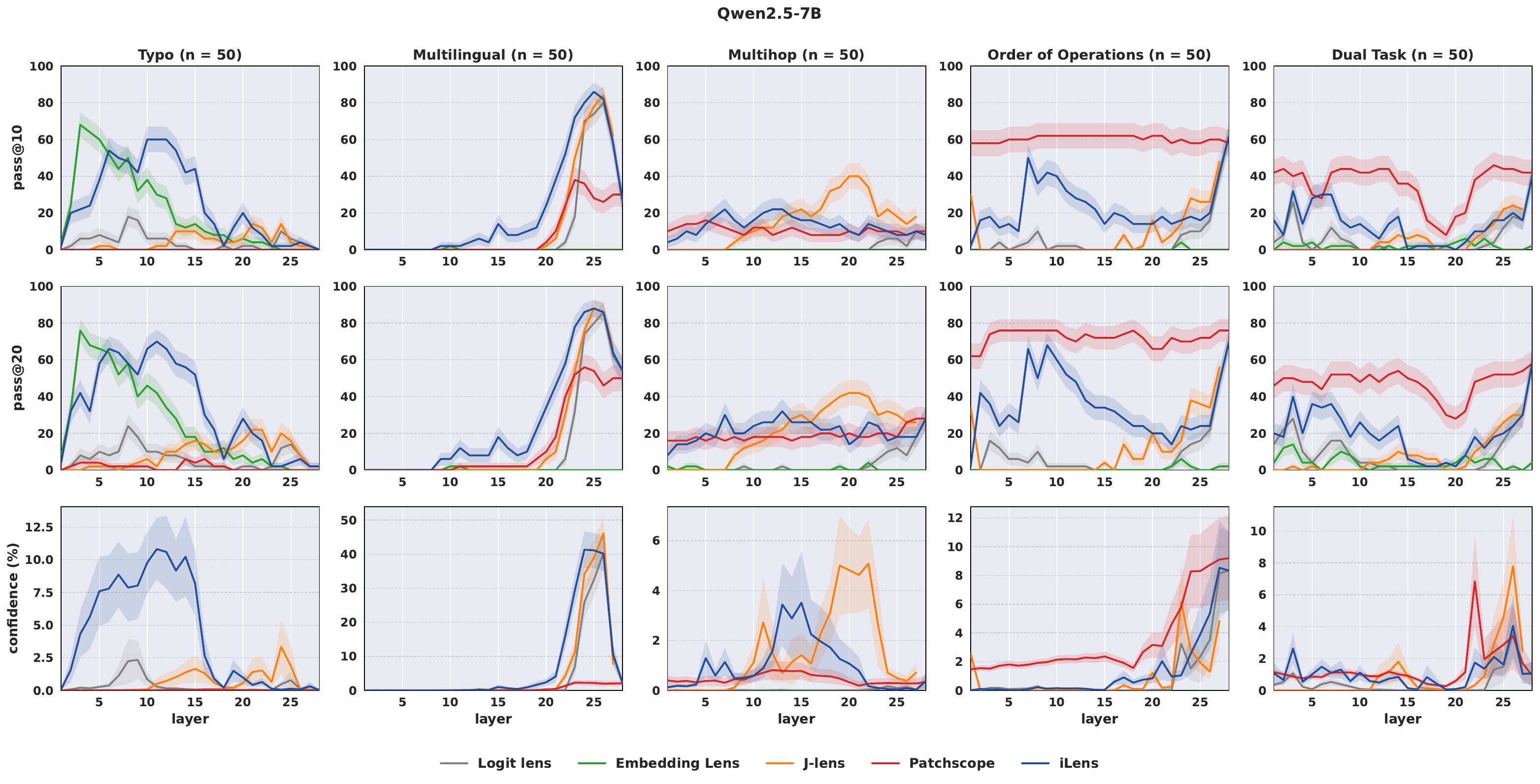}
    \caption{Layer-wise textual concept decoding performance of Qwen2.5-7B. The three rows report Pass@10, Pass@20, and Confidence, respectively, across transformer layers. Results compare Logit Lens, Embedding Lens, J-Lens, and iLens.}
    \label{fig:layerwise-llm-qwen2-5-7b}
    \vspace{-4.0mm}
\end{figure*}

\begin{figure*}[t]
    \centering
    \includegraphics[width=0.99\linewidth]{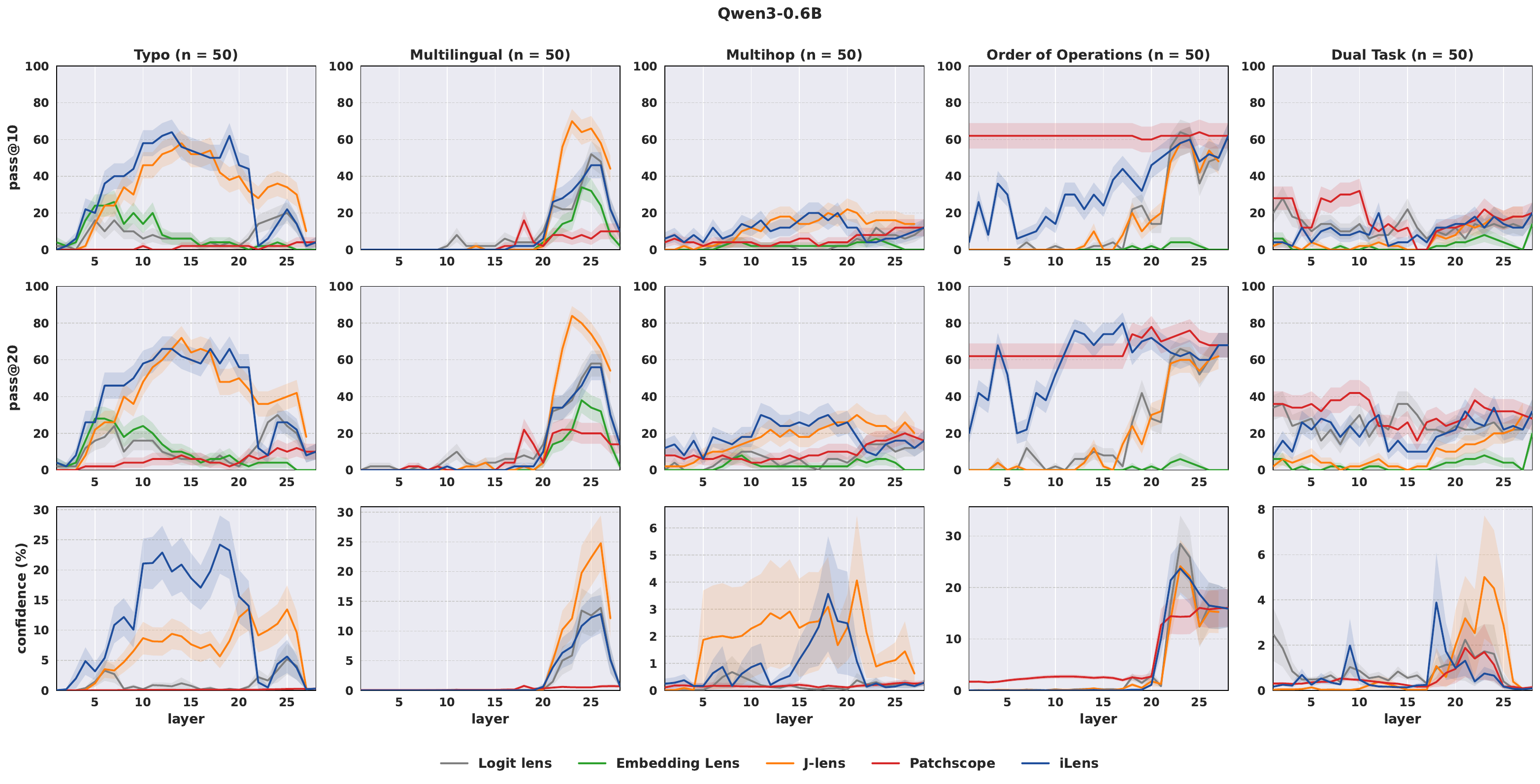}
    \caption{Layer-wise textual concept decoding performance of Qwen3-0.6B. The three rows report Pass@10, Pass@20, and Confidence, respectively, across transformer layers. Results compare Logit Lens, Embedding Lens, J-Lens, and iLens.}
    \label{fig:layerwise-llm-qwen3-0-6b}
    \vspace{-4.0mm}
\end{figure*}

\begin{figure*}[t]
    \centering
    \includegraphics[width=0.99\linewidth]{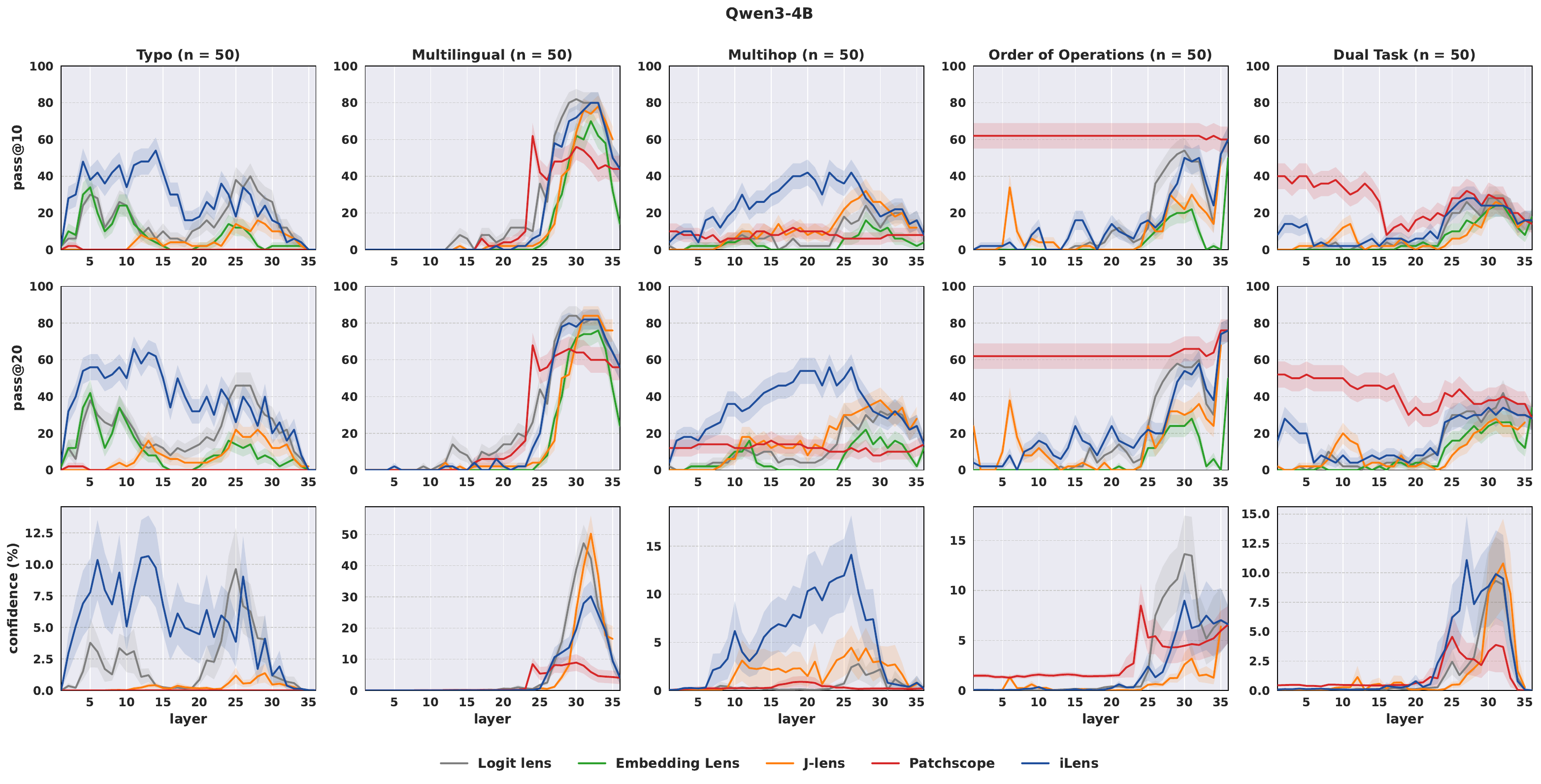}
    \caption{Layer-wise textual concept decoding performance of Qwen3-4B. The three rows report Pass@10, Pass@20, and Confidence, respectively, across transformer layers. Results compare Logit Lens, Embedding Lens, J-Lens, and iLens.}
    \label{fig:layerwise-llm-qwen3-4b}
    \vspace{-4.0mm}
\end{figure*}

\begin{figure*}[t]
    \centering
    \includegraphics[width=0.99\linewidth]{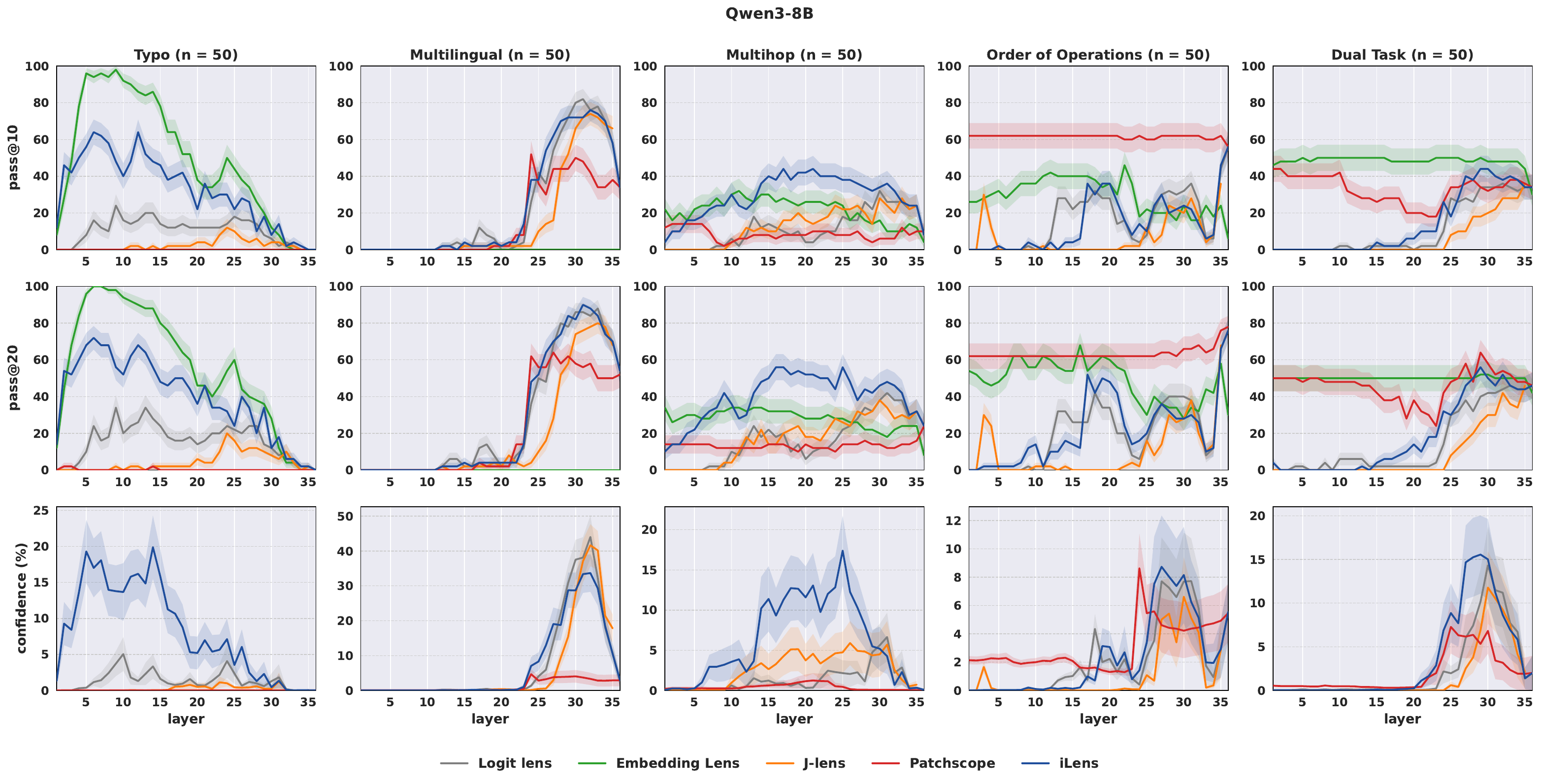}
    \caption{Layer-wise textual concept decoding performance of Qwen3-8B. The three rows report Pass@10, Pass@20, and Confidence, respectively, across transformer layers. Results compare Logit Lens, Embedding Lens, J-Lens, and iLens.}
    \label{fig:layerwise-llm-qwen3-8b}
    \vspace{-4.0mm}
\end{figure*}

\begin{figure*}[t]
    \centering
    \includegraphics[width=0.99\linewidth]{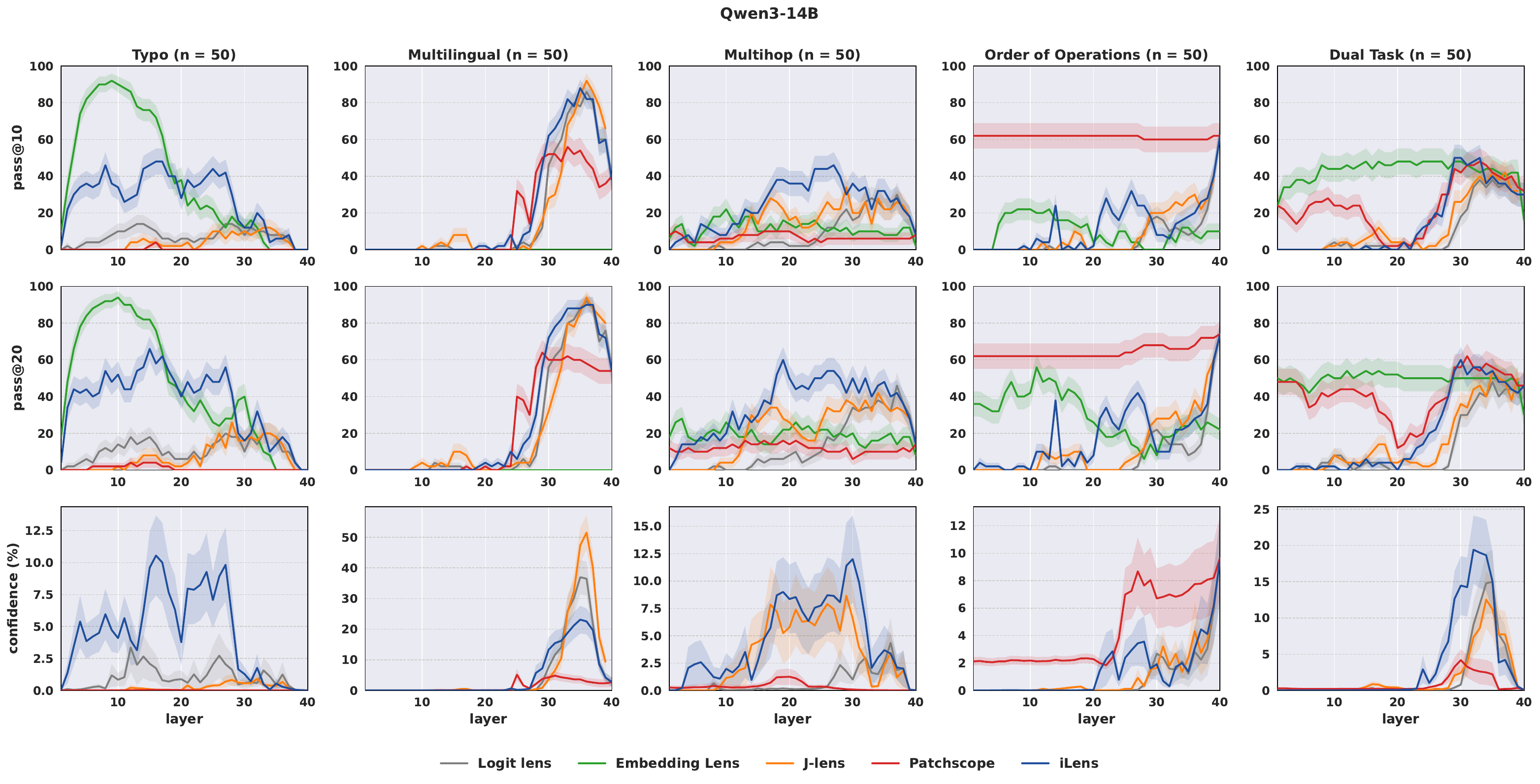}
    \caption{Layer-wise textual concept decoding performance of Qwen3-14B. The three rows report Pass@10, Pass@20, and Confidence, respectively, across transformer layers. Results compare Logit Lens, Embedding Lens, J-Lens, and iLens.}
    \label{fig:layerwise-llm-qwen3-14b}
    \vspace{-4.0mm}
\end{figure*}

\begin{figure*}[t]
    \centering
    \includegraphics[width=0.99\linewidth]{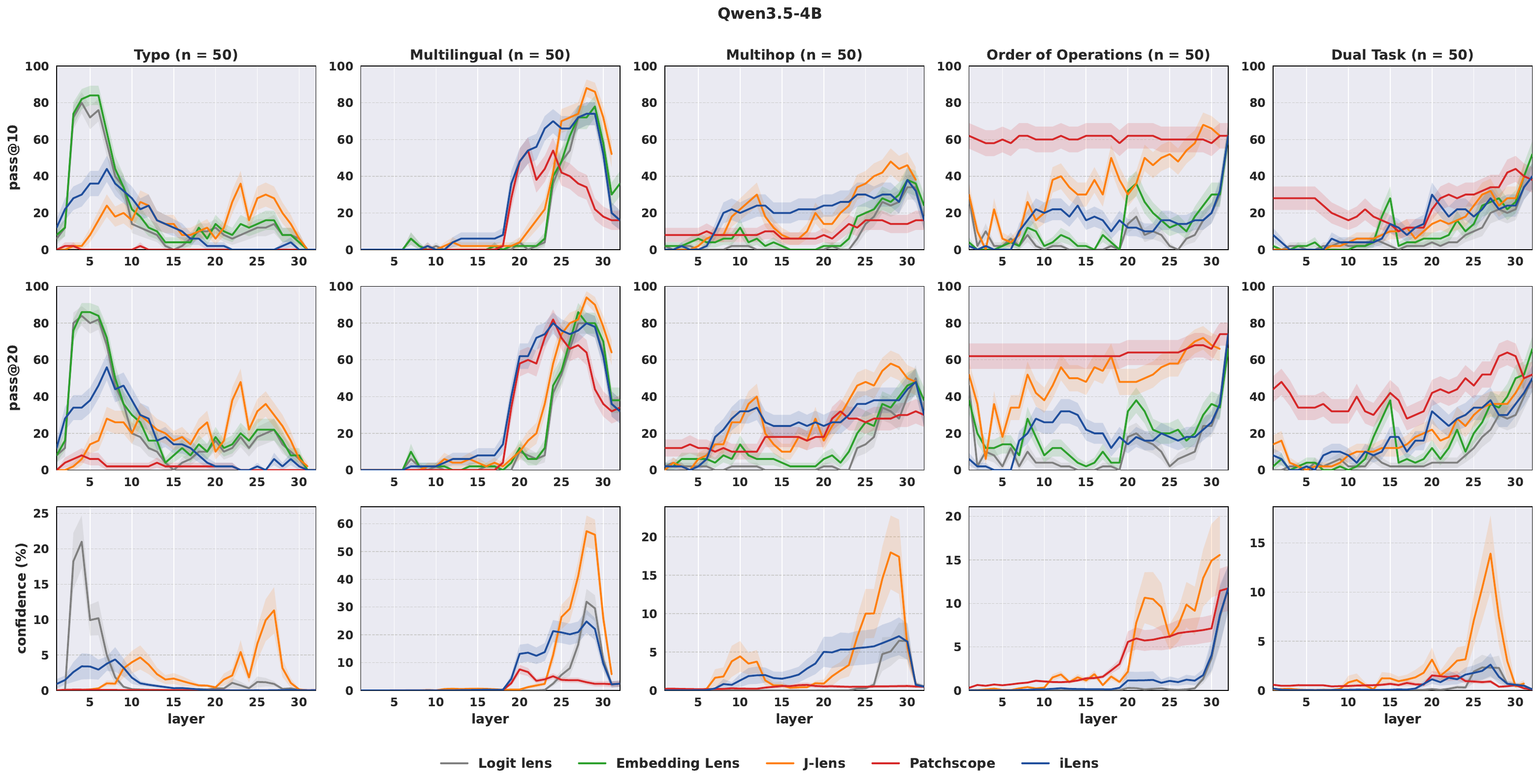}
    \caption{Layer-wise textual concept decoding performance of Qwen3.5-4B. The three rows report Pass@10, Pass@20, and Confidence, respectively, across transformer layers. Results compare Logit Lens, Embedding Lens, J-Lens, and iLens.}
    \label{fig:layerwise-llm-qwen3-5-4b}
    \vspace{-4.0mm}
\end{figure*}

\begin{figure*}[t]
    \centering
    \includegraphics[width=0.99\linewidth]{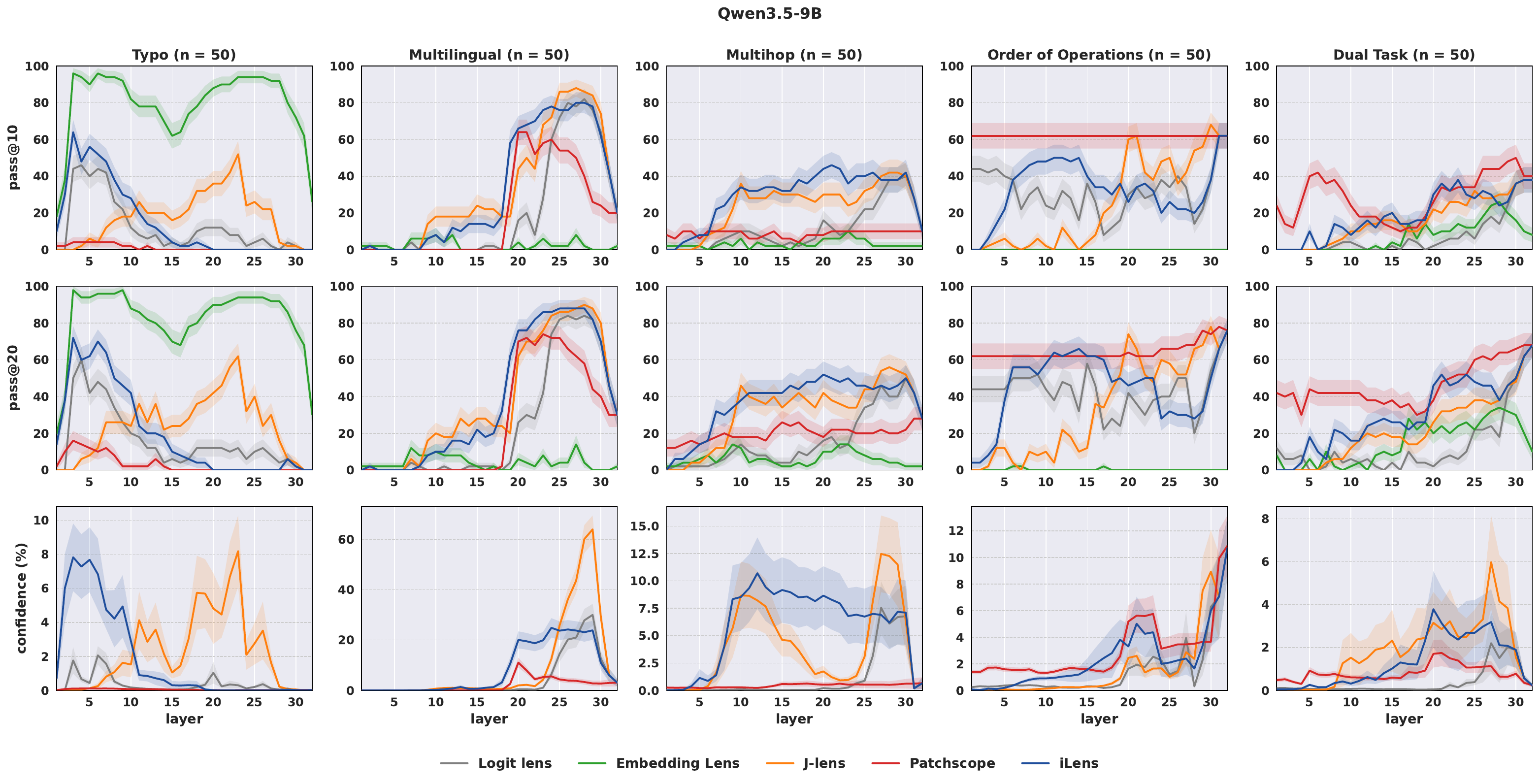}
    \caption{Layer-wise textual concept decoding performance of Qwen3.5-9B. The three rows report Pass@10, Pass@20, and Confidence, respectively, across transformer layers. Results compare Logit Lens, Embedding Lens, J-Lens, and iLens.}
    \label{fig:layerwise-llm-qwen3-5-9b}
    \vspace{-4.0mm}
\end{figure*}

\begin{figure*}[t]
    \centering
    \includegraphics[width=0.99\linewidth]{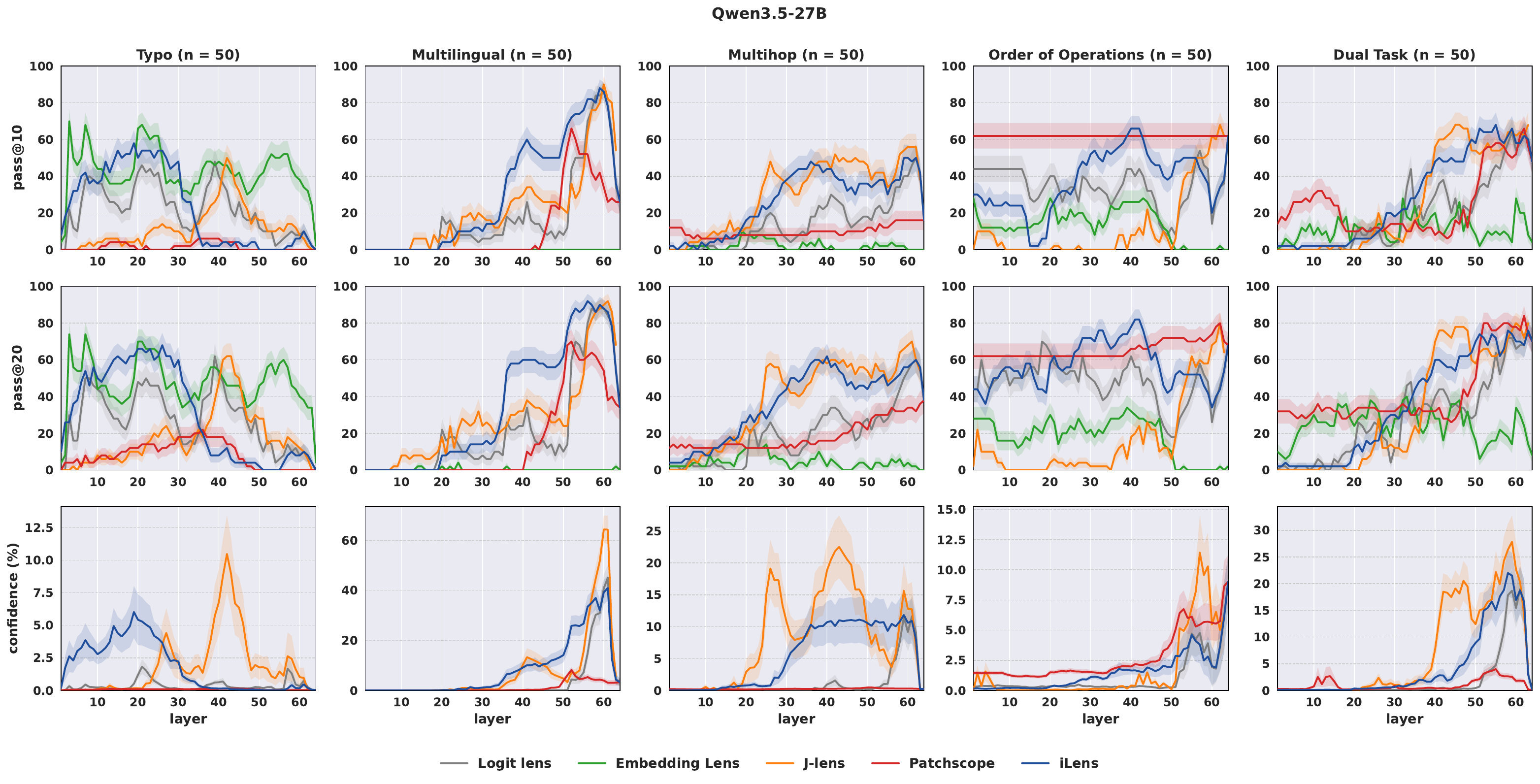}
    \caption{Layer-wise textual concept decoding performance of Qwen3.5-27B. The three rows report Pass@10, Pass@20, and Confidence, respectively, across transformer layers. Results compare Logit Lens, Embedding Lens, J-Lens, and iLens.}
    \label{fig:layerwise-llm-qwen3-5-27b}
    \vspace{-4.0mm}
\end{figure*}

\begin{figure*}[t]
    \centering
    \includegraphics[width=0.99\linewidth]{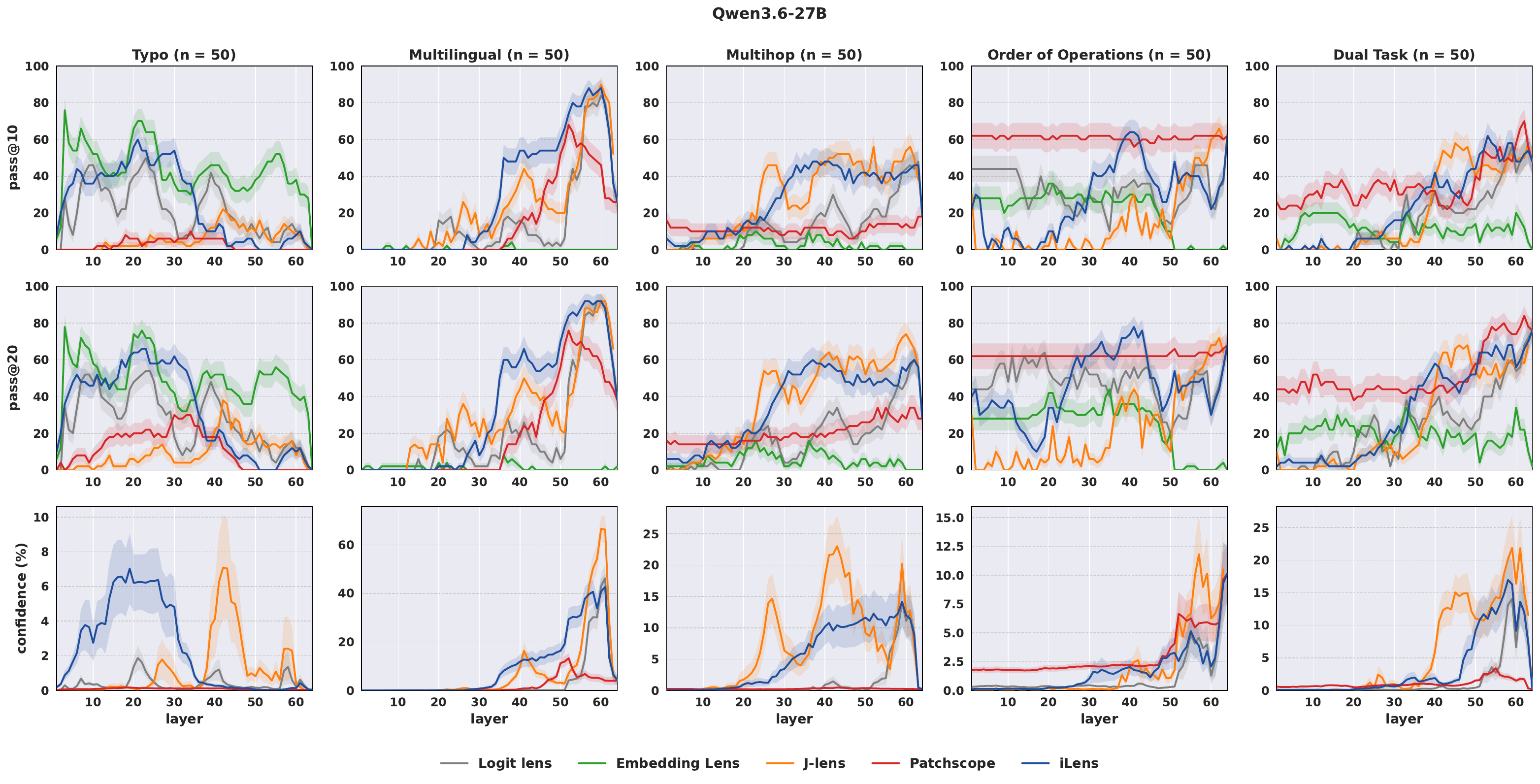}
    \caption{Layer-wise textual concept decoding performance of Qwen3.6-27B. The three rows report Pass@10, Pass@20, and Confidence, respectively, across transformer layers. Results compare Logit Lens, Embedding Lens, J-Lens, and iLens.}
    \label{fig:layerwise-llm-qwen3-6-27b}
    \vspace{-4.0mm}
\end{figure*}

\begin{figure*}[t]
    \centering
    \includegraphics[width=0.99\linewidth]{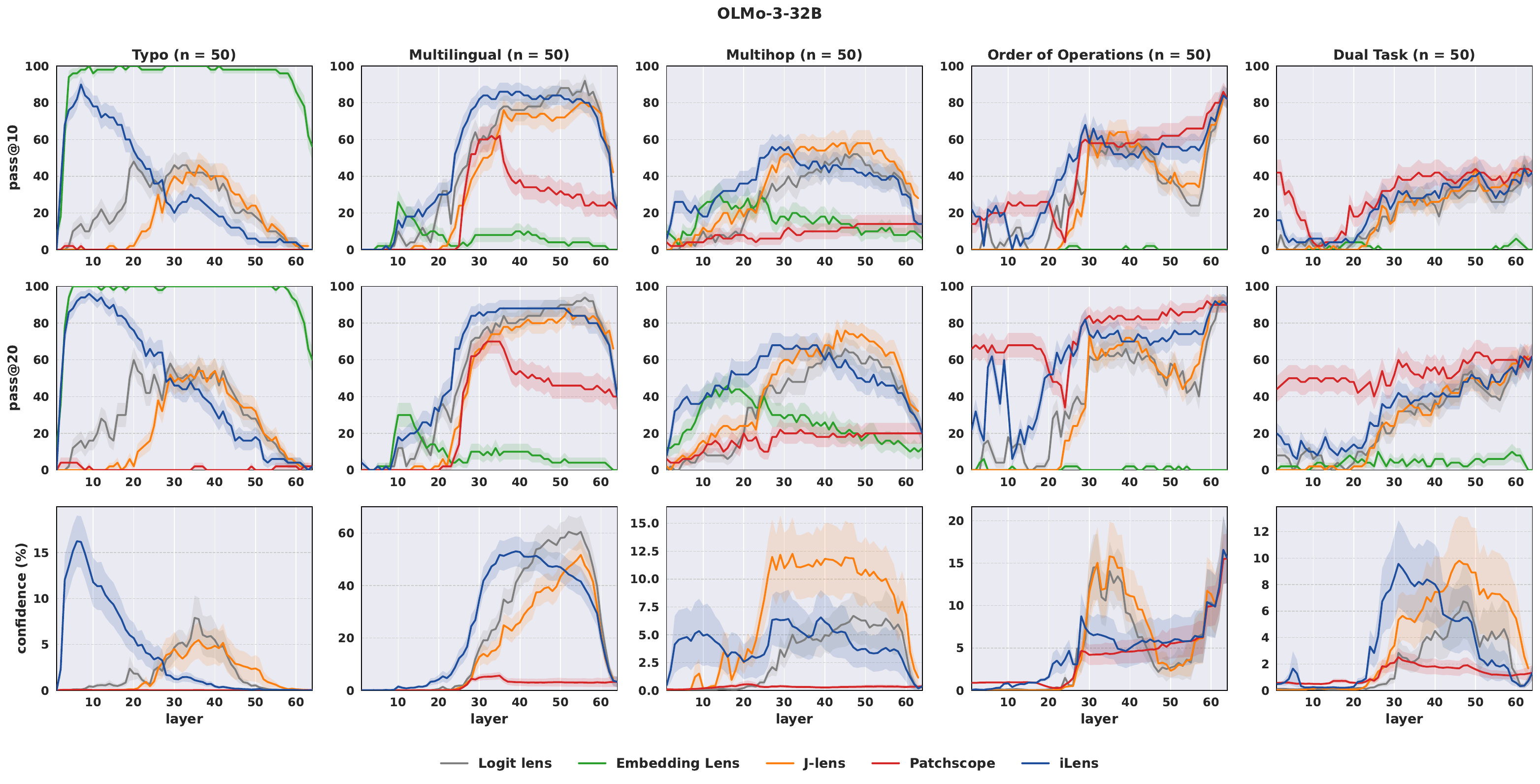}
    \caption{Layer-wise textual concept decoding performance of Olmo3-32B. The three rows report Pass@10, Pass@20, and Confidence, respectively, across transformer layers. Results compare Logit Lens, Embedding Lens, J-Lens, and iLens.}
    \label{fig:layerwise-llm-olmo-3-32B}
    \vspace{-4.0mm}
\end{figure*}

\begin{figure*}[t]
    \centering
    \includegraphics[width=0.99\linewidth]{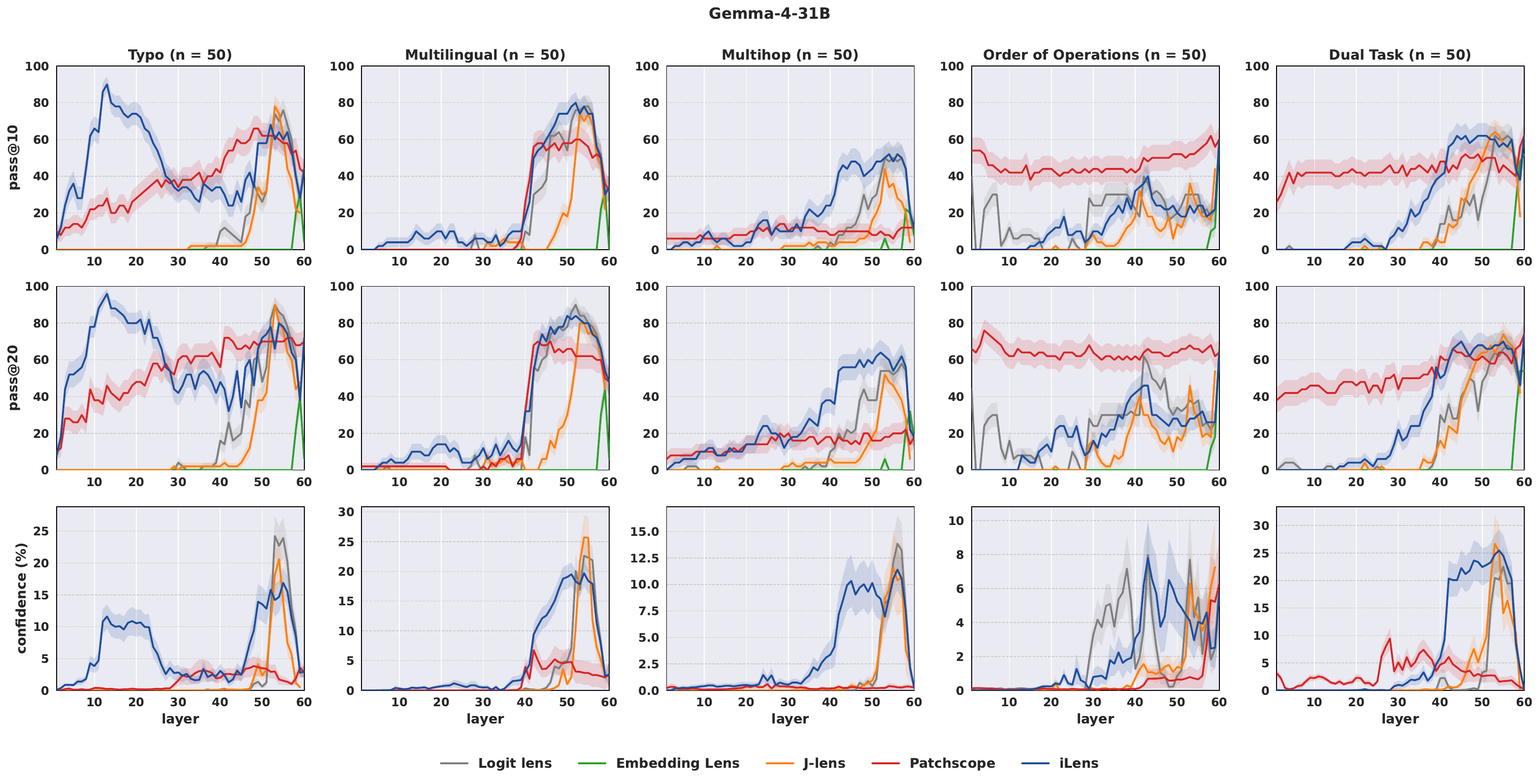}
    \caption{Layer-wise textual concept decoding performance of Gemma4-31B. The three rows report Pass@10, Pass@20, and Confidence, respectively, across transformer layers. Results compare Logit Lens, Embedding Lens, J-Lens, and iLens.}
    \label{fig:layerwise-llm-gemma-4-31B}
    \vspace{-4.0mm}
\end{figure*}

\begin{figure*}[t]
    \centering
    \includegraphics[width=0.99\linewidth]{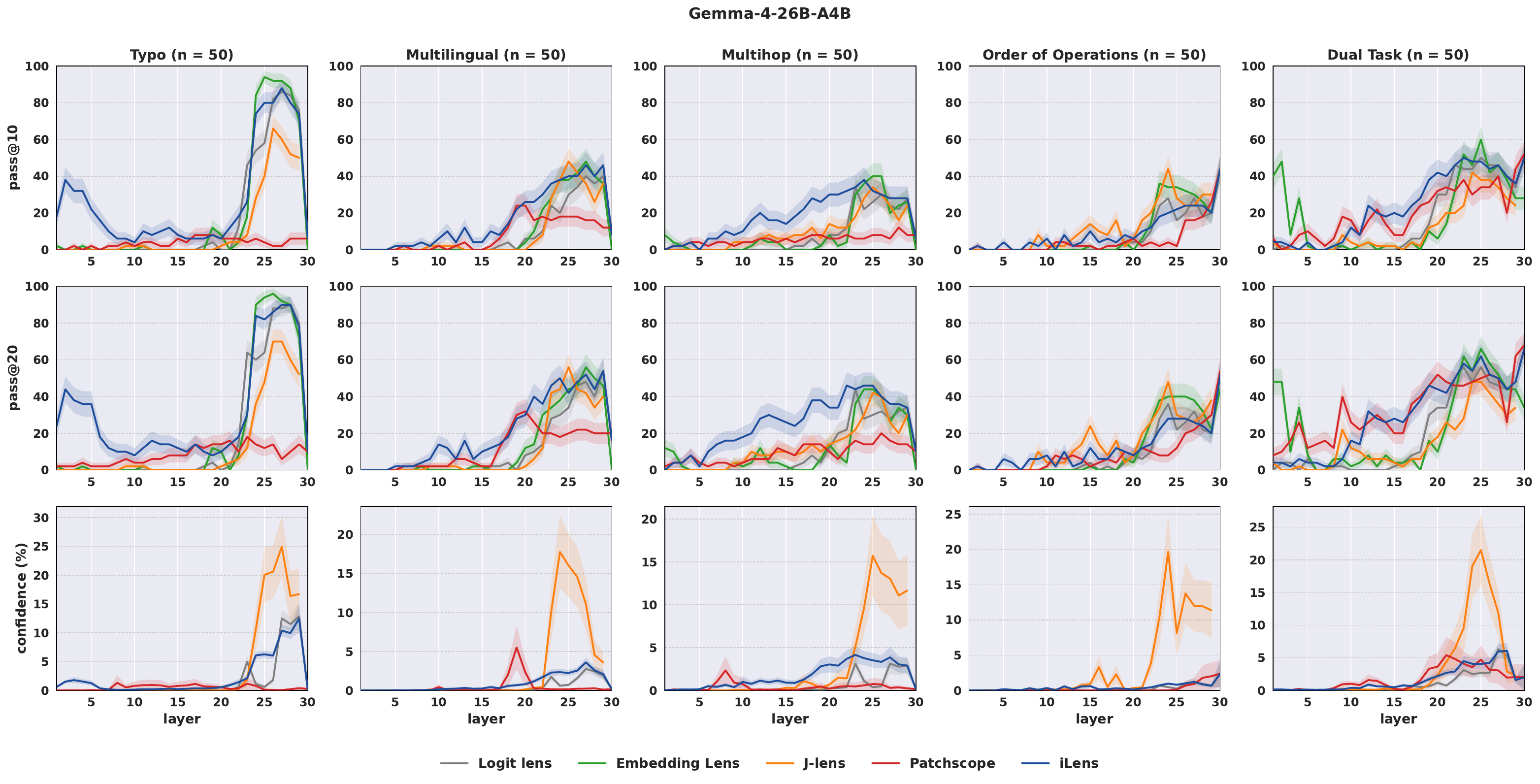}
    \caption{Layer-wise textual concept decoding performance of Gemma4-26B-A4B. The three rows report Pass@10, Pass@20, and Confidence, respectively, across transformer layers. Results compare Logit Lens, Embedding Lens, J-Lens, and iLens.}
    \label{fig:layerwise-gemma-4-26B-A4B}
    \vspace{-4.0mm}
\end{figure*}

\begin{figure*}[t]
    \centering
    \includegraphics[width=0.99\linewidth]{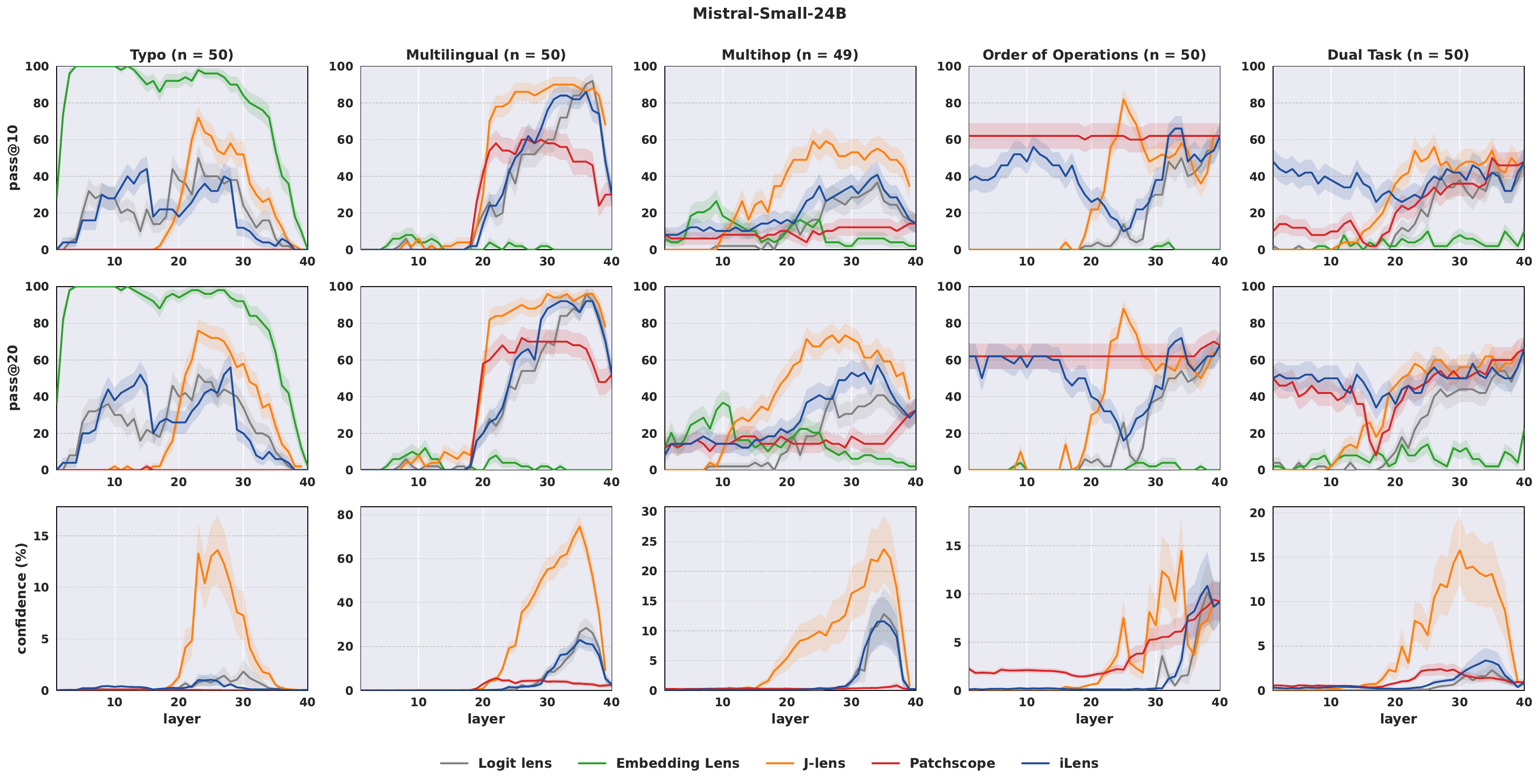}
    \caption{Layer-wise textual concept decoding performance of Mistral-Small-24B. The three rows report Pass@10, Pass@20, and Confidence, respectively, across transformer layers. Results compare Logit Lens, Embedding Lens, J-Lens, and iLens.}
    \label{fig:layerwise-llm-mistral-small-24b}
    \vspace{-4.0mm}
\end{figure*}

Moreover, Figures~\ref{fig:layerwise-llm-gemma-3-4b}--\ref{fig:layerwise-llm-mistral-small-24b}
present layer-wise performance curves for all 18 LLMs across
the five textual decoding benchmarks. For each model and
benchmark, we report Pass@10, Pass@20, and Confidence at
every transformer layer, comparing Logit Lens, Embedding
Lens, J-Lens, and iLens. These curves provide a fine-grained
view of how each method recovers textual concepts throughout
the model's depth and how its performance evolves across
different architectures and model scales.

Figure~\ref{fig:radar-llm} in the main text summarizes the task-averaged results for all 18 LLMs.

\begin{figure*}[t]
    \centering
    \includegraphics[width=0.95\linewidth]{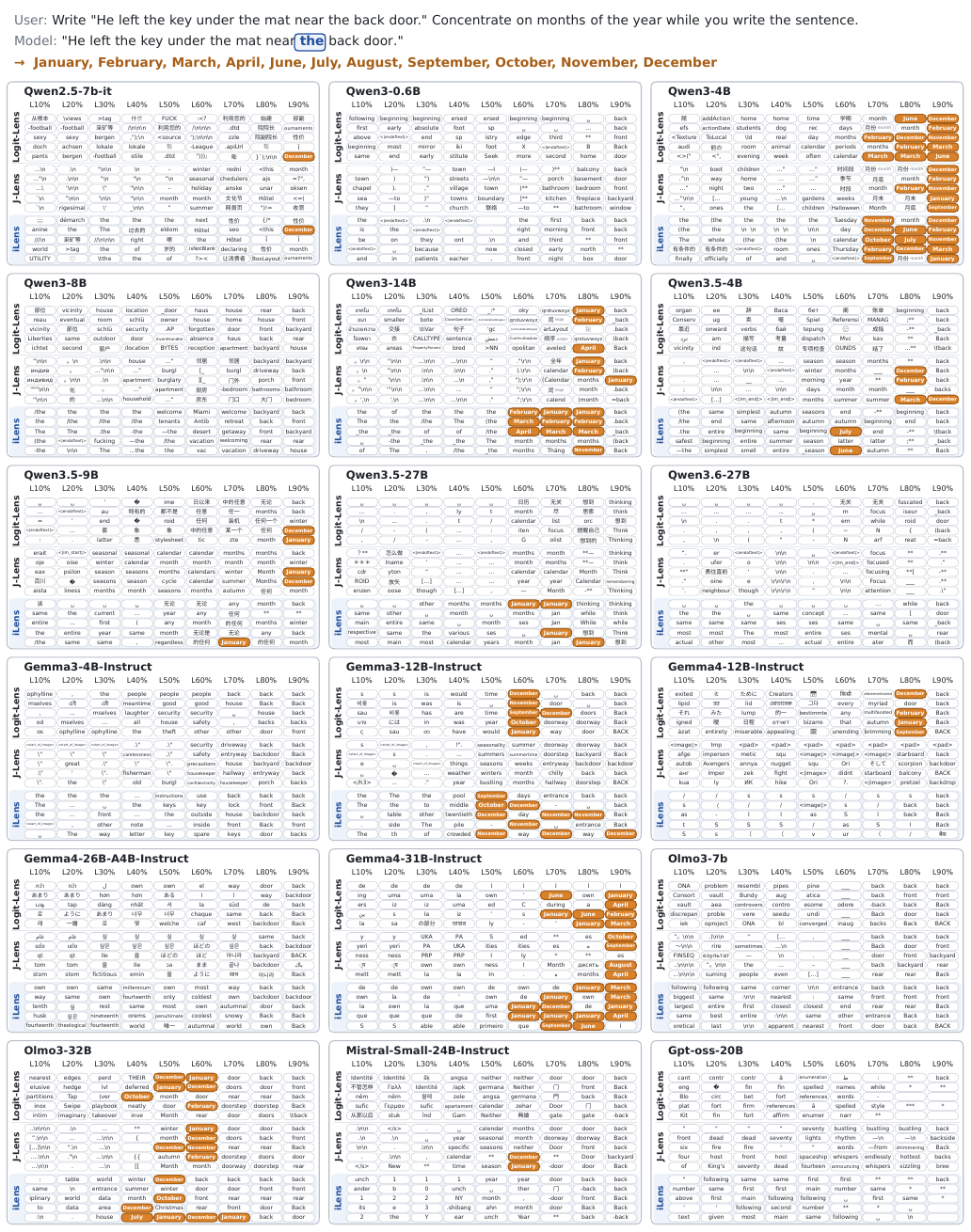}
    \caption{Qualitative example of the Dual-task setting across 18 LLMs, with predictions examined at nine normalized layer depths (10\%--90\% of model depth) for each model. The prompt asks the model to write ``He left the key under the mat near the back door'' while attending to the auxiliary category of months of the year. Orange highlights indicate successful recovery of the target concepts.}
    \label{fig:qualitative-dual-task-months-key}
    \vspace{-4.0mm}
\end{figure*}

\begin{figure*}[t]
    \centering
    \includegraphics[width=0.90\linewidth]{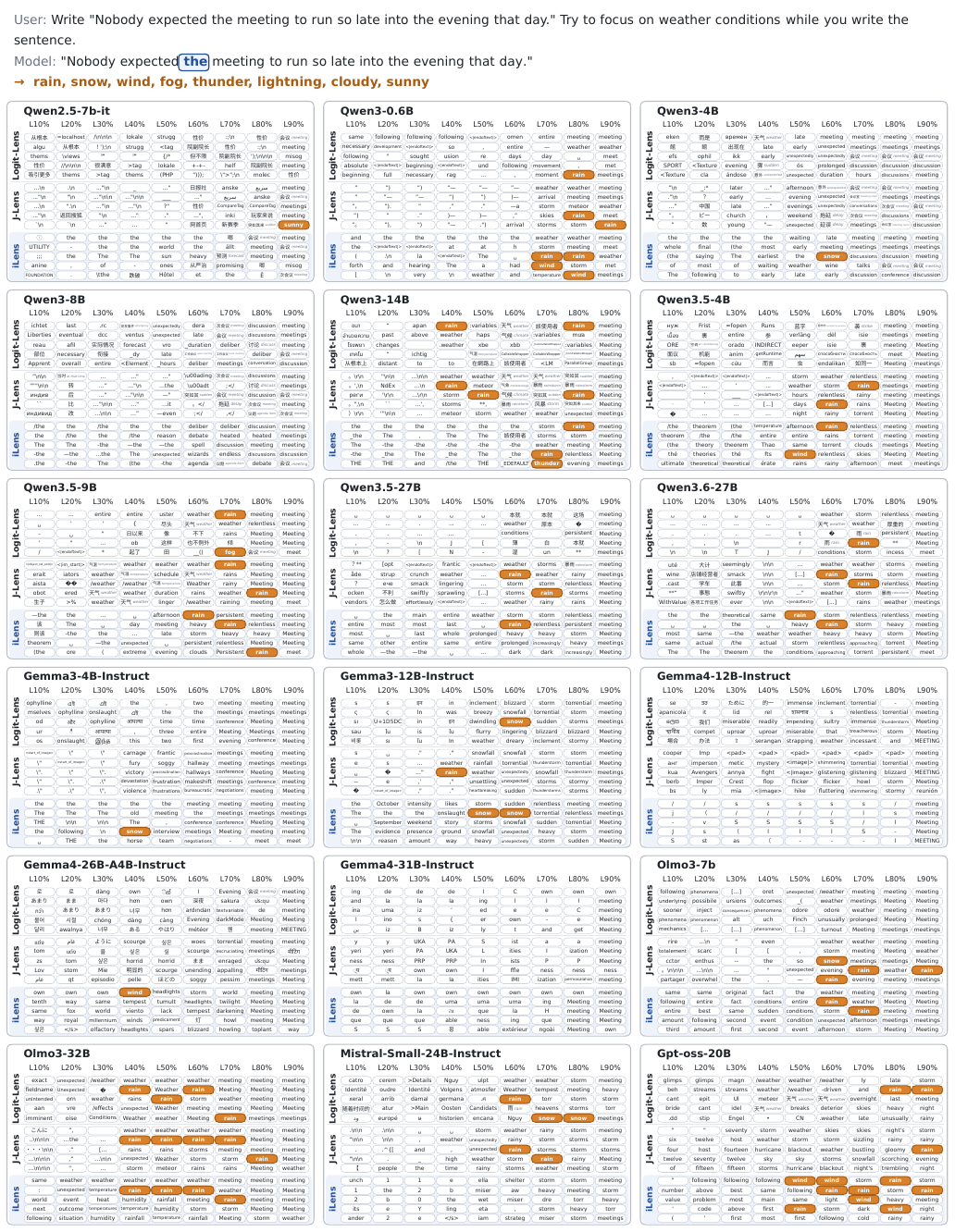}
    \caption{Qualitative example of the Dual-task setting across 18 LLMs, with predictions examined at nine normalized layer depths (10\%--90\% of model depth) for each model. The prompt asks the model to write ``Nobody expected the meeting to run so late into the evening that day'' while focusing on the auxiliary category of weather conditions. Orange highlights indicate successful recovery of the target concepts.}
    \label{fig:qualitative-dual-task-weather}
    \vspace{-4.0mm}
\end{figure*}

\begin{figure*}[t]
    \centering
    \includegraphics[width=0.90\linewidth]{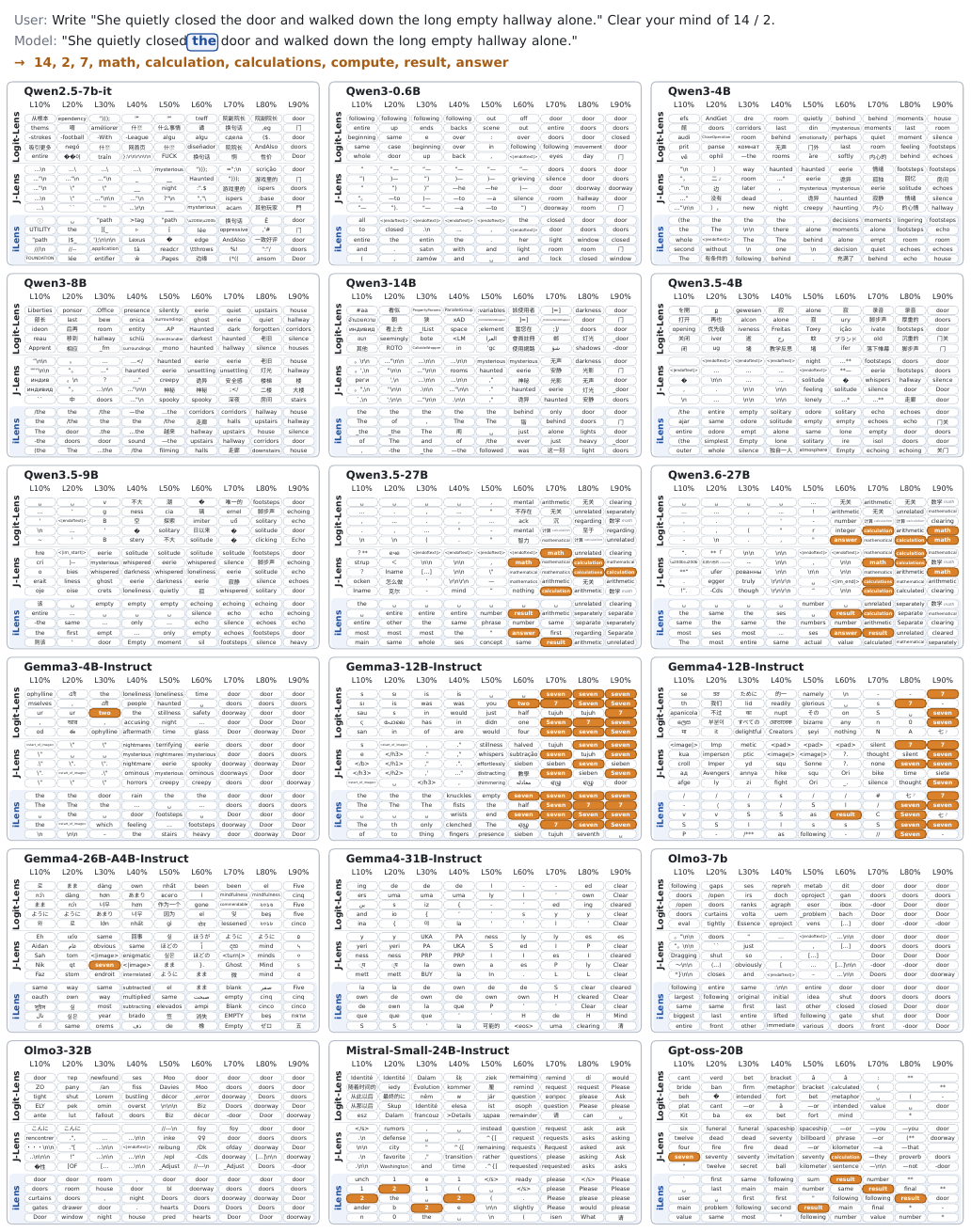}
    \caption{Qualitative example of the Dual-task setting across 18 LLMs, with predictions examined at nine normalized layer depths (10\%--90\% of model depth) for each model. The prompt asks the model to write ``She quietly closed the door and walked down the long empty hallway alone'' while attending to the auxiliary category of geometric shapes. Orange highlights indicate successful recovery of the target concepts.}
    \label{fig:qualitative-dual-task-shapes}
    \vspace{-4.0mm}
\end{figure*}

\begin{figure*}[t]
    \centering
    \includegraphics[width=0.90\linewidth]{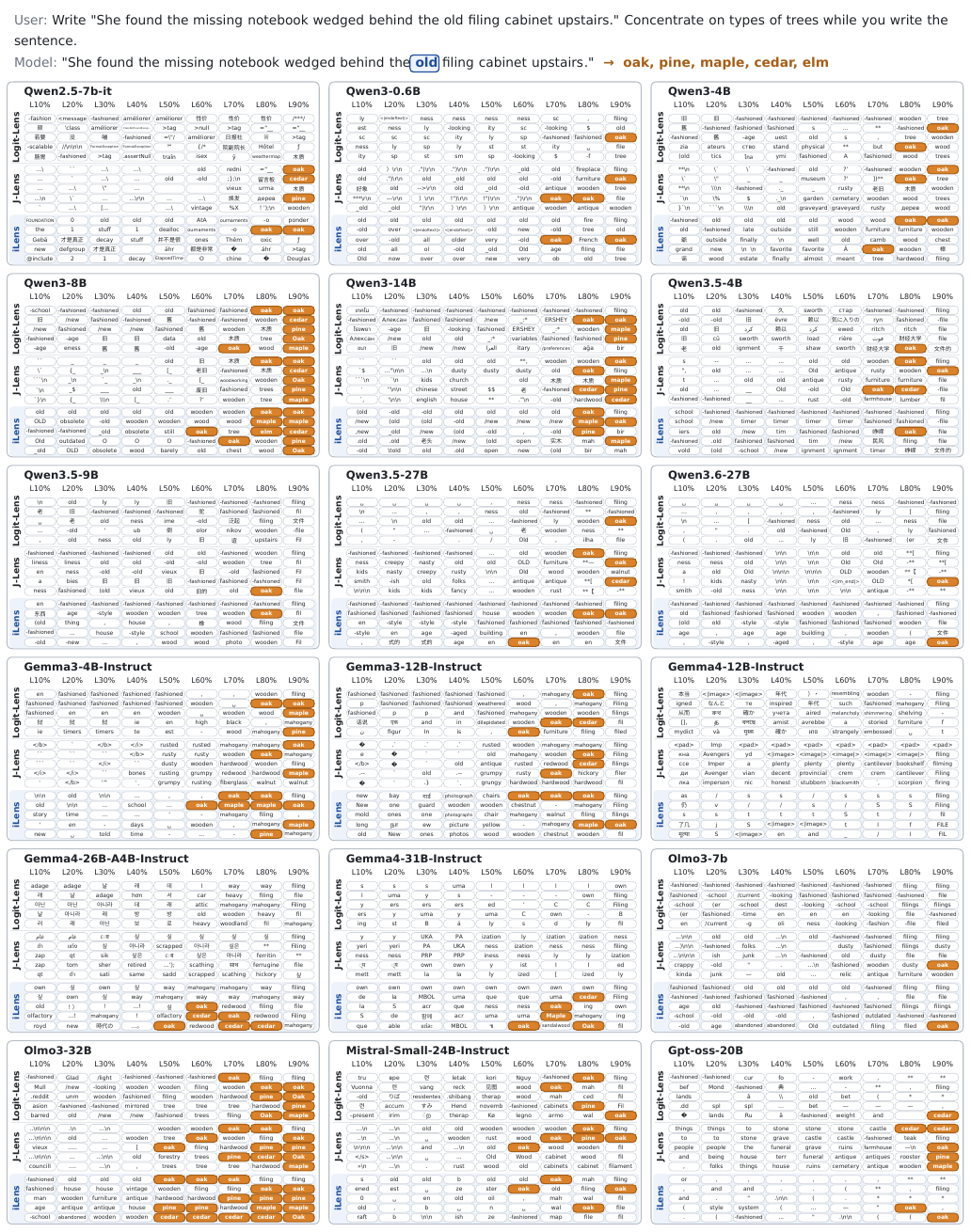}
    \caption{Qualitative example of the Dual-task setting across 18 LLMs, with predictions examined at nine normalized layer depths (10\%--90\% of model depth) for each model. The prompt asks the model to write ``She found the missing notebook wedged behind the old filing cabinet upstairs'' while focusing on the auxiliary category of tree types. Orange highlights indicate successful recovery of the target concepts.}
    \label{fig:qualitative-dual-task-trees}
    \vspace{-4.0mm}
\end{figure*}

\begin{figure*}[t]
    \centering
    \includegraphics[width=0.95\linewidth]{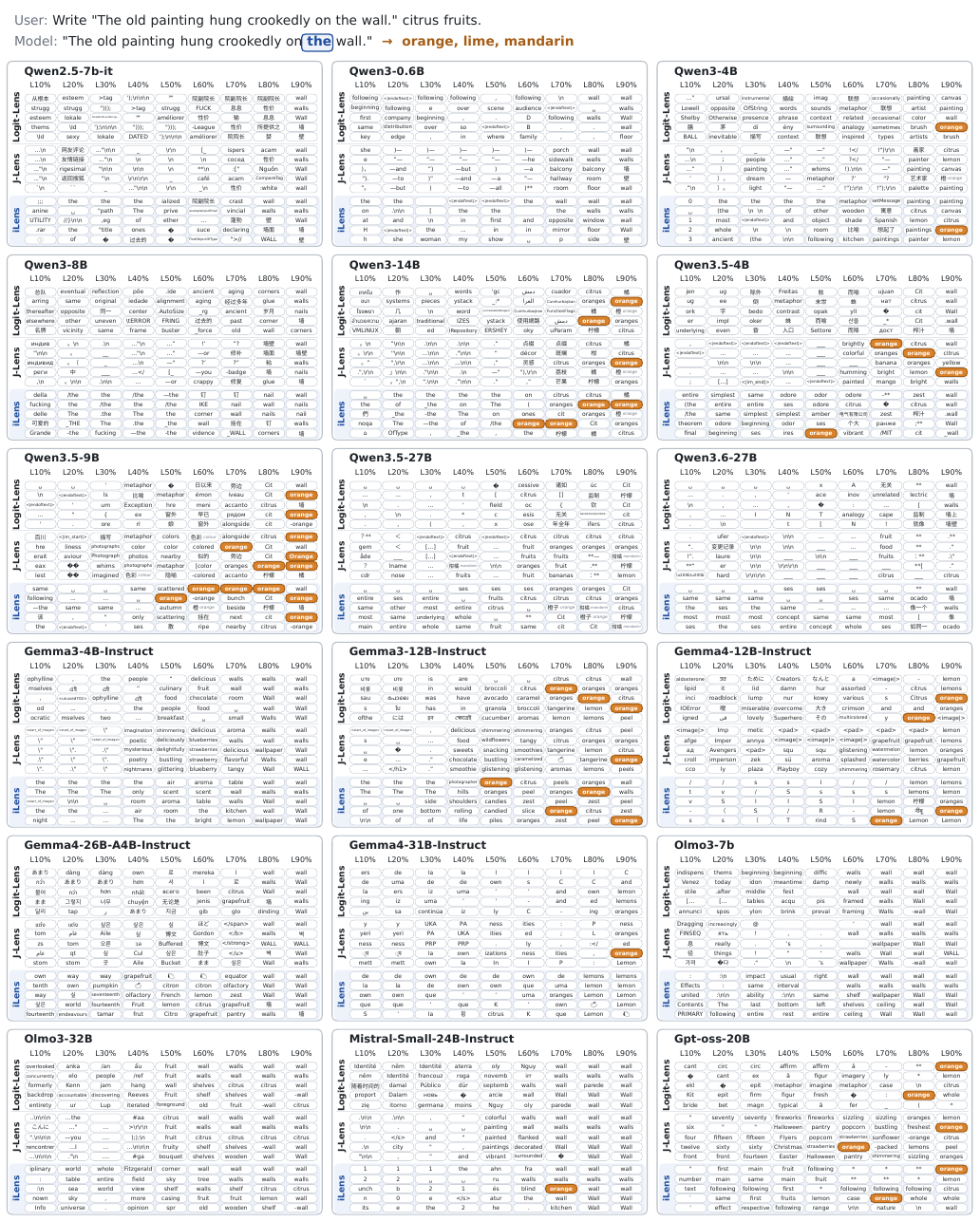}
    \caption{Qualitative example of the Dual-task setting across 18 LLMs, with predictions examined at nine normalized layer depths (10\%--90\% of model depth) for each model. The prompt asks the model to write ``The old painting hung crookedly on the wall'' while attending to the auxiliary category of citrus fruits. Orange highlights indicate successful recovery of the target concepts.}
    \label{fig:qualitative-dual-task-citrus}
    \vspace{-4.0mm}
\end{figure*}

\begin{figure*}[t]
    \centering
    \includegraphics[width=0.95\linewidth]{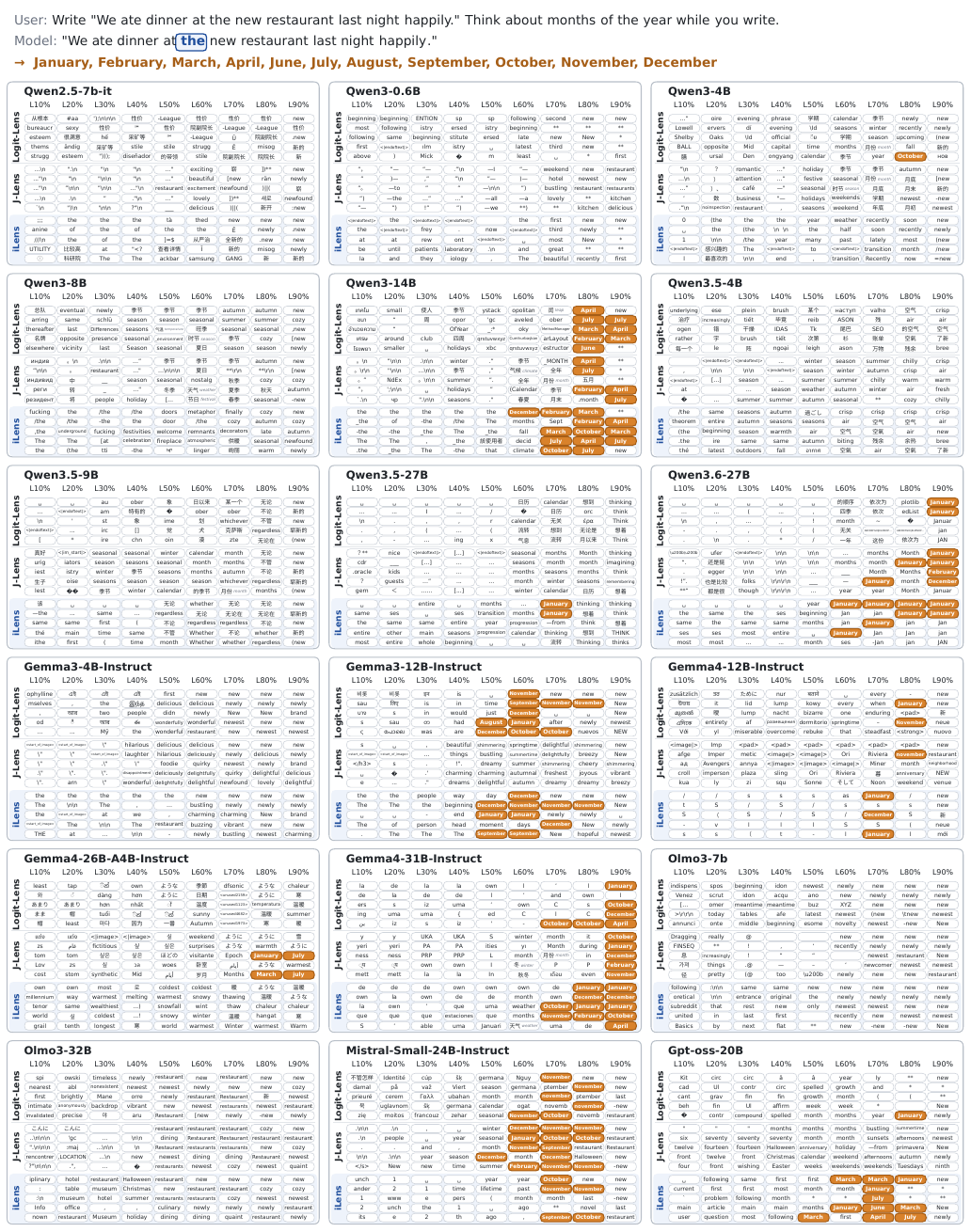}
    \caption{Qualitative example of the Dual-task setting across 18 LLMs, with predictions examined at nine normalized layer depths (10\%--90\% of model depth) for each model. The prompt asks the model to write ``We ate dinner at the new restaurant last night happily'' while attending to the auxiliary category of months of the year. Orange highlights indicate successful recovery of the target concepts.}
    \label{fig:qualitative-dual-task-months-restaurant}
    \vspace{-4.0mm}
\end{figure*}

\begin{figure*}[t]
    \centering
    \includegraphics[width=0.90\linewidth]{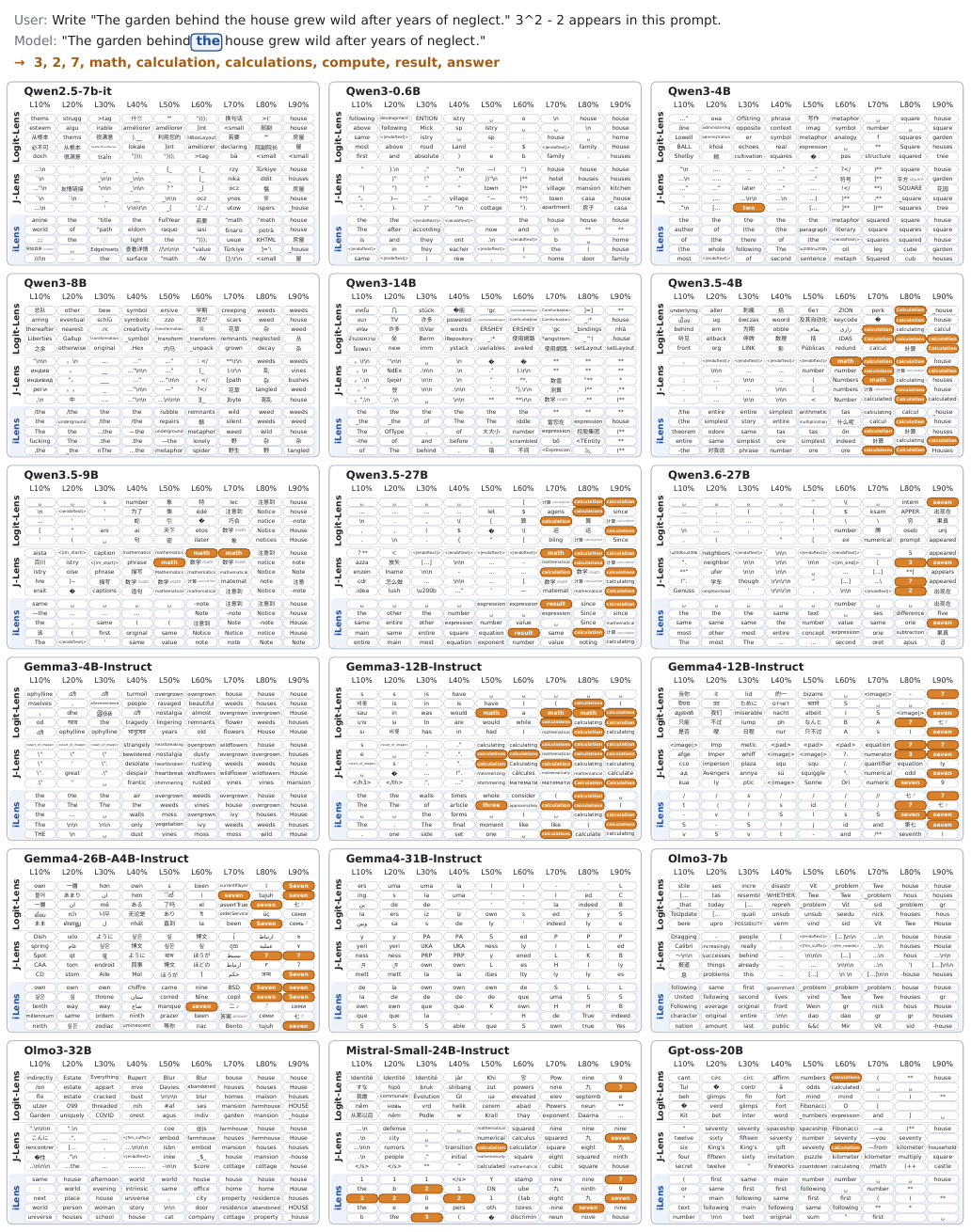}
    \caption{Qualitative example of the Dual-task setting across 18 LLMs, with predictions examined at nine normalized layer depths (10\%--90\% of model depth) for each model. The prompt asks the model to write ``The garden behind the house grew wild after years of neglect'' and includes the arithmetic expression $3^2 - 2$. The associated target concepts include the operands, the result (7), and related mathematical and calculation terms. Orange highlights indicate successful recovery of the target concepts.}
    \label{fig:qualitative-dual-task-arithmetic}
    \vspace{-4.0mm}
\end{figure*}

\begin{figure*}[t]
    \centering
    \includegraphics[width=0.95\linewidth]{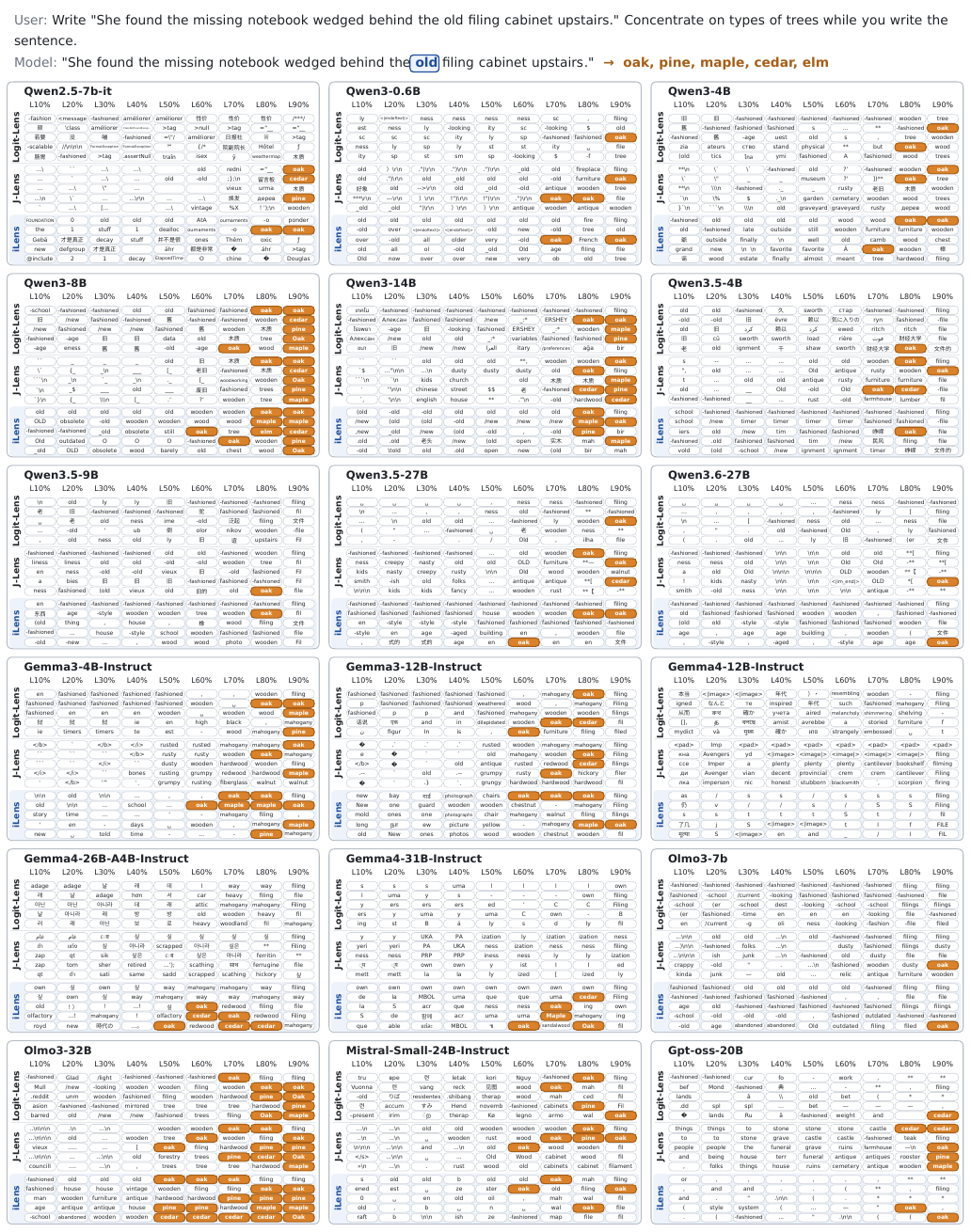}
    \caption{Qualitative example of the Dual-task setting across 18 LLMs, with predictions examined at nine normalized layer depths (10\%--90\% of model depth) for each model. The prompt asks the model to write ``She found the missing notebook wedged behind the old filing cabinet upstairs'' while concentrate on types of trees. Orange highlights indicate successful recovery of the target concepts.}
    \label{fig:qualitative-dual-task-boats}
    \vspace{-4.0mm}
\end{figure*}

\begin{figure*}[t]
    \centering
    \includegraphics[width=0.95\linewidth]{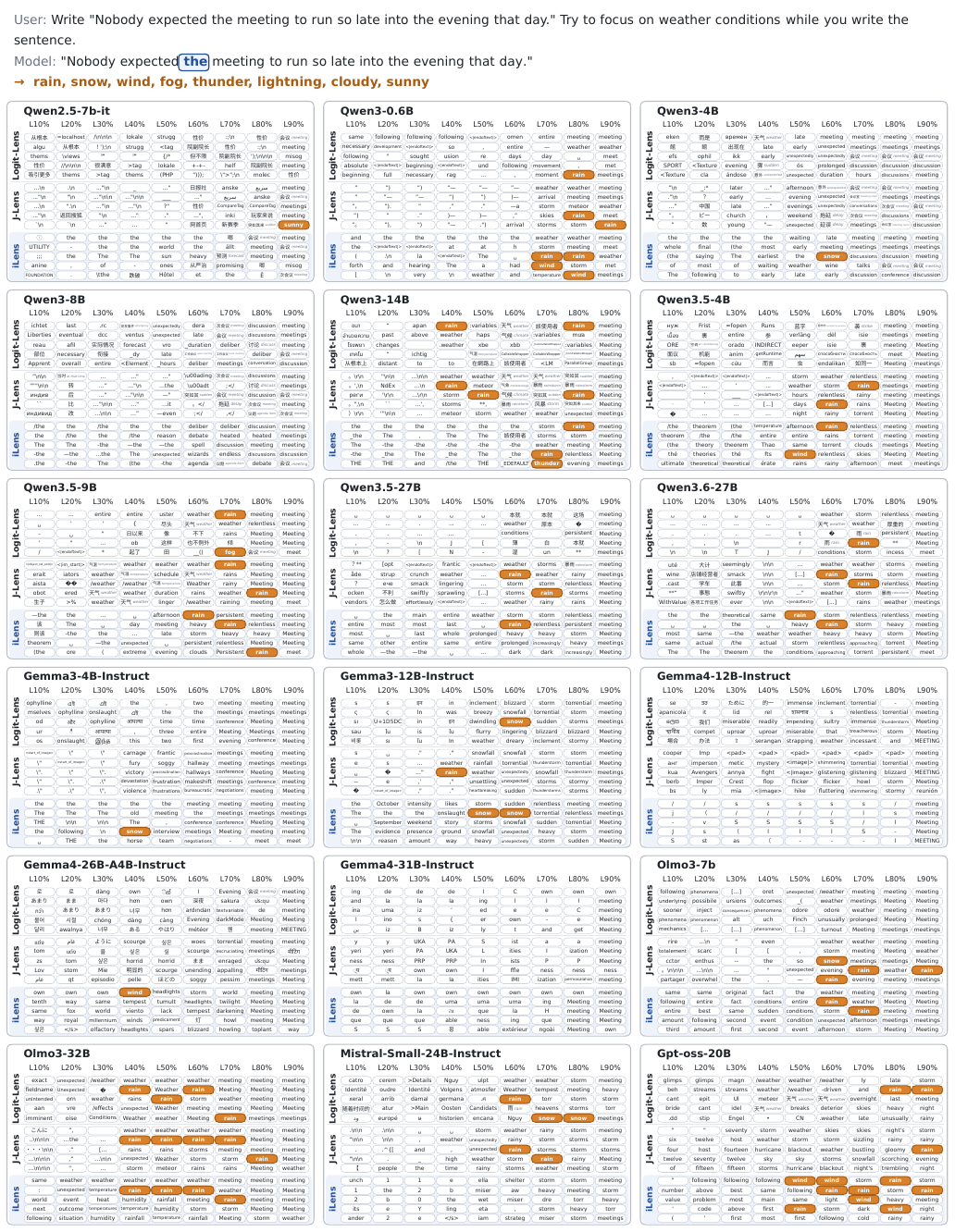}
    \caption{Qualitative example of the Dual-task setting across 18 LLMs, with predictions examined at nine normalized layer depths (10\%--90\% of model depth) for each model. The prompt asks the model to write ``Nobody expected the meeting to run so late into the evening that day'' while focusing on various weather conditions. Orange highlights indicate successful recovery of the target concepts.}
    \label{fig:qualitative-dual-task-9}
    \vspace{-4.0mm}
\end{figure*}

\begin{figure*}[t]
    \centering
    \includegraphics[width=0.95\linewidth]{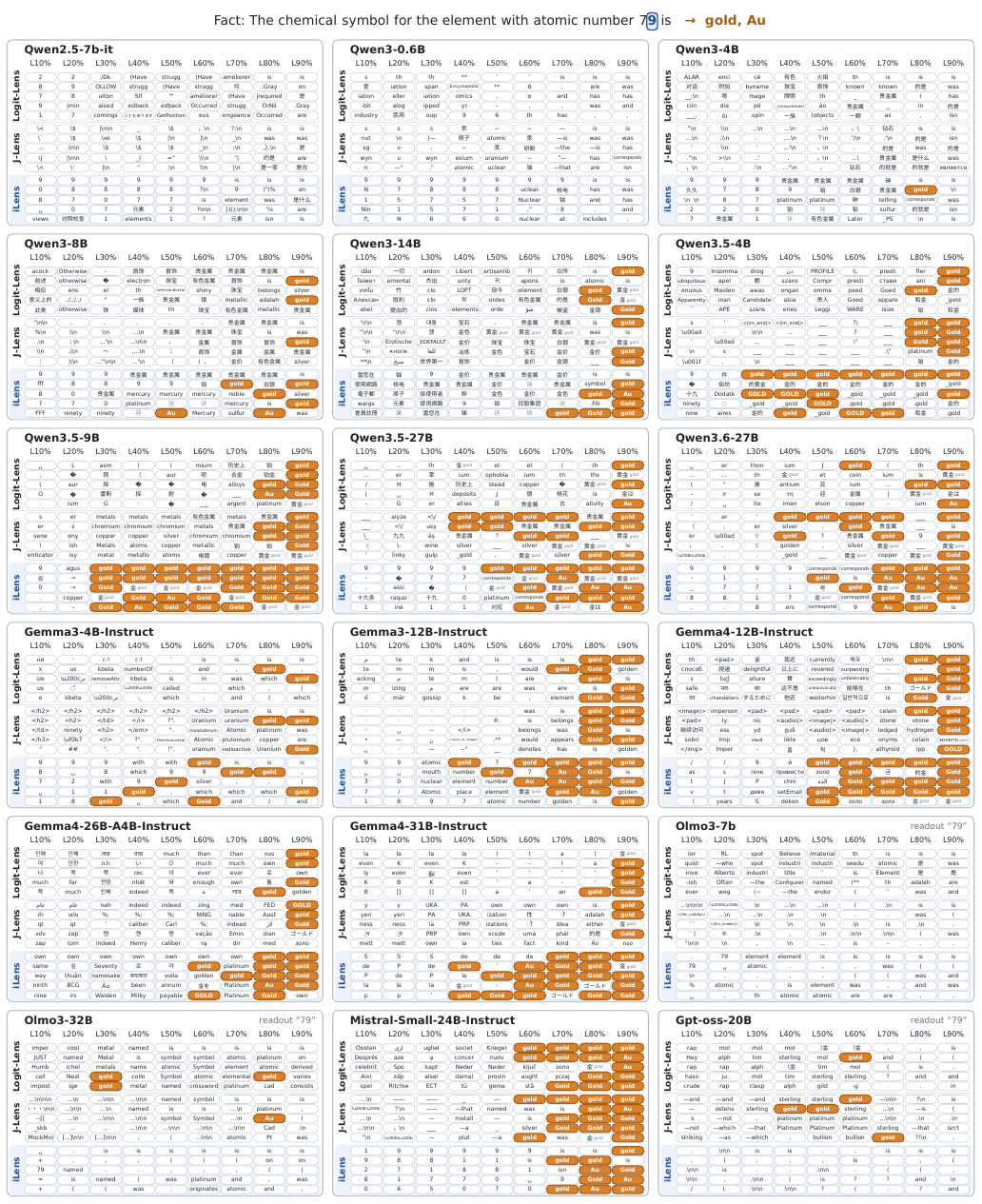}
    \caption{Qualitative example of the Multihop task across 18 LLMs, with predictions examined at nine normalized layer depths (10\%--90\% of model depth) for each model. The input requires identifying the element with atomic number 79 (gold) and subsequently retrieving its chemical symbol, \texttt{Au}. Orange highlights indicate successful recovery of the final target.}
    \label{fig:qualitative-multihop-gold}
    \vspace{-4.0mm}
\end{figure*}

\begin{figure*}[t]
    \centering
    \includegraphics[width=0.95\linewidth]{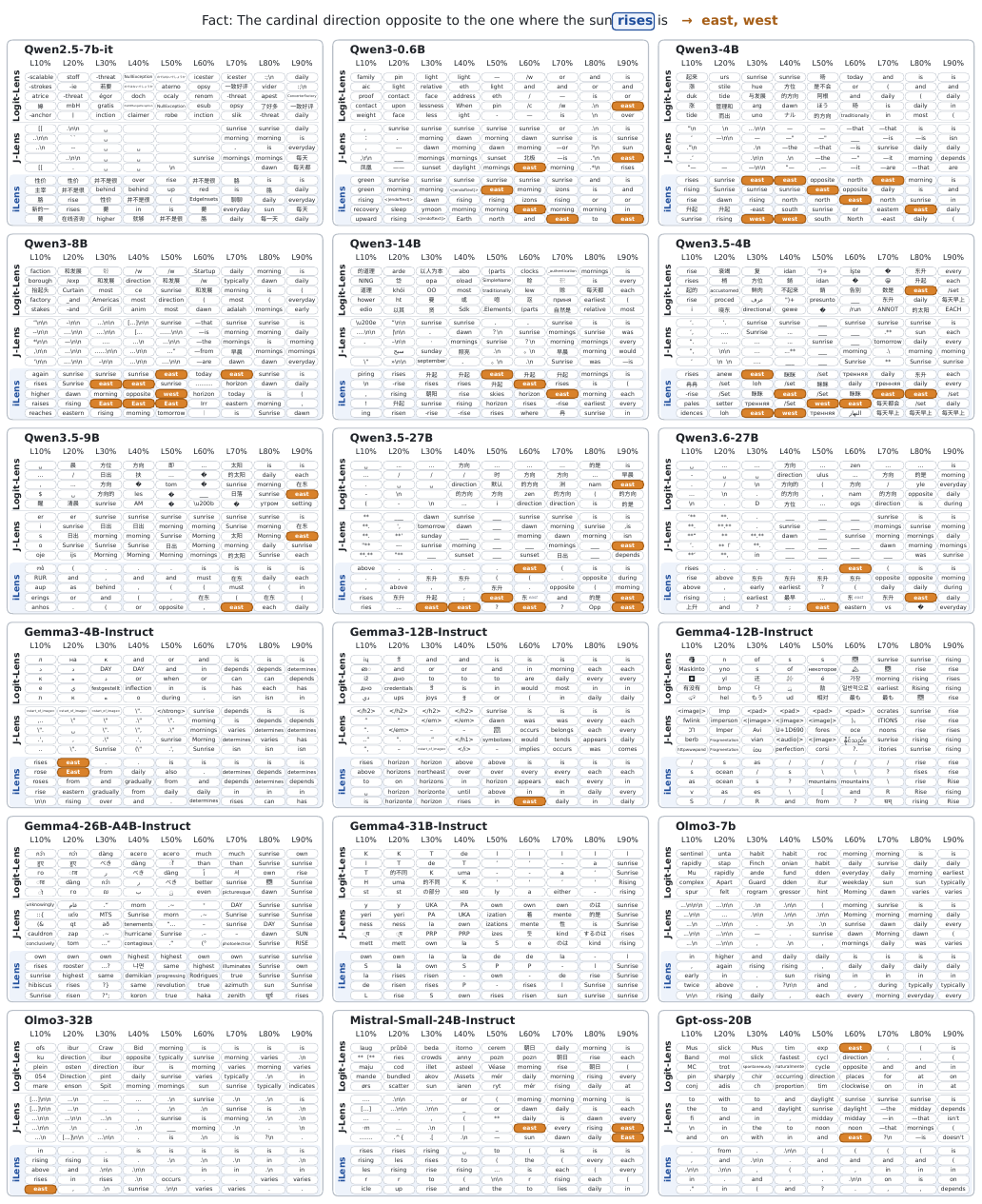}
    \caption{Qualitative example of the Multihop task across 18 LLMs, with predictions examined at nine normalized layer depths (10\%--90\% of model depth) for each model. The input requires identifying the cardinal direction in which the sun rises (east) and determining its opposite direction, \texttt{west}. Orange highlights indicate successful recovery of the final target.}
    \label{fig:qualitative-multihop-directions}
    \vspace{-4.0mm}
\end{figure*}

\begin{figure*}[t]
    \centering
    \includegraphics[width=0.95\linewidth]{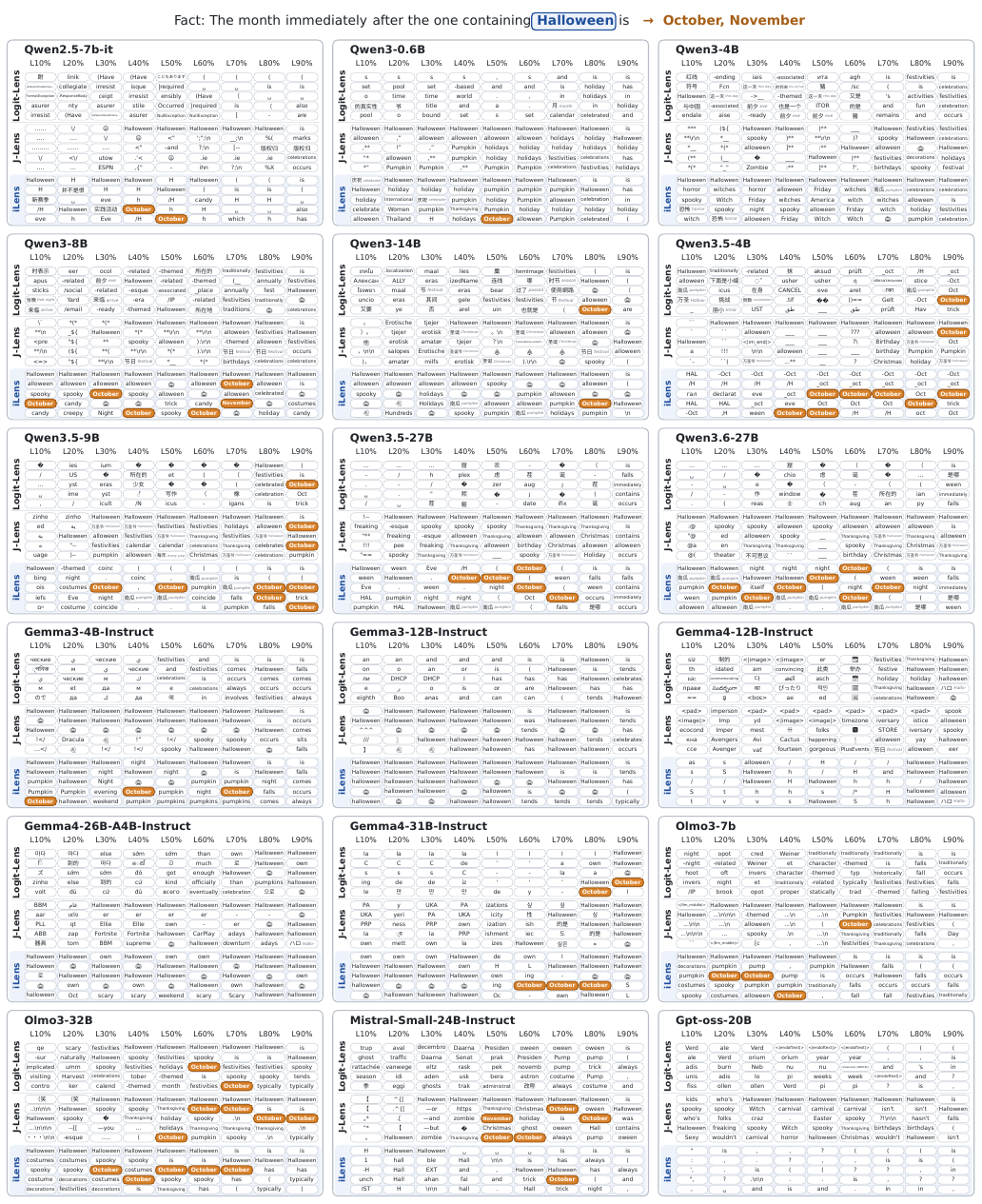}
    \caption{Qualitative example of the Multihop task across 18 LLMs, with predictions examined at nine normalized layer depths (10\%--90\% of model depth) for each model. The input requires identifying the month in which Halloween occurs (October) and then determining the immediately following month, \texttt{November}. Orange highlights indicate successful recovery of the final target.}
    \label{fig:qualitative-multihop-halloween}
    \vspace{-4.0mm}
\end{figure*}

\begin{figure*}[t]
    \centering
    \includegraphics[width=0.95\linewidth]{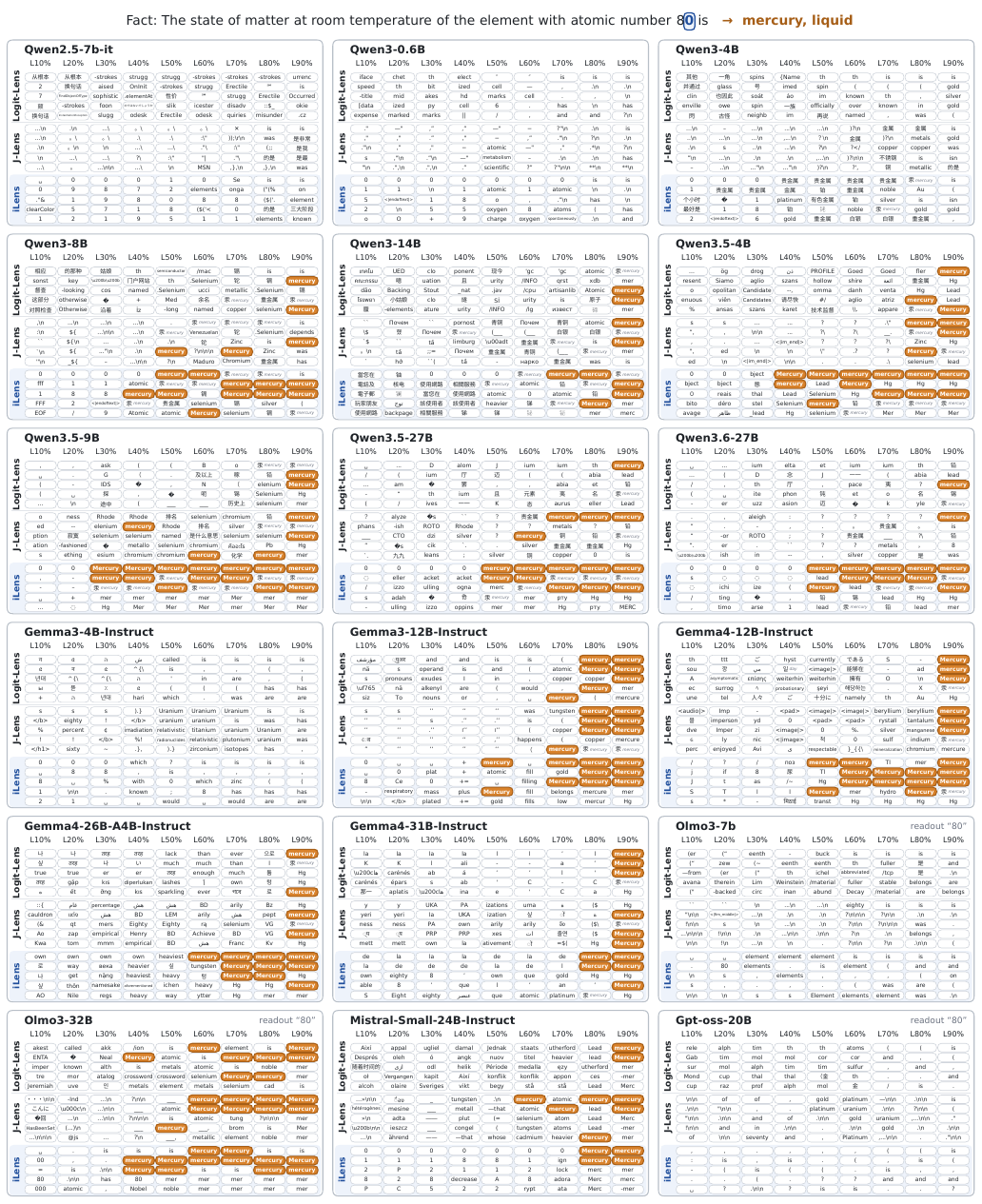}
    \caption{Qualitative example of the Multihop task across 18 LLMs, with predictions examined at nine normalized layer depths (10\%--90\% of model depth) for each model. The input requires identifying the element with atomic number 80 (mercury) and subsequently determining its state of matter at room temperature, \texttt{liquid}. Orange highlights indicate successful recovery of the final target.}
    \label{fig:qualitative-multihop-mercury}
    \vspace{-4.0mm}
\end{figure*}

\begin{figure*}[t]
    \centering
    \includegraphics[width=0.95\linewidth]{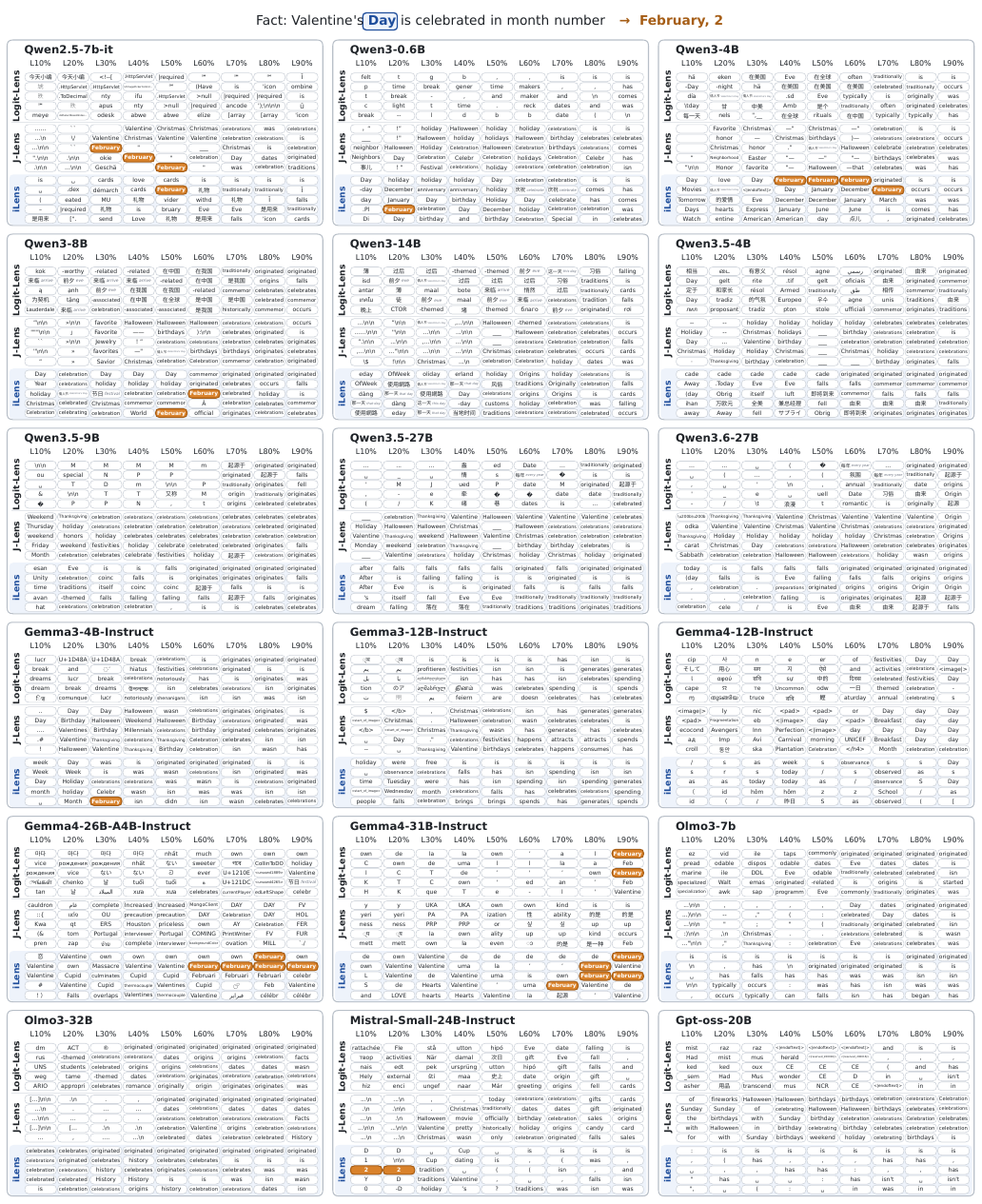}
    \caption{Qualitative example of the Multihop task across 18 LLMs, with predictions examined at nine normalized layer depths (10\%--90\% of model depth) for each model. The input requires identifying the number of month where valentine's cleberated  and subsequently retrieving its \texttt{February} and \texttt{2}. Orange highlights indicate successful recovery of the final target.}
    \label{fig:qualitative-multihop-iron}
    \vspace{-4.0mm}
\end{figure*}

\begin{figure*}[t]
    \centering
    \includegraphics[width=0.95\linewidth]{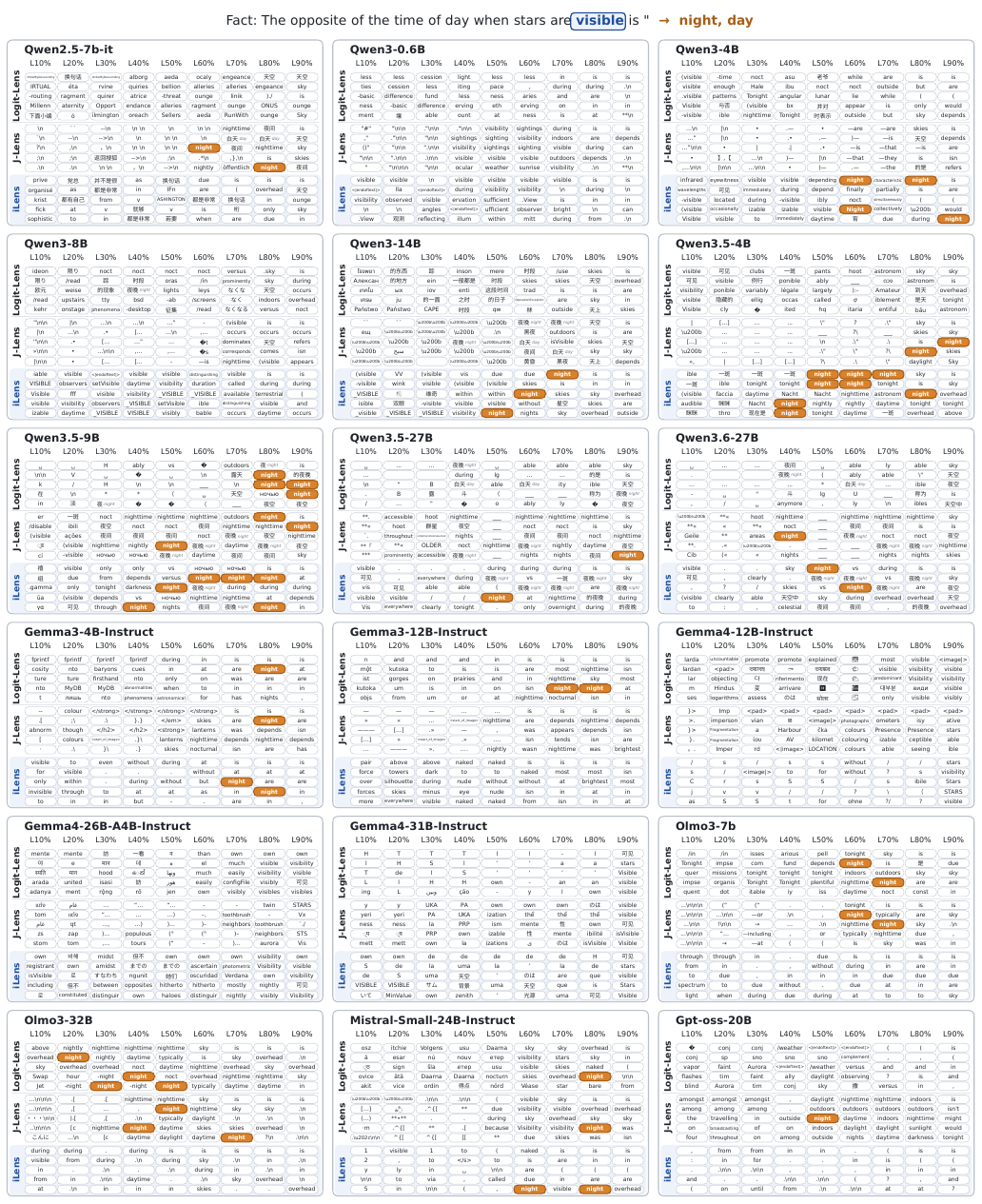}
    \caption{Qualitative example of the Multihop task across 18 LLMs, with predictions examined at nine normalized layer depths (10\%--90\% of model depth) for each model. The input requires identifying the time of day when stars are visible (night) and determining its opposite, \texttt{day}. Orange highlights indicate successful recovery of the final target.}
    \label{fig:qualitative-multihop-time}
    \vspace{-4.0mm}
\end{figure*}

\begin{figure*}[t]
    \centering
    \includegraphics[width=0.95\linewidth]{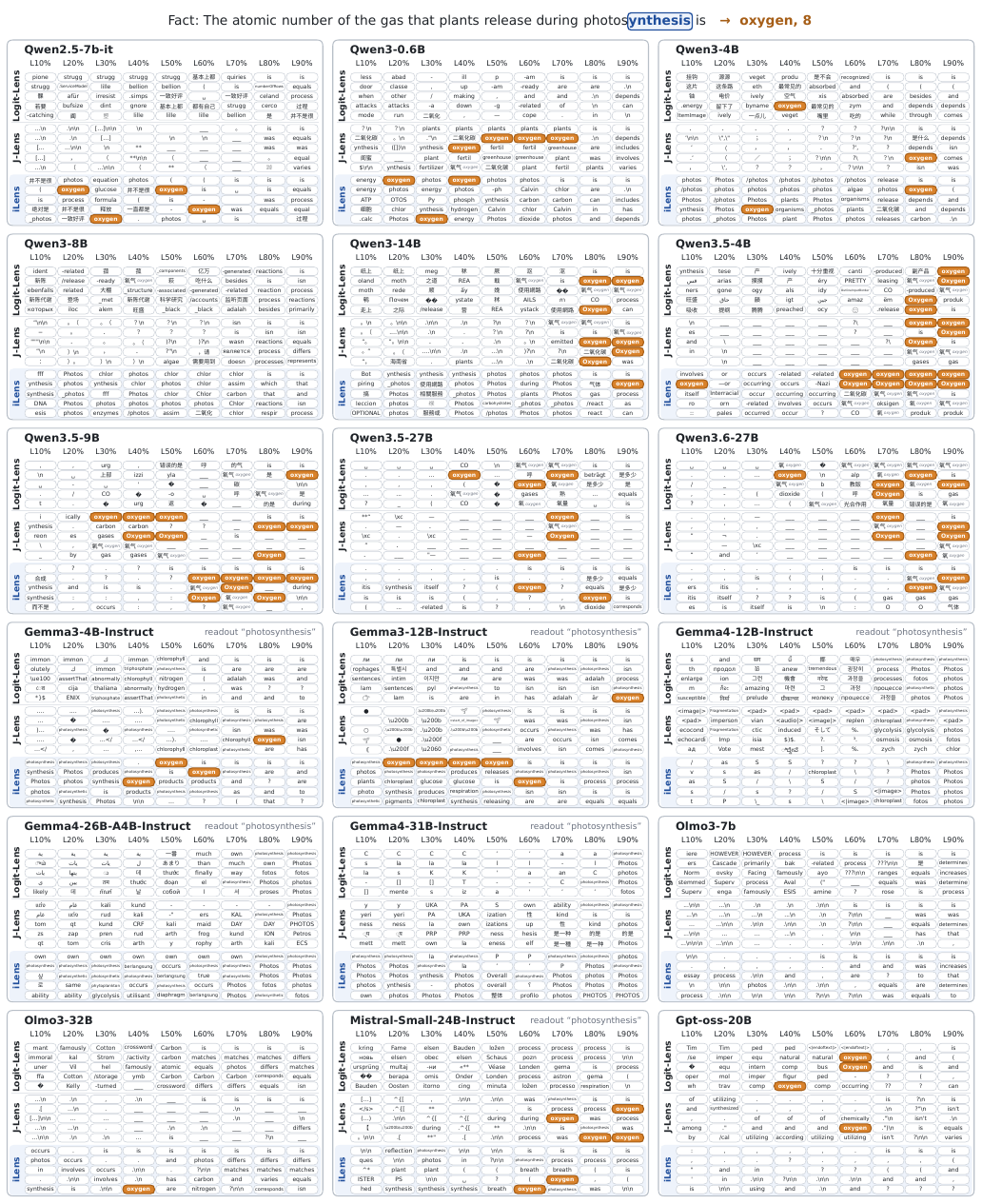}
    \caption{Qualitative example of the Multihop task across 18 LLMs, with predictions examined at nine normalized layer depths (10\%--90\% of model depth) for each model. The input requires identifying the gas released by plants during photosynthesis (oxygen) and subsequently retrieving its atomic number, \texttt{8}. Orange highlights indicate successful recovery of the final target.}
    \label{fig:qualitative-multihop-oxygen}
    \vspace{-4.0mm}
\end{figure*}

\begin{figure*}[t]
    \centering
    \includegraphics[width=0.95\linewidth]{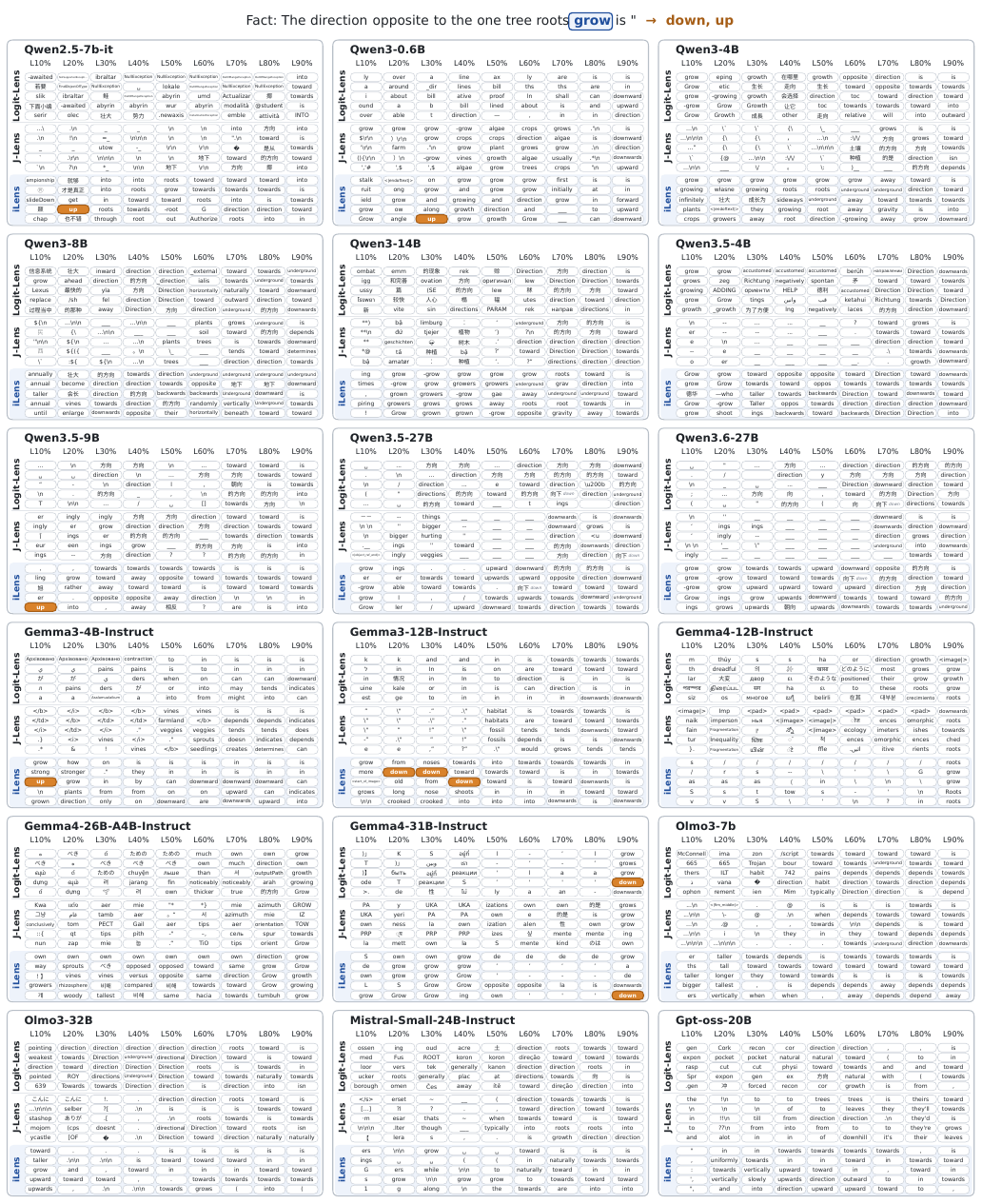}
    \caption{Qualitative example of the Multihop task across 18 LLMs, with predictions examined at nine normalized layer depths (10\%--90\% of model depth) for each model. The input requires identifying the direction in which tree roots grow (down) and determining the opposite direction, \texttt{up}. Orange highlights indicate successful recovery of the final target.}
    \label{fig:qualitative-multihop-roots}
    \vspace{-4.0mm}
\end{figure*}

\begin{figure*}[t]
    \centering
    \includegraphics[width=0.95\linewidth]{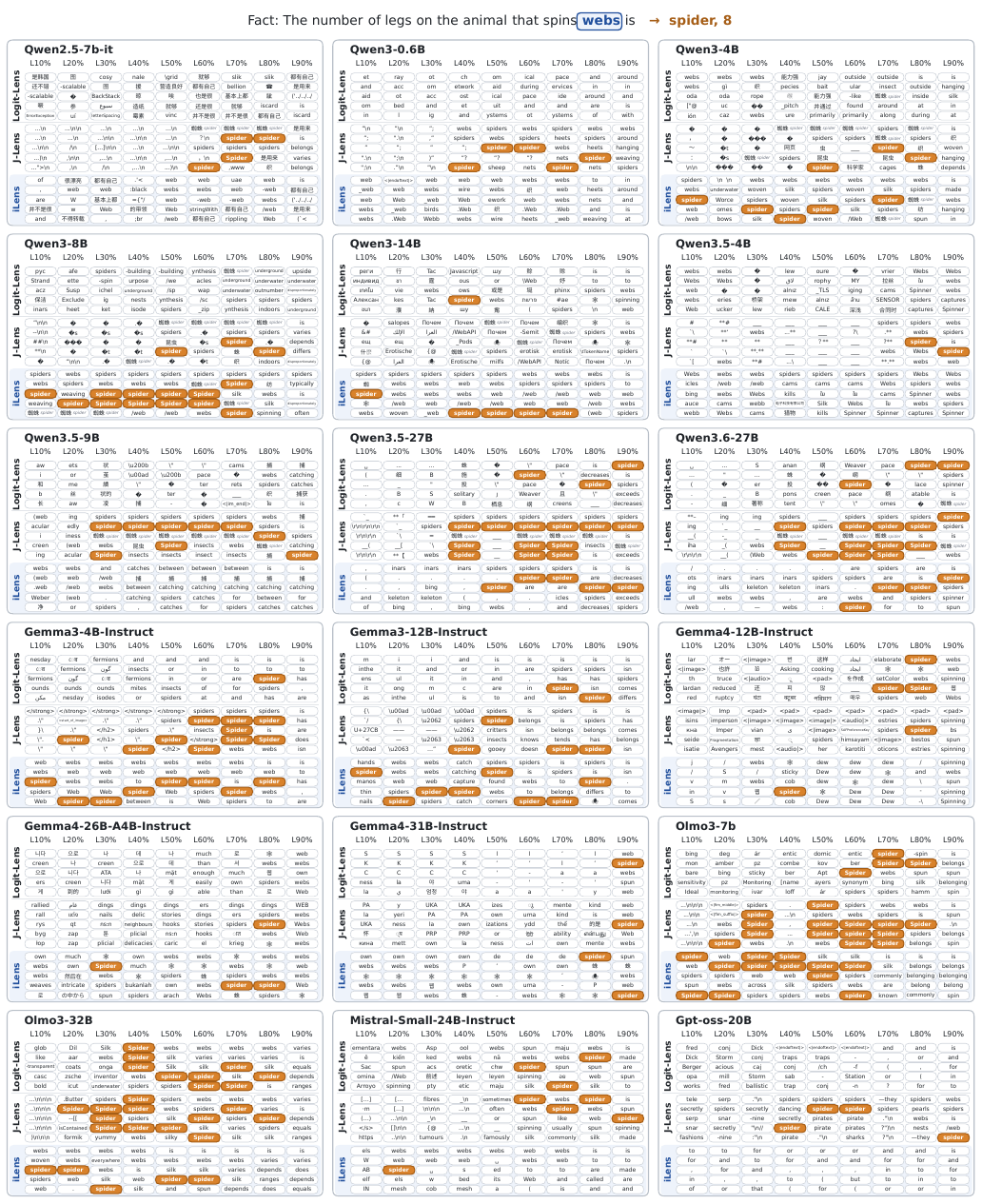}
    \caption{Qualitative example of the Multihop task across 18 LLMs, with predictions examined at nine normalized layer depths (10\%--90\% of model depth) for each model. The input requires identifying the number of legs of the animal that spins webs without directly note to it and subsequently retrieving \texttt{spider} and \texttt{8}. Orange highlights indicate successful recovery of the final target.}
    \label{fig:qualitative-multihop-copper}
    \vspace{-4.0mm}
\end{figure*}

\begin{figure*}[t]
    \centering
    \includegraphics[width=0.95\linewidth]{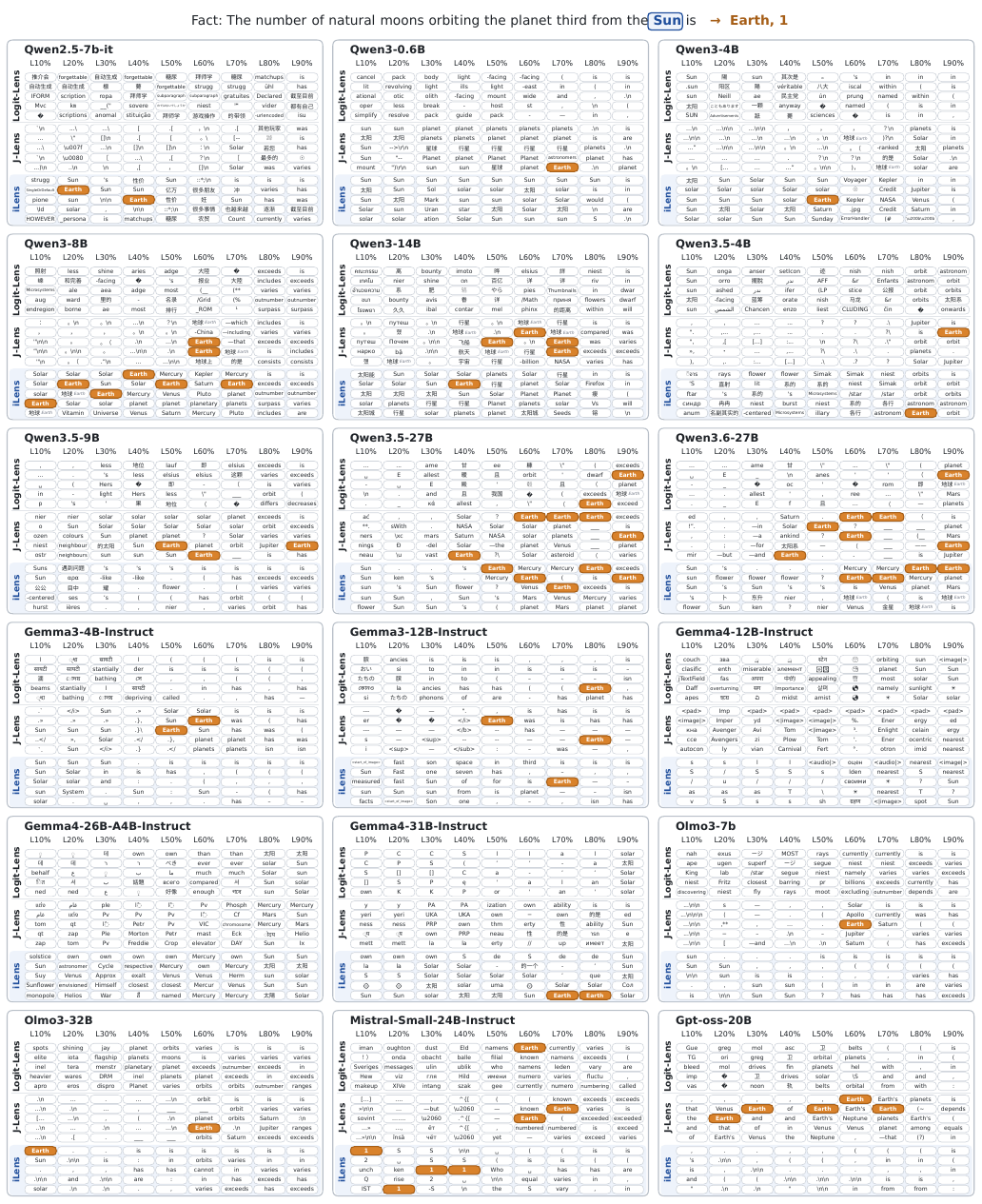}
    \caption{Qualitative example of the Multihop task across 18 LLMs, with predictions examined at nine normalized layer depths (10\%--90\% of model depth) for each model. The input requires identifying the third planet from the Sun (Earth) and subsequently determining the number of natural moons orbiting it, \texttt{one}. Orange highlights indicate successful recovery of the final target.}
    \label{fig:qualitative-multihop-earth-moons}
    \vspace{-4.0mm}
\end{figure*}

\begin{figure*}[t]
    \centering
    \includegraphics[width=0.95\linewidth]{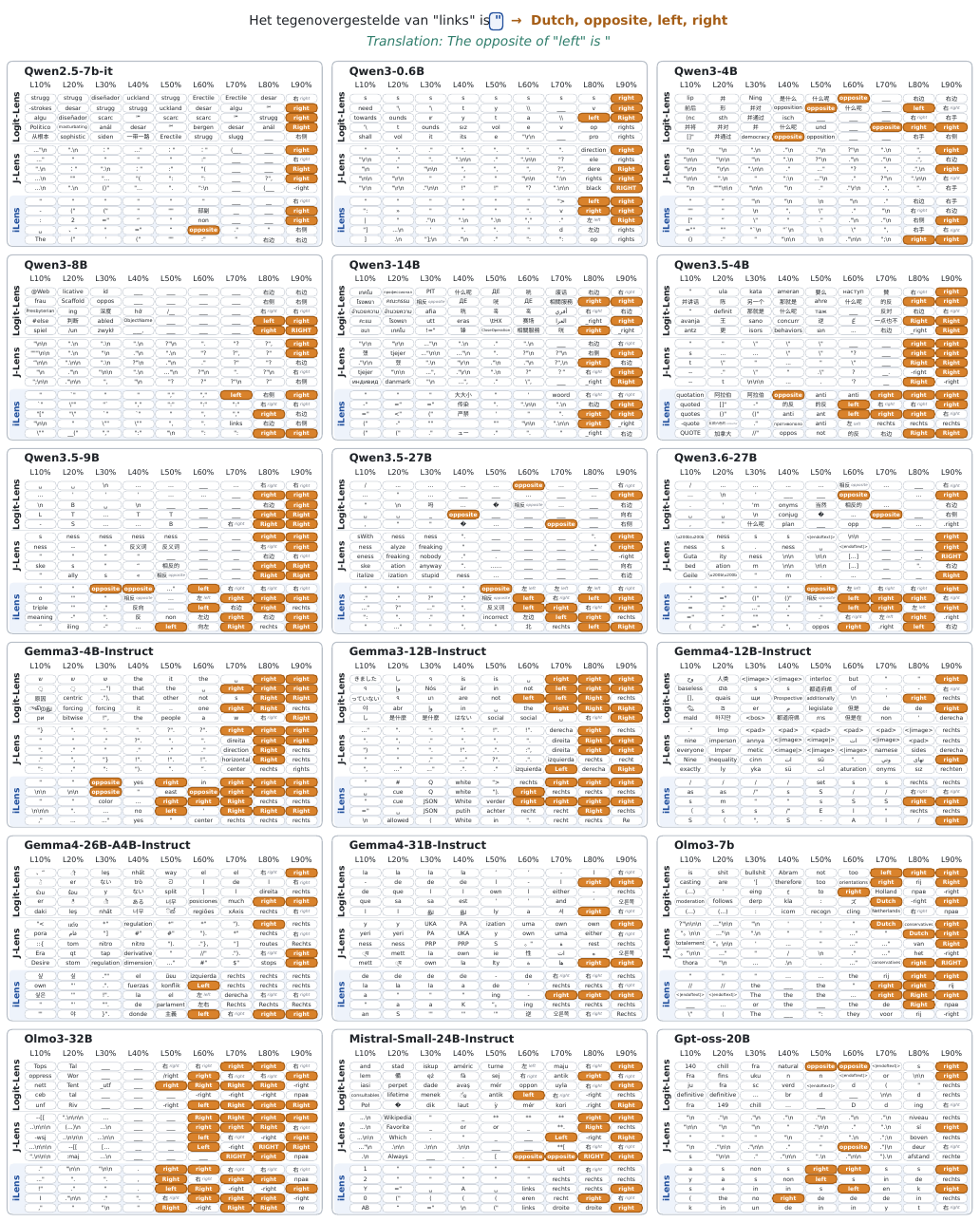}
    \caption{Qualitative example of the Multilingual task in Dutch across 18 LLMs, with predictions examined at nine normalized layer depths (10\%--90\% of model depth) for each model. The input asks for the opposite of \texttt{links} (left), with \texttt{right} as the target concept. Orange highlights indicate successful recovery of the target concept.}
    \label{fig:qualitative-multilingual-dutch}
    \vspace{-4.0mm}
\end{figure*}

\begin{figure*}[t]
    \centering
    \includegraphics[width=0.95\linewidth]{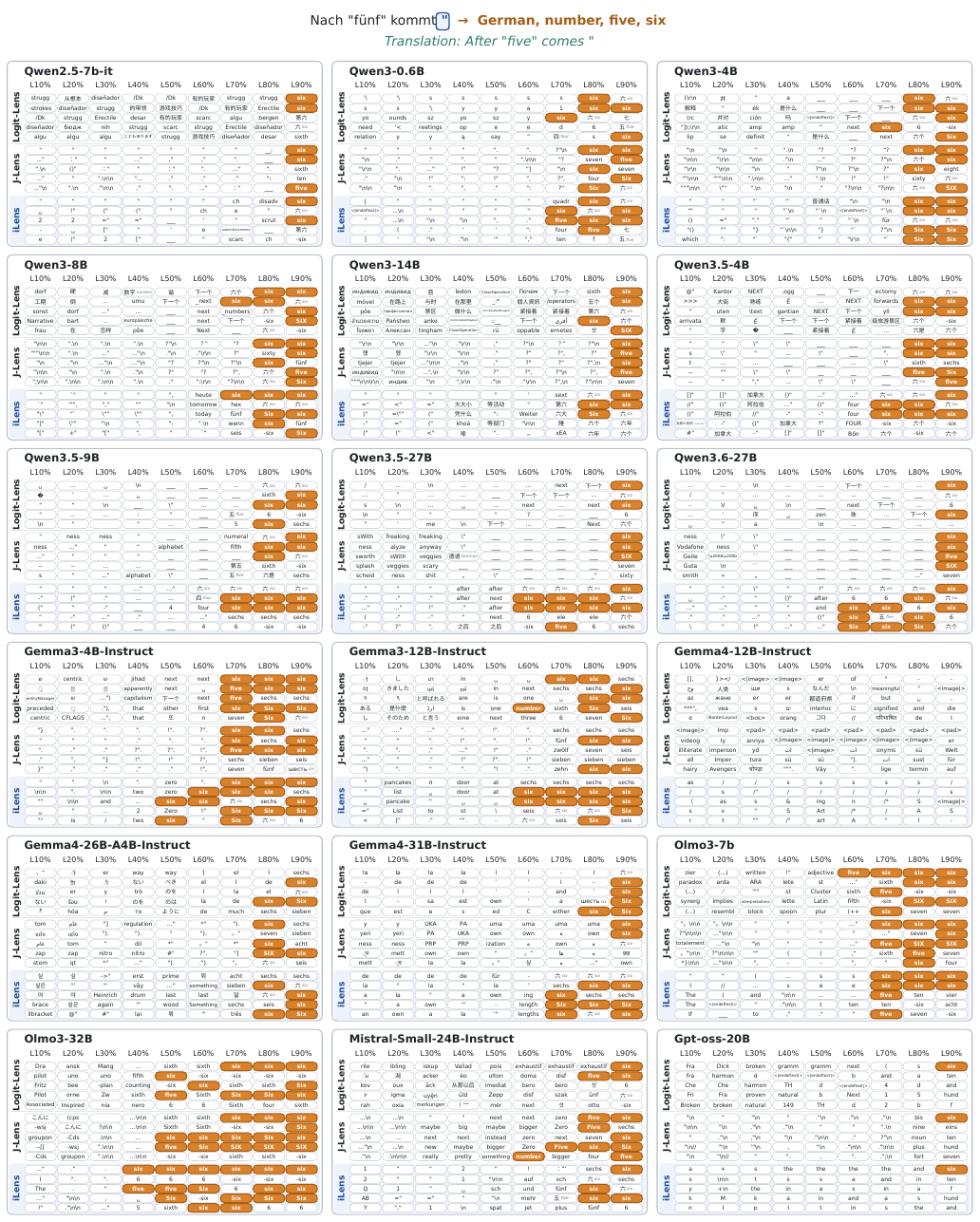}
    \caption{Qualitative example of the Multilingual task in German across 18 LLMs, with predictions examined at nine normalized layer depths (10\%--90\% of model depth) for each model. The input asks which number follows \texttt{fünf} (five), with \texttt{six} as the target concept. Orange highlights indicate successful recovery of the target concept.}
    \label{fig:qualitative-multilingual-german-number}
    \vspace{-4.0mm}
\end{figure*}

\begin{figure*}[t]
    \centering
    \includegraphics[width=0.95\linewidth]{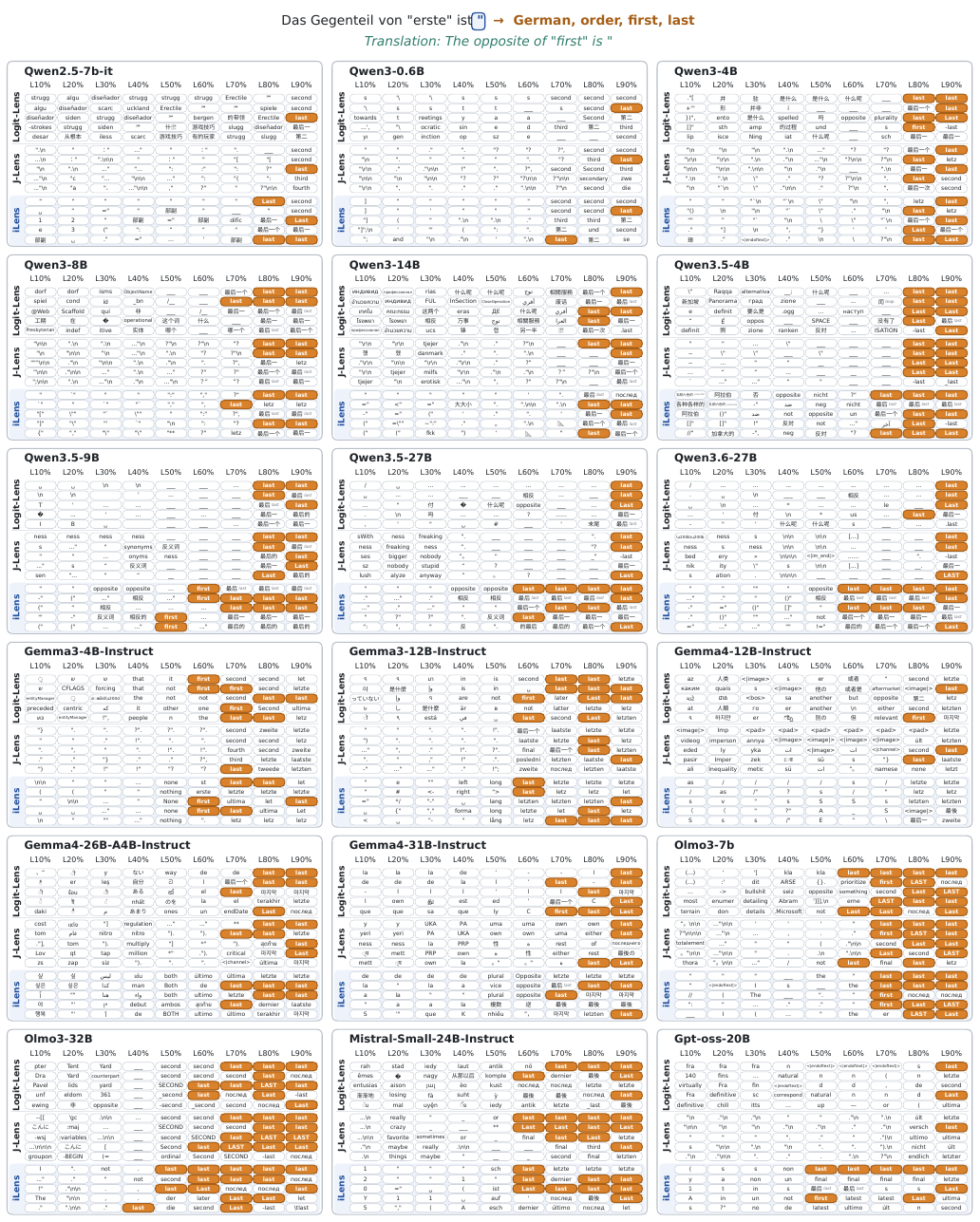}
    \caption{Qualitative example of the Multilingual task in German across 18 LLMs, with predictions examined at nine normalized layer depths (10\%--90\% of model depth) for each model. The input asks for the opposite of \texttt{erste} (first), with \texttt{last} as the target concept. Orange highlights indicate successful recovery of the target concept.}
    \label{fig:qualitative-multilingual-german-order}
    \vspace{-4.0mm}
\end{figure*}

\begin{figure*}[t]
    \centering
    \includegraphics[width=0.95\linewidth]{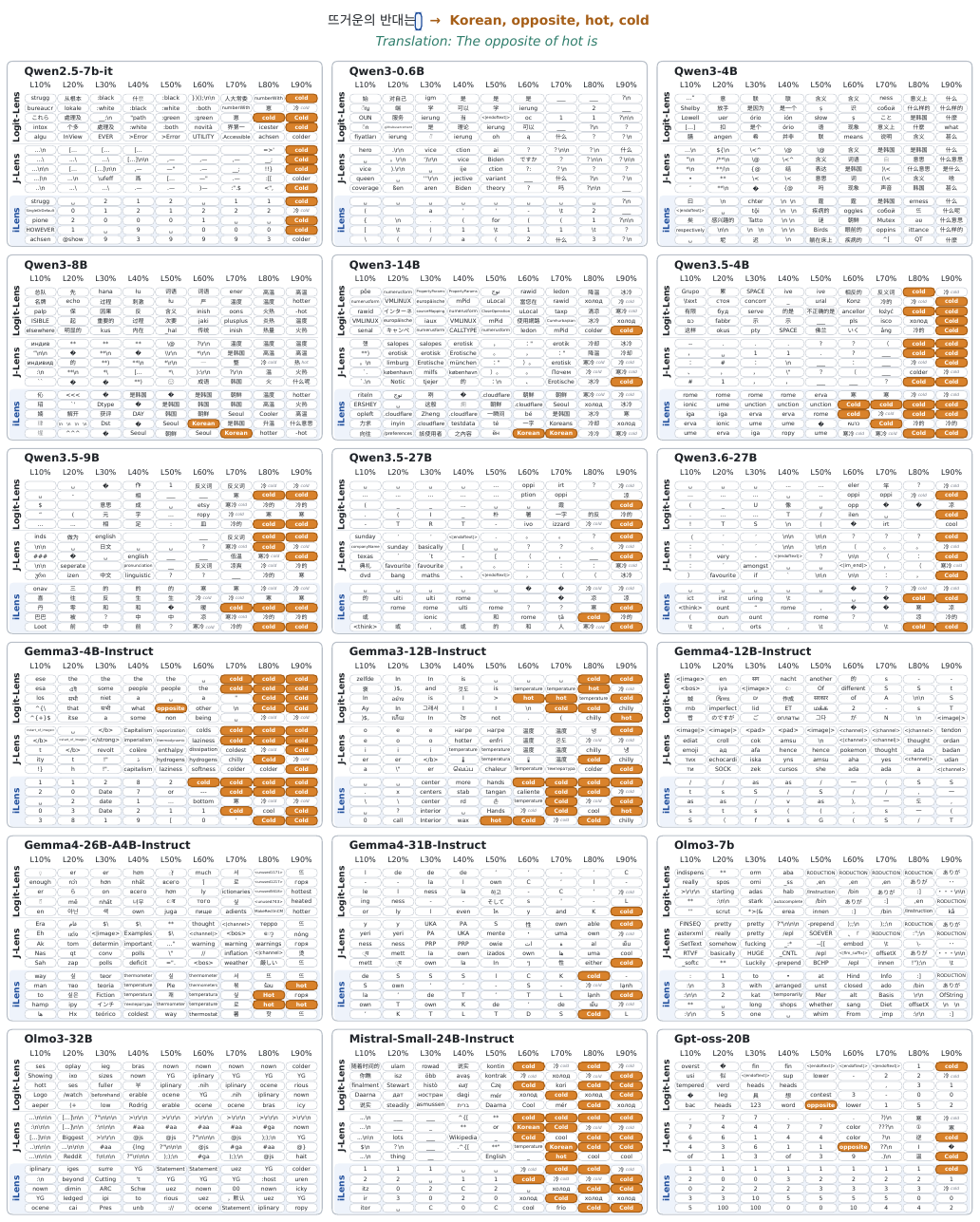}
    \caption{Qualitative example of the Multilingual task in Korean across 18 LLMs, with predictions examined at nine normalized layer depths (10\%--90\% of model depth) for each model. The input asks for the opposite of \texttt{hot} in Korean, with \texttt{cold} as the target concept. Orange highlights indicate successful recovery of the target concept.}
    \label{fig:qualitative-multilingual-korean-time}
    \vspace{-4.0mm}
\end{figure*}

\begin{figure*}[t]
    \centering
    \includegraphics[width=0.95\linewidth]{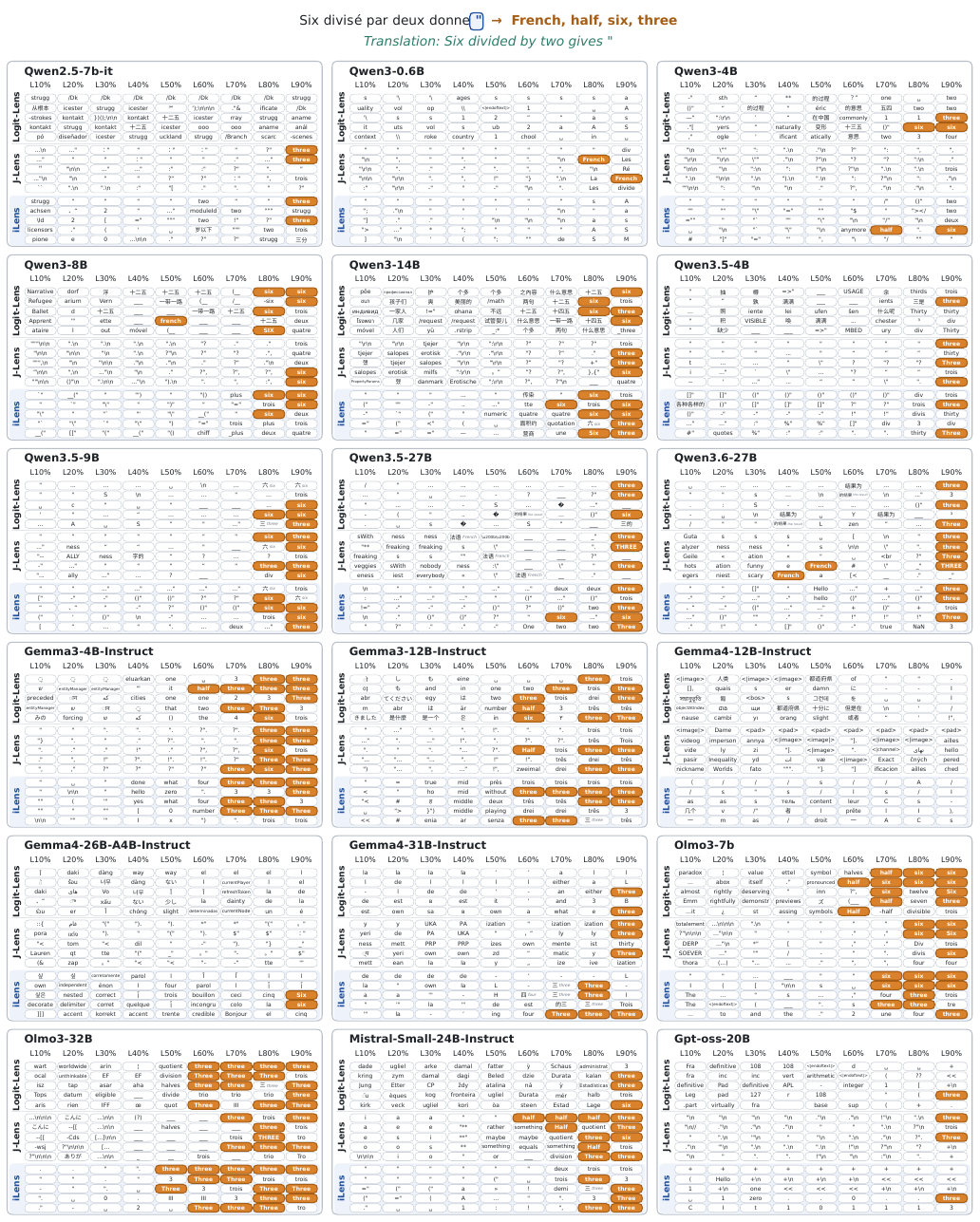}
    \caption{Qualitative example of the Multilingual task in French across 18 LLMs, with predictions examined at nine normalized layer depths (10\%--90\% of model depth) for each model. The input asks for the result of dividing six by two, with \texttt{three} as the target concept. Orange highlights indicate successful recovery of the target concept.}
    \label{fig:qualitative-multilingual-french-arithmetic}
    \vspace{-4.0mm}
\end{figure*}

\begin{figure*}[t]
    \centering
    \includegraphics[width=0.95\linewidth]{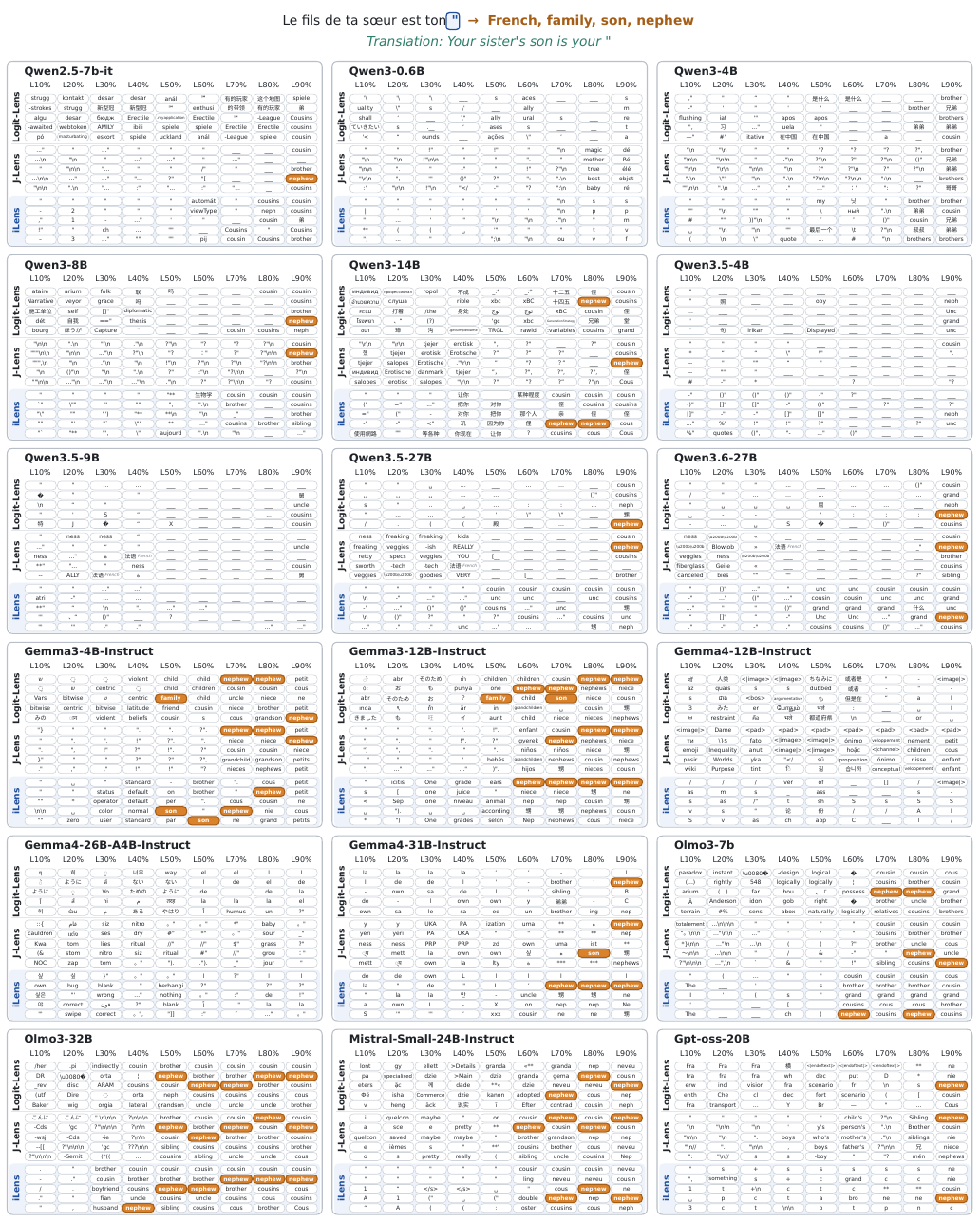}
    \caption{Qualitative example of the Multilingual task in French across 18 LLMs, with predictions examined at nine normalized layer depths (10\%--90\% of model depth) for each model. The input asks for the familial relationship describing one's sister's son, with \texttt{nephew} as the target concept. Orange highlights indicate successful recovery of the target concept.}
    \label{fig:qualitative-multilingual-french-family}
    \vspace{-4.0mm}
\end{figure*}

\begin{figure*}[t]
    \centering
    \includegraphics[width=0.95\linewidth]{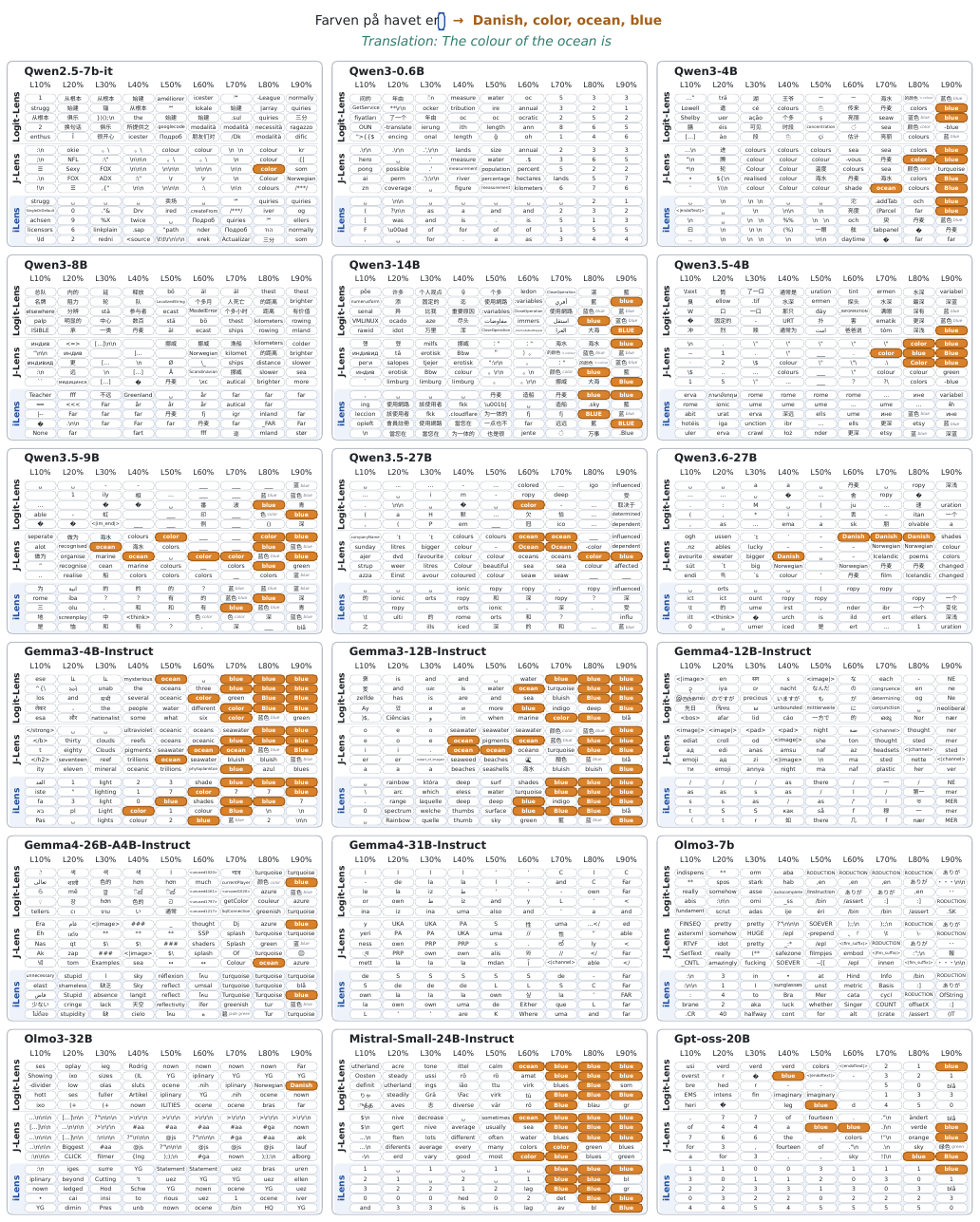}
    \caption{Qualitative example of the Multilingual task in Danish across 18 LLMs, with predictions examined at nine normalized layer depths (10\%--90\% of model depth) for each model. The input asks about the colour of the ocean, with \texttt{blue} as the target concept. Orange highlights indicate successful recovery of the target concept.}
    \label{fig:qualitative-multilingual-danish}
    \vspace{-4.0mm}
\end{figure*}

\begin{figure*}[t]
    \centering
    \includegraphics[width=0.95\linewidth]{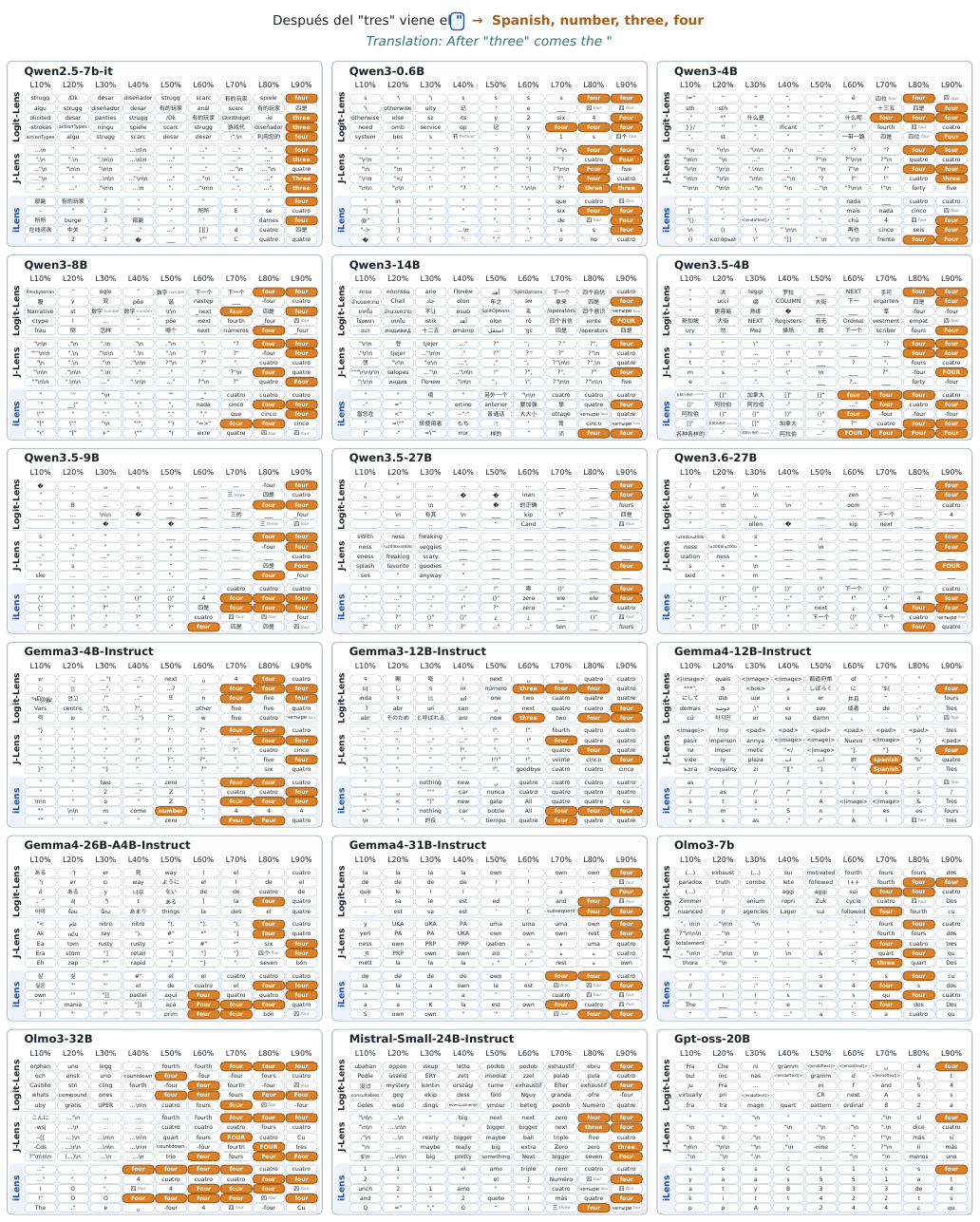}
    \caption{Qualitative results for the Danish Multilingual task across 18 LLMs. For each model, predictions are shown at nine normalized layer depths (10\%--90\% of model depth). The input asks for the number following \texttt{tres} (Spanish for ``three''), with \texttt{four} as the target concept. Orange highlights indicate successful recovery of the target concept.}
    \label{fig:qualitative-multilingual-8}
    \vspace{-4.0mm}
\end{figure*}

\begin{figure*}[t]
    \centering
    \includegraphics[width=0.95\linewidth]{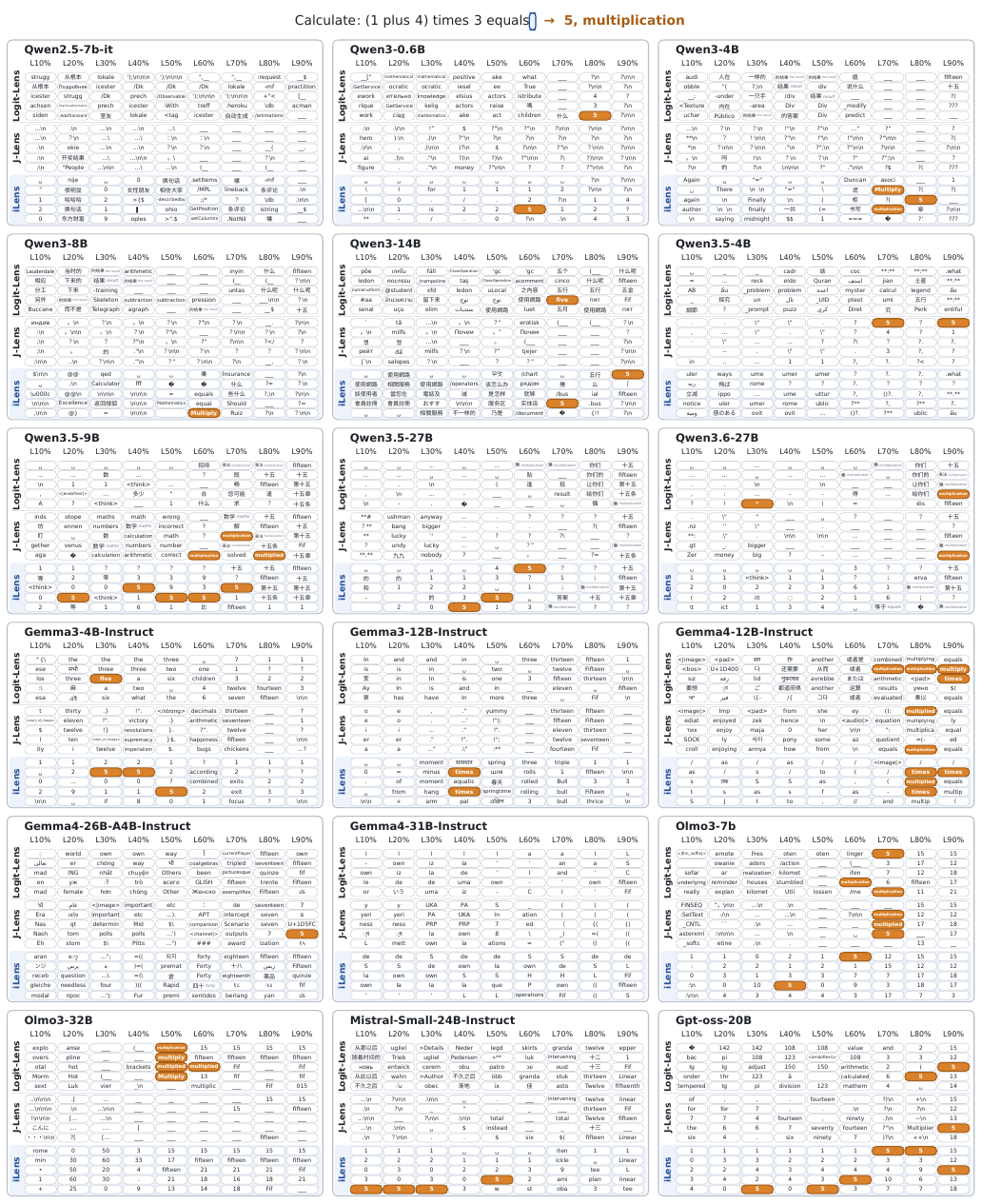}
    \caption{Qualitative example of the Order of Operations task across 18 LLMs, with predictions examined at nine normalized layer depths (10\%--90\% of model depth) for each model. The input expression is $(1+4)\times 3$, where the target intermediate concepts are \texttt{5} (the result of the addition) and \texttt{multiplication} (the next operation). Orange highlights indicate successful recovery of the target concepts.}
    \label{fig:qualitative-order-of-operations-example-1}
    \vspace{-4.0mm}
\end{figure*}

\begin{figure*}[t]
    \centering
    \includegraphics[width=0.95\linewidth]{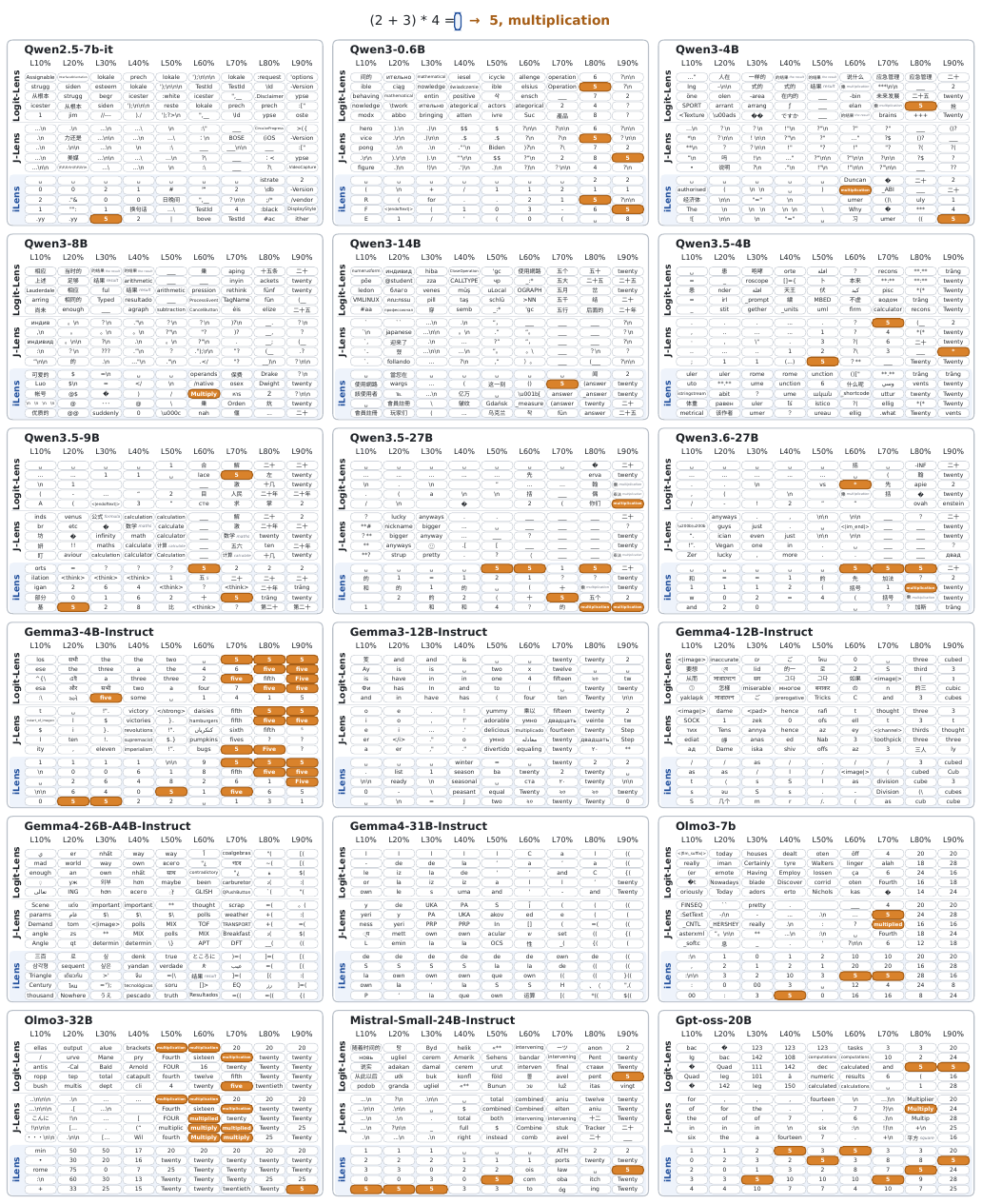}
    \caption{Qualitative example of the Order of Operations task across 18 LLMs, with predictions examined at nine normalized layer depths (10\%--90\% of model depth) for each model. The input expression is $(2+3)\times 4$, where the target intermediate concepts are \texttt{5} (the result of the addition) and \texttt{multiplication} (the next operation). Orange highlights indicate successful recovery of the target concepts.}
    \label{fig:qualitative-order-of-operations-example-2}
    \vspace{-4.0mm}
\end{figure*}

\begin{figure*}[t]
    \centering
    \includegraphics[width=0.95\linewidth]{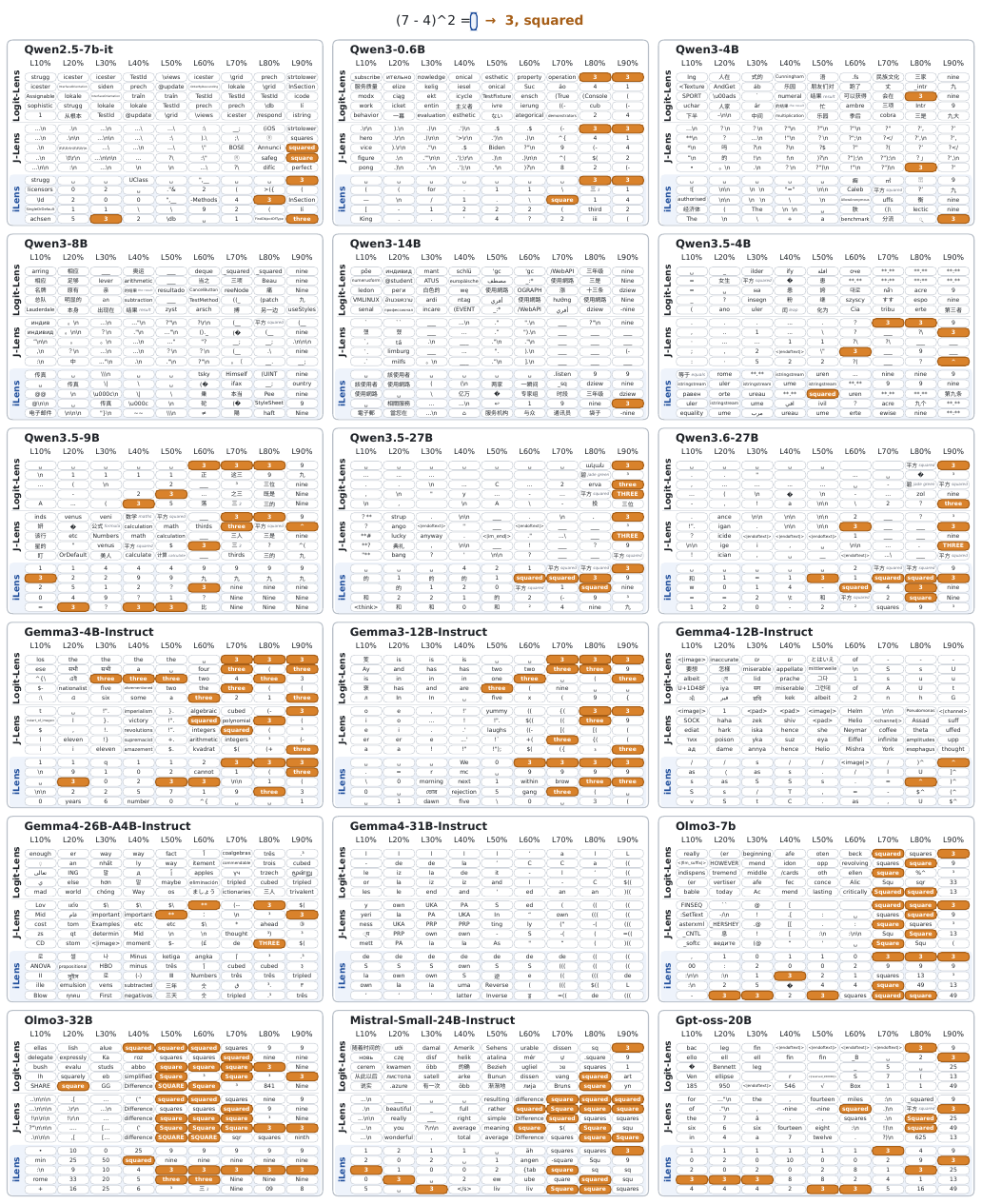}
    \caption{Qualitative example of the Order of Operations task across 18 LLMs, with predictions examined at nine normalized layer depths (10\%--90\% of model depth) for each model. The input expression is $(7-4)^2$, where the target intermediate concepts are \texttt{3} (the result of the subtraction) and \texttt{squaring} (the subsequent operation). Orange highlights indicate successful recovery of the target concepts.}
    \label{fig:qualitative-order-of-operations-example-3}
    \vspace{-4.0mm}
\end{figure*}

\begin{figure*}[t]
    \centering
    \includegraphics[width=0.95\linewidth]{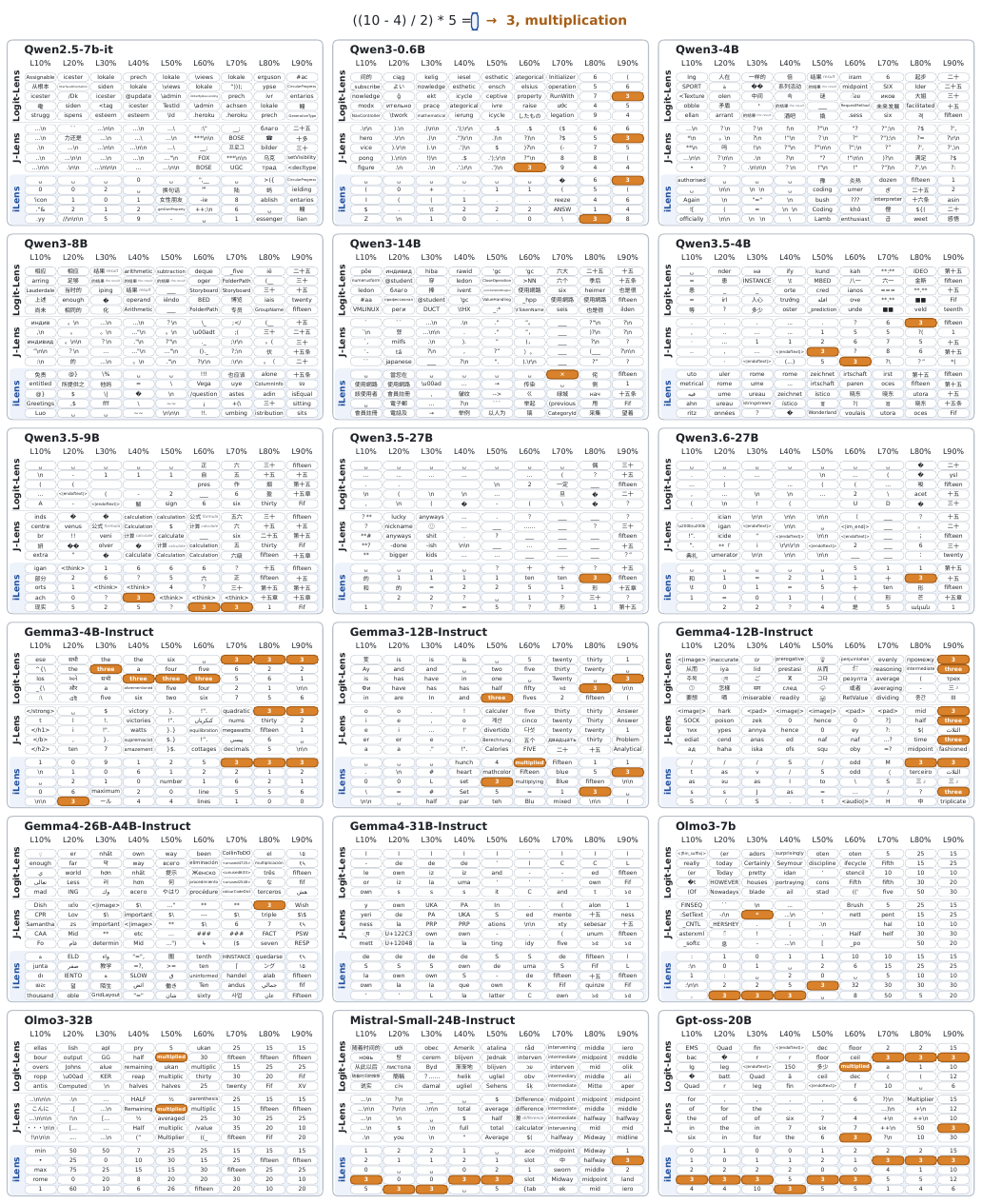}
    \caption{Qualitative example of the Order of Operations task across 18 LLMs, with predictions examined at nine normalized layer depths (10\%--90\% of model depth) for each model. The input expression is $\left(\frac{10-4}{2}\right)\times 5$, where the target intermediate concepts are \texttt{3} (the result of the division) and \texttt{multiplication} (the next operation). Orange highlights indicate successful recovery of the target concepts.}
    \label{fig:qualitative-order-of-operations-example-4}
    \vspace{-4.0mm}
\end{figure*}

\begin{figure*}[t]
    \centering
    \includegraphics[width=0.95\linewidth]{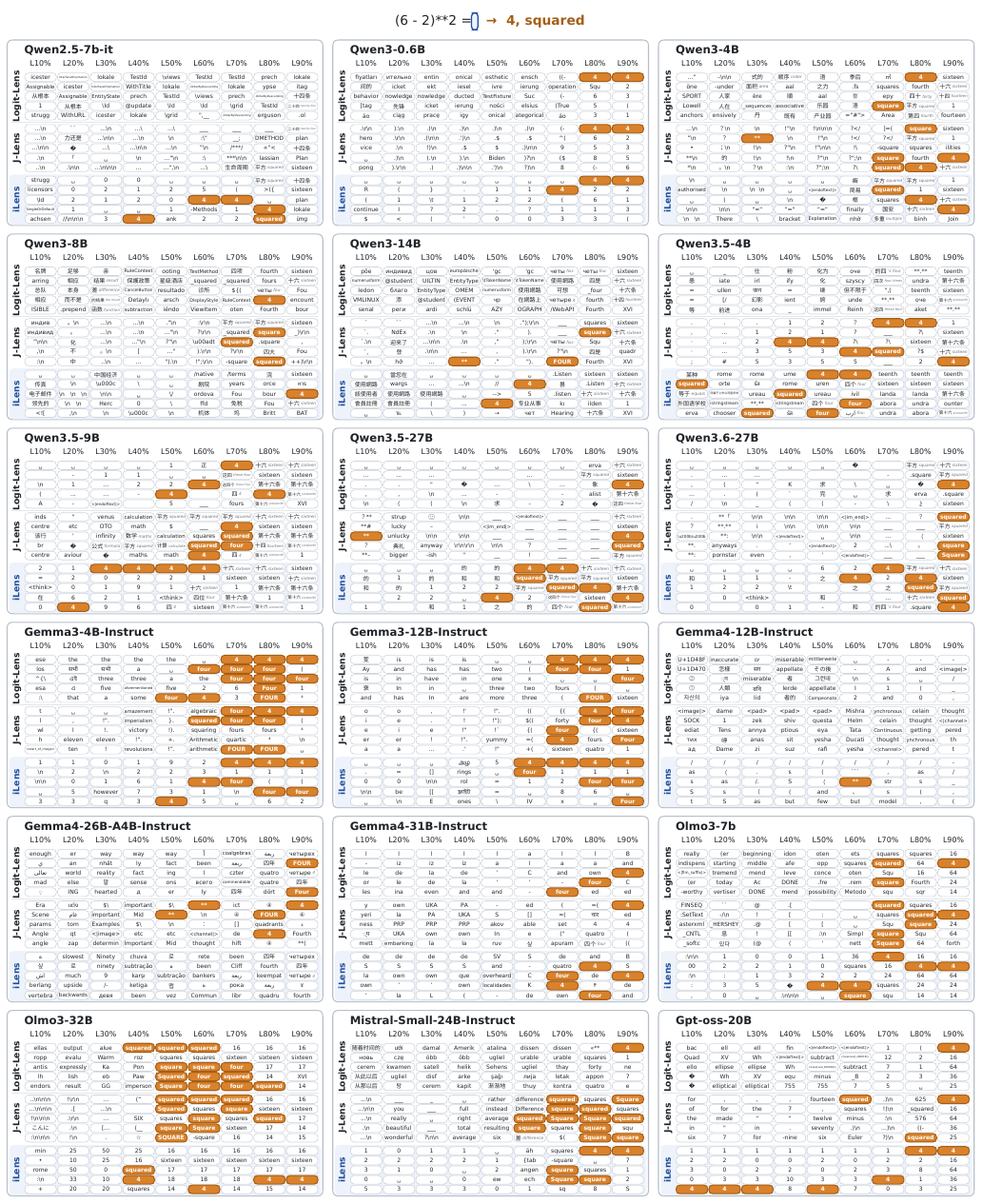}
    \caption{Qualitative example of the Order of Operations task across 18 LLMs, with predictions examined at nine normalized layer depths (10\%--90\% of model depth) for each model. The input expression is $(6-2)^2$, where the target intermediate concepts are \texttt{4} (the result of the subtraction) and \texttt{squaring} (the subsequent operation). Orange highlights indicate successful recovery of the target concepts.}
    \label{fig:qualitative-order-of-operations-example-5}
    \vspace{-4.0mm}
\end{figure*}

\begin{figure*}[t]
    \centering
    \includegraphics[width=0.95\linewidth]{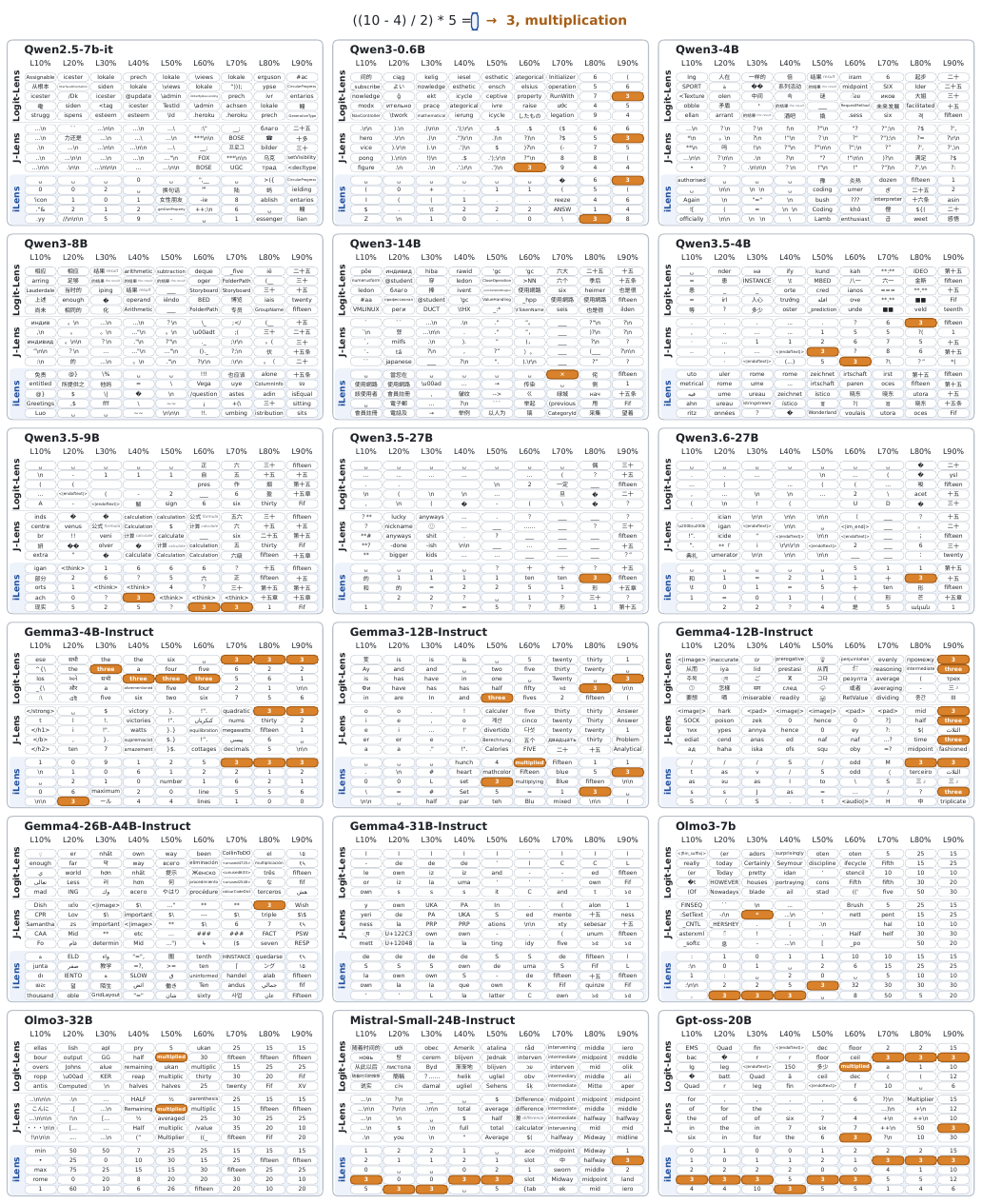}
    \caption{Qualitative example of the Order of Operations task across 18 LLMs, with predictions examined at nine normalized layer depths (10\%--90\% of model depth) for each model. The input expression is $\left(\frac{10-4}{2}\right)\times 5$, where the target intermediate concepts are \texttt{3} (the result of the division) and \texttt{multiplication} (the next operation). Orange highlights indicate successful recovery of the target concepts.}
    \label{fig:qualitative-order-of-operations-example-6}
    \vspace{-4.0mm}
\end{figure*}

\begin{figure*}[t]
    \centering
    \includegraphics[width=0.95\linewidth]{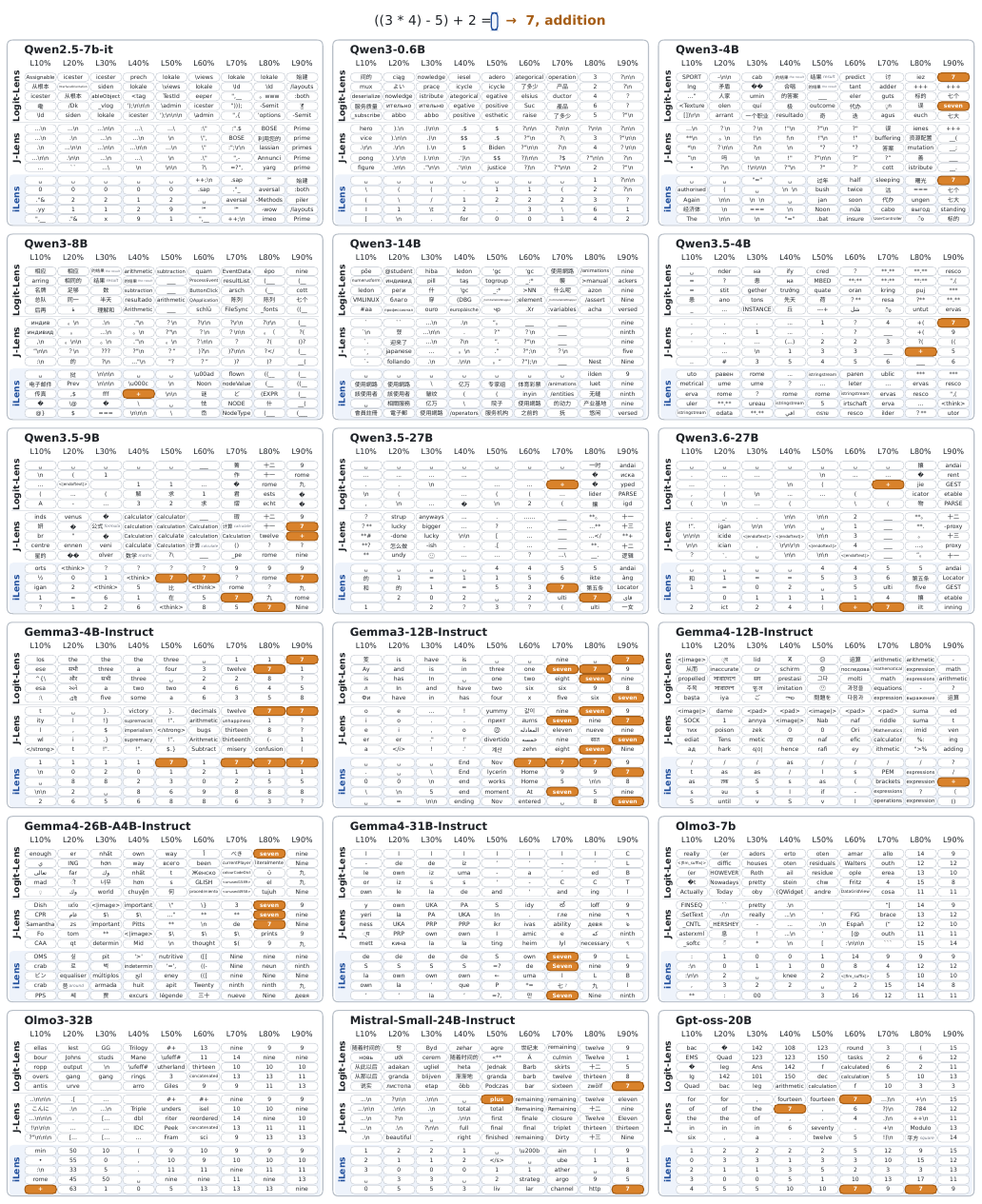}
    \caption{Qualitative example of the Order of Operations task across 18 LLMs, with predictions examined at nine normalized layer depths (10\%--90\% of model depth) for each model. The input expression is $((3\times 4)-5)+2$, where the target intermediate concepts are \texttt{7} (the result of the subtraction) and \texttt{addition} (the next operation). Orange highlights indicate successful recovery of the target concepts.}
    \label{fig:qualitative-order-of-operations-example-7}
    \vspace{-4.0mm}
\end{figure*}

\begin{figure*}[t]
    \centering
    \includegraphics[width=0.95\linewidth]{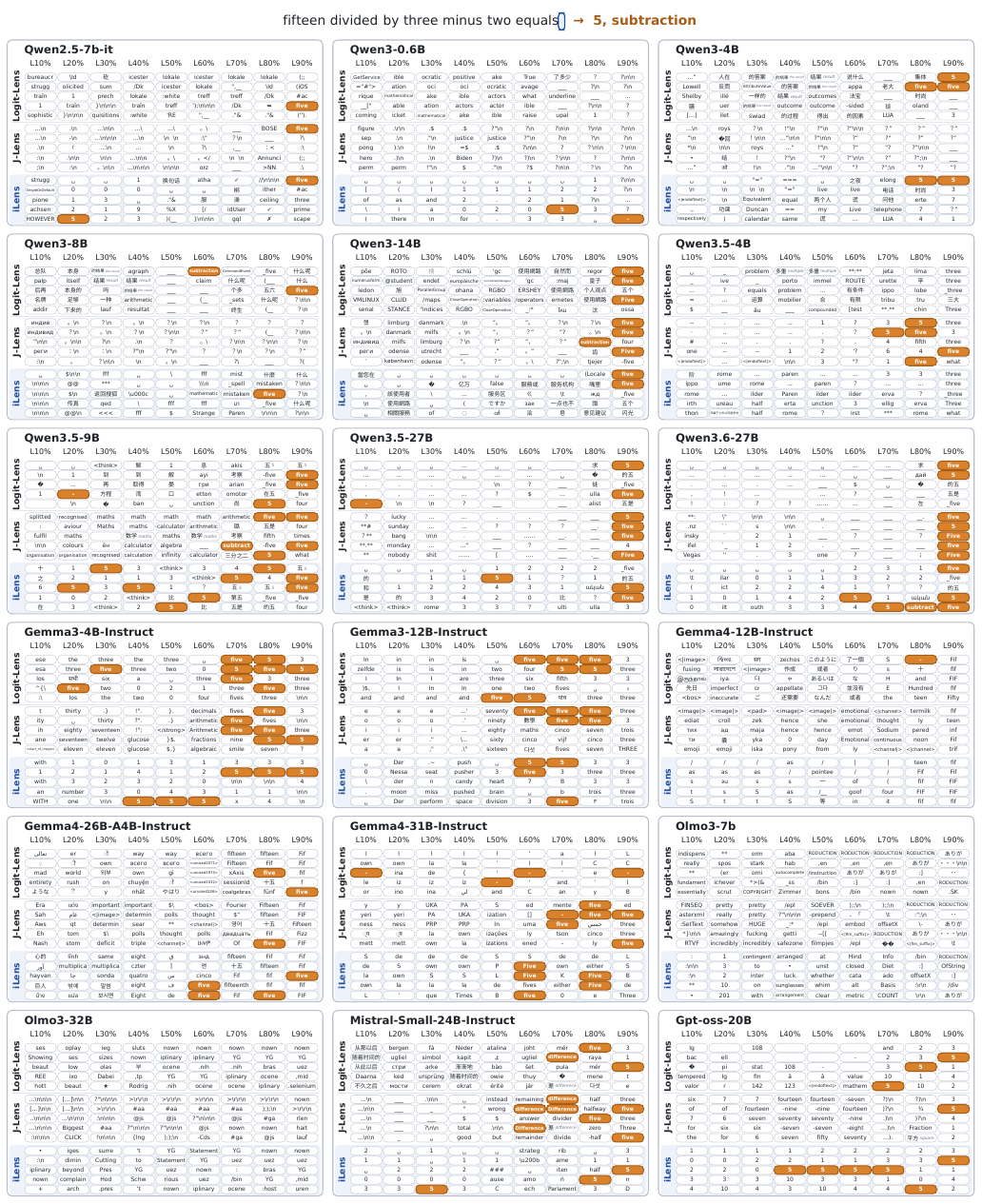}
    \caption{Qualitative example of the Order of Operations task across 18 LLMs, with predictions examined at nine normalized layer depths (10\%--90\% of model depth) for each model. The input expression is $\frac{15}{3}-2$, where the target intermediate concepts are \texttt{5} (the result of the division) and \texttt{subtraction} (the next operation). Orange highlights indicate successful recovery of the target concepts.}
    \label{fig:qualitative-order-of-operations-example-8}
    \vspace{-4.0mm}
\end{figure*}

\begin{figure*}[t]
    \centering
    \includegraphics[width=0.95\linewidth]{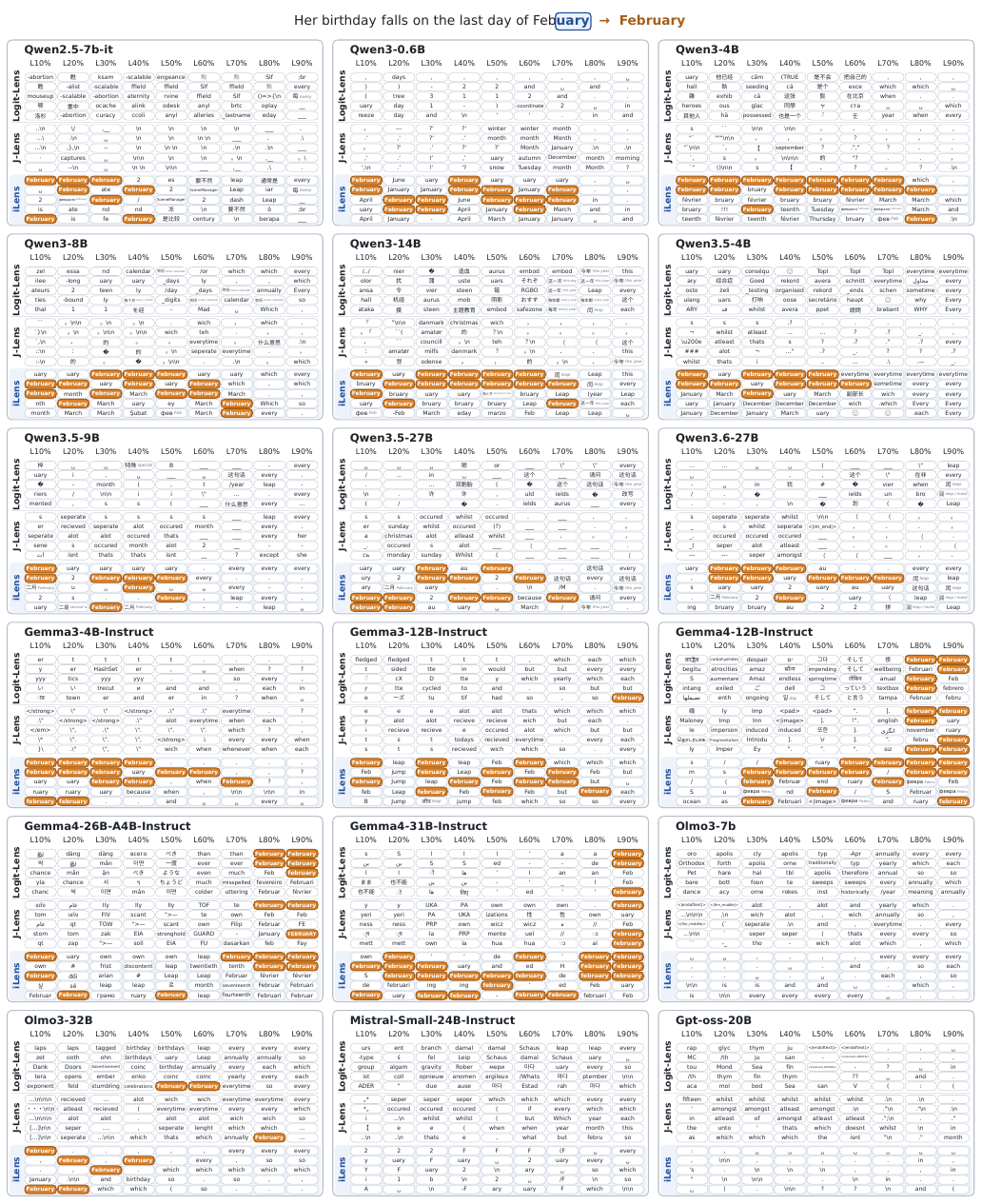}
    \caption{Qualitative example of the Typo task across 18 LLMs, with predictions examined at nine normalized layer depths (10\%--90\% of model depth) for each model. The input contains the misspelled word \texttt{Febuary}, whose intended form is \texttt{February}. Orange highlights indicate successful recovery of the target word.}
    \label{fig:qualitative-typo-1}
    \vspace{-4.0mm}
\end{figure*}

\begin{figure*}[t]
    \centering
    \includegraphics[width=0.95\linewidth]{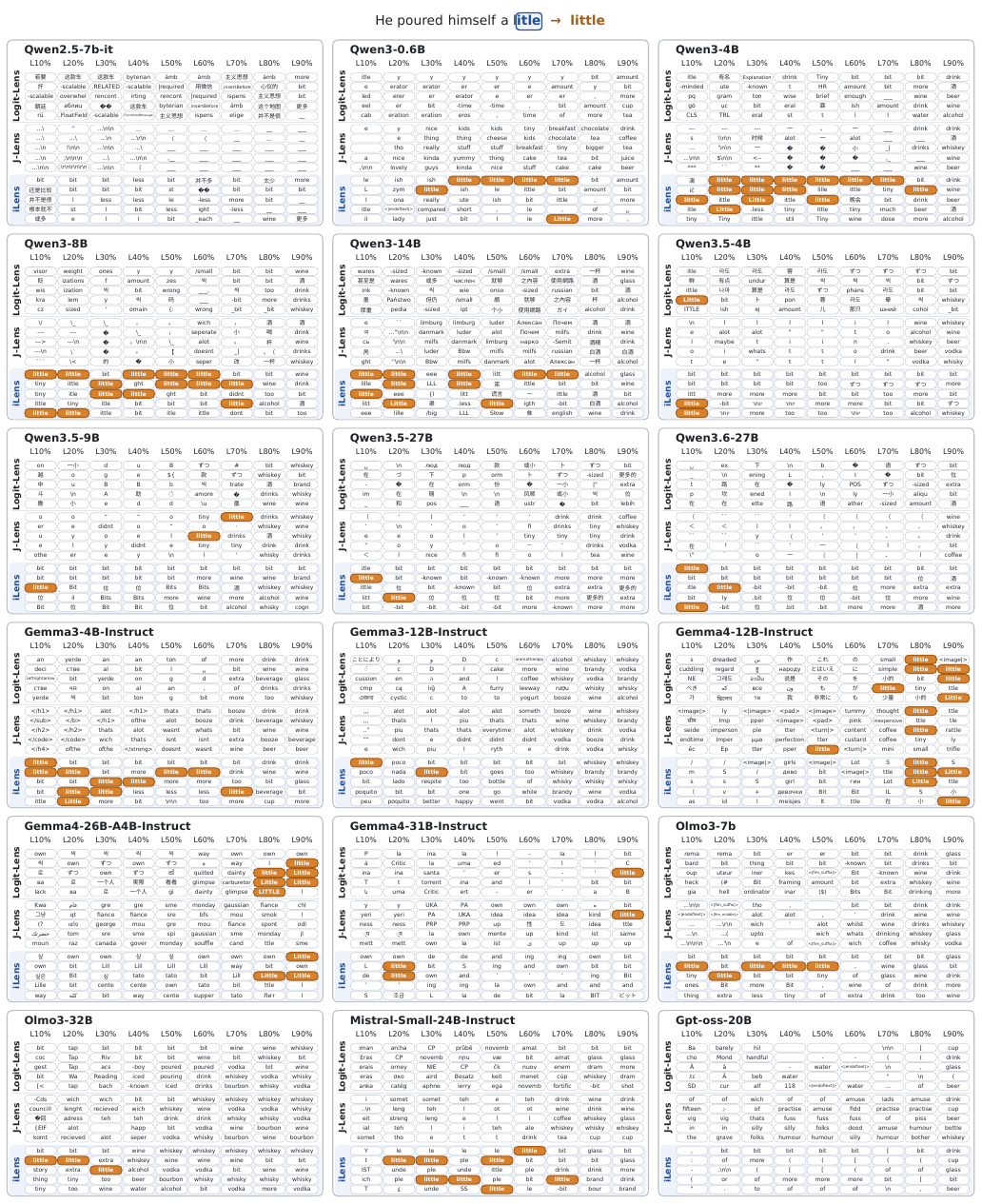}
    \caption{Qualitative example of the Typo task across 18 LLMs, with predictions examined at nine normalized layer depths (10\%--90\% of model depth) for each model. The input contains the misspelled word \texttt{litle}, whose intended form is \texttt{little}. Orange highlights indicate successful recovery of the target word.}
    \label{fig:qualitative-typo-2}
    \vspace{-4.0mm}
\end{figure*}

\begin{figure*}[t]
    \centering
    \includegraphics[width=0.95\linewidth]{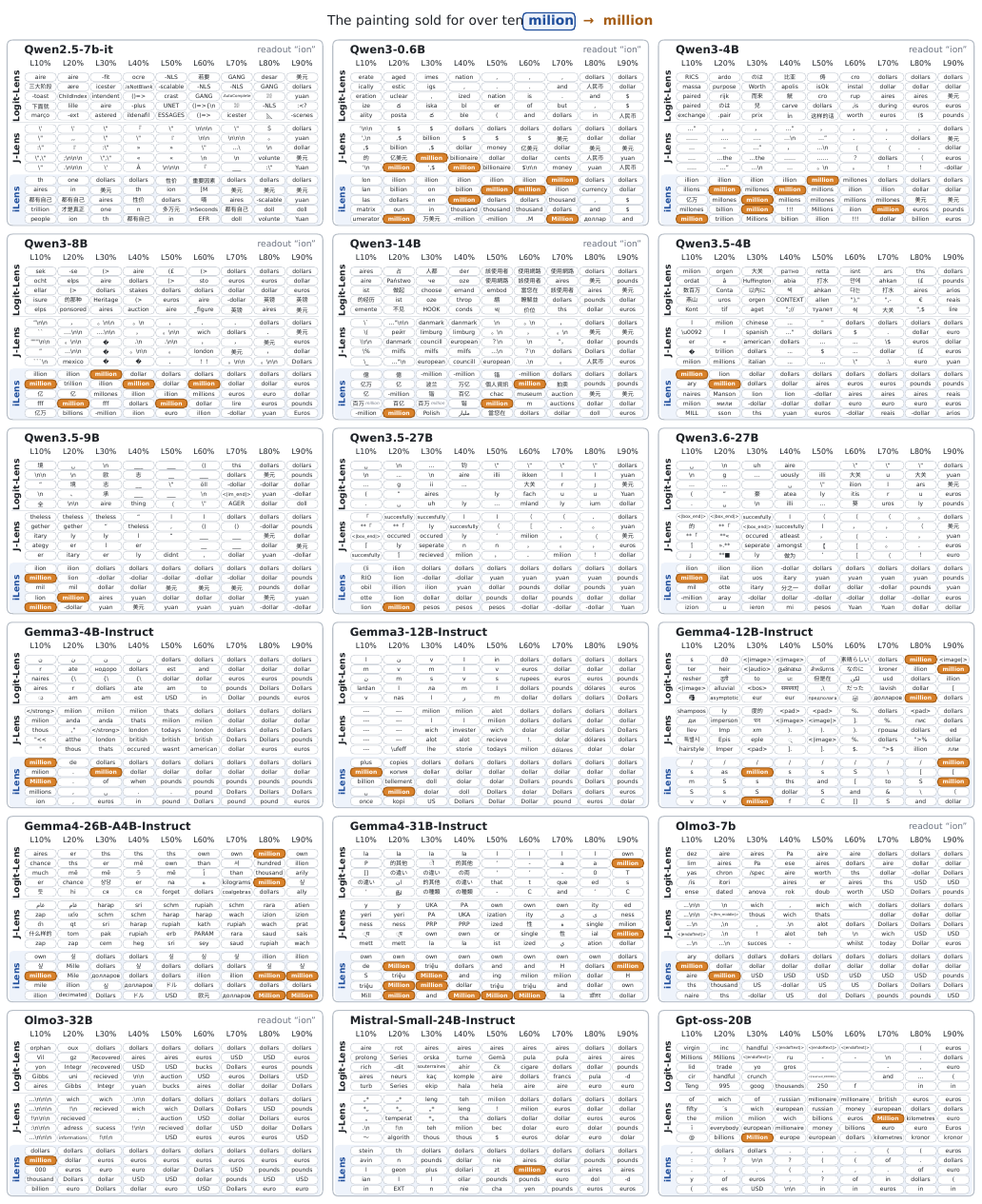}
    \caption{Qualitative example of the Typo task across 18 LLMs, with predictions examined at nine normalized layer depths (10\%--90\% of model depth) for each model. The input contains the misspelled word \texttt{milion}, whose intended form is \texttt{million}. Orange highlights indicate successful recovery of the target word.}
    \label{fig:qualitative-typo-3}
    \vspace{-4.0mm}
\end{figure*}

\begin{figure*}[t]
    \centering
    \includegraphics[width=0.95\linewidth]{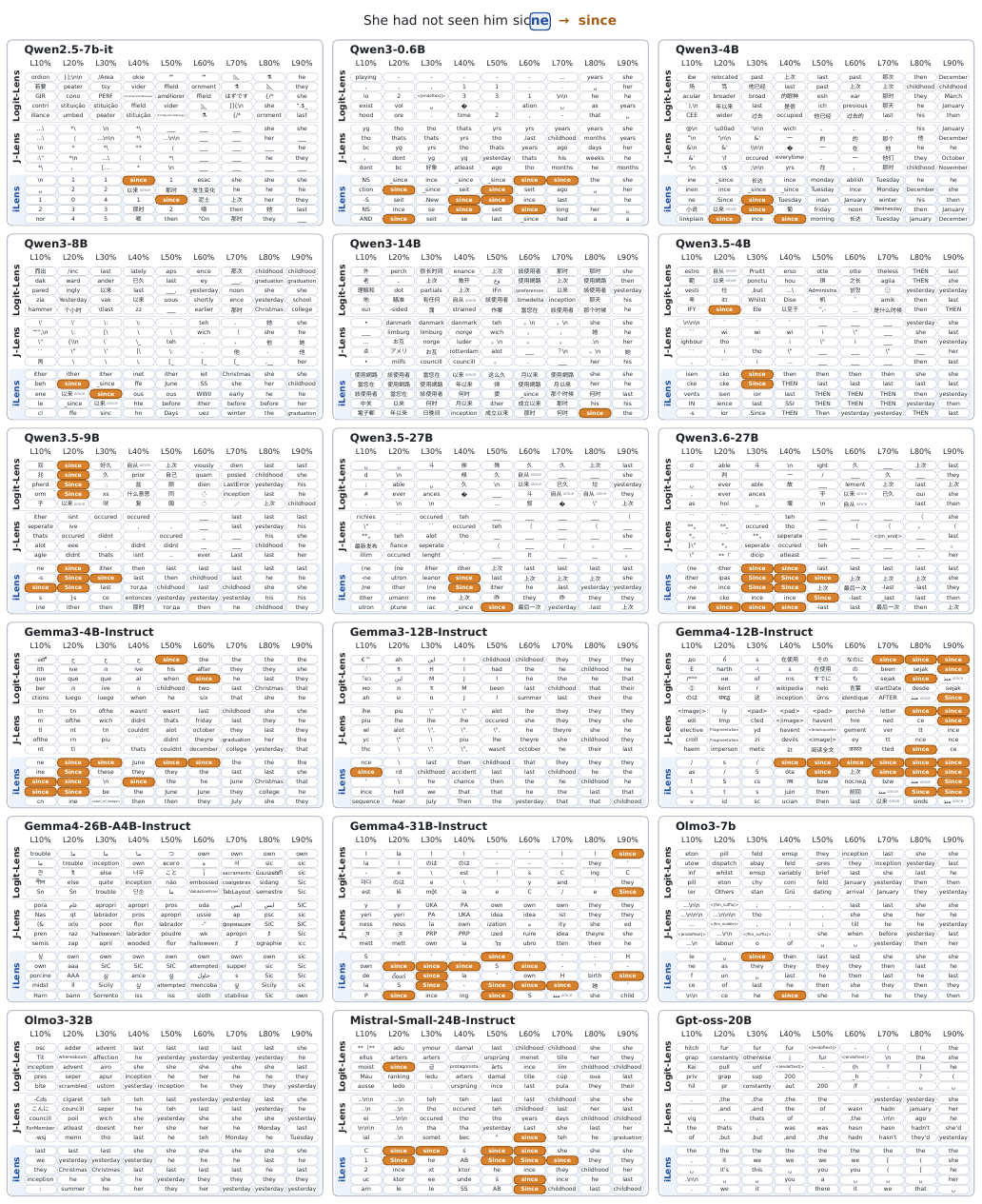}
    \caption{Qualitative example of the Typo task across 18 LLMs, with predictions examined at nine normalized layer depths (10\%--90\% of model depth) for each model. The input contains the misspelled word \texttt{sicne}, whose intended form is \texttt{since}. Orange highlights indicate successful recovery of the target word.}
    \label{fig:qualitative-typo-4}
    \vspace{-4.0mm}
\end{figure*}

\begin{figure*}[t]
    \centering
    \includegraphics[width=0.95\linewidth]{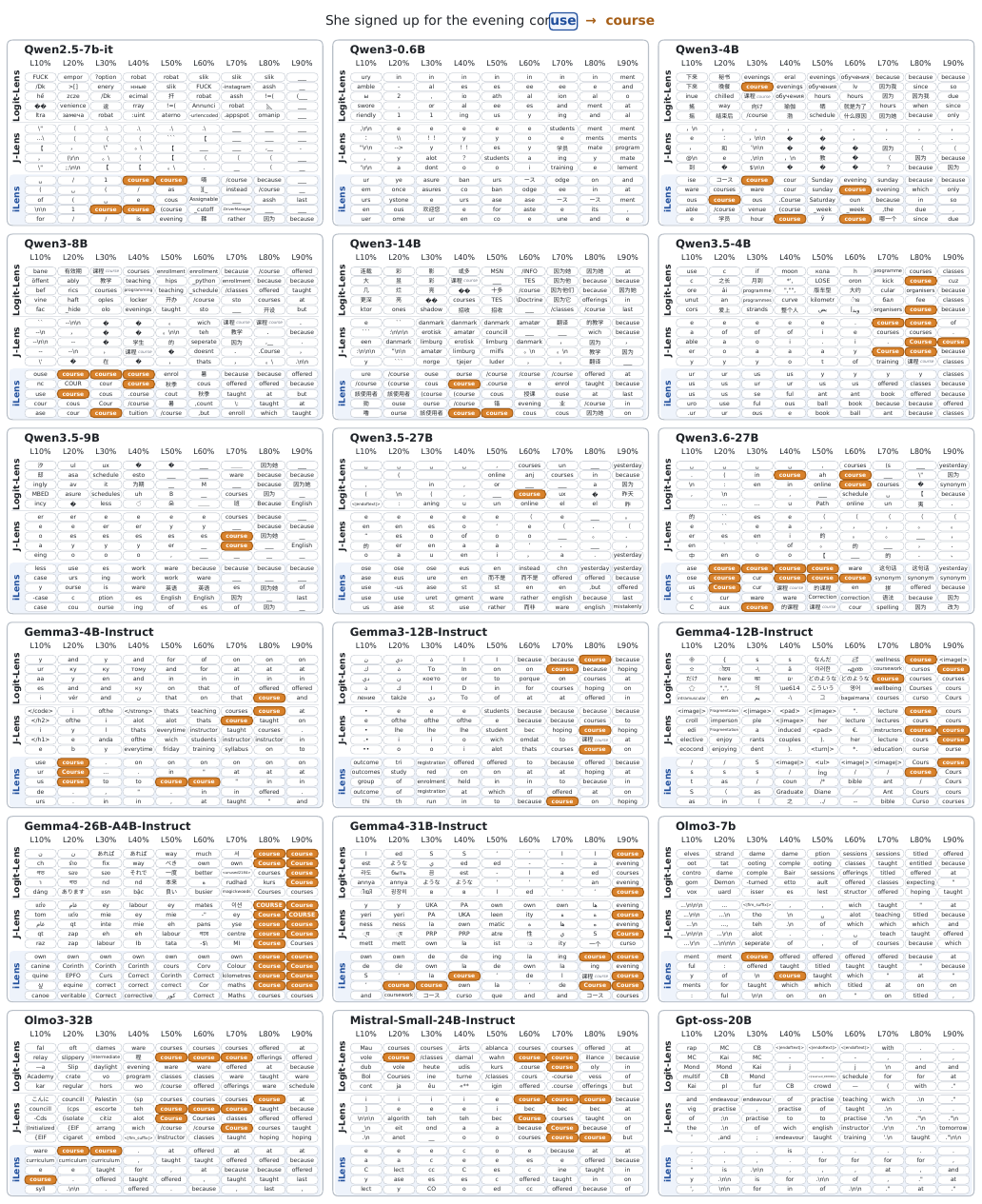}
    \caption{Qualitative example of the Typo task across 18 LLMs, with predictions examined at nine normalized layer depths (10\%--90\% of model depth) for each model. The input contains the misspelled word \texttt{coruse}, whose intended form is \texttt{course}. Orange highlights indicate successful recovery of the target word.}
    \label{fig:qualitative-typo-5}
    \vspace{-4.0mm}
\end{figure*}

\begin{figure*}[t]
    \centering
    \includegraphics[width=0.95\linewidth]{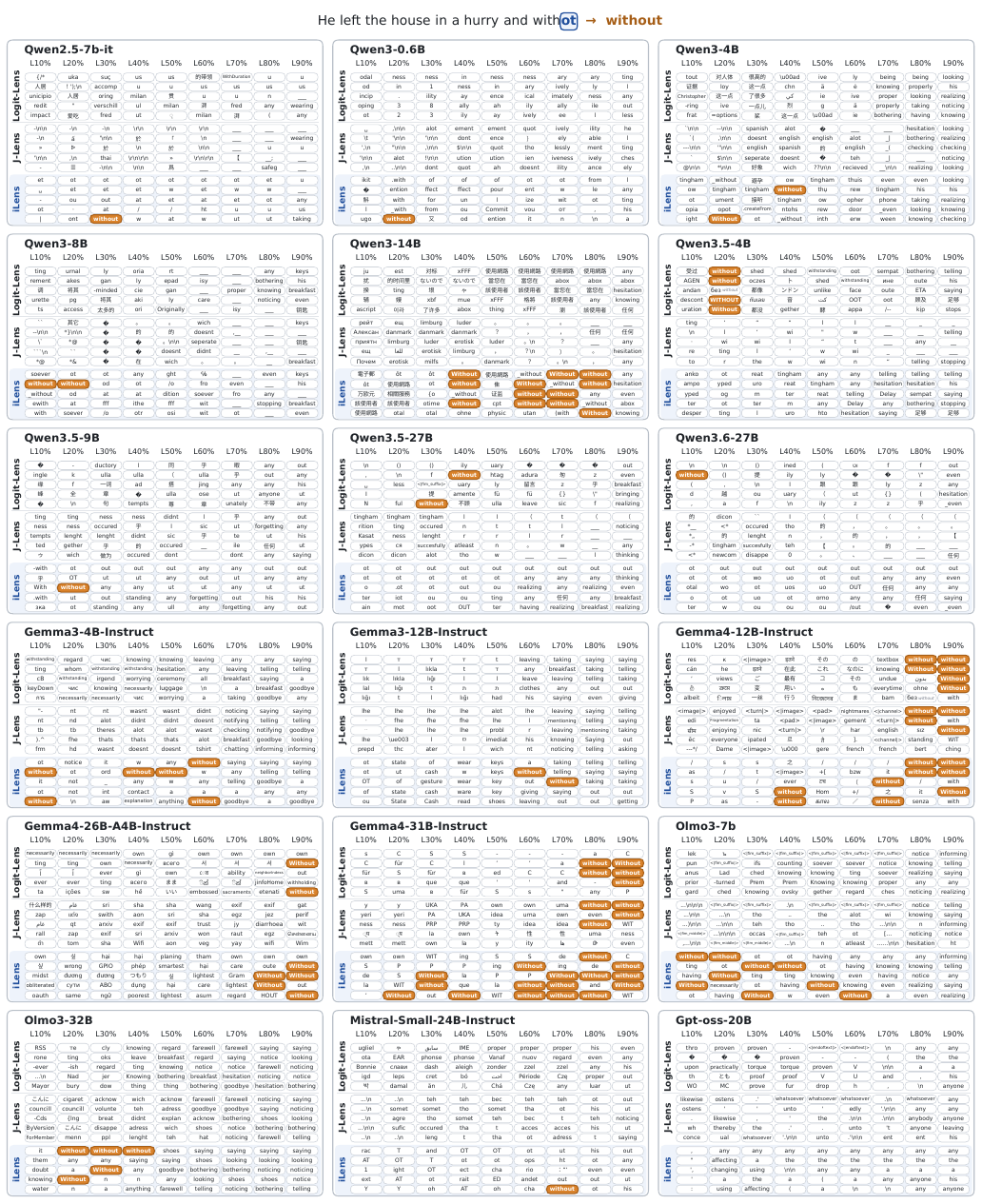}
    \caption{Qualitative example of the Typo task across 18 LLMs, with predictions examined at nine normalized layer depths (10\%--90\% of model depth) for each model. The input contains the misspelled word \texttt{withot}, whose intended form is \texttt{without}. Orange highlights indicate successful recovery of the target word.}
    \label{fig:qualitative-typo-6}
    \vspace{-4.0mm}
\end{figure*}

\begin{figure*}[t]
    \centering
    \includegraphics[width=0.95\linewidth]{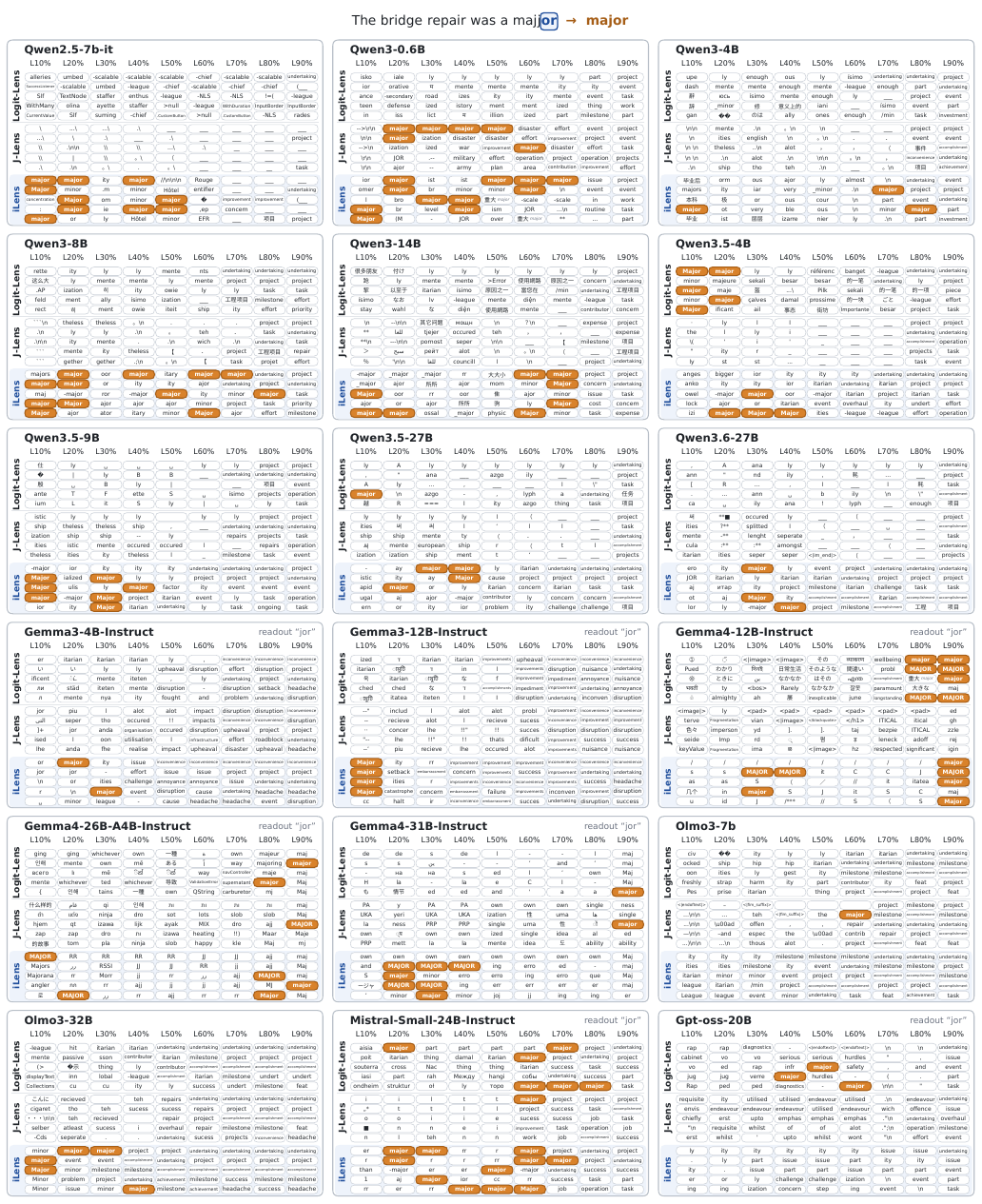}
    \caption{Qualitative example of the Typo task across 18 LLMs, with predictions examined at nine normalized layer depths (10\%--90\% of model depth) for each model. The input contains the misspelled word \texttt{majjor}, whose intended form is \texttt{major}. Orange highlights indicate successful recovery of the target word.}
    \label{fig:qualitative-typo-7}
    \vspace{-4.0mm}
\end{figure*}

\begin{figure*}[t]
    \centering
    \includegraphics[width=0.95\linewidth]{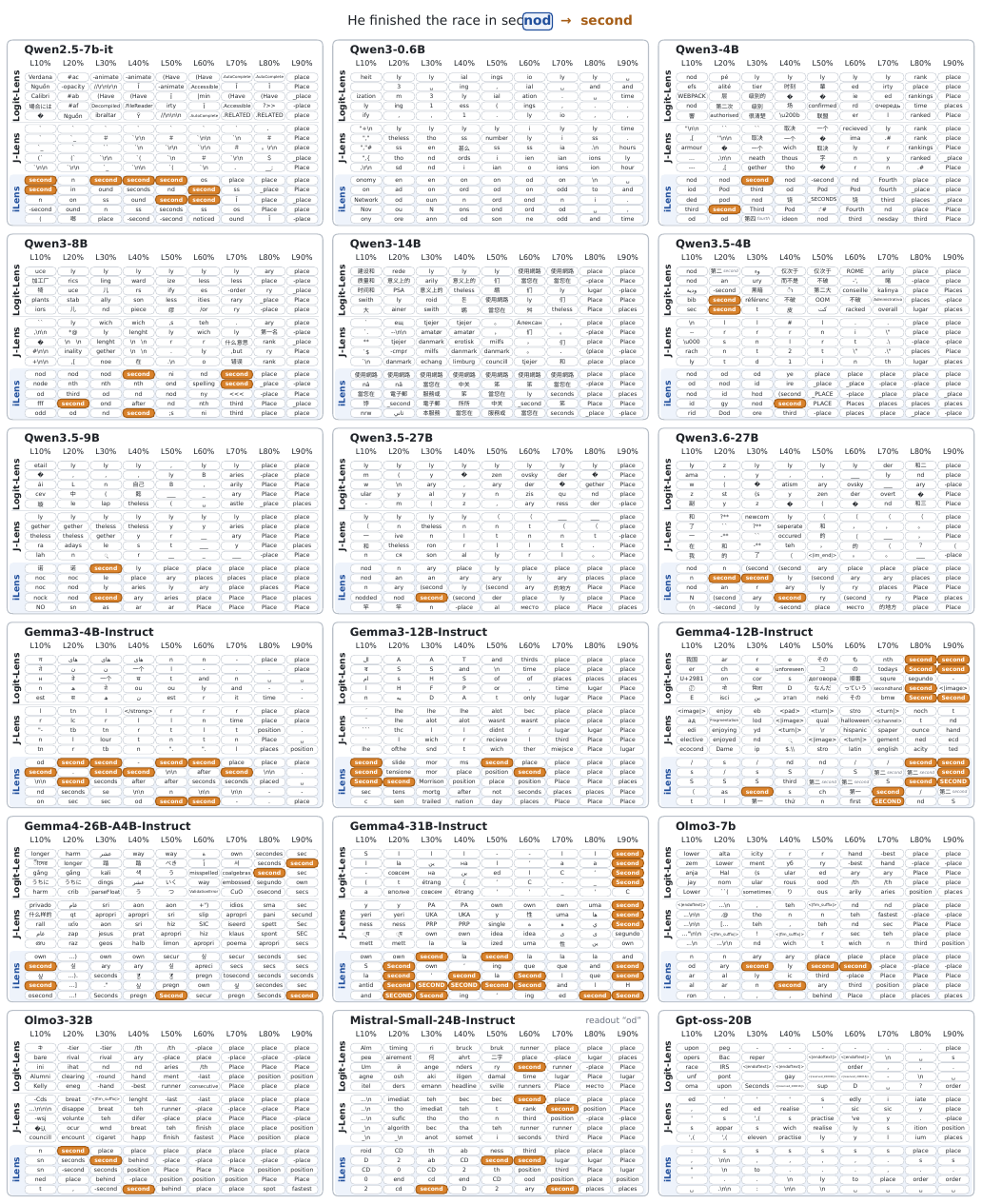}
    \caption{Qualitative example of the Typo task across 18 LLMs, with predictions examined at nine normalized layer depths (10\%--90\% of model depth) for each model. The input contains the misspelled word \texttt{secnod}, whose intended form is \texttt{sceond}. Orange highlights indicate successful recovery of the target word.}
    \label{fig:qualitative-typo-8}
    \vspace{-4.0mm}
\end{figure*}

\begin{figure*}[t]
    \centering
    \includegraphics[width=0.95\linewidth]{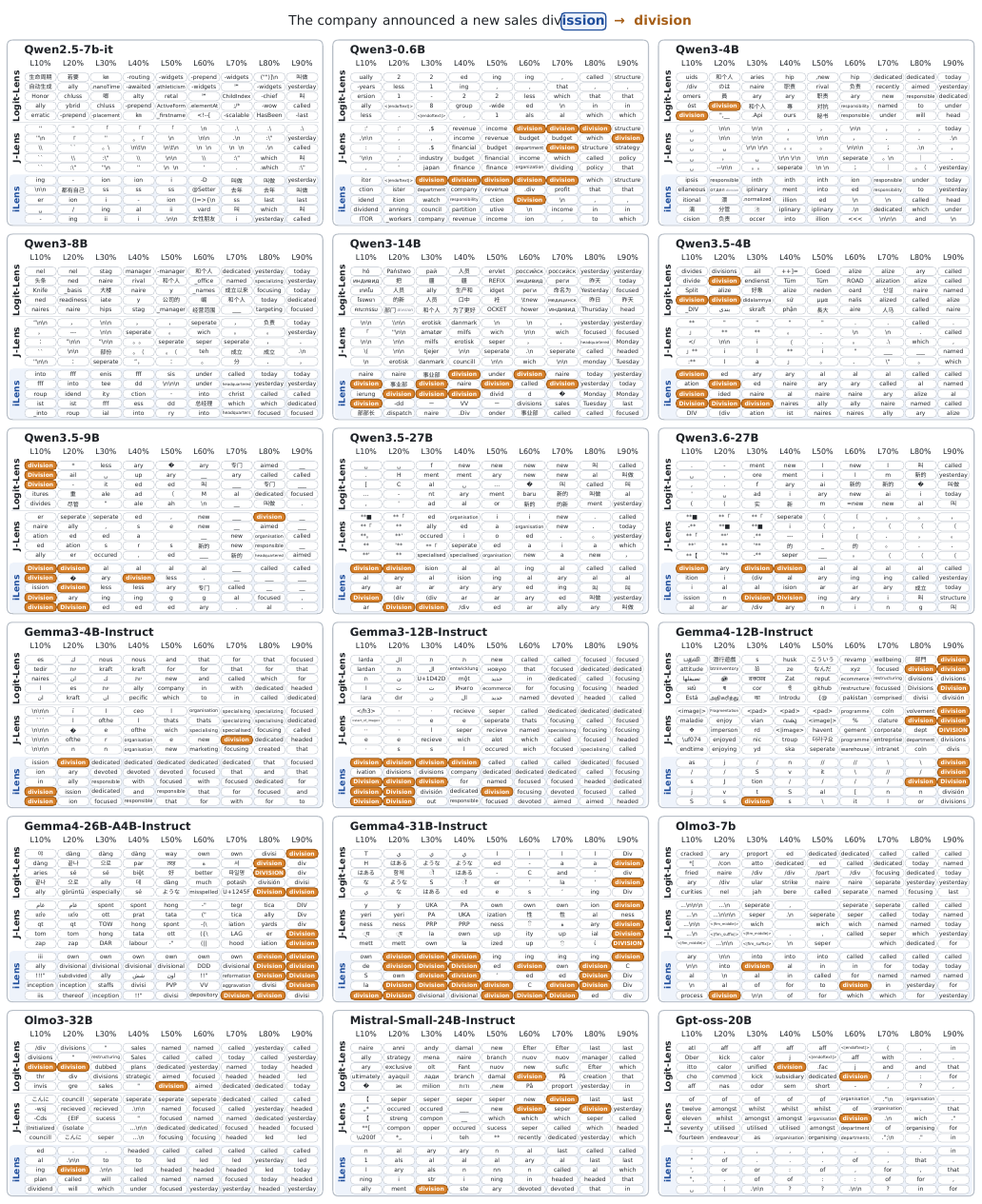}
    \caption{Qualitative example of the Typo task across 18 LLMs, with predictions examined at nine normalized layer depths (10\%--90\% of model depth) for each model. The input contains the misspelled word \texttt{divission}, whose intended form is \texttt{division}. Orange highlights indicate successful recovery of the target word.}
    \label{fig:qualitative-typo-9}
    \vspace{-4.0mm}
\end{figure*}

\begin{figure*}[t]
    \centering
    \includegraphics[width=0.95\linewidth]{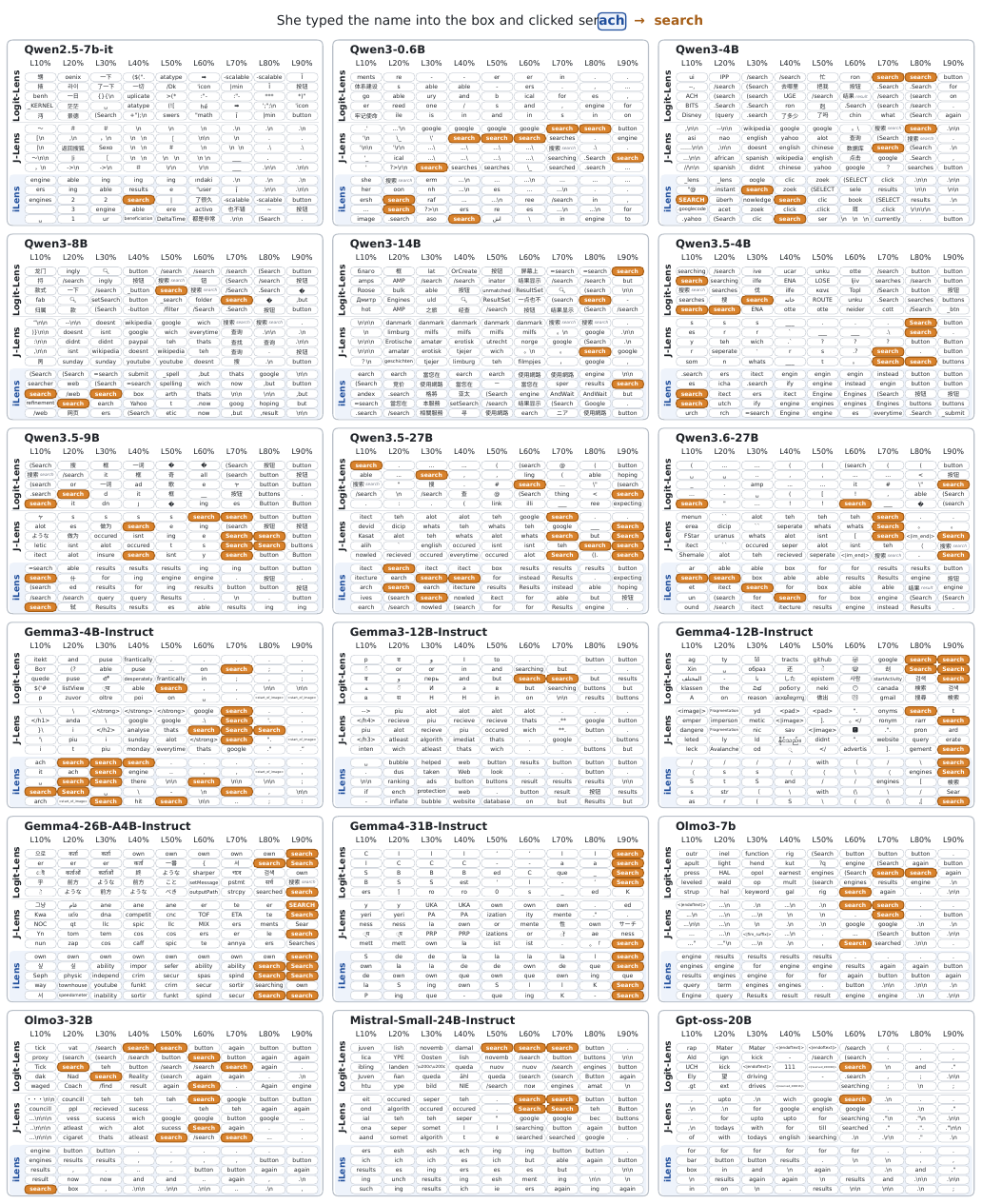}
    \caption{Qualitative example of the Typo task across 18 LLMs, with predictions examined at nine normalized layer depths (10\%--90\% of model depth) for each model. The input contains the misspelled word \texttt{serach}, whose intended form is \texttt{search}. Orange highlights indicate successful recovery of the target word.}
    \label{fig:qualitative-typo-10}
    \vspace{-4.0mm}
\end{figure*}

\paragraph{Qualitative Results.}
We provide additional qualitative examples for all five
textual decoding tasks in Figures~\ref{fig:qualitative-dual-task-months-key}--\ref{fig:qualitative-typo-10},
including both successful predictions and failure cases.
As illustrated, iLens not only successfully recovers target
concepts but also consistently recovers them at earlier
layers than the baseline methods. This behavior is consistent with
the lower FED values observed in the quantitative results.


\subsection{Additional Intervention Results}
\label{app:subsec:intervention_results}

\paragraph{Quantitative Results.}
We provide additional fine-grained results for concept suppression and injection across language and vision-language models. For LLMs, we report results separately for each of the 12 evaluated models, rather than aggregating them across model families. This allows us to examine model-dependent differences in intervention effectiveness across individual tasks and evaluation metrics. Table~\ref{tab:llm-intervention-source} reports source-word suppression, while Table~\ref{tab:llm-intervention-injected} reports injected-word gain. Both tables provide task-level results for $\Delta$Pass@10, $\Delta$Pass@20, and $\Delta$Confidence, together with per-model averages. Source suppression is evaluated on Probe Swap, Verbal Report, and Flexible Generalization, whereas injected-word gain additionally includes Verbal Introspection, for which no source concept is defined. All results are reported as normalized layer AUCs of the per-trial metric changes, computed before minus after editing for source suppression and after minus before editing for concept injection.

The LLM results show that iLens generally achieves substantially stronger source-word suppression than the competing methods across different model families and scales. This advantage is particularly evident in Probe Swap and Verbal Report, although its magnitude varies across architectures and tasks. For concept injection, iLens also achieves strong results across most models, particularly in Verbal Report, Flexible Generalization, and Verbal Introspection.

\begin{figure*}[t]
    \centering
    \includegraphics[width=0.95\textwidth]{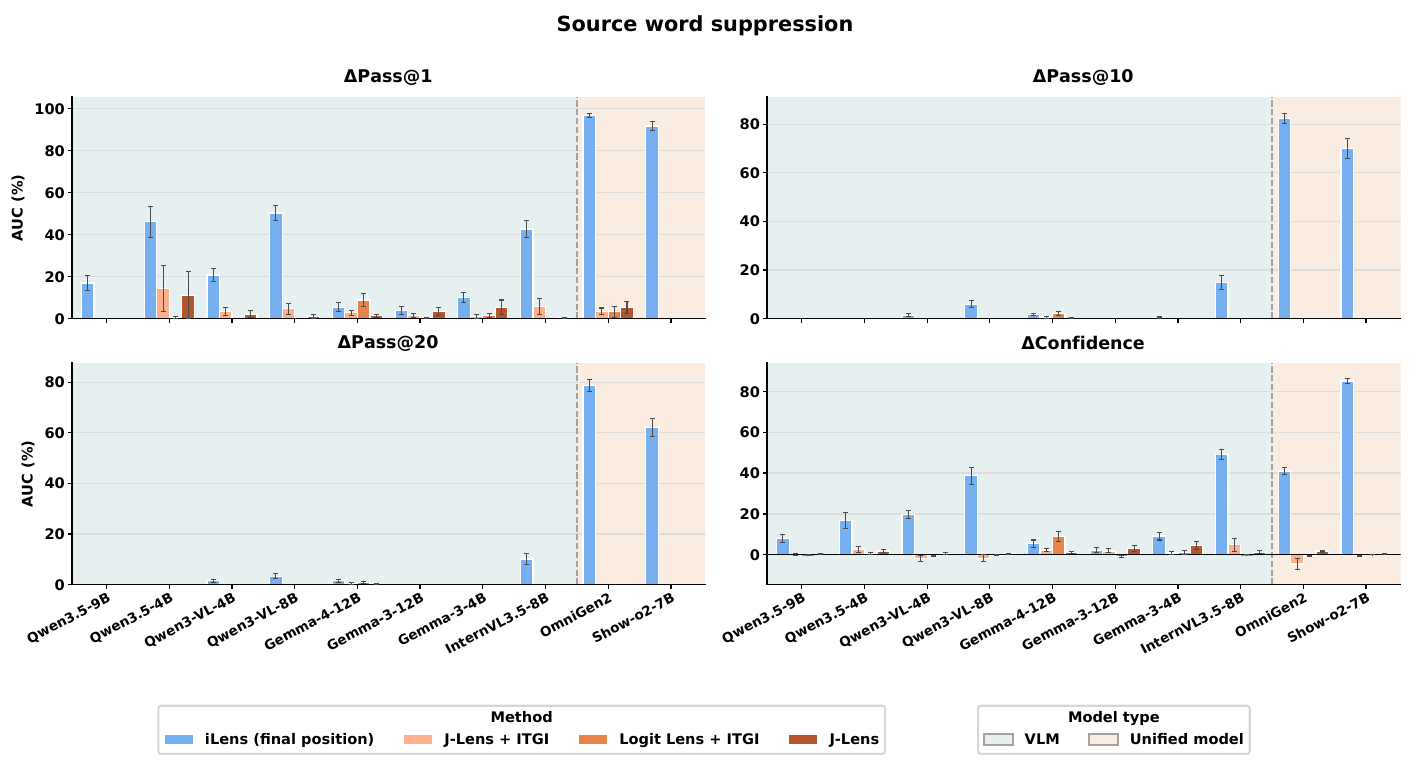}
    \caption{Source-word suppression across VLMs and unified vision models. We report normalized layer AUCs for $\Delta$Pass@1, $\Delta$Pass@10, $\Delta$Pass@20, and $\Delta$Confidence. Values correspond to the per-trial change before minus after editing; higher values indicate stronger source-concept suppression.}
    \label{fig:vlm-intervention-source}
    \vspace{-2mm}
\end{figure*}

\begin{figure*}[t]
    \centering
    \includegraphics[width=0.95\textwidth]{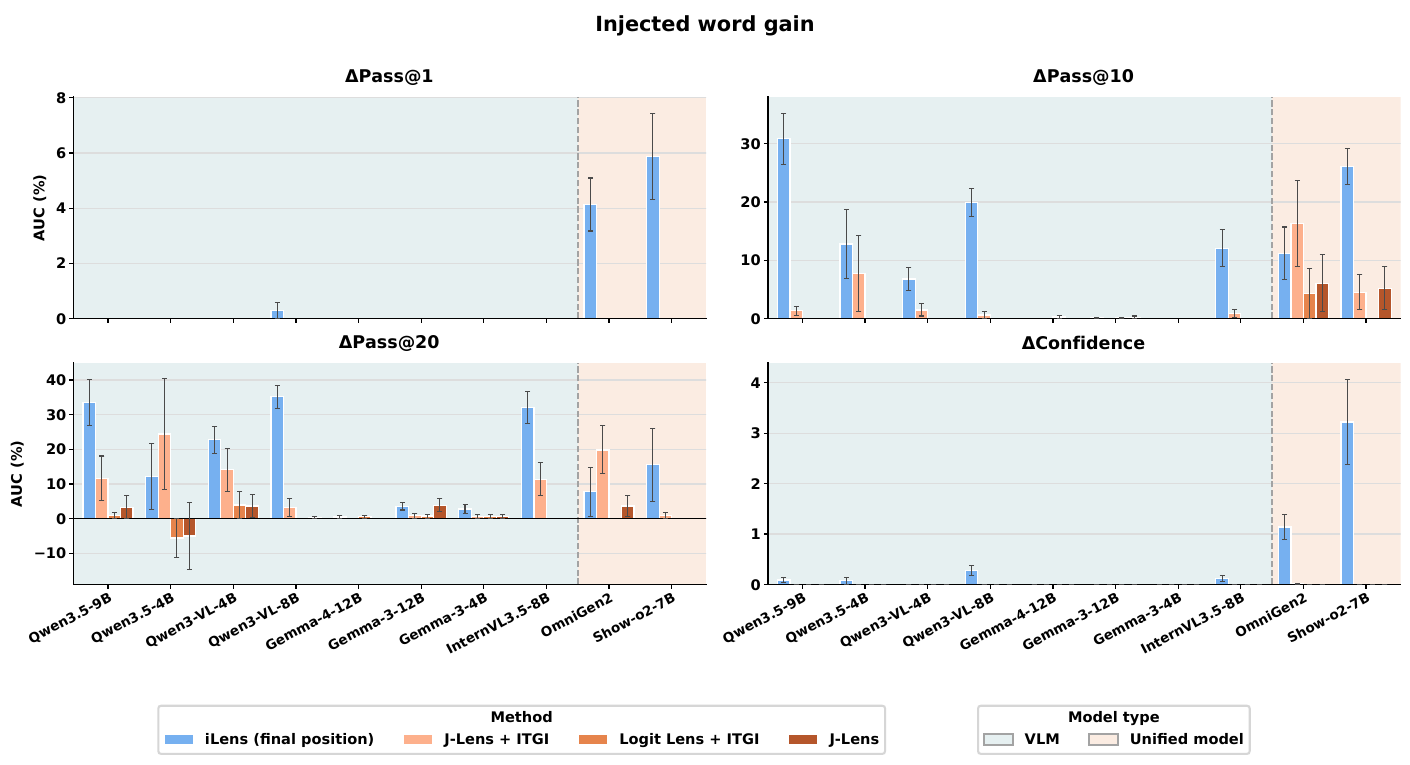}
    \caption{Injected-word gain across VLMs and unified vision models. We report normalized layer AUCs for $\Delta$Pass@1, $\Delta$Pass@10, $\Delta$Pass@20, and $\Delta$Confidence. Values correspond to the per-trial change after minus before editing; higher values indicate stronger target-concept injection.}
    \label{fig:vlm-intervention-injected}
    \vspace{-2mm}
\end{figure*}

For VLMs, we summarize the results graphically in Figures~\ref{fig:vlm-intervention-source} and~\ref{fig:vlm-intervention-injected}, reporting source-word suppression and injected-word gain separately. The figures cover the evaluated VLMs as well as the two unified vision models, OmniGen2 and Show-o2-7B. We report $\Delta$Pass@1, $\Delta$Pass@10, $\Delta$Pass@20, and $\Delta$Confidence as normalized layer AUCs, providing a complementary view of intervention effectiveness across visual model architectures.

The source-suppression results show a clear advantage for iLens across most evaluated VLMs and unified models, with particularly pronounced improvements in the higher-$k$ Pass metrics and confidence. Notably, iLens produces substantial suppression effects in both OmniGen2 and Show-o2-7B, demonstrating that the proposed intervention can also modify concept representations in unified vision models. The injected-word results exhibit greater variability across models and metrics. Nevertheless, iLens achieves notable improvements in several settings, particularly in Pass@10 and Pass@20, while some model-specific comparisons favor alternative methods. Taken together, these results demonstrate the effectiveness of iLens for concept intervention across both language and vision-language architectures, while also revealing differences in intervention behavior across model families and tasks.

\providecommand{\se}[1]{$_{\textcolor{gray}{\scalebox{0.65}{$\scriptstyle\scalebox{0.6}{$\scriptstyle\pm$}#1$}}}$}

\begin{table*}[t]
\centering
\caption{Source word suppression across 12 language models. Cells report normalized layer AUC of the per-trial change (before minus after editing). Bold marks the best method within each model and task.}
\label{tab:llm-intervention-source}
\begingroup
\fontsize{6}{7.2}\selectfont
\setlength{\tabcolsep}{1.2pt}
\renewcommand{\arraystretch}{1.05}
\resizebox{\textwidth}{!}{%
\begin{tabular}{@{}ll*{3}{rrr}|rrr@{}}
\toprule[0.9pt]
\multirow{2}{*}{Model} & \multirow{2}{*}{Method} & \multicolumn{3}{c}{Probe Swap} & \multicolumn{3}{c}{Verbal Report} & \multicolumn{3}{c}{Flexible Generalization} & \multicolumn{3}{|c}{Average} \\
\cmidrule[0.55pt](lr){3-14}
& & $\Delta$P@10 $\uparrow$ & $\Delta$P@20 $\uparrow$ & $\Delta$Conf. $\uparrow$ & $\Delta$P@10 $\uparrow$ & $\Delta$P@20 $\uparrow$ & $\Delta$Conf. $\uparrow$ & $\Delta$P@10 $\uparrow$ & $\Delta$P@20 $\uparrow$ & $\Delta$Conf. $\uparrow$ & $\Delta$P@10 $\uparrow$ & $\Delta$P@20 $\uparrow$ & $\Delta$Conf. $\uparrow$ \\
\midrule[0.65pt]
\multirow{4}{*}{Qwen3-0.6B-Inst.} & Logit Lens + ITGI & 28.46\se{11.58} & 8.46\se{3.08} & 15.01\se{4.19} & 0.15\se{0.15} & 0.00\se{0.00} & 2.00\se{0.74} & 2.45\se{2.09} & 4.11\se{2.32} & 0.90\se{0.53} & 10.35\se{3.92} & 4.19\se{1.28} & 5.97\se{1.43} \\
 & J-Lens & 16.92\se{10.43} & 2.31\se{2.31} & 11.55\se{4.37} & 1.23\se{0.71} & 0.00\se{0.00} & 4.73\se{1.01} & 5.07\se{2.50} & 11.89\se{3.36} & 2.95\se{1.73} & 7.74\se{3.58} & 4.73\se{1.36} & 6.41\se{1.60} \\
 & J-Lens + ITGI & 38.46\se{18.53} & 33.08\se{18.23} & 21.89\se{8.57} & 7.15\se{1.58} & 2.00\se{0.90} & 12.00\se{0.98} & 2.62\se{1.61} & 11.89\se{3.70} & 2.49\se{1.47} & 16.08\se{6.22} & 15.66\se{6.21} & 12.13\se{2.92} \\
\rowcolor{blue!7}
\cellcolor{white} & iLens  & \textbf{92.31}\se{3.85} & \textbf{86.92}\se{9.23} & \textbf{27.01}\se{8.02} & \textbf{41.31}\se{2.37} & \textbf{33.46}\se{2.34} & \textbf{21.75}\se{1.73} & \textbf{12.33}\se{4.96} & \textbf{22.90}\se{5.45} & \textbf{3.79}\se{2.19} & \textbf{48.65}\se{2.24} & \textbf{47.76}\se{3.66} & \textbf{17.52}\se{2.83} \\
\specialrule{0.55pt}{1.2pt}{1.2pt}
\multirow{4}{*}{Qwen3-4B-Inst.} & Logit Lens + ITGI & 5.20\se{1.49} & 1.66\se{0.50} & 18.59\se{2.40} & 0.00\se{0.00} & 0.00\se{0.00} & 1.58\se{0.86} & 0.00\se{0.00} & 0.00\se{0.00} & 0.25\se{0.19} & 1.73\se{0.50} & 0.55\se{0.17} & 6.81\se{0.85} \\
 & J-Lens & 6.79\se{2.29} & 4.68\se{1.87} & 22.34\se{3.10} & 0.00\se{0.00} & 0.00\se{0.00} & 2.12\se{1.05} & 0.13\se{0.13} & 0.00\se{0.00} & 3.72\se{1.92} & 2.31\se{0.77} & 1.56\se{0.62} & 9.39\se{1.27} \\
 & J-Lens + ITGI & 5.43\se{1.70} & 3.85\se{1.45} & 33.52\se{3.30} & 0.00\se{0.00} & 0.00\se{0.00} & 10.40\se{1.90} & 0.13\se{0.13} & 0.00\se{0.00} & 2.74\se{1.62} & 1.85\se{0.57} & 1.28\se{0.48} & 15.56\se{1.38} \\
\rowcolor{blue!7}
\cellcolor{white} & iLens  & \textbf{91.78}\se{1.15} & \textbf{85.14}\se{1.94} & \textbf{84.70}\se{2.83} & \textbf{16.94}\se{1.69} & \textbf{10.35}\se{1.44} & \textbf{55.09}\se{1.95} & \textbf{13.50}\se{3.14} & \textbf{7.02}\se{2.19} & \textbf{17.54}\se{2.76} & \textbf{40.74}\se{1.25} & \textbf{34.17}\se{1.09} & \textbf{52.44}\se{1.47} \\
\specialrule{0.55pt}{1.2pt}{1.2pt}
\multirow{4}{*}{Qwen3-8B-Inst.} & Logit Lens + ITGI & 3.74\se{0.57} & 0.80\se{0.21} & 12.98\se{1.78} & 0.00\se{0.00} & 0.00\se{0.00} & 1.88\se{0.64} & 0.00\se{0.00} & 0.00\se{0.00} & 0.20\se{0.12} & 1.25\se{0.19} & 0.27\se{0.07} & 5.02\se{0.63} \\
 & J-Lens & 6.43\se{2.17} & 3.43\se{1.45} & 14.92\se{2.57} & 0.00\se{0.00} & 0.00\se{0.00} & -0.67\se{0.67} & 0.13\se{0.13} & 0.00\se{0.00} & 2.30\se{1.04} & 2.19\se{0.72} & 1.14\se{0.48} & 5.52\se{0.95} \\
 & J-Lens + ITGI & 4.60\se{1.88} & 1.96\se{1.00} & 19.06\se{2.43} & 0.00\se{0.00} & 0.00\se{0.00} & 8.23\se{1.36} & 0.80\se{0.41} & 0.00\se{0.00} & 0.39\se{0.44} & 1.80\se{0.64} & 0.65\se{0.33} & 9.23\se{0.94} \\
\rowcolor{blue!7}
\cellcolor{white} & iLens  & \textbf{84.25}\se{2.25} & \textbf{76.16}\se{2.56} & \textbf{80.16}\se{3.24} & \textbf{10.82}\se{2.24} & \textbf{6.35}\se{1.65} & \textbf{45.12}\se{2.73} & \textbf{10.96}\se{2.74} & \textbf{6.62}\se{2.14} & \textbf{14.95}\se{2.62} & \textbf{35.35}\se{1.40} & \textbf{29.71}\se{1.24} & \textbf{46.75}\se{1.66} \\
\specialrule{0.55pt}{1.2pt}{1.2pt}
\multirow{4}{*}{Qwen3-14B-Inst.} & Logit Lens + ITGI & 2.96\se{0.48} & 1.54\se{0.40} & 11.92\se{1.35} & 0.00\se{0.00} & 0.00\se{0.00} & 0.57\se{0.37} & 0.00\se{0.00} & 0.00\se{0.00} & 0.10\se{0.22} & 0.99\se{0.16} & 0.51\se{0.13} & 4.20\se{0.47} \\
 & J-Lens & 2.96\se{1.26} & 2.08\se{1.05} & 10.68\se{2.39} & 0.00\se{0.00} & 0.00\se{0.00} & -0.10\se{0.43} & 0.00\se{0.00} & 0.00\se{0.00} & 0.97\se{0.49} & 0.99\se{0.42} & 0.69\se{0.35} & 3.85\se{0.82} \\
 & J-Lens + ITGI & 1.48\se{0.58} & 0.66\se{0.34} & 17.43\se{2.67} & 0.00\se{0.00} & 0.00\se{0.00} & 7.02\se{1.27} & 0.00\se{0.00} & 0.00\se{0.00} & -0.28\se{0.15} & 0.49\se{0.19} & 0.22\se{0.11} & 8.06\se{0.98} \\
\rowcolor{blue!7}
\cellcolor{white} & iLens  & \textbf{83.88}\se{2.53} & \textbf{74.34}\se{3.10} & \textbf{90.97}\se{2.29} & \textbf{6.89}\se{1.59} & \textbf{4.16}\se{1.17} & \textbf{35.86}\se{2.15} & \textbf{7.00}\se{1.98} & \textbf{3.17}\se{1.05} & \textbf{9.19}\se{2.51} & \textbf{32.59}\se{1.19} & \textbf{27.22}\se{1.16} & \textbf{45.34}\se{1.34} \\
\specialrule{0.55pt}{1.2pt}{1.2pt}
\multirow{4}{*}{Qwen2.5-7B-Inst.} & Logit Lens + ITGI & 1.63\se{0.53} & 0.58\se{0.24} & 7.89\se{1.37} & 0.00\se{0.00} & 0.00\se{0.00} & 1.52\se{0.67} & -1.05\se{0.59} & 0.00\se{0.00} & 0.60\se{0.23} & 0.19\se{0.26} & 0.19\se{0.08} & 3.34\se{0.51} \\
 & J-Lens & 4.90\se{1.87} & 2.80\se{1.38} & 16.77\se{3.48} & 0.00\se{0.00} & 0.00\se{0.00} & 4.23\se{0.97} & -2.45\se{1.75} & 0.00\se{0.00} & 2.23\se{1.21} & 0.82\se{0.85} & 0.93\se{0.46} & 7.74\se{1.27} \\
 & J-Lens + ITGI & 8.86\se{2.38} & 4.08\se{1.53} & 35.28\se{3.54} & 0.00\se{0.00} & 0.00\se{0.00} & 20.53\se{2.23} & -3.32\se{2.07} & 0.00\se{0.00} & 3.30\se{0.97} & 1.85\se{1.05} & 1.36\se{0.51} & 19.71\se{1.43} \\
\rowcolor{blue!7}
\cellcolor{white} & iLens  & \textbf{76.22}\se{2.97} & \textbf{70.75}\se{3.50} & \textbf{73.81}\se{3.73} & \textbf{23.31}\se{1.50} & \textbf{18.31}\se{1.27} & \textbf{45.14}\se{1.33} & \textbf{11.36}\se{2.82} & \textbf{13.72}\se{1.47} & \textbf{20.42}\se{3.58} & \textbf{36.97}\se{1.46} & \textbf{34.26}\se{1.33} & \textbf{46.46}\se{1.78} \\
\specialrule{0.55pt}{1.2pt}{1.2pt}
\multirow{4}{*}{Qwen3.5-4B-Inst.} & Logit Lens + ITGI & 0.15\se{0.10} & 0.00\se{0.00} & 8.64\se{0.93} & 0.00\se{0.00} & 0.00\se{0.00} & 0.66\se{0.37} & 0.00\se{0.00} & 0.00\se{0.00} & 0.08\se{0.18} & 0.05\se{0.03} & 0.00\se{0.00} & 3.13\se{0.34} \\
 & J-Lens & 3.78\se{1.54} & 1.48\se{0.99} & 16.92\se{2.44} & 0.00\se{0.00} & 0.00\se{0.00} & 0.66\se{0.57} & 0.61\se{0.61} & 0.00\se{0.00} & 1.88\se{0.66} & 1.46\se{0.55} & 0.49\se{0.33} & 6.48\se{0.86} \\
 & J-Lens + ITGI & 11.78\se{2.51} & 7.33\se{1.91} & 35.53\se{3.14} & 0.00\se{0.00} & 0.00\se{0.00} & 10.81\se{0.82} & 1.21\se{0.58} & 0.15\se{0.15} & 5.12\se{1.45} & 4.33\se{0.86} & 2.49\se{0.64} & 17.16\se{1.18} \\
\rowcolor{blue!7}
\cellcolor{white} & iLens  & \textbf{56.96}\se{2.57} & \textbf{45.19}\se{2.88} & \textbf{56.16}\se{3.28} & \textbf{20.00}\se{2.30} & \textbf{12.53}\se{1.78} & \textbf{36.14}\se{1.79} & \textbf{15.91}\se{3.01} & \textbf{9.70}\se{2.02} & \textbf{21.94}\se{2.77} & \textbf{30.96}\se{1.53} & \textbf{22.47}\se{1.31} & \textbf{38.08}\se{1.55} \\
\specialrule{0.55pt}{1.2pt}{1.2pt}
\multirow{4}{*}{Qwen3.5-9B-Inst.} & Logit Lens + ITGI & 1.93\se{0.51} & 1.07\se{0.36} & 8.57\se{0.83} & 0.00\se{0.00} & 0.00\se{0.00} & 0.86\se{0.32} & 2.42\se{1.74} & 0.00\se{0.00} & 0.27\se{0.14} & 1.45\se{0.60} & 0.36\se{0.12} & 3.23\se{0.30} \\
 & J-Lens & 2.60\se{1.37} & 1.73\se{1.04} & 10.98\se{1.69} & 0.00\se{0.00} & 0.00\se{0.00} & 2.52\se{0.50} & 1.06\se{1.06} & 0.00\se{0.00} & 1.02\se{0.72} & 1.22\se{0.58} & 0.58\se{0.35} & 4.84\se{0.63} \\
 & J-Lens + ITGI & 7.93\se{2.20} & 4.53\se{1.70} & 25.47\se{2.44} & 0.00\se{0.00} & 0.00\se{0.00} & 7.41\se{1.34} & 2.05\se{1.36} & 0.00\se{0.00} & 2.70\se{1.20} & 3.33\se{0.86} & 1.51\se{0.57} & 11.86\se{1.01} \\
\rowcolor{blue!7}
\cellcolor{white} & iLens  & \textbf{60.93}\se{2.68} & \textbf{49.00}\se{3.14} & \textbf{49.24}\se{2.95} & \textbf{22.87}\se{2.23} & \textbf{17.73}\se{2.03} & \textbf{45.57}\se{1.20} & \textbf{17.20}\se{3.10} & \textbf{10.61}\se{2.03} & \textbf{21.81}\se{3.62} & \textbf{33.67}\se{1.55} & \textbf{25.78}\se{1.42} & \textbf{38.87}\se{1.61} \\
\specialrule{0.55pt}{1.2pt}{1.2pt}
\multirow{4}{*}{Gemma-3-4B-Inst.} & Logit Lens + ITGI & 6.59\se{1.57} & 4.76\se{1.20} & 27.98\se{2.72} & 3.56\se{0.73} & 2.56\se{0.49} & 21.38\se{2.84} & 1.40\se{0.51} & 0.80\se{0.42} & -0.50\se{1.04} & 3.85\se{0.60} & 2.71\se{0.45} & 16.29\se{1.36} \\
 & J-Lens & 5.77\se{2.28} & 4.76\se{2.16} & 20.74\se{3.26} & 0.12\se{0.12} & 0.12\se{0.12} & 9.28\se{2.29} & 2.99\se{1.27} & \textbf{1.80}\se{0.85} & 1.22\se{1.98} & 2.96\se{0.87} & 2.23\se{0.77} & 10.41\se{1.48} \\
 & J-Lens + ITGI & 0.00\se{0.00} & 0.00\se{0.00} & 14.73\se{1.53} & 0.00\se{0.00} & 0.00\se{0.00} & 9.49\se{1.73} & 0.00\se{0.00} & 0.00\se{0.00} & -0.25\se{0.21} & 0.00\se{0.00} & 0.00\se{0.00} & 7.99\se{0.77} \\
\rowcolor{blue!7}
\cellcolor{white} & iLens  & \textbf{29.48}\se{2.08} & \textbf{20.45}\se{1.98} & \textbf{55.87}\se{2.25} & \textbf{6.75}\se{1.35} & \textbf{4.75}\se{1.06} & \textbf{40.21}\se{2.44} & \textbf{3.06}\se{0.97} & 1.73\se{0.68} & \textbf{5.13}\se{1.14} & \textbf{13.10}\se{0.89} & \textbf{8.98}\se{0.78} & \textbf{33.74}\se{1.17} \\
\specialrule{0.55pt}{1.2pt}{1.2pt}
\multirow{4}{*}{Gemma-3-12B-Inst.} & Logit Lens + ITGI & 4.39\se{1.04} & 3.09\se{0.73} & 10.54\se{1.57} & 0.57\se{0.22} & 0.43\se{0.21} & 6.79\se{1.82} & 0.28\se{0.21} & \textbf{0.19}\se{0.19} & 0.49\se{0.30} & 1.74\se{0.36} & 1.24\se{0.26} & 5.94\se{0.81} \\
 & J-Lens & 1.39\se{0.60} & 1.39\se{0.60} & 6.50\se{1.38} & 0.00\se{0.00} & 0.00\se{0.00} & 0.50\se{1.64} & 0.00\se{0.00} & 0.00\se{0.00} & 0.29\se{0.22} & 0.46\se{0.20} & 0.46\se{0.20} & 2.43\se{0.72} \\
 & J-Lens + ITGI & 0.26\se{0.19} & 0.09\se{0.09} & 6.58\se{0.85} & 0.00\se{0.00} & 0.00\se{0.00} & 3.52\se{0.95} & 0.00\se{0.00} & 0.00\se{0.00} & 0.10\se{0.12} & 0.09\se{0.06} & 0.03\se{0.03} & 3.40\se{0.43} \\
\rowcolor{blue!7}
\cellcolor{white} & iLens  & \textbf{6.87}\se{1.29} & \textbf{4.70}\se{0.97} & \textbf{23.27}\se{1.53} & \textbf{1.13}\se{0.39} & \textbf{0.87}\se{0.35} & \textbf{11.89}\se{1.81} & \textbf{0.37}\se{0.18} & 0.09\se{0.09} & \textbf{2.20}\se{0.66} & \textbf{2.79}\se{0.45} & \textbf{1.89}\se{0.34} & \textbf{12.45}\se{0.82} \\
\specialrule{0.55pt}{1.2pt}{1.2pt}
\multirow{4}{*}{Gemma-4-12B-Inst.} & Logit Lens + ITGI & \textbf{5.91}\se{0.52} & \textbf{5.48}\se{0.49} & \textbf{12.05}\se{1.04} & 4.78\se{0.31} & 4.52\se{0.30} & 11.06\se{0.99} & 4.07\se{0.84} & \textbf{3.15}\se{0.64} & 3.77\se{0.78} & \textbf{4.92}\se{0.35} & \textbf{4.38}\se{0.29} & 8.96\se{0.54} \\
 & J-Lens & 1.65\se{0.30} & 1.57\se{0.30} & 3.27\se{0.53} & 3.30\se{0.49} & 3.13\se{0.43} & 4.24\se{0.64} & 2.04\se{0.32} & 1.85\se{0.32} & 1.51\se{0.31} & 2.33\se{0.22} & 2.18\se{0.20} & 3.01\se{0.30} \\
 & J-Lens + ITGI & 1.83\se{0.33} & 1.39\se{0.32} & 4.82\se{0.78} & 4.09\se{0.44} & 3.91\se{0.42} & 7.53\se{0.78} & 1.57\se{0.48} & 1.30\se{0.44} & 0.59\se{0.39} & 2.50\se{0.24} & 2.20\se{0.23} & 4.31\se{0.39} \\
\rowcolor{blue!7}
\cellcolor{white} & iLens  & 2.17\se{0.57} & 1.30\se{0.36} & 11.29\se{1.50} & \textbf{5.70}\se{0.50} & \textbf{5.04}\se{0.51} & \textbf{20.18}\se{1.47} & \textbf{4.26}\se{1.30} & 2.87\se{0.84} & \textbf{4.30}\se{1.05} & 4.04\se{0.50} & 3.07\se{0.35} & \textbf{11.93}\se{0.78} \\
\specialrule{0.55pt}{1.2pt}{1.2pt}
\multirow{4}{*}{OLMo-3-1025-7B} & Logit Lens + ITGI & 5.81\se{1.27} & 3.42\se{0.94} & 9.02\se{1.02} & 0.00\se{0.00} & 0.00\se{0.00} & 0.66\se{0.24} & 0.00\se{0.00} & 0.00\se{0.00} & 0.53\se{0.27} & 1.94\se{0.42} & 1.14\se{0.31} & 3.40\se{0.36} \\
 & J-Lens & 2.22\se{1.05} & 0.77\se{0.69} & 10.07\se{1.41} & 0.00\se{0.00} & 0.00\se{0.00} & 1.62\se{0.34} & 0.00\se{0.00} & 0.00\se{0.00} & 1.44\se{0.49} & 0.74\se{0.35} & 0.26\se{0.23} & 4.38\se{0.51} \\
 & J-Lens + ITGI & 10.68\se{2.37} & 5.90\se{1.79} & 20.29\se{1.86} & 0.00\se{0.00} & 0.00\se{0.00} & 6.47\se{0.67} & 0.00\se{0.00} & 0.00\se{0.00} & 3.15\se{1.13} & 3.56\se{0.79} & 1.97\se{0.60} & 9.97\se{0.76} \\
\rowcolor{blue!7}
\cellcolor{white} & iLens  & \textbf{82.14}\se{2.97} & \textbf{69.57}\se{3.65} & \textbf{43.47}\se{3.45} & \textbf{44.20}\se{2.47} & \textbf{30.47}\se{2.16} & \textbf{44.20}\se{1.58} & \textbf{31.82}\se{3.72} & \textbf{23.03}\se{3.34} & \textbf{29.20}\se{3.39} & \textbf{52.72}\se{1.79} & \textbf{41.02}\se{1.80} & \textbf{38.96}\se{1.70} \\
\specialrule{0.55pt}{1.2pt}{1.2pt}
\multirow{4}{*}{Gpt-oss-20b} & Logit Lens + ITGI & 0.28\se{0.28} & 0.00\se{0.00} & 5.71\se{0.88} & 0.09\se{0.09} & 0.00\se{0.00} & 4.89\se{0.69} & 2.27\se{1.78} & 0.00\se{0.00} & 2.32\se{0.76} & 0.88\se{0.60} & 0.00\se{0.00} & 4.31\se{0.45} \\
 & J-Lens & 13.22\se{2.94} & 11.16\se{2.57} & 14.77\se{2.15} & 0.09\se{0.09} & 0.09\se{0.09} & 0.24\se{0.96} & 2.89\se{2.20} & 1.14\se{0.70} & 2.17\se{0.65} & 5.40\se{1.22} & 4.13\se{0.89} & 5.73\se{0.81} \\
 & J-Lens + ITGI & 7.58\se{2.27} & 3.86\se{1.62} & 19.63\se{2.14} & 0.09\se{0.09} & 0.00\se{0.00} & 13.95\se{1.45} & 9.92\se{3.73} & 6.30\se{2.51} & 7.01\se{1.51} & 5.86\se{1.46} & 3.39\se{1.00} & 13.53\se{1.00} \\
\rowcolor{blue!7}
\cellcolor{white} & iLens  & \textbf{78.51}\se{3.52} & \textbf{70.11}\se{3.76} & \textbf{42.09}\se{3.52} & \textbf{67.55}\se{3.04} & \textbf{58.18}\se{2.99} & \textbf{49.61}\se{3.00} & \textbf{70.97}\se{3.90} & \textbf{61.57}\se{3.86} & \textbf{60.11}\se{5.53} & \textbf{72.34}\se{2.02} & \textbf{63.29}\se{2.05} & \textbf{50.60}\se{2.40} \\
\bottomrule[0.9pt]
\end{tabular}%
}
\endgroup
\end{table*}

\providecommand{\se}[1]{$_{\textcolor{gray}{\pm#1}}$}

\begin{table*}[t]
\centering
\caption{Injected word gain across 12 language models. Cells report normalized layer AUC of the per-trial change (after minus before editing). Bold marks the best method within each model and task.}
\label{tab:llm-intervention-injected}
\begingroup
\fontsize{6}{7.2}\selectfont
\setlength{\tabcolsep}{1.2pt}
\renewcommand{\arraystretch}{1.05}
\resizebox{\textwidth}{!}{%
\begin{tabular}{@{}ll*{4}{rrr}|rrr@{}}
\toprule[0.9pt]
\multirow{2}{*}{Model} & \multirow{2}{*}{Method} & \multicolumn{3}{c}{Probe Swap} & \multicolumn{3}{c}{Verbal Report} & \multicolumn{3}{c}{Flexible Generalization} & \multicolumn{3}{c}{Verbal Introspection} & \multicolumn{3}{|c}{Average} \\
\cmidrule[0.55pt](lr){3-17}
& & $\Delta$P@10 $\uparrow$ & $\Delta$P@20 $\uparrow$ & $\Delta$Conf. $\uparrow$ & $\Delta$P@10 $\uparrow$ & $\Delta$P@20 $\uparrow$ & $\Delta$Conf. $\uparrow$ & $\Delta$P@10 $\uparrow$ & $\Delta$P@20 $\uparrow$ & $\Delta$Conf. $\uparrow$ & $\Delta$P@10 $\uparrow$ & $\Delta$P@20 $\uparrow$ & $\Delta$Conf. $\uparrow$ & $\Delta$P@10 $\uparrow$ & $\Delta$P@20 $\uparrow$ & $\Delta$Conf. $\uparrow$ \\
\midrule[0.65pt]
\multirow{4}{*}{Qwen3-0.6B-Inst.} & Logit Lens + ITGI & 26.15\se{16.02} & 32.31\se{19.10} & 2.23\se{3.39} & 9.77\se{3.15} & 8.23\se{2.78} & 0.62\se{0.19} & -0.96\se{0.96} & -2.01\se{1.29} & 0.01\se{0.01} & 0.00\se{0.00} & 0.00\se{0.00} & 0.00\se{0.00} & 8.74\se{4.09} & 9.63\se{4.84} & 0.71\se{0.85} \\
 & J-Lens & 34.62\se{24.57} & 36.15\se{25.44} & 4.66\se{2.93} & 4.15\se{2.26} & -1.00\se{2.86} & 0.30\se{0.18} & \textbf{1.31}\se{1.15} & \textbf{3.50}\se{2.20} & \textbf{0.22}\se{0.14} & 0.15\se{0.11} & 0.31\se{0.15} & 0.01\se{0.00} & 10.06\se{6.17} & 9.74\se{6.42} & 1.30\se{0.73} \\
 & J-Lens + ITGI & \textbf{36.15}\se{22.15} & \textbf{49.23}\se{21.53} & \textbf{16.08}\se{10.13} & 23.23\se{4.51} & 20.54\se{4.58} & 2.61\se{0.65} & 0.70\se{0.70} & 3.32\se{2.03} & 0.12\se{0.08} & 0.00\se{0.00} & 0.00\se{0.00} & 0.01\se{0.00} & \textbf{15.02}\se{5.65} & \textbf{18.27}\se{5.53} & \textbf{4.70}\se{2.54} \\
\rowcolor{blue!7}
\cellcolor{white} & iLens  & -16.15\se{24.79} & -6.15\se{20.55} & -3.52\se{3.36} & \textbf{40.23}\se{4.42} & \textbf{31.54}\se{5.22} & \textbf{11.54}\se{1.43} & 1.05\se{1.95} & 1.22\se{3.15} & 0.16\se{0.08} & \textbf{20.31}\se{2.26} & \textbf{22.62}\se{2.38} & \textbf{3.34}\se{0.60} & 11.36\se{6.34} & 12.31\se{5.39} & 2.88\se{0.93} \\
\specialrule{0.55pt}{1.2pt}{1.2pt}
\multirow{4}{*}{Qwen3-4B-Inst.} & Logit Lens + ITGI & 36.58\se{6.19} & \textbf{42.84}\se{6.80} & 4.12\se{1.44} & 6.59\se{3.18} & 1.65\se{1.12} & 0.46\se{0.17} & 0.00\se{0.19} & 0.00\se{0.00} & 0.13\se{0.13} & 0.00\se{0.00} & 0.00\se{0.00} & 0.00\se{0.00} & 10.79\se{1.74} & 11.12\se{1.72} & 1.18\se{0.36} \\
 & J-Lens & 25.64\se{5.00} & 30.09\se{5.50} & 6.40\se{2.27} & 1.00\se{3.58} & -0.41\se{1.72} & 0.01\se{0.06} & 2.41\se{1.17} & 2.67\se{1.29} & 1.91\se{1.33} & 0.00\se{0.00} & 0.00\se{0.00} & 0.00\se{0.00} & 7.26\se{1.56} & 8.09\se{1.48} & 2.08\se{0.66} \\
 & J-Lens + ITGI & \textbf{38.31}\se{5.74} & 42.08\se{6.76} & \textbf{9.98}\se{2.36} & 9.06\se{3.49} & 5.12\se{2.19} & 2.63\se{1.02} & 1.20\se{0.95} & 1.80\se{0.99} & 1.29\se{1.24} & 0.00\se{0.00} & 0.00\se{0.00} & 0.00\se{0.00} & 12.14\se{1.70} & 12.25\se{1.79} & 3.48\se{0.71} \\
\rowcolor{blue!7}
\cellcolor{white} & iLens  & 20.59\se{7.18} & 26.55\se{7.50} & 0.71\se{0.55} & \textbf{22.65}\se{4.30} & \textbf{17.94}\se{3.72} & \textbf{19.95}\se{3.28} & \textbf{5.21}\se{2.78} & \textbf{8.29}\se{2.63} & \textbf{2.81}\se{1.05} & \textbf{20.00}\se{1.95} & \textbf{22.59}\se{2.07} & \textbf{7.65}\se{1.13} & \textbf{17.11}\se{2.26} & \textbf{18.84}\se{2.25} & \textbf{7.78}\se{0.92} \\
\specialrule{0.55pt}{1.2pt}{1.2pt}
\multirow{4}{*}{Qwen3-8B-Inst.} & Logit Lens + ITGI & 17.71\se{4.22} & 22.49\se{4.61} & 2.53\se{1.01} & 1.76\se{1.37} & 0.71\se{0.49} & 0.91\se{0.43} & 1.67\se{1.69} & 0.53\se{0.42} & 0.00\se{0.00} & 0.00\se{0.00} & 0.00\se{0.00} & 0.00\se{0.00} & 5.29\se{1.19} & 5.93\se{1.16} & 0.86\se{0.27} \\
 & J-Lens & 24.51\se{4.09} & 27.94\se{4.44} & \textbf{7.08}\se{1.79} & -1.18\se{1.06} & -0.71\se{0.71} & -0.28\se{0.44} & 5.08\se{2.25} & 4.95\se{1.75} & 1.35\se{0.96} & 0.00\se{0.00} & 0.00\se{0.00} & 0.00\se{0.00} & 7.10\se{1.20} & 8.05\se{1.21} & 2.04\se{0.52} \\
 & J-Lens + ITGI & 19.91\se{4.24} & 24.51\se{4.70} & 4.32\se{1.26} & 2.82\se{1.90} & 4.12\se{1.87} & 2.50\se{1.02} & 4.21\se{2.73} & 1.87\se{0.89} & 0.16\se{0.13} & 0.00\se{0.00} & 0.00\se{0.00} & 0.00\se{0.00} & 6.74\se{1.35} & 7.62\se{1.28} & 1.74\se{0.41} \\
\rowcolor{blue!7}
\cellcolor{white} & iLens  & \textbf{31.19}\se{5.79} & \textbf{37.32}\se{6.78} & 1.76\se{0.49} & \textbf{11.88}\se{3.53} & \textbf{14.53}\se{3.87} & \textbf{14.11}\se{2.55} & \textbf{14.24}\se{3.68} & \textbf{15.71}\se{3.24} & \textbf{3.75}\se{1.15} & \textbf{22.00}\se{2.18} & \textbf{24.00}\se{2.17} & \textbf{4.73}\se{0.84} & \textbf{19.83}\se{2.00} & \textbf{22.89}\se{2.18} & \textbf{6.08}\se{0.74} \\
\specialrule{0.55pt}{1.2pt}{1.2pt}
\multirow{4}{*}{Qwen3-14B-Inst.} & Logit Lens + ITGI & 12.77\se{3.42} & 18.26\se{4.37} & 1.72\se{0.69} & 1.47\se{1.04} & 0.11\se{0.40} & 0.96\se{0.39} & 0.00\se{0.00} & -1.50\se{1.77} & 0.00\se{0.00} & 0.00\se{0.00} & 0.00\se{0.00} & 0.00\se{0.00} & 3.56\se{0.89} & 4.22\se{1.18} & 0.67\se{0.20} \\
 & J-Lens & 19.79\se{4.05} & 22.75\se{4.15} & \textbf{7.12}\se{2.05} & -0.11\se{1.92} & -2.53\se{2.13} & 0.04\se{0.22} & 0.12\se{0.12} & -0.54\se{1.84} & 0.03\se{0.03} & 0.00\se{0.00} & 0.00\se{0.00} & 0.00\se{0.00} & 4.95\se{1.12} & 4.92\se{1.25} & 1.80\se{0.52} \\
 & J-Lens + ITGI & 20.45\se{4.51} & 22.86\se{4.89} & 5.25\se{1.54} & 4.05\se{2.12} & 3.63\se{1.97} & 3.05\se{1.13} & 0.36\se{0.36} & 0.12\se{2.14} & 0.00\se{0.00} & 0.00\se{0.00} & 0.00\se{0.00} & 0.00\se{0.00} & 6.22\se{1.25} & 6.65\se{1.42} & 2.08\se{0.48} \\
\rowcolor{blue!7}
\cellcolor{white} & iLens  & \textbf{29.88}\se{6.24} & \textbf{35.25}\se{6.64} & 2.85\se{0.61} & \textbf{13.89}\se{3.51} & \textbf{16.42}\se{3.56} & \textbf{14.70}\se{2.83} & \textbf{16.81}\se{3.33} & \textbf{13.10}\se{3.43} & \textbf{3.00}\se{1.28} & \textbf{21.37}\se{2.31} & \textbf{24.16}\se{2.38} & \textbf{6.20}\se{1.07} & \textbf{20.49}\se{2.06} & \textbf{22.23}\se{2.15} & \textbf{6.69}\se{0.84} \\
\specialrule{0.55pt}{1.2pt}{1.2pt}
\multirow{4}{*}{Qwen2.5-7B-Inst.} & Logit Lens + ITGI & 11.89\se{4.17} & 12.47\se{5.31} & 1.88\se{0.66} & 4.00\se{1.72} & 1.77\se{1.18} & 0.62\se{0.36} & 2.62\se{1.32} & 2.27\se{1.63} & 0.00\se{0.00} & 0.00\se{0.00} & 0.00\se{0.00} & 0.00\se{0.00} & 4.63\se{1.17} & 4.13\se{1.42} & 0.62\se{0.19} \\
 & J-Lens & 26.22\se{5.57} & 23.78\se{5.75} & 7.44\se{2.64} & 4.15\se{1.97} & 3.62\se{1.92} & 0.84\se{0.48} & 7.52\se{2.44} & 6.29\se{2.47} & 1.27\se{1.11} & 0.00\se{0.00} & 0.00\se{0.00} & 0.00\se{0.00} & 9.47\se{1.60} & 8.42\se{1.64} & 2.39\se{0.73} \\
 & J-Lens + ITGI & \textbf{38.69}\se{5.88} & \textbf{38.11}\se{6.90} & \textbf{14.09}\se{3.45} & 14.69\se{3.90} & \textbf{14.77}\se{3.82} & 6.22\se{2.05} & 10.40\se{3.26} & 7.17\se{3.26} & 1.40\se{0.80} & 0.00\se{0.00} & 0.15\se{0.15} & 0.00\se{0.00} & 15.95\se{1.94} & 15.05\se{2.13} & 5.43\se{1.02} \\
\rowcolor{blue!7}
\cellcolor{white} & iLens  & 18.07\se{5.65} & 18.41\se{7.58} & 2.72\se{0.75} & \textbf{19.62}\se{4.02} & 14.23\se{4.05} & \textbf{12.83}\se{1.71} & \textbf{17.48}\se{3.35} & \textbf{10.84}\se{3.25} & \textbf{3.34}\se{0.90} & \textbf{21.46}\se{1.30} & \textbf{23.00}\se{1.41} & \textbf{4.82}\se{0.59} & \textbf{19.16}\se{1.95} & \textbf{16.62}\se{2.33} & \textbf{5.93}\se{0.54} \\
\specialrule{0.55pt}{1.2pt}{1.2pt}
\multirow{4}{*}{Qwen3.5-4B-Inst.} & Logit Lens + ITGI & 11.56\se{2.98} & 16.67\se{4.04} & 2.20\se{0.49} & 1.53\se{1.42} & 2.80\se{1.50} & 0.76\se{0.28} & 0.61\se{0.42} & 2.73\se{1.86} & 0.04\se{0.03} & 0.00\se{0.00} & 1.53\se{1.53} & 0.00\se{0.00} & 3.42\se{0.83} & 5.93\se{1.23} & 0.75\se{0.14} \\
 & J-Lens & 24.89\se{5.23} & 27.04\se{5.36} & 7.94\se{1.61} & -0.13\se{0.30} & 1.60\se{0.92} & 0.06\se{0.13} & 4.24\se{1.87} & 5.61\se{2.53} & 0.87\se{0.45} & 0.20\se{0.11} & 1.67\se{1.40} & 0.00\se{0.00} & 7.30\se{1.39} & 8.98\se{1.54} & 2.22\se{0.42} \\
 & J-Lens + ITGI & \textbf{37.56}\se{5.97} & \textbf{38.37}\se{6.09} & \textbf{16.53}\se{2.39} & 7.27\se{2.57} & 7.07\se{2.55} & 3.78\se{0.90} & 14.62\se{3.28} & 15.23\se{3.83} & 3.05\se{1.19} & 0.27\se{0.27} & 1.47\se{1.47} & 0.01\se{0.00} & 14.93\se{1.82} & 15.53\se{1.94} & 5.84\se{0.70} \\
\rowcolor{blue!7}
\cellcolor{white} & iLens  & 15.78\se{6.76} & 20.67\se{6.84} & 3.16\se{0.99} & \textbf{16.20}\se{4.55} & \textbf{17.07}\se{4.41} & \textbf{12.79}\se{2.14} & \textbf{25.15}\se{4.43} & \textbf{24.77}\se{5.04} & \textbf{7.91}\se{1.44} & \textbf{36.07}\se{3.31} & \textbf{39.13}\se{3.35} & \textbf{10.14}\se{1.41} & \textbf{23.30}\se{2.46} & \textbf{25.41}\se{2.53} & \textbf{8.50}\se{0.78} \\
\specialrule{0.55pt}{1.2pt}{1.2pt}
\multirow{4}{*}{Qwen3.5-9B-Inst.} & Logit Lens + ITGI & 14.47\se{2.96} & 26.33\se{4.79} & 1.80\se{0.44} & 1.60\se{1.26} & 5.20\se{2.64} & 0.54\se{0.24} & 4.32\se{2.44} & 0.53\se{2.04} & 0.03\se{0.02} & 0.00\se{0.00} & 0.00\se{0.00} & 0.00\se{0.00} & 5.10\se{1.01} & 8.02\se{1.46} & 0.59\se{0.12} \\
 & J-Lens & 25.73\se{4.28} & 31.73\se{5.16} & 6.05\se{1.43} & 1.33\se{1.17} & 2.93\se{1.38} & 0.66\se{0.27} & 7.12\se{2.32} & 4.09\se{2.48} & 0.54\se{0.45} & 0.07\se{0.07} & 0.07\se{0.07} & 0.00\se{0.00} & 8.56\se{1.25} & 9.71\se{1.47} & 1.81\se{0.38} \\
 & J-Lens + ITGI & \textbf{44.20}\se{5.32} & \textbf{49.60}\se{5.92} & \textbf{13.77}\se{2.06} & 9.33\se{2.68} & 13.00\se{3.75} & 5.56\se{1.37} & 18.26\se{3.82} & 8.41\se{3.05} & 2.14\se{1.04} & 0.00\se{0.00} & 0.80\se{0.80} & 0.02\se{0.01} & 17.95\se{1.77} & 17.95\se{1.92} & 5.37\se{0.67} \\
\rowcolor{blue!7}
\cellcolor{white} & iLens  & 33.33\se{5.84} & 42.20\se{6.00} & 5.05\se{1.00} & \textbf{17.60}\se{4.37} & \textbf{21.67}\se{4.46} & \textbf{16.15}\se{2.53} & \textbf{35.38}\se{4.52} & \textbf{22.42}\se{4.65} & \textbf{8.60}\se{1.89} & \textbf{35.60}\se{2.97} & \textbf{39.80}\se{3.13} & \textbf{6.89}\se{0.89} & \textbf{30.48}\se{2.27} & \textbf{31.52}\se{2.34} & \textbf{9.17}\se{0.86} \\
\specialrule{0.55pt}{1.2pt}{1.2pt}
\multirow{4}{*}{Gemma-3-4B-Inst.} & Logit Lens + ITGI & 12.50\se{4.01} & 18.27\se{4.26} & 4.53\se{1.09} & \textbf{12.44}\se{3.56} & \textbf{8.50}\se{3.11} & 12.32\se{2.73} & 2.93\se{2.20} & 1.66\se{2.12} & 0.79\se{0.39} & 0.00\se{0.00} & 0.06\se{0.06} & 0.00\se{0.00} & 6.97\se{1.45} & 7.12\se{1.42} & 4.41\se{0.74} \\
 & J-Lens & \textbf{17.05}\se{5.38} & 22.49\se{5.54} & \textbf{7.70}\se{2.17} & 0.94\se{1.11} & 1.12\se{1.49} & 2.44\se{1.52} & 9.18\se{3.24} & 5.52\se{3.46} & 1.70\se{0.96} & 0.00\se{0.00} & 0.00\se{0.00} & 0.00\se{0.00} & 6.79\se{1.59} & 7.28\se{1.67} & 2.96\se{0.70} \\
 & J-Lens + ITGI & 5.84\se{2.11} & 12.70\se{3.92} & 1.66\se{0.61} & 4.69\se{2.22} & 3.06\se{1.71} & 4.21\se{1.45} & -0.13\se{1.47} & 0.00\se{0.77} & 0.00\se{0.00} & 0.00\se{0.00} & 0.00\se{0.00} & 0.00\se{0.00} & 2.60\se{0.85} & 3.94\se{1.09} & 1.47\se{0.39} \\
\rowcolor{blue!7}
\cellcolor{white} & iLens  & 16.03\se{3.95} & \textbf{26.56}\se{4.66} & 4.66\se{1.02} & 12.38\se{3.21} & \textbf{8.50}\se{3.54} & \textbf{16.01}\se{2.60} & \textbf{12.83}\se{2.68} & \textbf{8.38}\se{2.92} & \textbf{3.45}\se{0.88} & \textbf{10.25}\se{1.74} & \textbf{18.12}\se{2.33} & \textbf{0.42}\se{0.15} & \textbf{12.87}\se{1.50} & \textbf{15.39}\se{1.74} & \textbf{6.13}\se{0.73} \\
\specialrule{0.55pt}{1.2pt}{1.2pt}
\multirow{4}{*}{Gemma-3-12B-Inst.} & Logit Lens + ITGI & 6.22\se{2.20} & 10.87\se{2.97} & 3.69\se{0.99} & 7.00\se{2.97} & 9.74\se{3.55} & 4.63\se{1.77} & 3.19\se{1.11} & 2.73\se{2.01} & 0.23\se{0.23} & 0.30\se{0.18} & 0.30\se{0.18} & 0.04\se{0.04} & 4.18\se{0.97} & 5.91\se{1.26} & 2.15\se{0.51} \\
 & J-Lens & 8.57\se{2.10} & 13.26\se{2.51} & 2.81\se{1.09} & 0.26\se{1.19} & 6.00\se{2.49} & 0.25\se{0.50} & 3.70\se{1.60} & 4.35\se{2.41} & 0.24\se{0.17} & 0.00\se{0.00} & 0.00\se{0.00} & 0.00\se{0.00} & 3.13\se{0.72} & 5.90\se{1.07} & 0.83\se{0.30} \\
 & J-Lens + ITGI & 1.00\se{1.37} & 5.00\se{2.22} & 1.08\se{0.58} & 3.57\se{2.22} & 8.04\se{3.11} & 2.00\se{0.91} & 1.02\se{0.56} & 2.41\se{2.07} & 0.00\se{0.00} & 0.00\se{0.00} & 0.09\se{0.09} & 0.00\se{0.00} & 1.40\se{0.67} & 3.88\se{1.09} & 0.77\se{0.27} \\
\rowcolor{blue!7}
\cellcolor{white} & iLens  & \textbf{12.09}\se{2.48} & \textbf{19.43}\se{3.06} & \textbf{4.61}\se{0.95} & \textbf{12.30}\se{3.45} & \textbf{12.26}\se{3.71} & \textbf{5.69}\se{1.55} & \textbf{6.48}\se{1.98} & \textbf{6.11}\se{2.30} & \textbf{0.54}\se{0.24} & \textbf{10.17}\se{1.91} & \textbf{15.91}\se{2.58} & \textbf{0.64}\se{0.30} & \textbf{10.26}\se{1.27} & \textbf{13.43}\se{1.48} & \textbf{2.87}\se{0.47} \\
\specialrule{0.55pt}{1.2pt}{1.2pt}
\multirow{4}{*}{Gemma-4-12B-Inst.} & Logit Lens + ITGI & -0.04\se{1.44} & -1.70\se{1.99} & 0.33\se{0.17} & -2.35\se{0.94} & -3.96\se{2.26} & 1.06\se{0.38} & -0.69\se{1.37} & 1.39\se{0.85} & 0.25\se{0.14} & 0.35\se{0.35} & -0.78\se{0.78} & 0.00\se{0.00} & -0.68\se{0.56} & -1.26\se{0.81} & 0.41\se{0.11} \\
 & J-Lens & 0.17\se{0.50} & -1.13\se{0.95} & 0.01\se{0.00} & -1.74\se{0.37} & \textbf{-2.04}\se{0.91} & 0.24\se{0.17} & -0.97\se{0.98} & 0.56\se{0.49} & 0.03\se{0.02} & 0.17\se{0.17} & -0.70\se{0.70} & 0.00\se{0.00} & -0.59\se{0.29} & -0.83\se{0.39} & 0.07\se{0.04} \\
 & J-Lens + ITGI & 2.83\se{1.06} & -0.09\se{1.62} & 0.58\se{0.23} & -1.65\se{0.76} & -2.65\se{1.21} & 0.82\se{0.34} & -1.39\se{1.64} & 1.06\se{0.55} & 0.15\se{0.09} & 0.09\se{0.09} & -0.61\se{0.61} & 0.00\se{0.00} & -0.03\se{0.53} & -0.57\se{0.54} & 0.39\se{0.11} \\
\rowcolor{blue!7}
\cellcolor{white} & iLens  & \textbf{3.91}\se{1.55} & \textbf{4.30}\se{2.67} & \textbf{0.59}\se{0.23} & \textbf{-0.09}\se{1.82} & -3.39\se{2.70} & \textbf{3.36}\se{0.97} & \textbf{1.99}\se{1.55} & \textbf{5.64}\se{1.36} & \textbf{0.72}\se{0.38} & \textbf{0.78}\se{0.70} & \textbf{-0.35}\se{0.64} & \textbf{0.02}\se{0.01} & \textbf{1.65}\se{0.73} & \textbf{1.55}\se{1.02} & \textbf{1.17}\se{0.27} \\
\specialrule{0.55pt}{1.2pt}{1.2pt}
\multirow{4}{*}{OLMo-3-1025-7B} & Logit Lens + ITGI & 15.13\se{4.02} & 11.97\se{3.67} & 1.57\se{0.46} & 7.53\se{2.57} & 4.93\se{2.20} & 0.95\se{0.23} & 2.58\se{1.76} & 0.00\se{0.57} & 0.44\se{0.23} & 0.00\se{0.00} & 0.00\se{0.00} & 0.00\se{0.00} & 6.31\se{1.27} & 4.22\se{1.08} & 0.74\se{0.14} \\
 & J-Lens & 25.98\se{6.27} & 24.53\se{6.55} & 3.99\se{1.25} & 5.13\se{2.37} & 3.87\se{2.18} & 0.50\se{0.25} & 4.47\se{2.58} & 4.47\se{1.86} & 0.90\se{0.33} & 0.00\se{0.00} & 0.00\se{0.00} & 0.00\se{0.00} & 8.90\se{1.80} & 8.22\se{1.79} & 1.35\se{0.33} \\
 & J-Lens + ITGI & \textbf{34.19}\se{6.53} & \textbf{33.68}\se{6.33} & \textbf{6.93}\se{1.72} & 23.93\se{4.98} & 14.33\se{4.38} & 6.36\se{0.94} & 6.59\se{3.01} & 3.41\se{1.70} & 2.08\se{0.95} & 0.00\se{0.00} & 0.67\se{0.67} & 0.01\se{0.00} & 16.18\se{2.19} & 13.02\se{1.98} & 3.84\se{0.54} \\
\rowcolor{blue!7}
\cellcolor{white} & iLens  & 27.44\se{7.42} & 24.96\se{7.13} & 1.29\se{0.76} & \textbf{41.80}\se{6.65} & \textbf{18.60}\se{5.59} & \textbf{48.63}\se{1.56} & \textbf{30.91}\se{4.91} & \textbf{31.06}\se{6.05} & \textbf{6.11}\se{0.99} & \textbf{32.60}\se{3.44} & \textbf{35.33}\se{3.58} & \textbf{2.42}\se{0.46} & \textbf{33.19}\se{2.91} & \textbf{27.49}\se{2.87} & \textbf{14.61}\se{0.51} \\
\specialrule{0.55pt}{1.2pt}{1.2pt}
\multirow{4}{*}{Gpt-oss-20b} & Logit Lens + ITGI & 17.36\se{5.70} & 10.61\se{5.07} & 0.69\se{0.17} & 8.36\se{3.50} & 1.27\se{1.09} & 1.15\se{0.32} & 1.86\se{2.20} & 0.93\se{1.99} & 0.54\se{0.31} & 0.00\se{0.00} & 0.00\se{0.00} & 0.00\se{0.00} & 6.89\se{1.76} & 3.20\se{1.39} & 0.60\se{0.12} \\
 & J-Lens & 18.32\se{6.80} & 6.06\se{4.46} & 0.85\se{0.32} & -0.55\se{2.44} & -0.73\se{1.88} & -0.17\se{0.42} & 5.06\se{2.37} & 3.31\se{2.11} & 0.53\se{0.21} & 0.00\se{0.00} & 0.00\se{0.00} & 0.00\se{0.00} & 5.71\se{1.90} & 2.16\se{1.32} & 0.30\se{0.14} \\
 & J-Lens + ITGI & \textbf{36.64}\se{7.19} & \textbf{24.79}\se{6.90} & \textbf{3.70}\se{0.75} & \textbf{13.73}\se{4.27} & \textbf{6.09}\se{2.77} & 4.81\se{1.22} & \textbf{8.88}\se{3.02} & \textbf{9.61}\se{3.53} & \textbf{1.33}\se{0.48} & 0.18\se{0.13} & 0.36\se{0.18} & 0.02\se{0.00} & \textbf{14.86}\se{2.22} & \textbf{10.21}\se{2.06} & \textbf{2.47}\se{0.38} \\
\rowcolor{blue!7}
\cellcolor{white} & iLens  & 3.17\se{6.83} & -8.68\se{7.69} & 0.35\se{0.27} & -1.45\se{5.27} & -12.36\se{5.20} & \textbf{5.63}\se{1.11} & -10.43\se{5.76} & -13.74\se{6.02} & 0.69\se{0.43} & \textbf{4.18}\se{1.00} & \textbf{7.91}\se{1.70} & \textbf{0.75}\se{0.22} & -1.13\se{2.60} & -6.72\se{2.80} & 1.85\se{0.31} \\
\bottomrule[0.9pt]
\end{tabular}%
}
\endgroup
\end{table*}

\section{Runtime Comparison}
\label{app:runtime}

We compare the runtime of the decoding methods using Qwen3.5-9B~\cite{qwen3.5}. We use the first 20 samples from the Typo task and decode the anchor token at layers $1,\ldots,L-1$ (31 layers for Qwen3.5-9B), since J-Lens does not provide a translator for the final layer. The initial forward pass used to obtain the residual-stream representations is shared across all methods and is therefore excluded from the measurements. Model loading and loading the precomputed J-Lens matrices are also excluded.

For each sample, we perform five timed runs after three warm-up runs and take the median runtime. We report the mean and standard error (SE) of these per-sample medians across the 20 samples.

We consider two execution settings. In the \emph{batched} setting, all 31 layers are decoded in a single call. The lens-based methods apply their respective readout operations to the stacked residual representations, while iLens propagates all 31 isolated states through their corresponding downstream computations together. This is the implementation used in the main experiments. In the \emph{non-batched} setting, each layer is decoded in a separate call. For iLens, this means independently propagating the selected state at layer $\ell$ through its remaining $L-\ell$ transformer blocks. We report both the total runtime across all 31 layers (\emph{All layers}) and the average runtime per layer (\emph{Per layer}), obtained by dividing the total by 31.

Table~\ref{tab:runtime} presents the results. As expected, iLens incurs additional computational overhead compared with the direct readout methods because it retains and evaluates the model's downstream nonlinear transformations. Nevertheless, the batched implementation requires only approximately $0.063$ seconds to decode one token across all 31 layers. 

Although iLens incurs higher computational overhead than the baseline methods, its absolute runtime remains negligible for the intended use case: decoding a token across all 31 layers takes only approximately 0.063 seconds with the batched implementation. More importantly, in interpretability research, the reliability and accuracy of the recovered representations are more important than minimizing the computational cost of the readout. Since the primary goal is to understand what information is encoded in intermediate representations, prioritizing accurate and faithful concept recovery is well justified, particularly when the additional runtime remains modest in absolute terms.

Furthermore, iLens requires no additional training, fitted parameters, or data-dependent readout mappings. In contrast, J-Lens relies on precomputed layer-wise translators constructed from data, introducing an additional preprocessing requirement whose cost is not reflected in the reported decoding times.

\begin{table*}[t]
\centering
\caption{Runtime comparison on Qwen3.5-9B~\cite{qwen3.5}, measured on a single NVIDIA L40S GPU in bfloat16. Values are reported in milliseconds (mean $\pm$ SE across 20 Typo samples). Batched execution decodes all 31 layers in one call, whereas non-batched execution uses a separate call for each layer. Per-layer runtime is calculated by dividing the total runtime by 31.}
\label{tab:runtime}
\small
\begin{tabular}{lcccc}
\toprule
 & \multicolumn{2}{c}{Batched} & \multicolumn{2}{c}{Non-batched} \\
\cmidrule(lr){2-3} \cmidrule(lr){4-5}
Method & Per layer & All layers & Per layer & All layers \\
\midrule
Logit Lens & 0.107\,{\scriptsize\textcolor{gray}{$\pm{<}0.001$}} & 3.32\,{\scriptsize\textcolor{gray}{$\pm{<}0.01$}} & 2.837\,{\scriptsize\textcolor{gray}{$\pm{<}0.001$}} & 87.96\,{\scriptsize\textcolor{gray}{$\pm$0.01}} \\
Embedding Lens & 0.107\,{\scriptsize\textcolor{gray}{$\pm{<}0.001$}} & 3.33\,{\scriptsize\textcolor{gray}{$\pm{<}0.01$}} & 2.825\,{\scriptsize\textcolor{gray}{$\pm{<}0.001$}} & 87.56\,{\scriptsize\textcolor{gray}{$\pm{<}0.01$}} \\
J-Lens & 0.196\,{\scriptsize\textcolor{gray}{$\pm{<}0.001$}} & 6.08\,{\scriptsize\textcolor{gray}{$\pm{<}0.01$}} & 2.938\,{\scriptsize\textcolor{gray}{$\pm{<}0.001$}} & 91.08\,{\scriptsize\textcolor{gray}{$\pm{<}0.01$}} \\
Patchscopes & 4.605\,{\scriptsize\textcolor{gray}{$\pm$0.002}} & 142.77\,{\scriptsize\textcolor{gray}{$\pm$0.08}} & 36.975\,{\scriptsize\textcolor{gray}{$\pm$0.002}} & 1146.23\,{\scriptsize\textcolor{gray}{$\pm$0.05}} \\
iLens (ours) & 2.041\,{\scriptsize\textcolor{gray}{$\pm{<}0.001$}} & 63.28\,{\scriptsize\textcolor{gray}{$\pm{<}0.01$}} & 18.544\,{\scriptsize\textcolor{gray}{$\pm$0.006}} & 574.87\,{\scriptsize\textcolor{gray}{$\pm$0.18}} \\
\bottomrule
\end{tabular}
\end{table*}

\clearpage

\clearpage


\clearpage



\clearpage

\clearpage

\end{document}